\documentclass[runningheads]{llncs}

\usepackage{eccv}

\usepackage{eccvabbrv}

\usepackage{graphicx}
\usepackage{booktabs}

\usepackage[accsupp]{axessibility}  

\usepackage{microtype}

\newcommand{\refsec}[1]{Sec.\,\ref{sec:#1}}
\newcommand{\figref}[1]{Fig.\,\ref{fig:#1}}
\newcommand{\tabref}[1]{Tab.\,\ref{tab:#1}}

\newcommand{\PAR}[1]{\vskip4pt \noindent {\bf #1~}}

\usepackage{soul}
\usepackage{colortbl}
\usepackage{amssymb}

\newcommand{\method}{\mbox{MICDrop}}

\newcommand{\degree}{70}

\usepackage{xcolor}
\usepackage{floatrow}

\usepackage{pifont} \newcommand{\xmark}{\ding{55}}
\newcommand{\cmark}{\ding{51}}

\definecolor{resnet}{HTML}{FAFBC9}
\definecolor{segformer}{HTML}{C9E9FB}
\definecolor{ours}{HTML}{C9FBD2}

\usepackage{hyperref}

\usepackage{orcidlink}

\begin{document}

\title{MICDrop: Masking Image and Depth Features via Complementary Dropout for Domain-Adaptive Semantic Segmentation} 

\titlerunning{MICDrop}
\author{Linyan Yang\inst{1, 3},
Lukas Hoyer\inst{2}\and
Mark Weber\inst{1, 3}\and
Tobias Fischer\inst{2}\and \\
Dengxin Dai\inst{2}\and
Laura Leal-Taixé\inst{4}\and
Marc Pollefeys\inst{2,5}\and
Daniel Cremers\inst{1,3}\and \\
Luc Van Gool\inst{2}
}

\authorrunning{L. Yang et al.}

\makeatletter
\let\oldand\and
\renewcommand{\and}{\unskip~\textbullet~}
\makeatother

\institute{$^{1}$ TU Munich, $^{2}$ ETH Zurich, $^{3}$ Munich Center for Machine Learning,\\ $^{4}$ NVIDIA, $^{5}$ Microsoft}

\maketitle

\begin{abstract}
Unsupervised Domain Adaptation (UDA) is the task of bridging the domain gap between a labeled source domain, e.g., synthetic data, and an unlabeled target domain. We observe that current UDA methods show inferior results on fine structures and tend to oversegment objects with ambiguous appearance. To address these shortcomings, we propose to leverage geometric information, i.e., depth predictions, as depth discontinuities often coincide with segmentation boundaries. We show that naively incorporating depth into current UDA methods does not fully exploit the potential of this complementary information. To this end, we present MICDrop, which learns a joint feature representation by masking image encoder features while inversely masking depth encoder features. With this simple yet effective complementary masking strategy, we enforce the use of both modalities when learning the joint feature representation. To aid this process, we propose a feature fusion module to improve both global as well as local information sharing while being robust to errors in the depth predictions. We show that our method can be plugged into various recent UDA methods and consistently improve results across standard UDA benchmarks, obtaining new state-of-the-art performances. Code: \url{https://github.com/ly-muc/MICDrop}
\keywords{Domain Adaptation \and Semantic Segmentation \and Depth Guidance}
\end{abstract}

\begin{figure*}[t]
    \centering
    \footnotesize
    \begin{tabular}{@{}cc@{}}
    \includegraphics[width=0.4\linewidth]{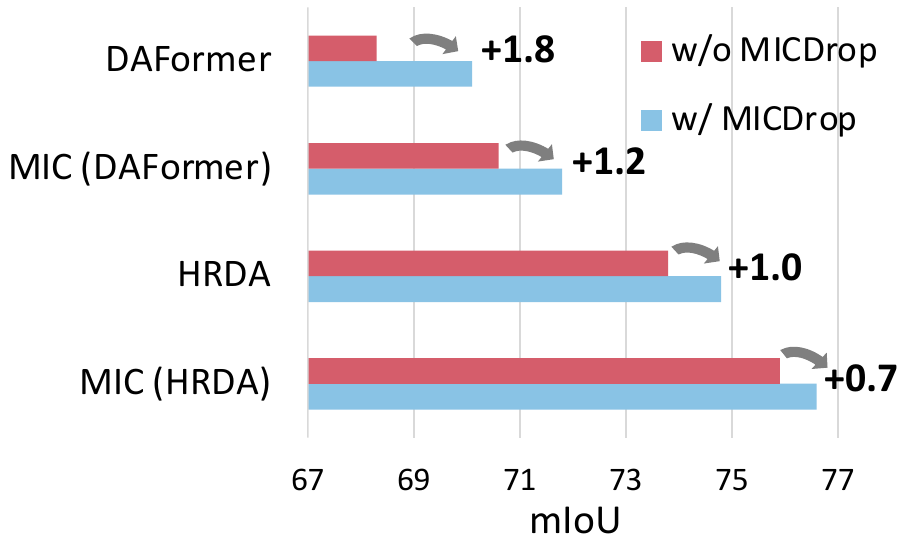} &
    \includegraphics[width=0.55\linewidth]{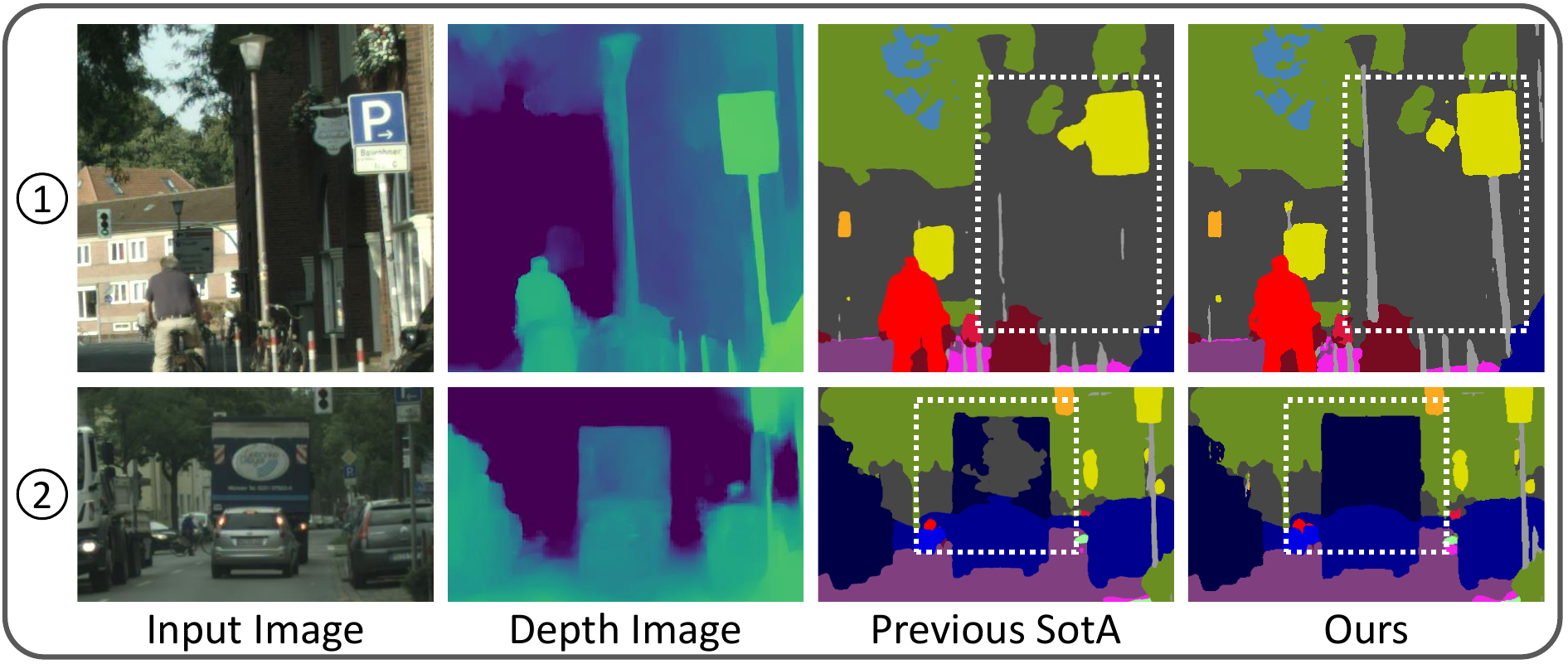} \\
    a) Quantitative improvement & b) Qualitative Examples \\
    \end{tabular}
    \caption{Previous UDA methods such as MIC~\cite{MIC} struggle with the segmentation of fine structures (top row) and oversegmentation of difficult objects (bottom row). Therefore, we propose \method\ to improve semantic segmentation UDA with depth estimates, which can capture fine structures and are consistent within object boundaries.
   We apply \method\ to four different methods on the GTA$\to$Cityscapes benchmark and show consistent improvements.}
    \label{fig:teaser}
\end{figure*}
\section{Introduction}
\label{sec:intro}
The computer vision community has seen tremendous success in recognition tasks over the years, yet the issue of efficiently sourcing large volumes of labeled images for supervised training of neural networks persists. 
This problem is especially pronounced in semantic segmentation, where manually creating labels is particularly labor-intensive~\cite{CityScapes, sakaridis2021acdc}.
Alternatively, images can be obtained at a large scale from a simulator, which can also easily generate the corresponding segmentation labels.
In that scenario, models are often trained on synthetic datasets and later applied to real-world data. 
This transition frequently results in a noticeable performance decline due to the variance in data distribution between the synthetic source and real-world target sets (\eg appearance of objects), a phenomenon known as domain shift. 
Therefore, conventional training mostly considers datasets where the training and test data are drawn from the same distribution. 
However, this assumption often breaks in real-world applications under domain shifts.
Recognizing these challenges, researchers have been exploring ways to minimize or eliminate the need for annotated data from the target domain.
This paper focuses on Unsupervised Domain Adaptation (UDA), where a model is trained using labeled data from a synthetic source domain and unlabeled data from a real-world target domain.

\PAR{Current challenges in UDA.} Recent UDA methods~\cite{DAFormer,HRDA,SePiCo,MIC,chen2023pipa} are able to significantly reduce the gap to methods trained in a fully supervised fashion on the target domain. 
However, state-of-the-art methods struggle with two main aspects shown in \figref{teaser}: (1) Despite using high-resolution strategies such as HRDA~\cite{HRDA}, they still face problems with fine structures and high-frequency details. (2) UDA methods are prone to oversegmentation when visual appearance clues are ambiguous.
These issues motivate us to look into scene representations that are more robust to appearance changes and provide precise boundaries to strengthen the existing UDA models.

\PAR{Complementary representation.}
An appearance-based image representation is essential to our task, however, a \emph{geometric representation} could provide complementary cues when it comes to segmentation.
In particular, the correlation of depth and segmentation boundaries can help to address challenges (1) and (2), as shown in \figref{teaser}. 
First, a pole might blend with a building behind it in color, but its depth profile is distinct, simplifying its segmentation.
Second, the back of the truck might have visual features that could also be part of a building.
However, the depth is smooth within the boundaries of the truck, suggesting that the semantic class should be consistent.
While actively measured depth might not be available, advances in image-based depth estimation~\cite{UniMatch, MonoDepth2} enable us to explore the task in a general setting.
Previous works~\cite{xMUDA, GUDA, CorDA} in UDA have focused on improving the learning process via an auxiliary depth prediction task.
In such multi-task learning settings, a network is trained to predict both depth and semantics from RGB inputs.
However, multi-task learning adds additional complexity, including balancing multiple network branches and their corresponding losses, to the already challenging UDA setting. 

\PAR{Contributions.} We propose a more streamlined approach: Instead of engaging in multi-task learning, we treat it as a modality fusion problem. 
Rather than producing multiple outputs from a single input (one-to-many), we redefine and simplify the task as a many-to-one prediction problem.
With semantic segmentation as our output and readily available depth estimates, we
study two research questions:
First, what is the most effective method to \emph{fuse features from two modalities} in a UDA context? 
Second, how can we \emph{utilize existing work} and design our method as a \emph{plugin network} that seamlessly integrates with pretrained models, thereby eliminating the need for extensive retraining?

We integrate our findings into \method, a novel framework for leveraging depth in domain-adaptive semantic segmentation.
Our framework is based on a novel \emph{cross-modality complementary dropout} technique along with a tailored masking schedule.
Our masking strategy mitigates the tendency of the network to underutilize additional depth features, as is prevalent in multi-modal learning, and becomes more pronounced with pretrained networks.
In particular, we foster cross-modal feature learning by strategically corrupting both RGB and depth features in a complementary manner, enforcing the utilization of the different modalities to fill in masked information.
To integrate information from both modalities effectively, we also propose a \emph{cross-modality feature fusion} module. It is designed to integrate global and local cues from one modality to the other.
First, it computes depth feature similarities to aggregate RGB features based on the resulting attention map, aiding the RGB feature aggregation with global depth cues.
This is particularly beneficial for segmenting objects that the RGB encoder struggles to represent accurately but have a smooth depth profile.
Second, it applies local self-attention to depth features, leveraging the discontinuity in local depth for describing boundaries, a critical factor in identifying thin structures.
This approach yields significant improvements over various recent UDA methods on two standard benchmarks while only requiring the training of a light-weight plugin network for a low number of iterations. Thus, MICDrop (w/ DAFormer) can be trained within 11 hours on a single GTX Titan X GPU (12 GB).

\noindent In summary, our key contributions are:
\begin{itemize}
    \item A \textbf{complementary feature masking strategy} for depth and RGB, fostering cross-modal feature learning.
    \item A \textbf{cross-modality fusion module} to improve segmentation based on depth by using global and local cues.
    \item Comprehensive ablations demonstrating \method's efficacy, with \textbf{improvements ranging from 0.7 to 1.8 mIoU} across four recent UDA methods on the GTA\textrightarrow Cityscapes benchmark.
\end{itemize}

By showing that complementary geometric information even improves modern high-resolution, Transformer-based UDA methods, we hope to lay the foundation for future research exploring the merits of auxiliary modalities for semantic segmentation UDA.

\section{Related Work}
\label{sec:related}
\PAR{Unsupervised Domain Adaptation (UDA).}
In UDA, methods have access to labeled source and unlabeled target data at training time and can mostly be categorized into two primary groups.
The first one utilizes a Generative Adversarial Network (GAN)~\cite{GANs} to align input images~\cite{CyCADA}, image features~\cite{hoffman2016fcn}, or output features~\cite{AdaptSegNet, Advent, Saito2018} across domains.
The second stream of works are built on self-training~\cite{grandvalet2004semi, lee2013pseudo}.
Here, pseudo labels are created using a teacher network~\cite{EMA, SAC}.
These labels can be further refined using confidence thresholds~\cite{Mei2020, Zhang2018, Zou2018} and pseudo label prototypes~\cite{Pan2019, ProDA, CAGUDA}.
The student model then receives an image version with cross-domain class mix~\cite{DACS, ContextMixup} and color augmentations~\cite{SAC}. The self-training can further strengthened by domain-robust Transformers~\cite{DAFormer,xu2021cdtrans,saha2023edaps}, class-balanced sampling~\cite{Zou2018,DAFormer}, multi-resolution adaptation~\cite{HRDA}, or contrastive learning~\cite{SePiCo, chen2023pipa}.
Our proposed \method\ builds on the self-training paradigm.

\PAR{Depth in Semantic Segmentation.}
Several works in semantic segmentation have shown the merits of leveraging geometric cues. In one branch of work~\cite{Mti-Net, CorDA, hoyer2021three, SPIGAN, DADA, CTRL-DA}, depth estimation is only used as an auxiliary task.
Different from that, some methods~\cite{Papnet, PadNet, XTAM} explore multi-task learning from RGB input, in which depth is another output target.
Similar multi-task studies~\cite{xMUDA, GUDA, CorDA, hoyer2023improving} have also been made in the context of UDA.
In both cases, this requires a bi-directional feature exchange across modalities.
Our method, however, is more closely related to RGB-D semantic segmentation~\cite{ACNet, SAGate}, in which both RGB and depth are input, while semantic segmentation is the only output, and hence we focus on a \emph{uni-directional} feature refinement from depth to RGB.
In our case, depth also serves the purpose of reducing the domain gap further.
While some methods~\cite{PadNet, ResidualExcite, ACNet} uses variants and extensions of the Squeeze-and-Excitation Block~\cite{SEBlock} for cross-modal feature fusion, more recent methods~\cite{XTAM, TokenFusion, SAGate, CMX} propose softmax attention-based aggregation.
Inspired by their success, we propose a combination of an excitation block for local windows and a cross-attention block for global reasoning.
Crucially, we use a \emph{depth-guided attention map}, thereby enhancing uni-directional guidance using geometric data.
In contrast to previous RGB-D works such as \cite{CMX, SAGate, liu2020S2MA}, we leverage both local and global dependencies for domain-robust depth-to-segmentation refinement. We show in Tab.~\ref{tab:ablation}b that leveraging geometric cues is not trivial in the context of UDA and conventional cross-attention fails here.

\PAR{Masked Image Modeling (MIM).}
MIM is a powerful method for self-supervised pretraining.
In this approach, information is withheld in order to train the network to recover certain targets.
Such reconstruction targets can range from RGB inputs~\cite{MAE, SimMIM}, to HOG features~\cite{HOGMIM}, to visual tokens~\cite{BEiT, PECO}.
MultiMAE~\cite{MultiMAE} shows the benefits of using masking of input patches in \emph{supervised} multi-task learning by using a shared encoder and modality-specific decoder. 
In contrast to their work, we propose complementary masking in a UDA setup on a \emph{multi-resolution feature level} in separate encoders (instead of input masking), enabling the use of pretrained RGB encoders. 
MIC~\cite{MIC} applies MIM to UDA to improve context reasoning.
Different from MIC, we propose a novel complementary multi-modal feature dropout to facilitate cross-modality learning instead of only masking RGB inputs for context enhancement.
In Sec.~\ref{sec:ablation}, we show that complementary feature dropout is orthogonal and further boosts networks trained with MIC.

\section{Method}
\label{sec:method}

\begin{figure*}[t]
\centering
\footnotesize
\begin{tabular}{@{}c|c@{}}
\includegraphics[width=0.49\linewidth]{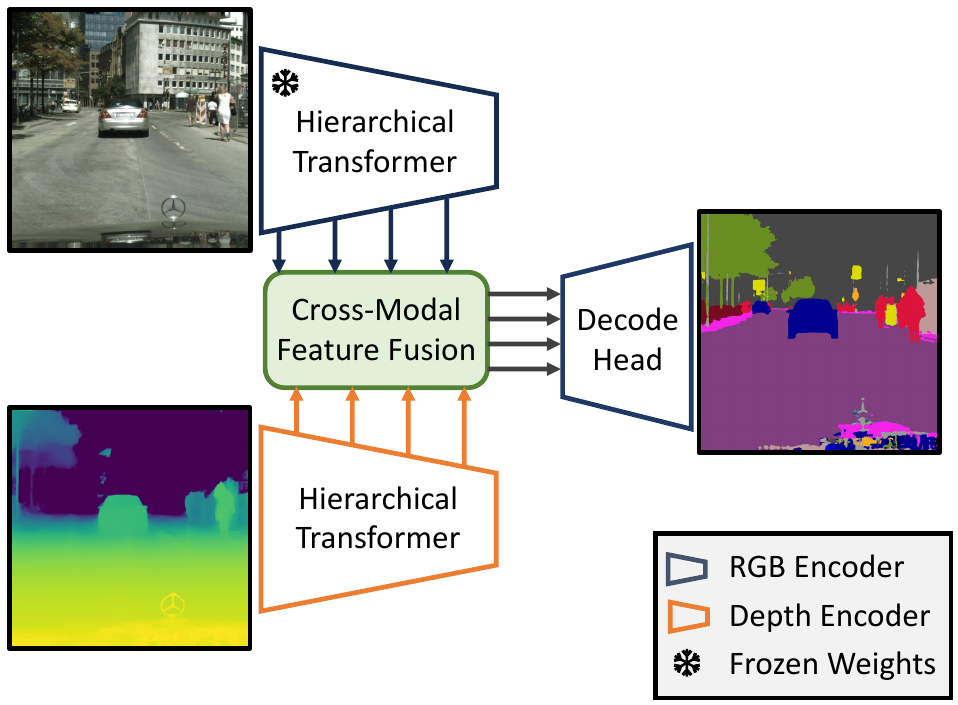} &
\includegraphics[width=0.49\linewidth]{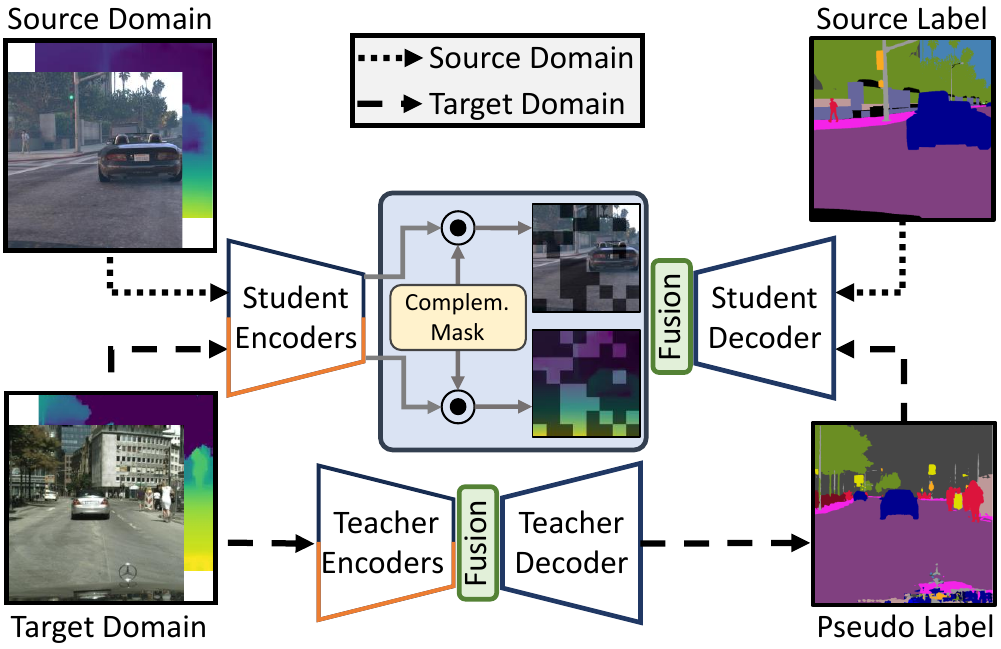} \\
a) Architecture & b) Training Scheme \\
\end{tabular}
\caption{\textbf{Method overview.} Our proposed architecture is visualized on the left side. We use a light-weight hierarchical depth encoder and process the features in our proposed cross-modal feature fusion module.
On the right side, we illustrate our training pipeline, in which source and target images are fed through the student encoders.
Then, our proposed cross-modality complementary dropout is applied to the corresponding features on each feature resolution.
Finally, we feed them through our fusion block, followed by the decoder, to make a final prediction. 
}

\label{fig:my_arch}
\end{figure*}

\PAR{Overview.} In~\figref{my_arch}, we present our method, featuring two novel modules that can be plugged into various UDA methods to leverage geometric cues. Our feature fusion module (\refsec{methods_feature_fusion}) integrates auxiliary inputs, \eg, depth, into RGB features. It fuses global and local information via attention-based aggregation. Our masking module (\refsec{methods_masking}) ensures balanced input use, avoiding pure reliance on a single input modality such as RGB or depth. We outline UDA training essentials before diving into the details of the multi-modal feature fusion module and the masking strategy.

\PAR{Problem Definition.} We tackle the problem of unsupervised domain adaptation, in which we have access to labeled source data ($\textbf{X}_s$, $\textbf{Y}_s$) and unlabeled target data ($\textbf{X}_t$) to train a neural network. The goal is to bridge the domain gap between $\textbf{X}_s$ and $\textbf{X}_t$. The performance is measured on a labeled hold-out validation set of the target domain. In this work, we focus on RGB images and depth images as input and semantic segmentation as output.

\PAR{Preliminaries.} 
Training a network on a source domain typically follows standard supervised methods. However, overcoming the domain gap with the target domain requires leveraging the unlabeled target data.
Recent approaches, such as those in~\cite{SAC, DACS, DAFormer, SePiCo, HRDA, MIC, Fredom}, adopt a student-teacher framework.
In this framework, the teacher network is updated each training iteration as an exponential-moving average (EMA)~\cite{EMA} of the student network.
This EMA teacher generates pseudo labels on the target images, which in turn act as a supervisory signal ($\textbf{X}_t$, $\tilde{\textbf{Y}}_t$) to the student.
We follow standard practice~\cite{DACS} and present the student with a heavily augmented view while presenting the teacher with a weakly augmented view of an image.
Additionally, we note that most methods use hierarchical encoders to produce multi-resolution feature maps, enhancing fine-grained segmentation.
We study the effectiveness of our proposed method by extending existing pretrained hierarchical encoders~\cite{DAFormer,HRDA} to leverage depth.
The depth estimates are obtained from RGB images. If stereo image pairs are available, we utilize UniMatch~\cite{UniMatch}. If not, we use the monocular method MonoDepth2~\cite{MonoDepth2}.

\subsection{Multi-Modal Feature Fusion}
\label{sec:methods_feature_fusion}
\begin{figure}
    \floatbox[{\capbeside\thisfloatsetup{capbesideposition={right,center},capbesidewidth=0.49\linewidth}}]{figure}[\FBwidth]
    {\caption{\textbf{Feature fusion of RGB and depth.} The presented method comprises two key components: a \textcolor[HTML]{557bc2}{global} and a \textcolor[HTML]{74992e}{local} attention module. 
    The local attention module refines information coming from depth within a local window by using sigmoid gates.
    In contrast to that, the global attention module aggregates image features based on similarity in their corresponding depth features, and thus providing more global context.
    Finally, the residual feature fusion block fuses all features.}
    \label{fig:feature_fusion}}
    {\includegraphics[width=0.9\linewidth]{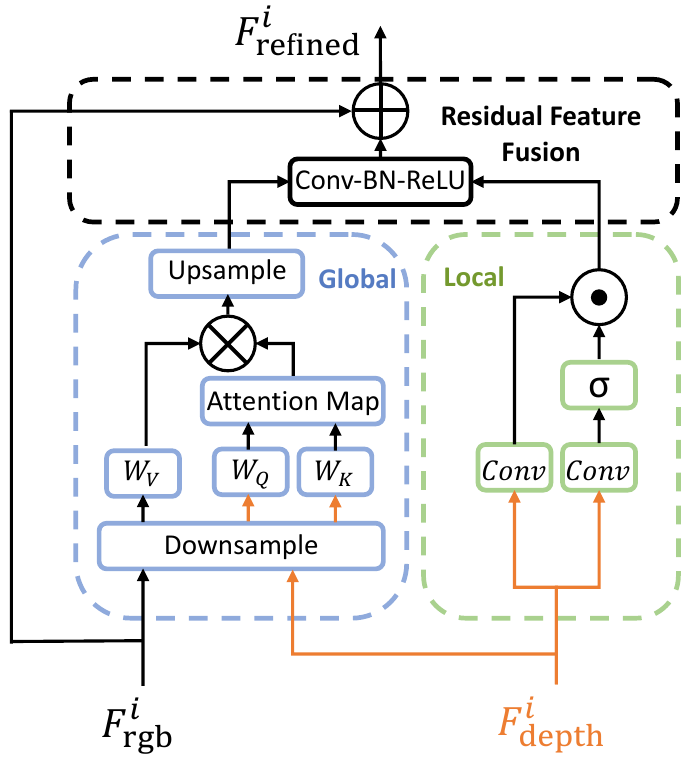}}
\end{figure}

First, we study the fusion of features from different modalities.
Our goal is to have a \emph{light-weight} training pipeline, which can make use of already \emph{existing work} in UDA.
For that purpose, we construct a multi-modality encoder that contains two individual encoders, one for RGB features and one for depth features.
The depth features come from a newly trained, light-weight depth encoder, while the RGB features come from a pretrained RGB feature encoder.
As state-of-the-art encoders typically output multi-scale feature levels, we perform feature fusion separately on each level. 
Different from multi-task learning, in which features from different modalities are \emph{all} refined, our goal is \emph{solely} to improve semantic segmentation. 
Thus, we focus on a unidirectional refinement, \ie, the depth features are used to enrich the RGB features but not vice-versa.
As can be seen in Fig.~\ref{fig:feature_fusion}, we divide our feature fusion block into (1) \emph{global} depth-guided cross-attention, (2) \emph{local} self-attention and (3) final \emph{residual fusion}.

\PAR{Global Depth-Guided Cross-Attention.}
Intuitively, similarities in depth features can provide a strong cue towards the same semantic class.
For example, large objects like bus or train exhibit similar gradual changes within their object, while thin structures such as pole or sign typically exhibit rapid depth changes relative to their surroundings.
Such additional cues could serve as a \emph{correctional and complementary signal} to the RGB features when predicting the semantic class.
Thus, the purpose of this branch is to aggregate RGB features globally based on their corresponding depth feature similarity.
For such a global aggregation across different tasks, one natural choice would be cross-attention.

However, directly using (global) attention usually exhibits problematic scaling behavior.
Given an input $x \in \mathbb{R}^{H\times W\times C}$, the standard attention~\cite{Attention} has a complexity of $\mathcal{O}((HW)^2C+HWC^2)$ making it computationally infeasible for our case. 
We, therefore, bilinearly downsample high-resolution feature maps to reduce the spatial dimensions before applying cross-attention.
During training, we sample feature maps with a pooling factor of \{4, 2, 1, 1\} for low- and high-level features, respectively. Conversely, during inference and pseudo-label generation, this pooling is adjusted to \{2, 1, 1, 1\} and thus only applied to low-level features.

Given potentially downscaled depth features $\mathbf{F}_{\text{depth}}^i$ at level $i$, we obtain depth-based queries $\mathbf{Q}_{\text{depth}}$ and keys $\mathbf{K}_{\text{depth}}$ by using projection weights $\textbf{W}_q^i$ and $\textbf{W}_k^i$.
The corresponding RGB features $\mathbf{F}_{\text{rgb}}^i$ are downscaled in similar fashion and serve as values $\mathbf{V}_{\text{rgb}}$ after being projected by $\textbf{W}_k^i$.
Formally, the cross-attention for the aggregation is:
\begin{equation}
    \mathbf{F}^i_{\text{global}} = \text{softmax}\left(\frac{\mathbf{Q}_{\text{depth}}\mathbf{K}_{\text{depth}}^T,}{\sqrt{d_k}}\right)\, \mathbf{V}_{\text{rgb}}
\end{equation}
\PAR{Local Self-Attention.}
Modeling global interactions on a downsized resolution might not be enough to capture the fine details of objects like sign or pole.
Keeping the same computational complexity problem for the self-attention of depth features in mind, we argue that many important interactions for segmentation happen within a \emph{local} window.
Specifically, \emph{depth discontinuities} provide strong cues for \emph{boundary regions} among semantic classes, while \emph{smooth and continuous depth} indicate \emph{no change} in semantics.
Thus, we hypothesize that restricting the self-attention to a local window would still capture important \emph{complementary signals} to the global information.

To model such dynamics, we draw inspiration from earlier work~\cite{LSTM,PadNet}, in which sigmoid gates were used successfully to control the local flow of information without adding a large computational overhead.
In particular, we use two $3\times 3$ convolutions.
The first convolution outputs features into a sigmoid function $\sigma$ to obtain a \emph{local} attention map.
We note that \emph{no pooling} has been used, enabling the network to model a precise local control flow.
The second convolution is used to refine the depth features, which are fed into a pointwise multiplication together with the local attention map.
As this branch is used to model \emph{complementary} features, we exclusively use depth features.
Formally, we compute the local self-attention as:
\begin{equation}
    \mathbf{F}^i_{\text{local}} = \sigma \left(\text{Conv}_{3\times 3}\left(\mathbf{F}^i_{\text{depth}}\right)\right) \odot \text{Conv}_{3\times 3}\left(\mathbf{F}^i_{\text{depth}}\right)
\end{equation}

\PAR{Residual Feature Fusion.}
After aggregating global and local features, we propose a simple two-step feature fusion block to fuse all aggregated features as well as the original RGB features.
At first, the depth-guided global features $\mathbf{F}^i_{\text{global}}$ and local features $\mathbf{F}^i_{\text{local}}$ are concatenated ($||$) and fused through a $1\times1$-Conv-BN-ReLU block.
After that, the original RGB features $\mathbf{F}_{\text{rgb}}$ are added, resulting in the refined features:
\begin{equation}
    \mathbf{F}^i_{\text{refined}} = \mathbf{F}^i_{\text{rgb}} + \text{ReLU}(\text{BN}(\text{Conv}(\mathbf{F}^i_{\text{global}} || \mathbf{F}^i_{\text{local}}))) 
\end{equation}
The refined features are then fed to a DAFormer head~\cite{DAFormer} for the final predictions.

\subsection{Complementary Feature Masking}
\label{sec:methods_masking}
During initial experiments, we observe that simply providing estimated depth and RGB images to the network does not enable the network to leverage the full potential of all provided information.
We refer the reader to \refsec{ablation} for details of that analysis.
We hypothesize that the network grows too confident in the RGB encoder and thus dismisses complementary information from the depth encoder, which limits adaptability to the target domain.
To improve cross-modal information exchange, we therefore introduce a \emph{cross-modal masking strategy}.
In contrast to Masked Image Modeling in UDA~\cite{MIC}, our method involves masking the learned representation of \emph{different modalities} on \emph{feature-level} rather than a single modality on input-level.
Moreover, our masking strategy and schedule are specifically designed to improve redundancy across modalities and prevent getting stuck in local minima due to one encoder already being pretrained.

\PAR{Complementary Dropout.} For this, we propose using \emph{blockwise dropout}~\cite{DropBlock} to generate masked features.
When masking only individual pixels, the information could be easily restored from the neighborhood in the same modality without requiring the other modality.
When masking larger blocks, the network has to understand the semantics of the other modality to recover the missing information.
We, therefore, opt to mask out whole blocks within the feature map according to a predefined schedule.
Furthermore, we hypothesize that learning complementary features across modalities can be achieved best by masking the feature maps of different modalities in \emph{complementary} fashion.
For example, if we mask 70\% of the RGB features, we would mask the remaining 30\% in the depth features.
This intuitively corrupts the information across modalities and forces the network to rely on features from all modalities.
\figref{my_arch} illustrates the idea of complementary dropout. 
Formally, we define the complementary masking as:
\begin{equation}
    \mathbf{M}_{\text{rgb}}(u,v) = [\gamma > m_r^t], \quad \gamma \sim \text{Uniform}(0, 1)
\end{equation}
\begin{equation}
    \mathbf{M}_{\text{depth}}(u, v) = 1 - \mathbf{M}_{\text{rgb}}(u,v)
\end{equation}
where $m_r^t$ denotes the masking ratio at iteration $t$ and ($u,v$) the block index of the $i$-th feature map.
To fulfill our goal of true cross-modal complementary masking, we use the same masking across all feature map levels and experimentally validate that design in~\refsec{ablation}.
Conceptually, this avoids the recovery of features within the feature pyramid of the same modality.
Therefore, our method is designed to foster the \emph{transfer of complementary information} and to promote the \emph{learning of potentially redundant information}, which in turn increases robustness and reduces sensitivity to domain-specific appearance changes.

\PAR{Masking Schedule.}
Prior studies~\cite{DropBlock, DropKey} have highlighted the limitations of a static masking ratio.
In response, we adopt a dynamic masking ratio schedule for RGB and depth features.
This approach is particularly effective when using a pretrained encoder for one modality and an untrained encoder for the other, as it compensates for the initial disparity in feature quality.
At the beginning of the training, we keep a high proportion of depth features to accelerate the training of the depth encoder and improve its feature quality.
As training progresses, the schedule is adjusted to gradually reduce depth feature retention, thereby increasing the reliance on the RGB encoder.
We note that this masking is only applied during training but not during inference.
This method not only promotes an efficient exchange of information between modalities but also capitalizes on the depth data to bolster semantic learning in the early stages of training.

\section{Experiments}
\label{sec:exp}

\PAR{Datasets.} We perform our experimental evaluation on two widely used UDA benchmarks. The first one uses synthetic source data from the GTA~\cite{GTA} dataset, which contains 24,966 images with a resolution of $1914{\times}1052$.
The second benchmark uses SYNTHIA~\cite{Synthia}, which consists of 9,400 synthetic images with a resolution of $1280{\times}760$.
In both cases, the target dataset is Cityscapes~\cite{CityScapes}, which includes 500 validation images, each having a resolution of $2048{\times}1024$.

\PAR{Depth Estimates.}
\label{par:depth_estimation}
We obtain depth estimations for the source domain from  MonoDepth2~\cite{MonoDepth2} via self-supervised monocular depth estimation trained on image sequences from VIPER(GTA)~\cite{VIPER} and SYNTHIA-SEQ.
For Cityscapes, we obtain disparity estimations from UniMatch~\cite{UniMatch} using stereo images trained on a large synthetic dataset~\cite{SceneFlow}. 

\PAR{Metrics.} Following previous studies, we report the mean Intersection over Union (mIoU) in \% over the 19 common categories shared by GTA and Cityscapes and the 16 common categories shared by SYNTHIA and Cityscapes.

\PAR{Network Architecture.} We use DAFormer~\cite{DAFormer} as our baseline model for ablation studies, as it achieves a strong  performance at high 
training and inference speed.
As depth feature extractor, we use the light-weight MiT-B3~\cite{SegFormer}.
To demonstrate the \emph{plugin} capability of our method, we additionally apply \method\ to the state-of-the-art methods HRDA~\cite{HRDA} and MIC~\cite{MIC}.

\PAR{Training details. }
We use an AdamW~\cite{AdamW} optimizer with a learning rate of $6{\times}10^{-5}$ for the depth encoder and $6{\times}10^{-4}$ for the decode head and feature fusion module. To address the limited scale of SYNTHIA, we align the learning rate for all modules to the depth encoder.
As learning rate schedule, we use linear warm-up in the first 1.5k iterations and polynomial decay with factor 0.9 afterward. 
The EMA~\cite{EMA} teacher is updated with a momentum of $\alpha{=}0.999$ at each step.
Following prior works~\cite{DACS,DAFormer,MIC}, the batch size is set to 2, with data augmentations such as color jitter, Gaussian blur, and cross-domain class mixing.
We initialize the RGB encoder and decode head with the publicly available pre-trained weights~\cite{DAFormer,HRDA,MIC} for our experiments. The depth encoder is initialized with ImageNet weights.
We keep the RGB encoder frozen and train the rest of the network 20k iterations on both GTA~\cite{GTA} and SYNTHIA~\cite{Synthia}.
We use a cross-entropy loss for both source and target images. We additionally apply a forward pass without masking to reduce the feature distribution shift between training and (unmasked) inference time.
\method\ can be trained in 11 hours using a single GTX Titan X GPU (12 GB) with DAFormer and in 17 hours on two Titan GPUs with HRDA.

\subsection{Main Results}
\label{sec:sota}

\begin{table}[t!]
\centering
\setlength{\tabcolsep}{1.5pt}
\resizebox{\textwidth}{!}{
\footnotesize
\begin{tabular}{l|c|ccccccccccccccccccc}
\toprule[0.2em]
Method & \rotatebox{\degree}{mIoU} & \rotatebox{\degree}{Road} & \rotatebox{\degree}{S.walk} & \rotatebox{\degree}{Build.} & \rotatebox{\degree}{Wall} & \rotatebox{\degree}{Fence} & \rotatebox{\degree}{Pole} & \rotatebox{\degree}{Tr.Light} & \rotatebox{\degree}{Sign} & \rotatebox{\degree}{Vege.} & \rotatebox{\degree}{Terrain} & \rotatebox{\degree}{Sky} & \rotatebox{\degree}{Person} & \rotatebox{\degree}{Rider} & \rotatebox{\degree}{Car} & \rotatebox{\degree}{Truck} & \rotatebox{\degree}{Bus} & \rotatebox{\degree}{Train} & \rotatebox{\degree}{M.bike} & \rotatebox{\degree}{Bike}\\
\toprule[0.2em]
\multicolumn{21}{c}{\textbf{Synthetic-to-Real: GTA$\to$Cityscapes (Val.)}} \\
\midrule
\cellcolor{resnet}AdaptSeg~\cite{AdaptSegNet} & \cellcolor{resnet}41.4 & 86.5 & 25.9 & 79.8 & 22.1 & 20.0 & 23.6 & 33.1 & 21.8 & 81.8 & 25.9 & 75.9 & 57.3 & 26.2 & 76.3 & 29.8 & 32.1 & 7.2 & 29.5 & 32.5\\
\cellcolor{resnet}ADVENT~\cite{Advent} & \cellcolor{resnet}45.5 & 89.4 & 33.1 & 81.0 & 26.6 & 26.8 & 27.2 & 33.5 & 24.7 & 83.9 & 36.7 & 78.8 & 58.7 & 30.5 & 84.8 & 38.5 & 44.5 & 1.7 & 31.6 & 32.4 \\
\cellcolor{resnet}DACS~\cite{DACS} & \cellcolor{resnet}52.1 & 89.9 & 39.7 & 87.9 & 30.7 & 39.5 & 38.5 & 46.4 & 52.8 & 88.0 & 44.0 & 88.8 & 67.2 & 35.8 & 84.5 & 45.7 & 50.2 & 0.0 & 27.3 & 34.0\\
\cellcolor{resnet}CorDA~\cite{CorDA} & \cellcolor{resnet}56.6 & 94.7 & 63.1 & 87.6 & 30.7 & 40.6 & 40.2 & 47.8 & 51.6 & 87.6 & 47.0 & 89.7 & 66.7 & 35.9 & 90.2 & 48.9 & 57.5 & 0.0 & 39.8 & 56.0 \\
\cellcolor{resnet}ProDA~\cite{ProDA} & \cellcolor{resnet}57.5 &  87.8 & 56.0 & 79.7 & 46.3 & 44.8 & 45.6 & 53.5 & 53.5 & 88.6 & 45.2 & 82.1 & 70.7 & 39.2 & 88.8 & 45.5 & 59.4 & 1.0 & 48.9 & 56.4 \\
\midrule
\cellcolor{resnet}DAFormer~\cite{DAFormer} & \cellcolor{resnet}54.2 & 85.7 & 66.8 & 81.5 & 27.3 & 20.4 & 46.4 & 53.2 & 63.0 & 84.5 & 32.1 & 72.9 & 71.9 & 45.0 & 90.5 & 60.7 & 58.8 & 0.1 & 23.2 & 46.4 \\
\cellcolor{resnet}\textit{\enspace + \method} & \cellcolor{resnet}58.3 & 95.2 & 69.1 & 88.1 & 26.0 & 27.7 & 48.8 & 55.2 & 63.6 & 89.6 & 49.5 & 90.3 & 72.0 & 45.4 & 91.4 & 63.3 & 61.1 & 0.0 & 23.8 & 46.7 \\

\midrule
\cellcolor{segformer}AdaptSeg$^\dagger$~\cite{AdaptSegNet} & \cellcolor{segformer}47.8 & 85.2 & 20.4 & 85.5 & 38.2 & 30.9 & 34.5 & 43.0 & 26.2 & 87.4 & 40.3 & 86.4 & 63.6 & 23.7 & 88.6 & 48.5 & 50.6 & 5.8 & 33.1 & 16.2 \\
\cellcolor{segformer}DACS$^\dagger$~\cite{DACS} & \cellcolor{segformer}58.2 & 88.9 & 50.0 & 88.4 & 46.4 & 43.9 & 43.1 & 53.4 & 54.8 & {89.9} & {51.2} & {92.8} & 64.2 & 9.4 & 91.4 & {77.3} & 63.3 & 0.0 & 47.4 & 49.8 \\
\midrule
\cellcolor{segformer}DAFormer~\cite{DAFormer} & \cellcolor{segformer}{68.3} & {95.7} & {70.2} & {89.4} & {53.5} & {48.1} & {49.6} & 55.8 & {59.4} & {89.9} & 47.9 & 92.5 & 72.2 & {44.7} & {92.3} & 74.5 & {78.2} & {65.1} & {55.9} & {61.8} \\
\cellcolor{segformer}\textit{\enspace + \method} &  \cellcolor{segformer} 70.1 &  96.0 &    71.8 &    90.3 &  53.3 &   46.4 &  54.8 &     57.8 &    66.7 &    90.0 &     49.2 & 92.2 &    73.6 &   46.3 & 92.8 &   78.1 & 80.6 &   70.7 &   57.5 &  63.2 \\
\midrule
\cellcolor{segformer}MIC$_\mathit{DAFormer}$~\cite{MIC} & \cellcolor{segformer}{70.6} & 96.7 & 75.0 & 90.0 & 58.2 & 50.4 & 51.1 & 56.7 & 62.1 & 90.2 & 51.3 & 92.9 & 72.4 & 47.1 & 92.8 & 78.9 & 83.4 & 75.6 & 54.2 & 62.6 \\
\cellcolor{segformer}\textit{\enspace + \method} & \cellcolor{segformer} 71.8 &  96.5 &   74.2 &   90.8 &  60.5 &  52.0 &  55.8 &    59.9 &   65.6 &   90.3 &    51.8 &  93.0 &   73.1 &  46.9 &  93.4 &  82.0 &  85.8 &  74.3 &   56.6 &  62.8    \\
\midrule
\cellcolor{segformer}HRDA~\cite{HRDA} & \cellcolor{segformer}{73.8} & {96.4} & {74.4} & {91.0} & \textit{61.6} & {51.5} & {57.1} & {63.9} & {69.3} & {91.3} & {48.4} & \textit{94.2} & {79.0} & {52.9} & {93.9} & {84.1} & {85.7} & {75.9} & {63.9} & {67.5} \\
\cellcolor{segformer} \textit{\enspace + \method} & \cellcolor{segformer} 74.8 &  95.8 &   71.1 &   91.5 & \textbf{62.8} &  55.0 & \textit{60.8} & \textit{64.0} & \textit{73.4} &   91.3 & 49.1 &  94.0 &   79.2 &  54.6 &  94.4 &  84.8 &  88.5 &  79.0 &   \textbf{65.9} &  65.5  \\
\midrule
\cellcolor{segformer}MIC$_\mathit{HRDA}$~\cite{MIC} & \cellcolor{segformer}\textit{75.9} & \textit{97.4} & \textit{80.1} & \textit{91.7} & 61.2 & \textit{56.9} & 59.7 & \textbf{66.0} & 71.3 & \textbf{91.7} & \textit{51.4} & \textbf{94.3} & \textit{79.8} & \textit{56.1} & \textbf{95.6} & \textit{85.4} & \textit{90.3} & \textit{80.4} & 64.5 & \textbf{68.5} \\
\cellcolor{segformer}\textit{\enspace + \method} & \cellcolor{segformer}\textbf{76.6} &  \textbf{97.6} & \textbf{81.5} & \textbf{92.0} & \textbf{62.8} & \textbf{59.4} & \textbf{62.6} & 62.9 & \textbf{73.6} & \textit{91.6} & \textbf{52.6} & 94.1 & \textbf{80.2} & \textbf{57.0} & \textit{94.8} & \textbf{87.4} & \textbf{90.7} & \textbf{81.6} & \textit{65.3} & \textit{67.8} \\
\midrule
\multicolumn{21}{c}{\textbf{Synthetic-to-Real: Synthia$\to$Cityscapes (Val.)}} \\
\midrule

\cellcolor{resnet}ADVENT~\cite{Advent} & \cellcolor{resnet}41.2 & 85.6 & 42.2 & 79.7 & 8.7 & 0.4 & 25.9 & 5.4 & 8.1 & 80.4 & -- & 84.1 & 57.9 & 23.8 & 73.3 & -- & 36.4 & -- & 14.2 & 33.0 \\
\cellcolor{resnet}DACS~\cite{DACS} & \cellcolor{resnet}48.3 &  80.6 & 25.1 & 81.9 & 21.5 & 2.9 & 37.2 & 22.7 & 24.0 & 83.7 & -- & {90.8} & 67.6 & 38.3 & 82.9 & -- & 38.9 & -- & 28.5 & 47.6 \\
\cellcolor{resnet}CorDA~\cite{CorDA} & \cellcolor{resnet}55.0 & \textbf{93.3} & \textbf{61.6} & 85.3 & 19.6 & 5.1 & 37.8 & 36.6 & 42.8 & 84.9 & -- & 90.4 & 69.7 & 41.8 & 85.6 & -- & 38.4 & -- & 32.6 & 53.9 \\
\cellcolor{resnet}ProDA~\cite{ProDA} & \cellcolor{resnet}55.5 & \textit{87.8} & 45.7 & 84.6 & 37.1 & 0.6 & 44.0 & 54.6 & 37.0 & \textit{88.1} & -- & 84.4 & {74.2} & 24.3 & 88.2 & -- & 51.1 & -- & 40.5 & 45.6 \\
\midrule
\cellcolor{segformer}DACS$^\dagger$~\cite{DACS} &\cellcolor{segformer}52.2 & 58.0 & 46.0 & 84.8 & 37.7 & {5.2} & 38.6 & 20.9 & 47.3 & 85.9 & -- & 81.6 & 73.0 & 43.9 & 86.9 & -- & 55.6 & -- & 51.1 & 18.6 \\
\midrule
\cellcolor{segformer}DAFormer~\cite{DAFormer} & \cellcolor{segformer}{61.3} & 82.2 &   37.2 &   88.6 &  42.9 &   8.5 &  50.1 &    55.1 &   54.5 &   85.7 & -- &  88.0 &   73.6 &  48.6 &  87.6 &  -- &  62.8 &   -- &   53.1 &  62.4 \\
\cellcolor{segformer}\textit{\enspace + \method} & \cellcolor{segformer} 62.4 &  81.0 &   37.1 &   89.4 &  45.7 &  \textit{9.5} &  51.8 &    57.3 &   58.0 &   86.7 &      -- &  85.0 &   73.6 &  50.4 &  88.2 &    -- &  64.7 &    -- &   56.8 &  62.8  \\
\midrule
\cellcolor{segformer}HRDA~\cite{HRDA} & \cellcolor{segformer}{65.8} & 85.2 & 47.7 & {88.8} & {49.5} & 4.8 & {57.2} & {65.7} & {60.9} & 85.3 & -- & {92.9} & {79.4} & {52.8} & {89.0} & -- & {64.7} & -- & {63.9} & {64.9} \\
\cellcolor{segformer}\textit{\enspace + \method} & \cellcolor{segformer}{66.8} &  86.3 &   49.6 & \textit{89.3} & \textbf{53.7} &   5.1 &  57.6 & \textit{66.4} & \textit{63.8} &   86.1 &      -- &  94.1 &   79.1 &  56.0 &  87.8 &    -- & \textit{65.0} &    -- &   64.2 &  \textit{65.0} \\
\midrule
\cellcolor{segformer}MIC$_\mathit{HRDA}$~\cite{MIC} & \cellcolor{segformer}\textit{67.3} & 86.6 & \textit{50.5} & \textit{89.3} & 47.9 & 7.8 & \textit{59.4} & \textbf{66.7} & 63.4 & 87.1 & -- & \textit{94.6} & \textit{81.0} & \textit{58.9} & \textit{90.1} & -- & 61.9 & -- & \textit{67.1} & 64.3 \\
\cellcolor{segformer}\textit{\enspace + \method} & \cellcolor{segformer}\textbf{67.9} &  82.8 &   42.6 & \textbf{90.5} & \textit{51.6} & \textbf{9.6} & \textbf{61.0} &    65.7 &   \textbf{65.0} & \textbf{89.1} &      -- &  \textbf{95.0} & \textbf{81.1} &  \textbf{59.7} &  \textbf{90.6} &    -- &  \textbf{68.3} &    -- & \textbf{67.4} & \textbf{66.5} \\
\bottomrule[0.1em]
\end{tabular}
}
\caption{\textbf{Comparison of \method\ with state-of-the-art UDA methods.} The performance is reported as IoU in \%. We group methods based on \colorbox{resnet}{ResNet}~\cite{he2016deep} and \colorbox{segformer}{Segformer}~\cite{SegFormer} backbones. $^\dagger$ denotes results obtained with a Segformer backbone from~\cite{hoyer2023domain}.
On both GTA and SYNTHIA, \method\ achieves consistent improvements, demonstrating the effectiveness of our masking strategy and fusion module.
}
\label{tab:main_results}
\end{table}

To validate the effectiveness of \method\ and its capabilities as a \emph{plugin}, we evaluate its performance across the three state-of-the-art methods DAFormer~\cite{DAFormer}, HRDA~\cite{HRDA}, and MIC~\cite{MIC}. The results are shown in \tabref{main_results}.

Starting with applying \method\ to DAFormer~\cite{DAFormer} on GTA, the results improve by 1.8 mIoU.
Using the recent MIC pretrained model, we obtain improvements by 1.2 mIoU.
Remarkably, our method still improves over the strong HRDA method by 1.0 mIoU.
When we build on top of the currently best performing model MIC$_\mathit{HRDA}$, we can further boost results by 0.7 mIoU, setting a new state of the art in UDA semantic segmentation. Considering that the improvement is on top of the best-performing SOTA approach on a saturating benchmark (94\% of the oracle performance), this gain can be considered significant.
By plugging our light-weight modules into each of these architectures and adding complementary dropout, we achieve consistent improvements, clearly showing that leveraging depth helps in closing the domain gap. 
Furthermore, the comparison to MIC supports our hypothesis that our contributions are orthogonal to the successes of input masking (MIM) in UDA.

Diving into the details of these improvements, we notice predominantly gains in two types of objects.
First, we see consistent improvements in classes of thin structures such as poles, signs, or motorbikes.
This is enabled by our design of aggregating local depth features without using any pooling, as these local depth continuities at boundary regions serve as a strong cue.
Second, larger classes of lower prevalence in the dataset, such as truck, bus, or train show generally improved performance when adding \method.
In these cases, both global as well as local depth features can help.
Due to their size, global reasoning can improve the consistency of their segmentation, but also the locally smooth, continuous depth lower the likelihood of changes in the semantics within a local window.

We also benchmark MICDrop with a ResNet-101 architecture in the DAFormer framework in Tab.~\ref{tab:main_results}. It shows a significant gain of 4.1 mIoU over the baseline and outperforms the previous SOTA depth-guided UDA method CorDA~\cite{CorDA}.

\begin{table*}[t!]
\centering
\setlength{\tabcolsep}{1.5pt}
\resizebox{\textwidth}{!}{
\footnotesize
\begin{tabular}{l|c|ccccccccccccccccccc}
\toprule[0.2em]
Method & \rotatebox{\degree}{mIoU} & \rotatebox{\degree}{Road} & \rotatebox{\degree}{S.walk} & \rotatebox{\degree}{Build.} & \rotatebox{\degree}{Wall} & \rotatebox{\degree}{Fence} & \rotatebox{\degree}{Pole} & \rotatebox{\degree}{Tr.Light} & \rotatebox{\degree}{Sign} & \rotatebox{\degree}{Vege.} & \rotatebox{\degree}{Terrain} & \rotatebox{\degree}{Sky} & \rotatebox{\degree}{Person} & \rotatebox{\degree}{Rider} & \rotatebox{\degree}{Car} & \rotatebox{\degree}{Truck} & \rotatebox{\degree}{Bus} & \rotatebox{\degree}{Train} & \rotatebox{\degree}{M.bike} & \rotatebox{\degree}{Bike}\\
\toprule[0.2em]
\cellcolor{segformer}MIC$_{HRDA}$~\cite{MIC} & \cellcolor{segformer}{52.0} & 41.9 & 59.1 & 54.0 & 36.6 & 31.7 & 58.1 & 53.3 & 56.1 & 64.9 & 34.6 & 66.6 & 63.3 & 44.3 & 72.5 & 49.2 & 61.0 & 49.4 & 41.2 & 50.3  \\
\cellcolor{segformer}\textit{\enspace + \method} & \cellcolor{segformer} 53.6 & 41.1 & 60.2 & 58.2 & 36.9 & 33.8 & 61.0 & 51.9 & 59.9 & 65.3 & 35.0 & 66.3 & 65.6 & 46.6 & 73.6 & 54.1 & 64.1 & 49.3 & 43.6 & 52.3 \\
\cellcolor{segformer}\textit{\enspace $\Delta$} & \cellcolor{segformer} +1.6 & -0.8 & +1.1 & \textbf{+4.2} & +0.3 & +2.1 & \textbf{+2.9} & -1.4 & \bf{+3.8} & +0.4 & +0.4 & -0.3 & +2.3 & +2.3 & +1.1 & \bf{+4.9} & \bf{+3.1} & -0.1 & +2.4 & +2.0 \\
\bottomrule
\end{tabular}
}
\caption{\textbf{Boundary IoU} on GTA$\to$Cityscapes with a dilation factor of 0.005.}
\label{tab:boundary_iou}
\end{table*}
\begin{figure}[t]
\centering
\makebox[\linewidth][c]{    \begin{subfigure}{\linewidth}
        \centering
        \label{fig:palette}
        \includegraphics[width=\linewidth]{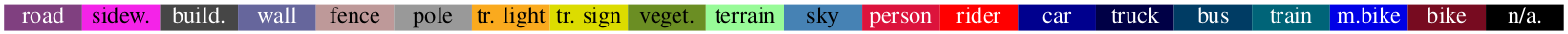}
    \end{subfigure}}
\makebox[\linewidth][c]{    \begin{subfigure}{.2\textwidth}
        \centering
        \includegraphics[width=.98\linewidth]{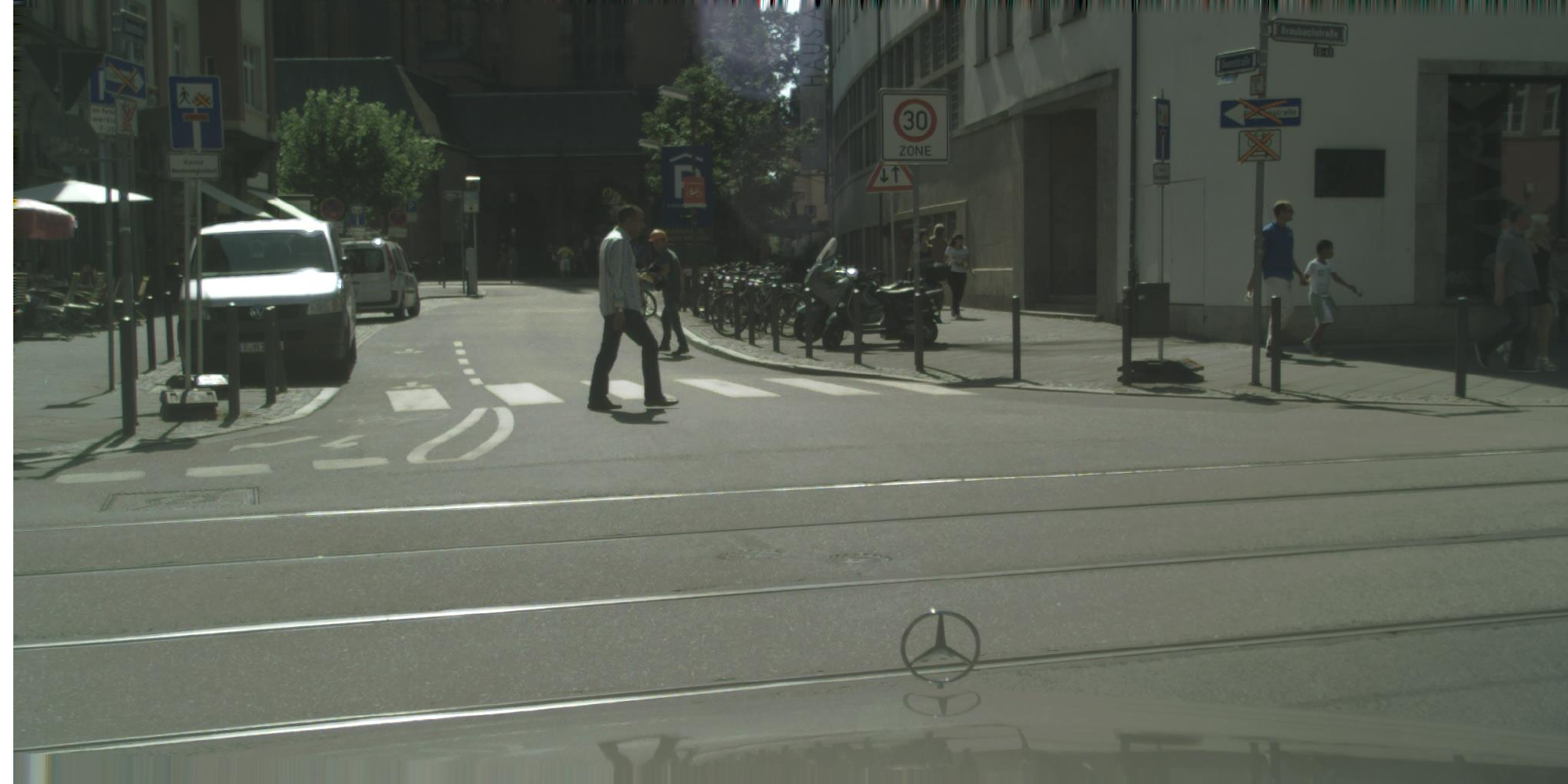}
    \end{subfigure}    \begin{subfigure}{.2\textwidth}
        \centering
        \includegraphics[width=.98\linewidth]{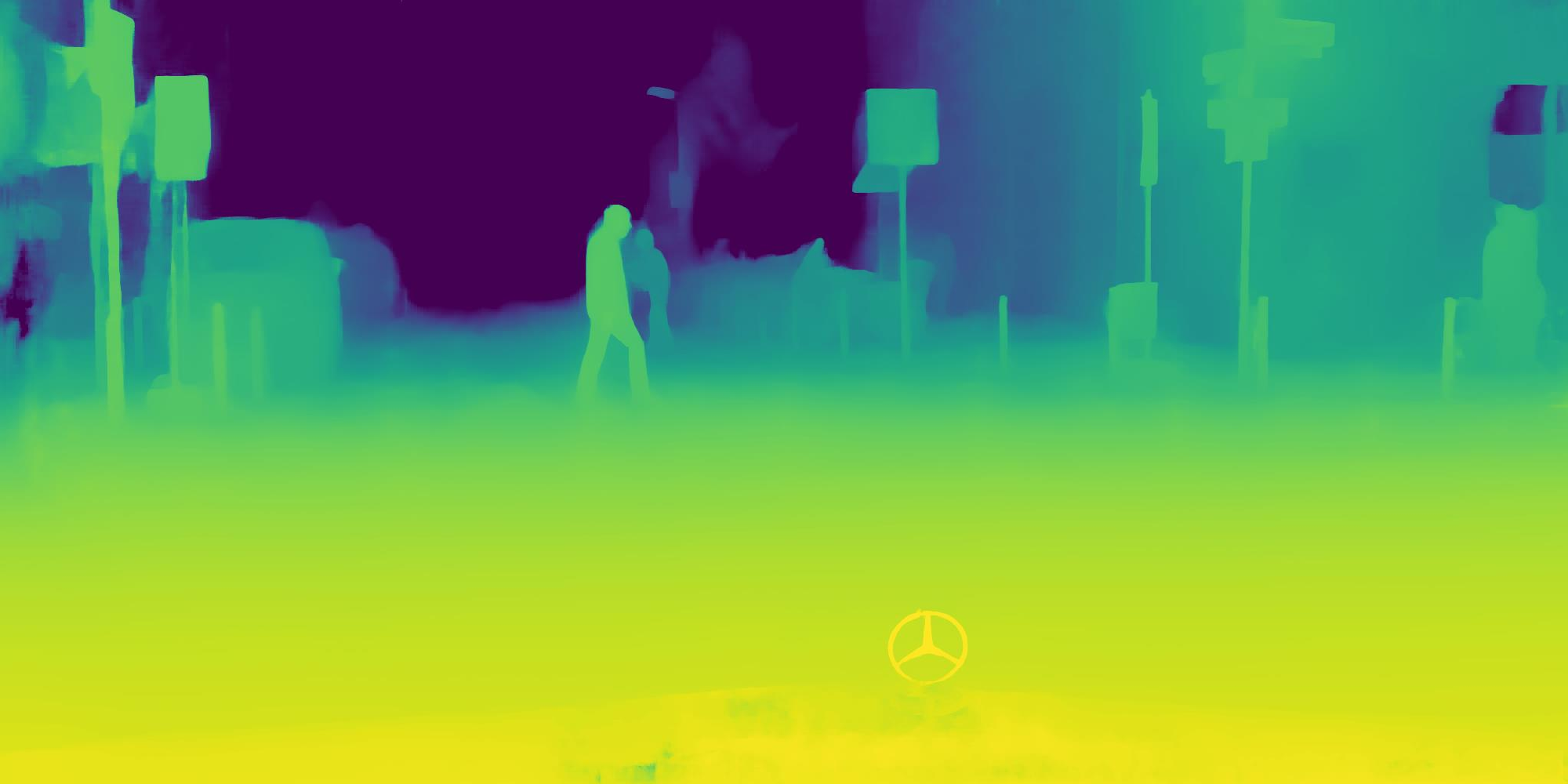}
    \end{subfigure}    \begin{subfigure}{.2\textwidth}
        \centering
        \includegraphics[width=.98\linewidth]{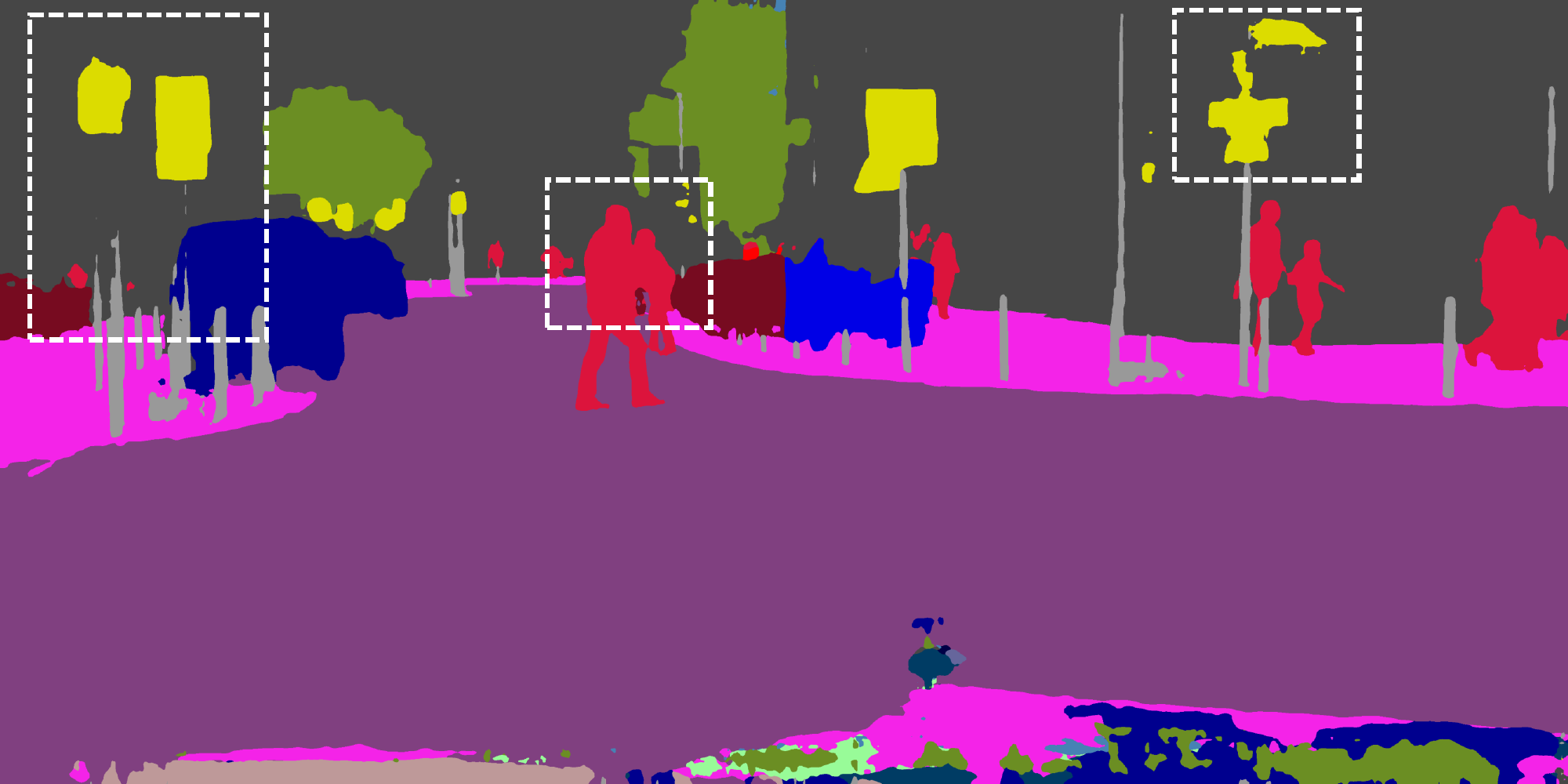}
    \end{subfigure}    \begin{subfigure}{.2\textwidth}
        \centering
        \includegraphics[width=.98\linewidth]{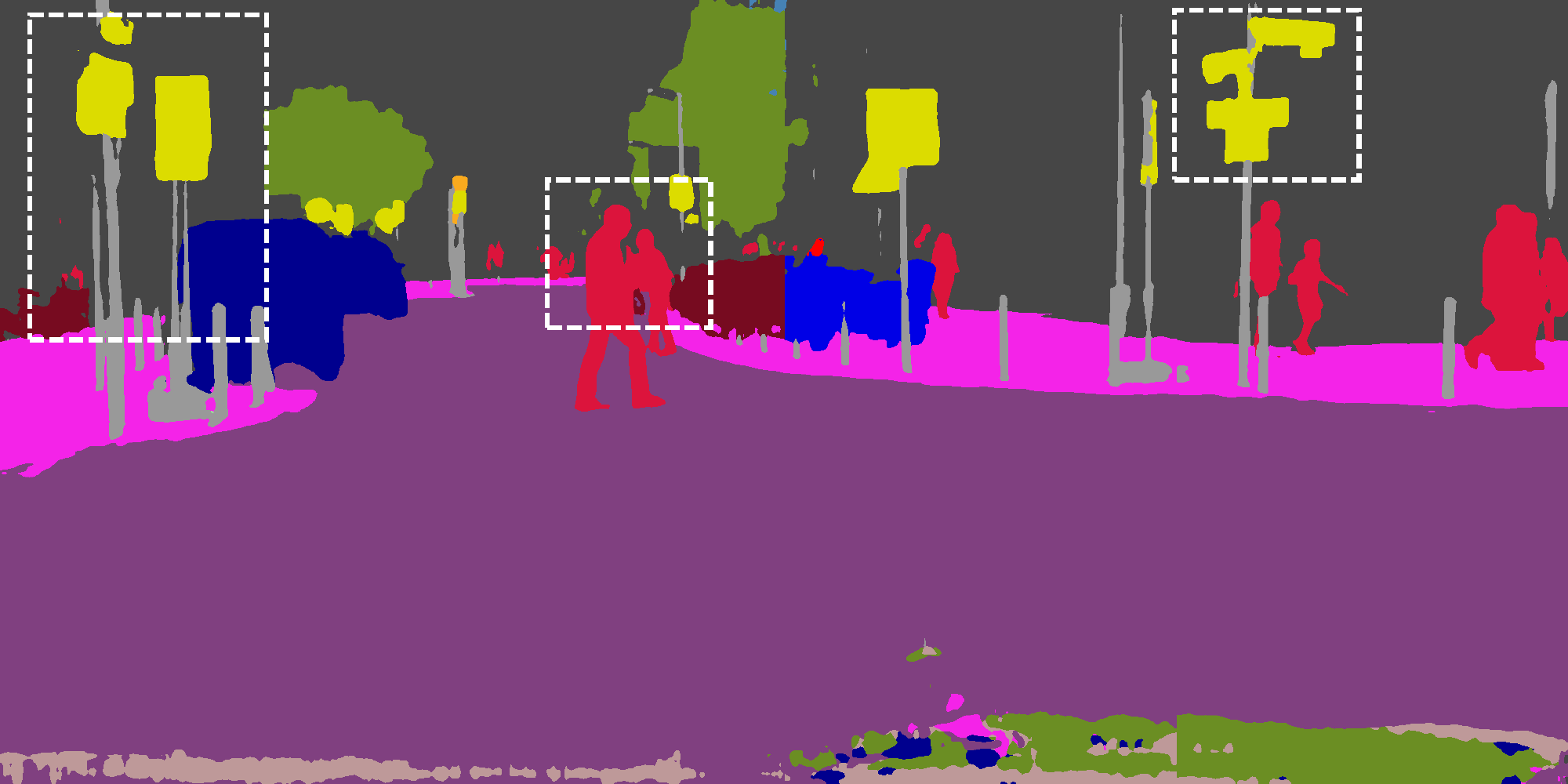}
    \end{subfigure}    \begin{subfigure}{.2\textwidth}
        \centering
        \includegraphics[width=.98\linewidth]{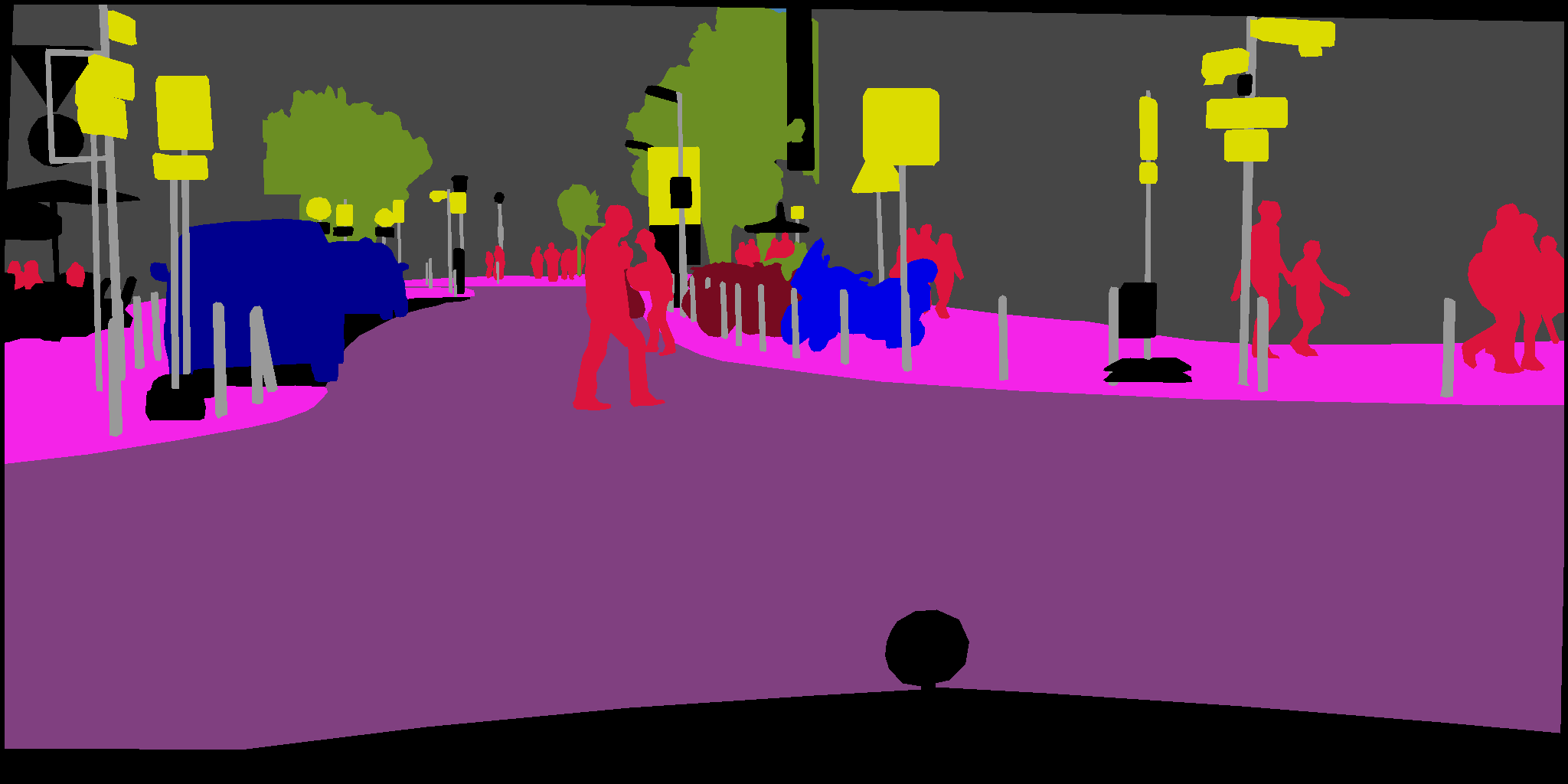}
    \end{subfigure}}
\makebox[\linewidth][c]{    \begin{subfigure}{.2\textwidth}
        \centering
        \includegraphics[width=.98\linewidth]{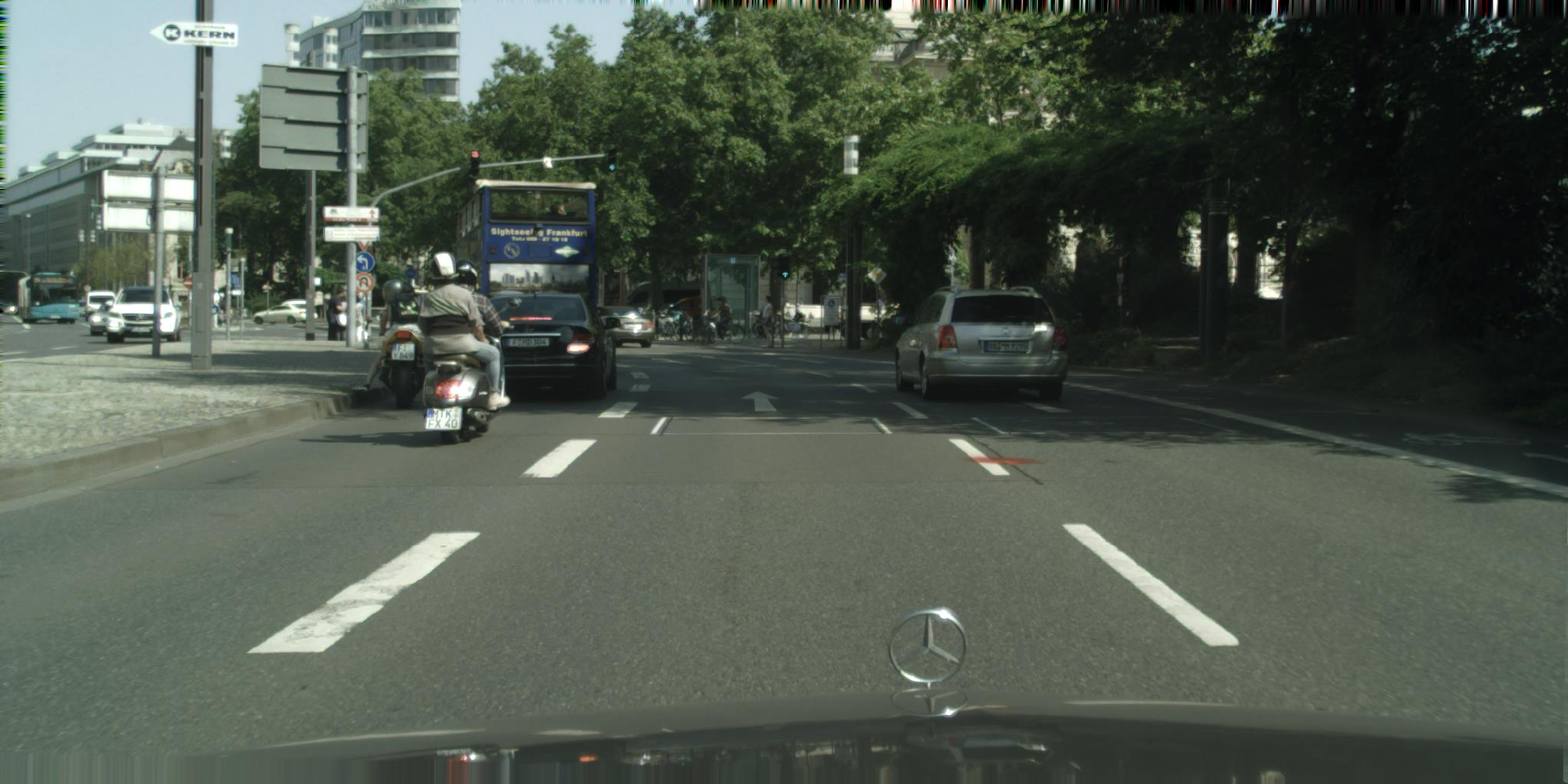}
    \end{subfigure}    \begin{subfigure}{.2\textwidth}
        \centering
        \includegraphics[width=.98\linewidth]{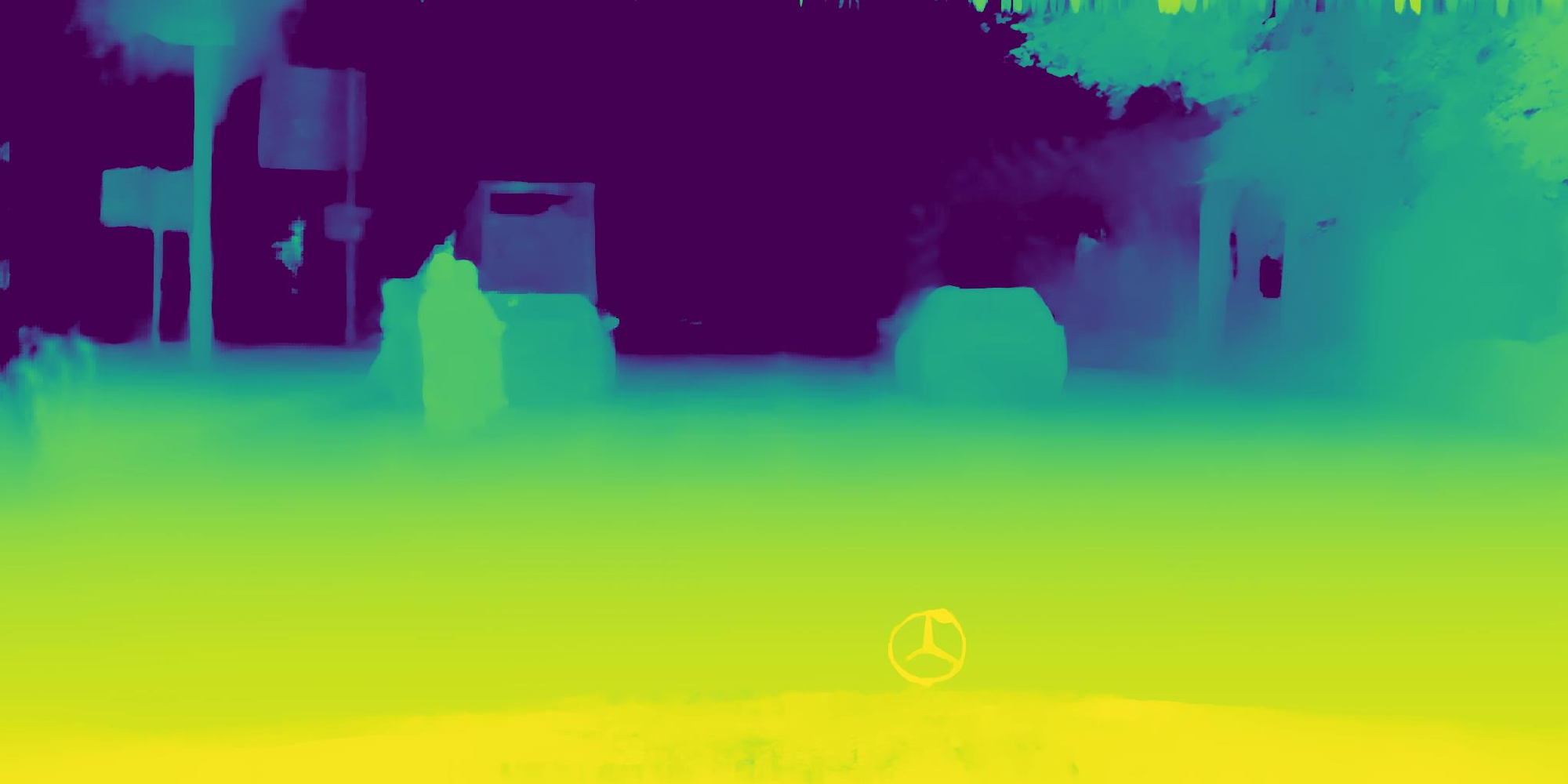}
    \end{subfigure}    \begin{subfigure}{.2\textwidth}
        \centering
        \includegraphics[width=.98\linewidth]{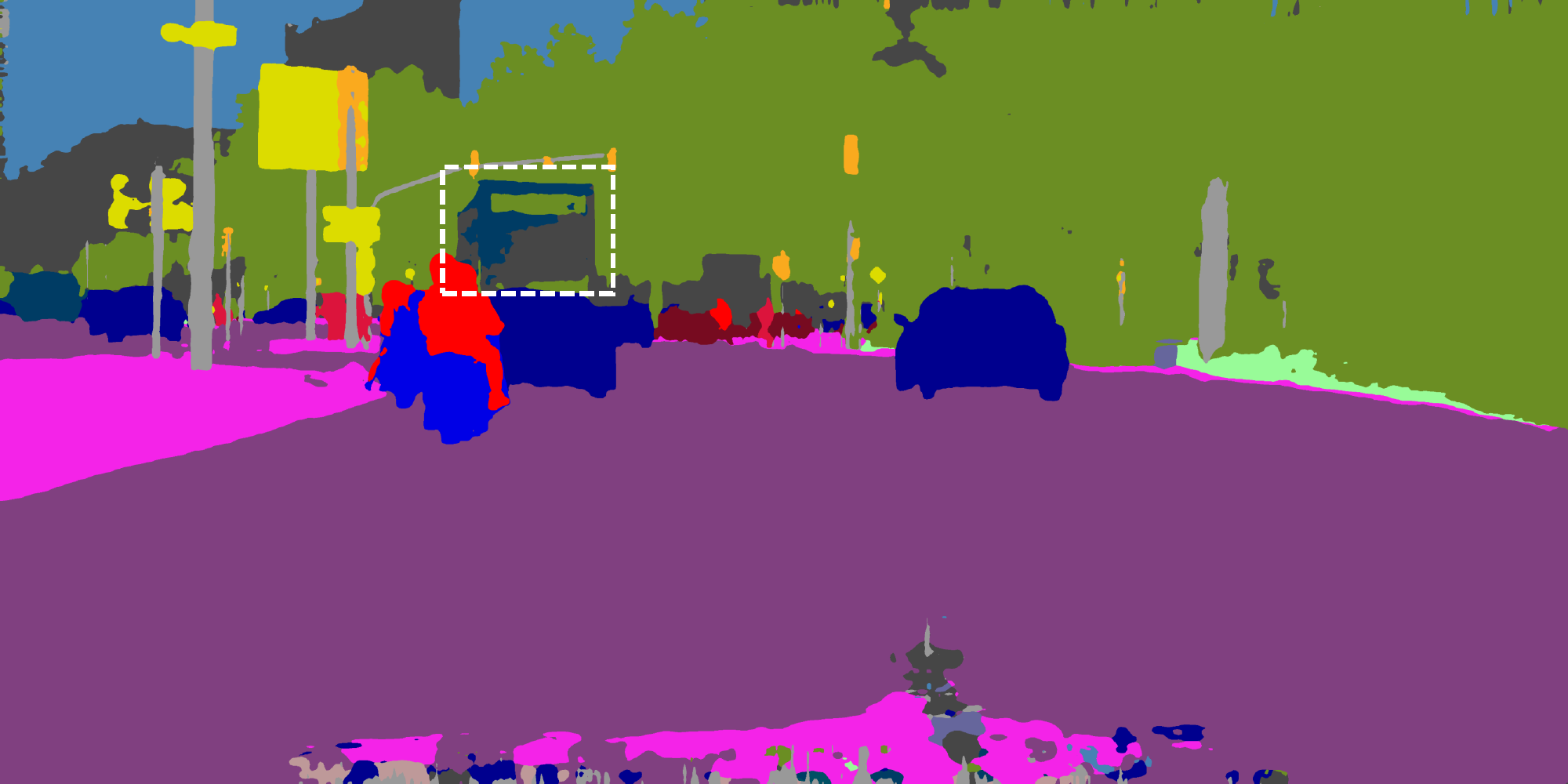}
    \end{subfigure}    \begin{subfigure}{.2\textwidth}
        \centering
        \includegraphics[width=.98\linewidth]{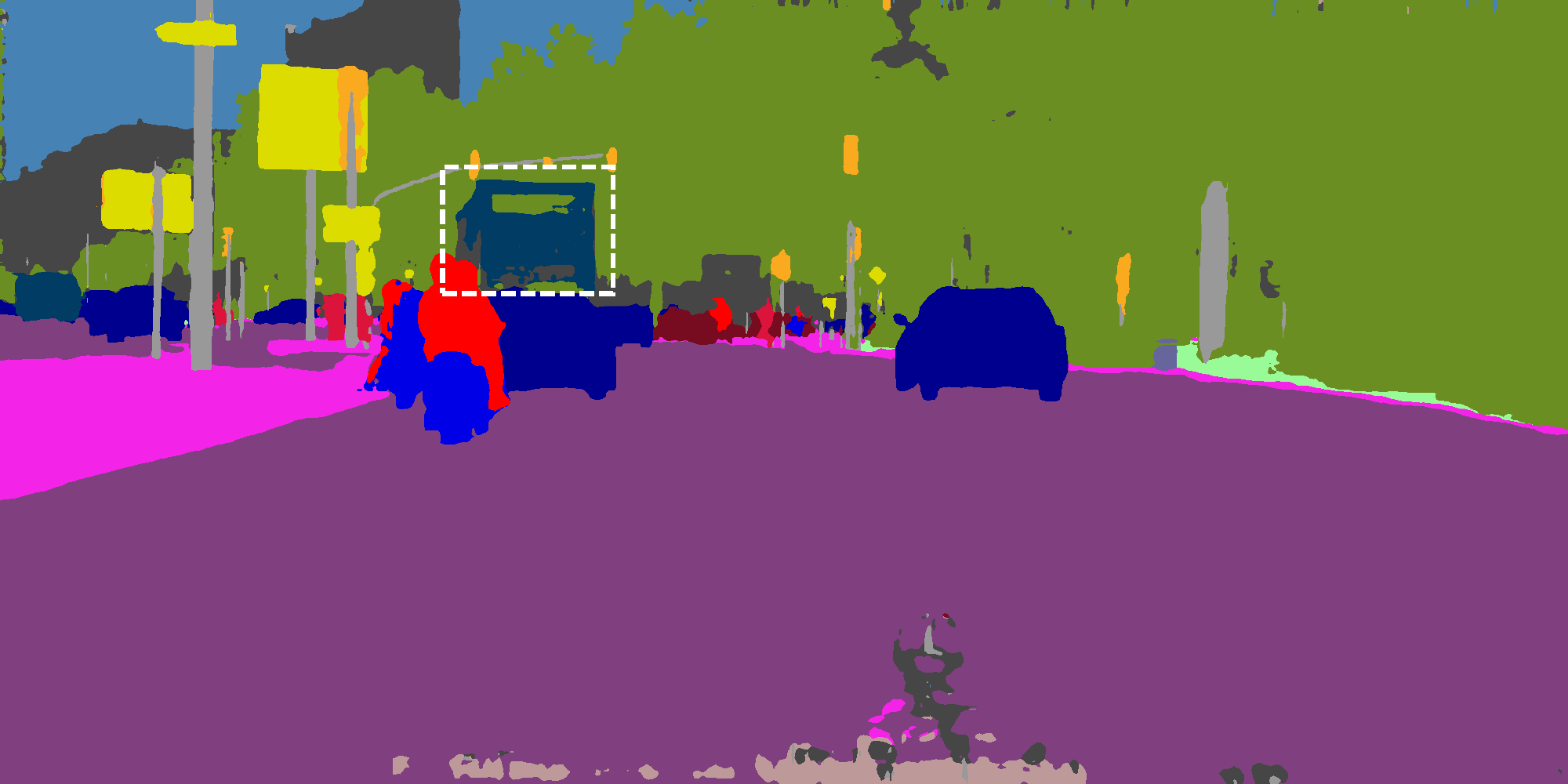}
    \end{subfigure}    \begin{subfigure}{.2\textwidth}
        \centering
        \includegraphics[width=.98\linewidth]{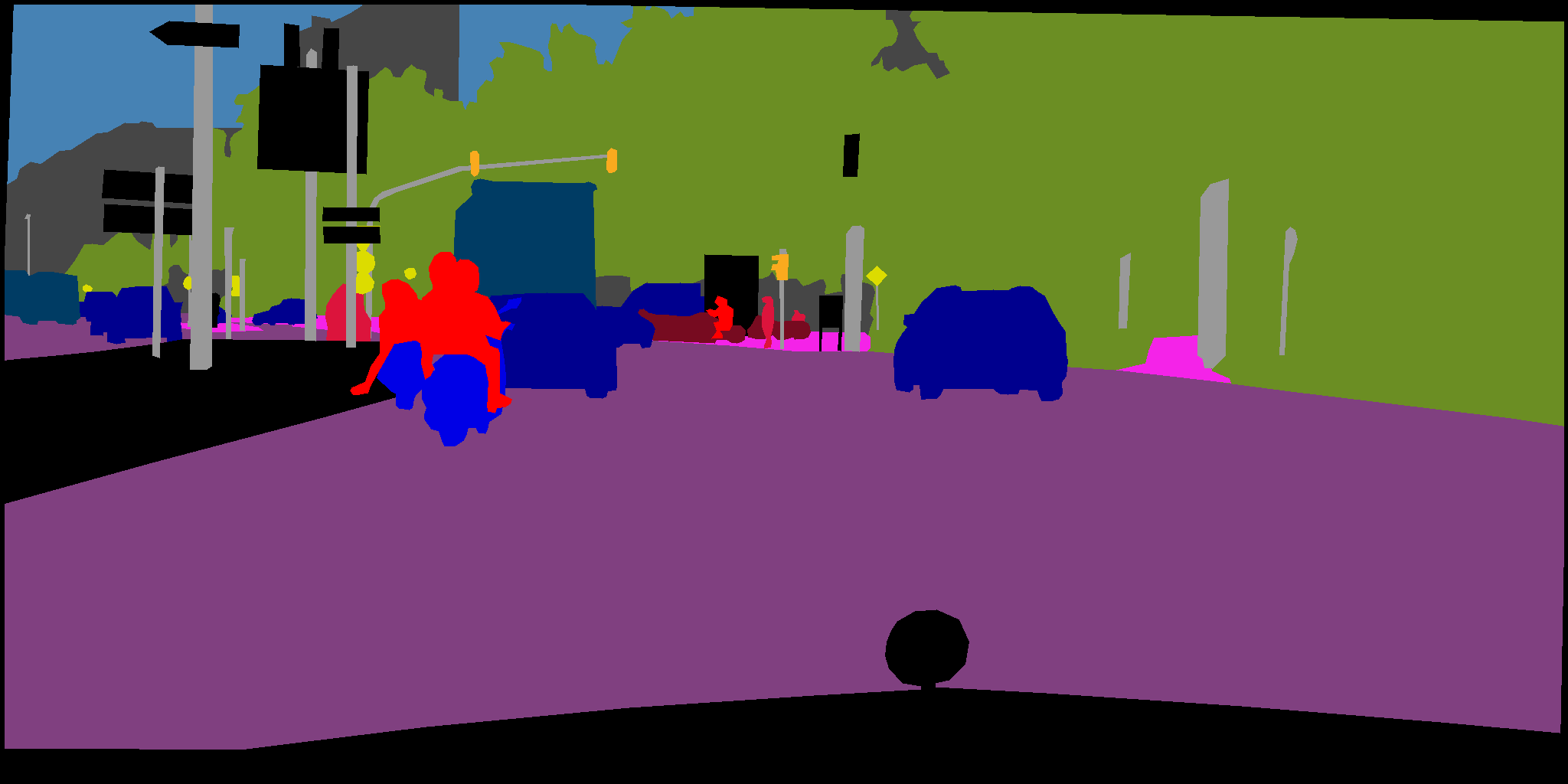}
    \end{subfigure}}
\makebox[\linewidth][c]{    \begin{subfigure}{.2\textwidth}
        \centering
        \includegraphics[width=.98\linewidth]{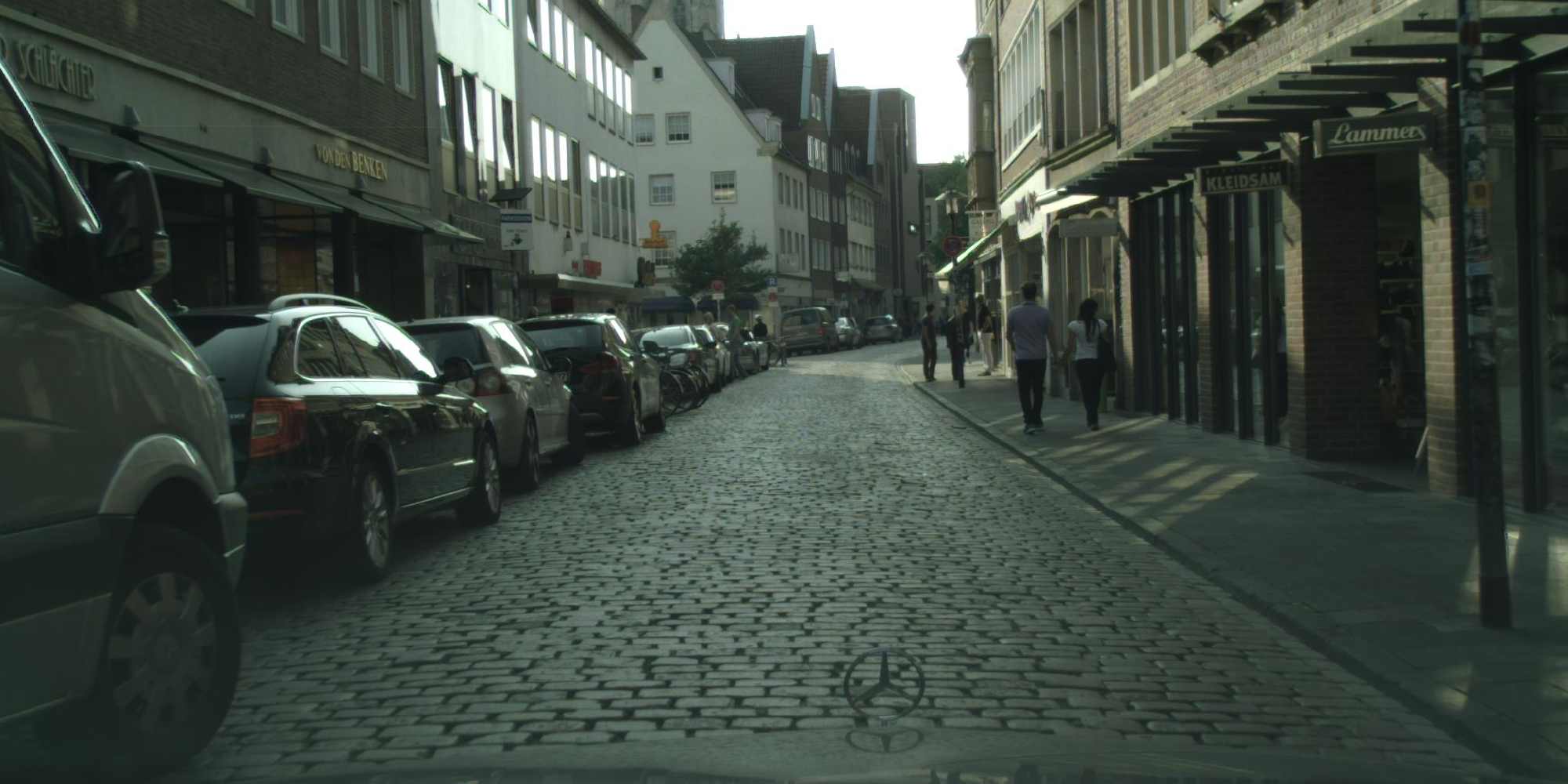}
    \end{subfigure}    \begin{subfigure}{.2\textwidth}
        \centering
        \includegraphics[width=.98\linewidth]{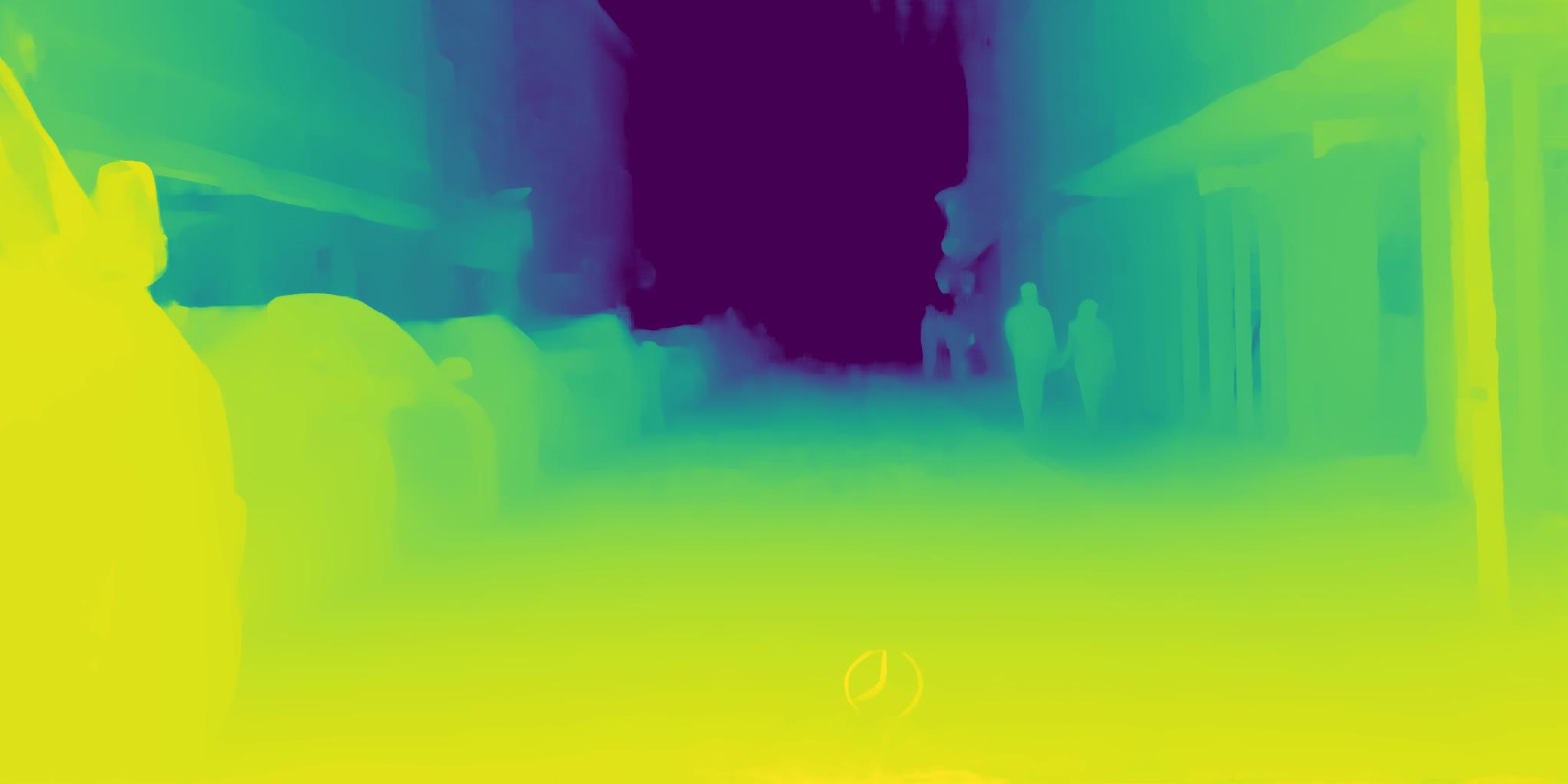}
    \end{subfigure}    \begin{subfigure}{.2\textwidth}
        \centering
        \includegraphics[width=.98\linewidth]{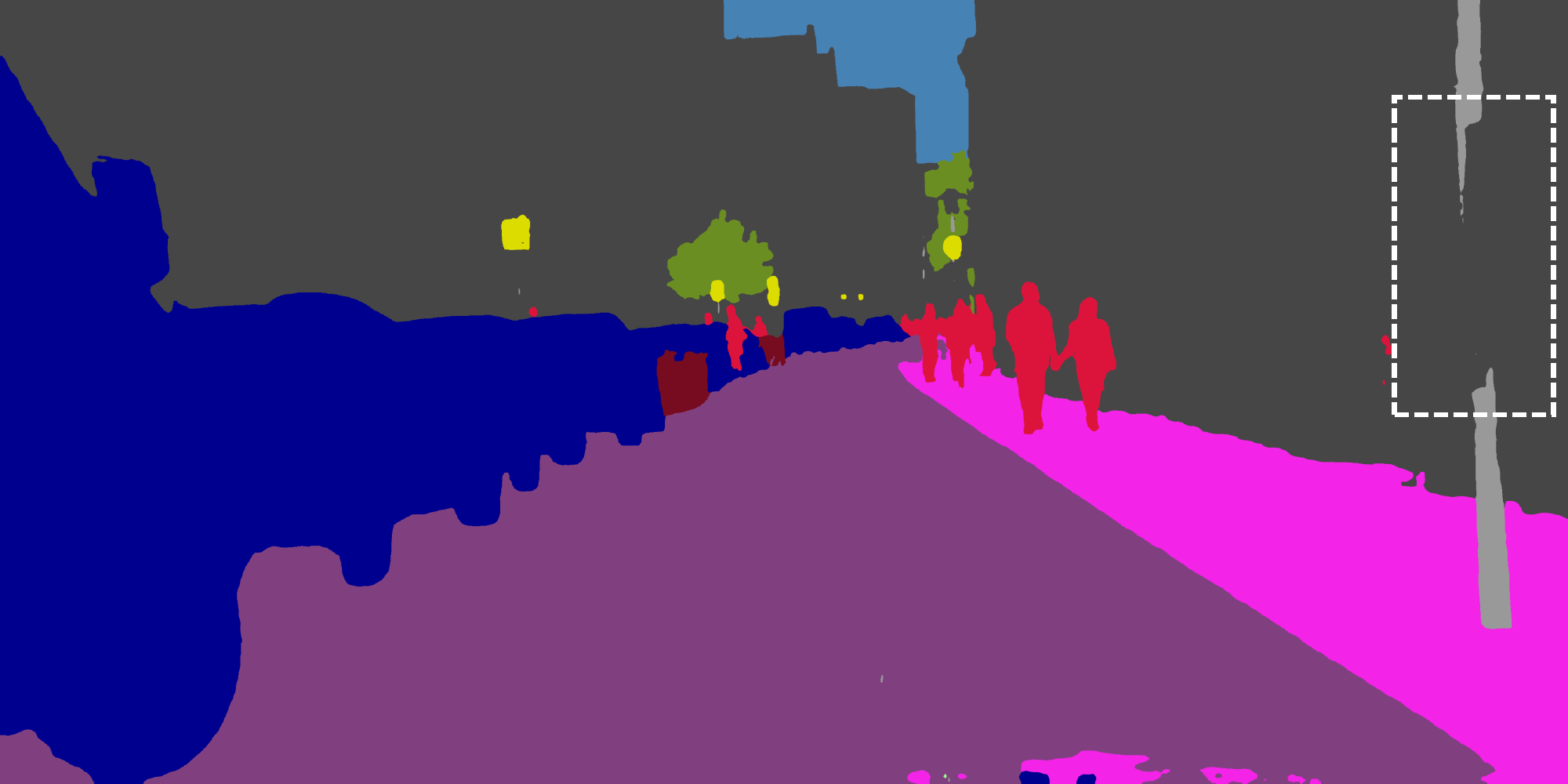}
    \end{subfigure}    \begin{subfigure}{.2\textwidth}
        \centering
        \includegraphics[width=.98\linewidth]{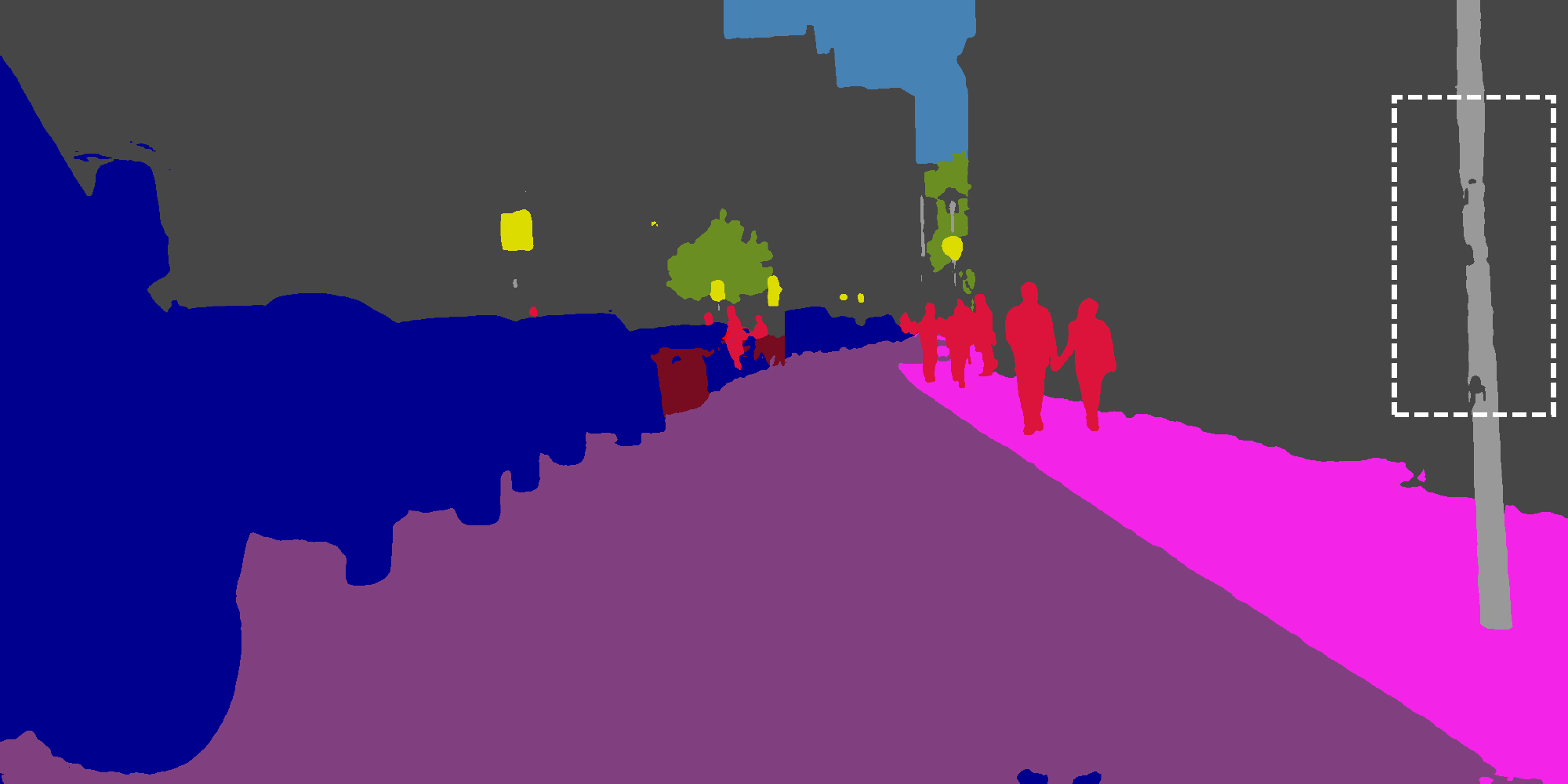}
    \end{subfigure}    \begin{subfigure}{.2\textwidth}
        \centering
        \includegraphics[width=.98\linewidth]{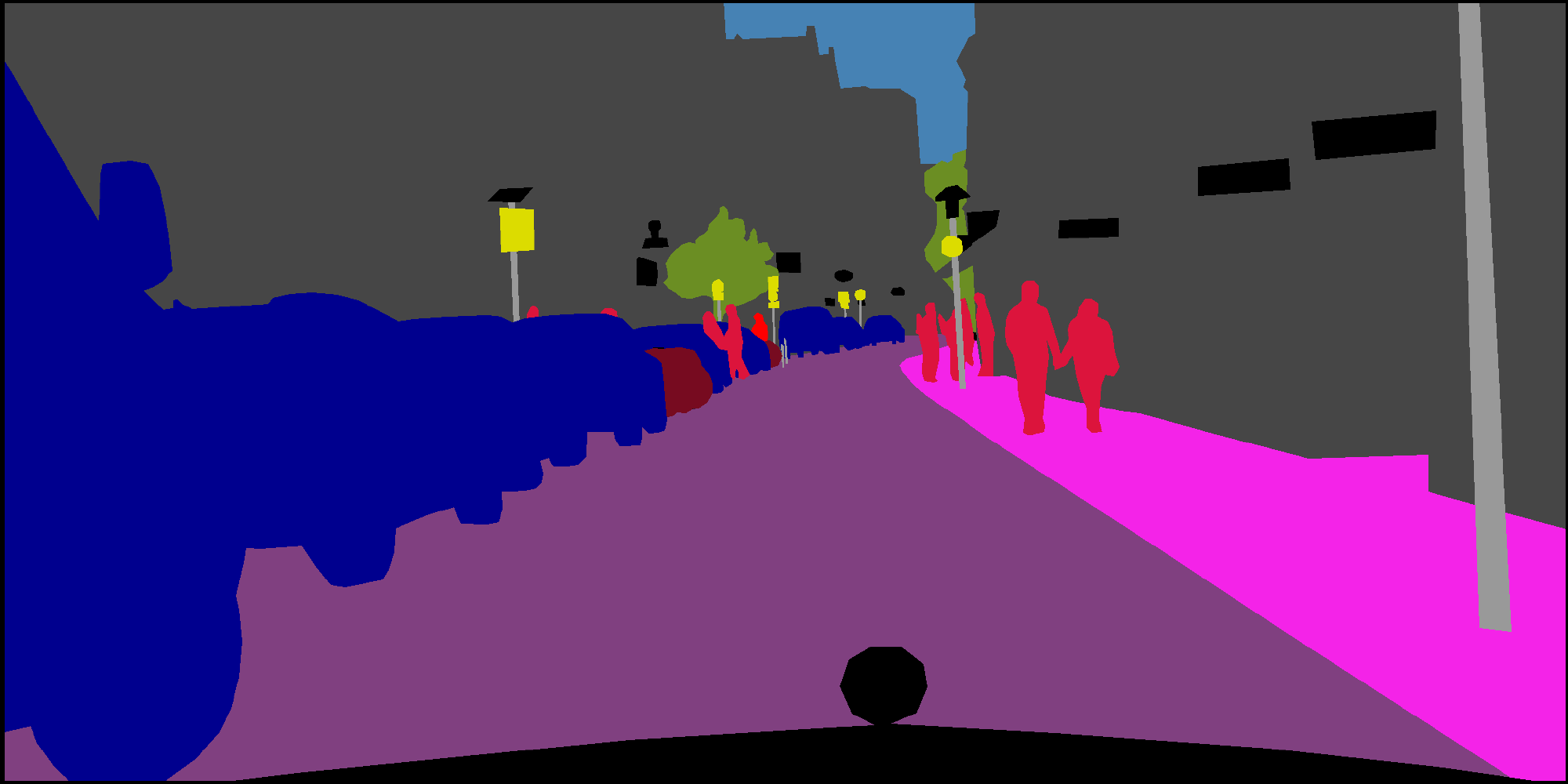}
    \end{subfigure}}
\makebox[\linewidth][c]{    \begin{subfigure}{.2\textwidth}
        \centering
        \includegraphics[width=.98\linewidth]{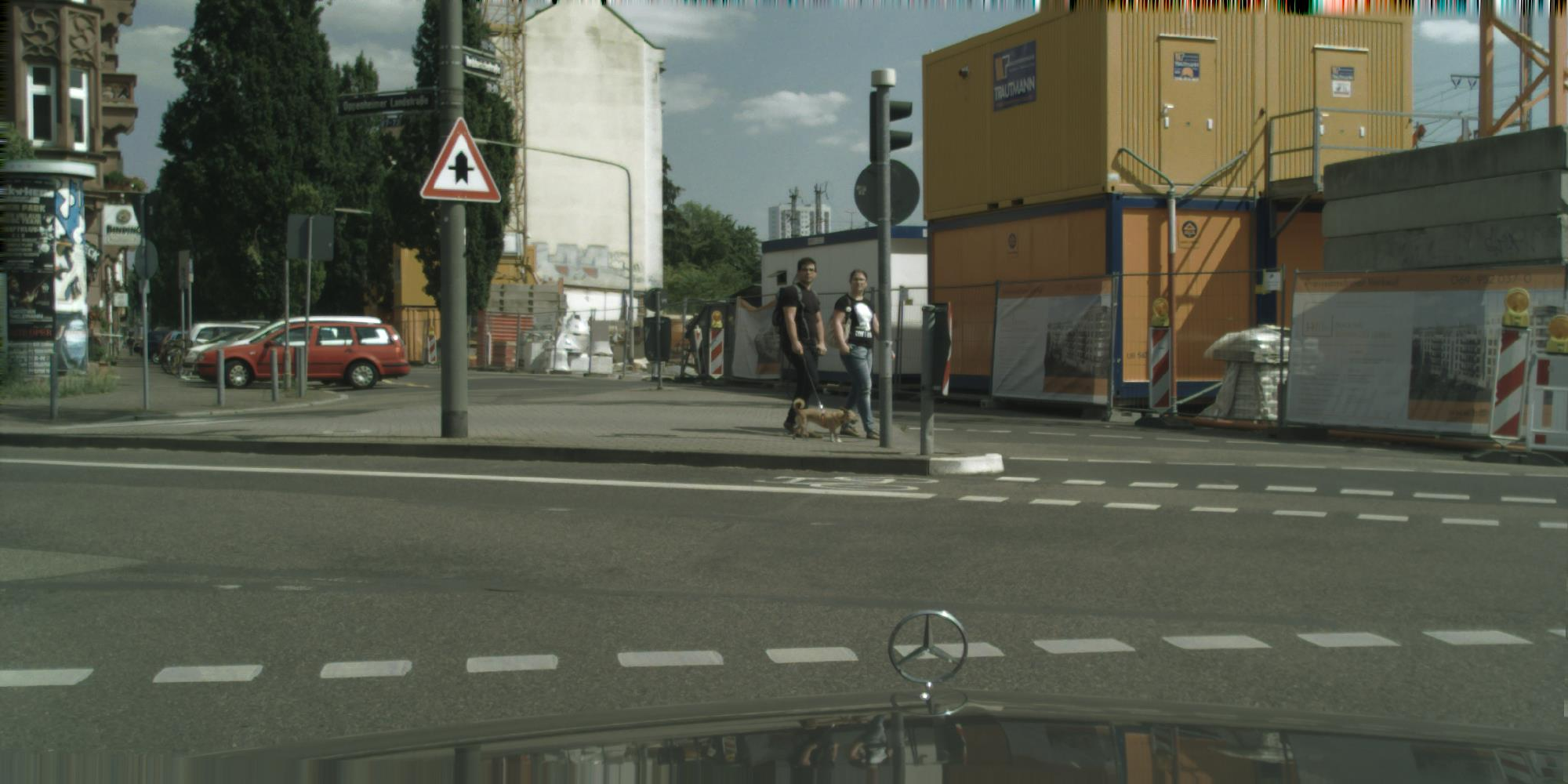}
    \end{subfigure}    \begin{subfigure}{.2\textwidth}
        \centering
        \includegraphics[width=.98\linewidth]{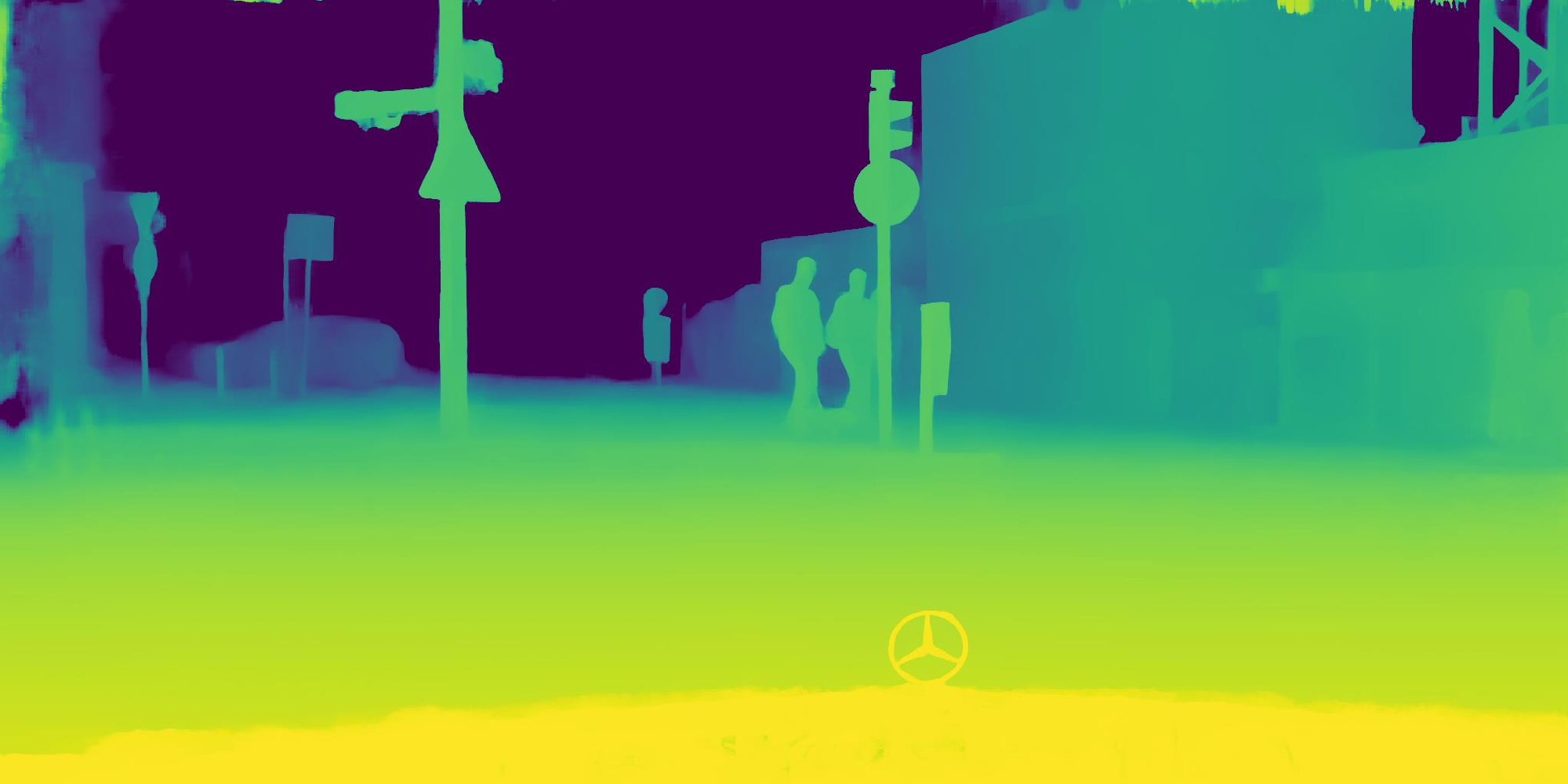}
    \end{subfigure}    \begin{subfigure}{.2\textwidth}
        \centering
        \includegraphics[width=.98\linewidth]{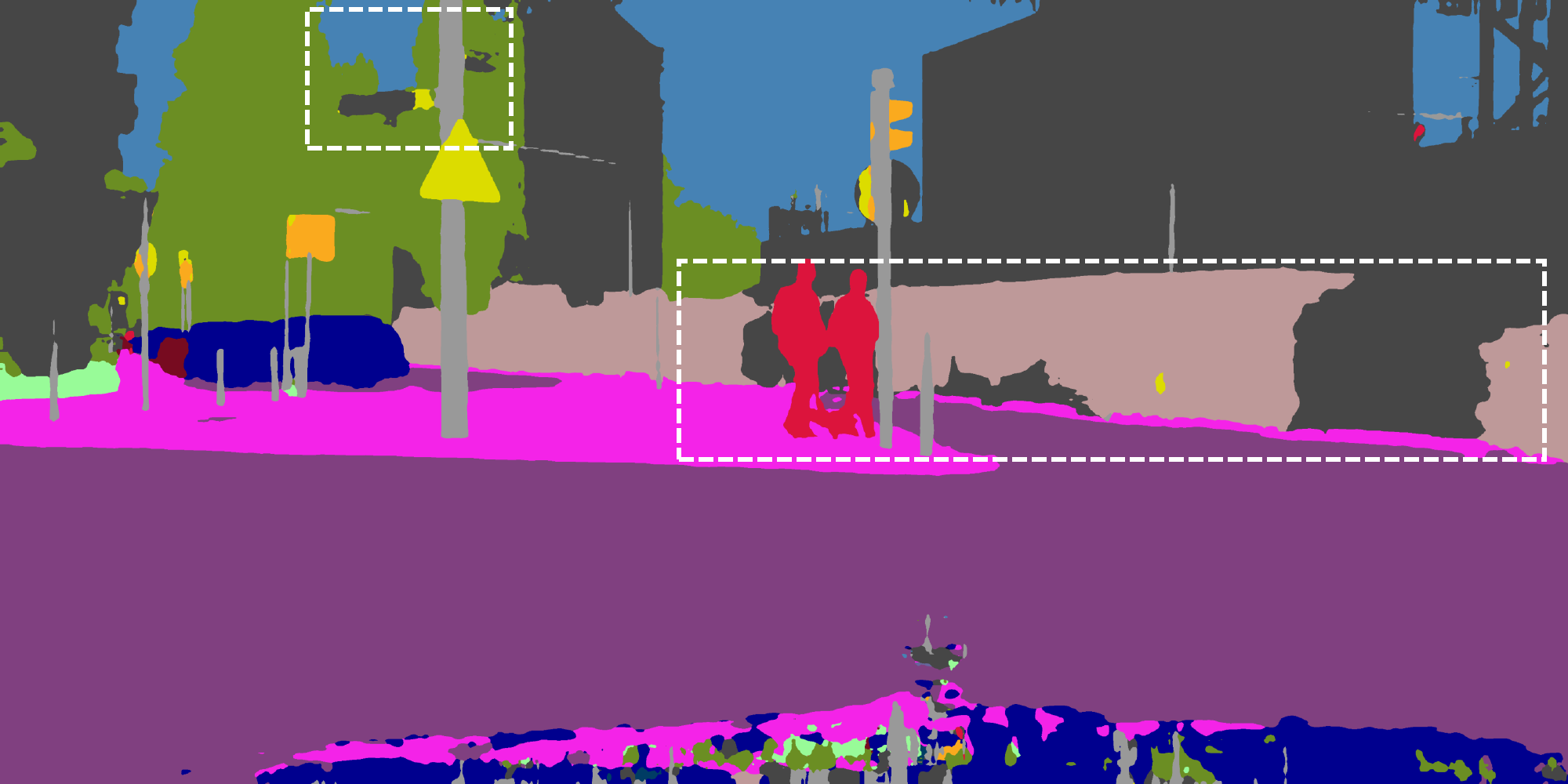}
    \end{subfigure}    \begin{subfigure}{.2\textwidth}
        \centering
        \includegraphics[width=.98\linewidth]{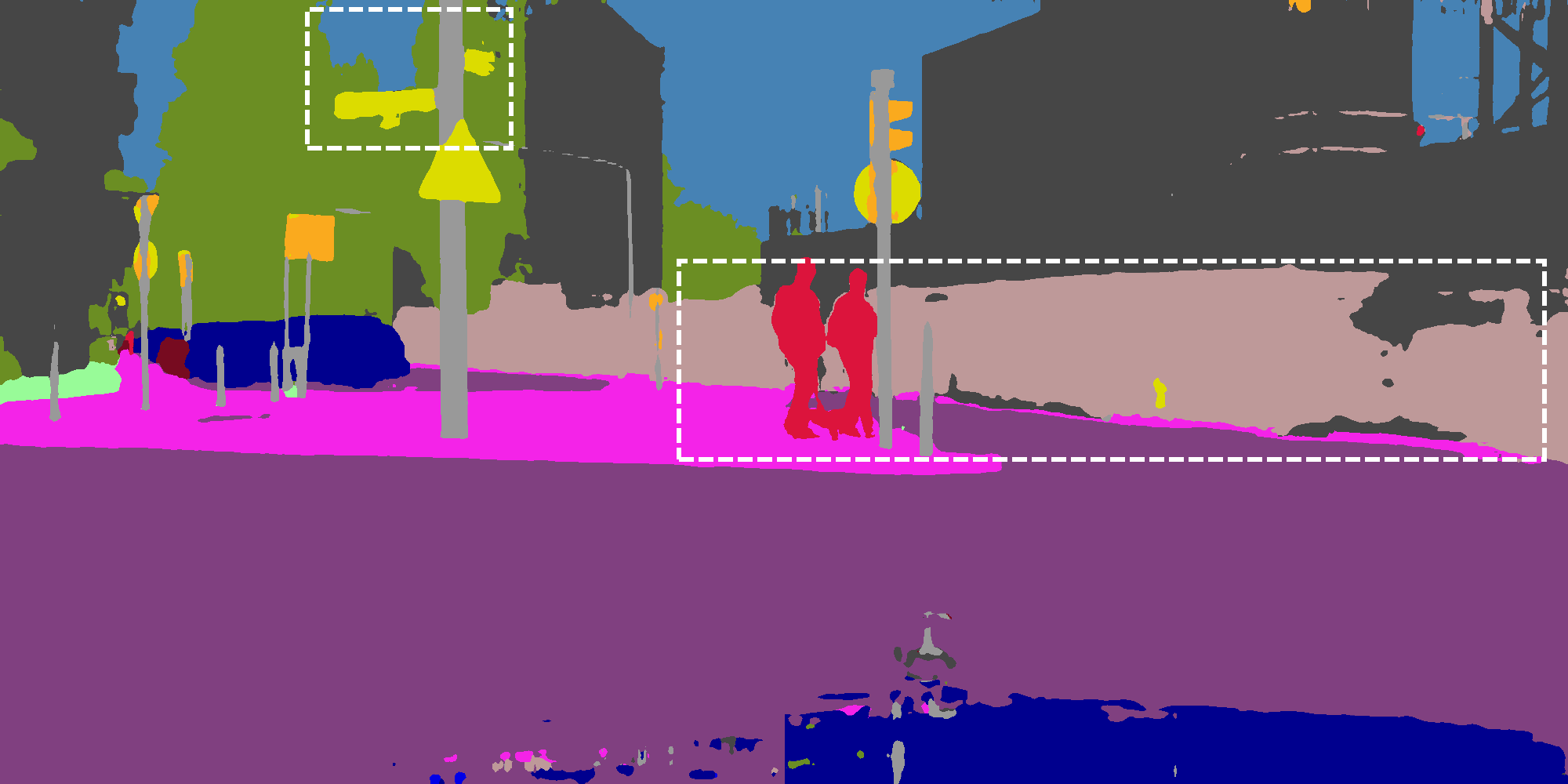}
    \end{subfigure}    \begin{subfigure}{.2\textwidth}
        \centering
        \includegraphics[width=.98\linewidth]{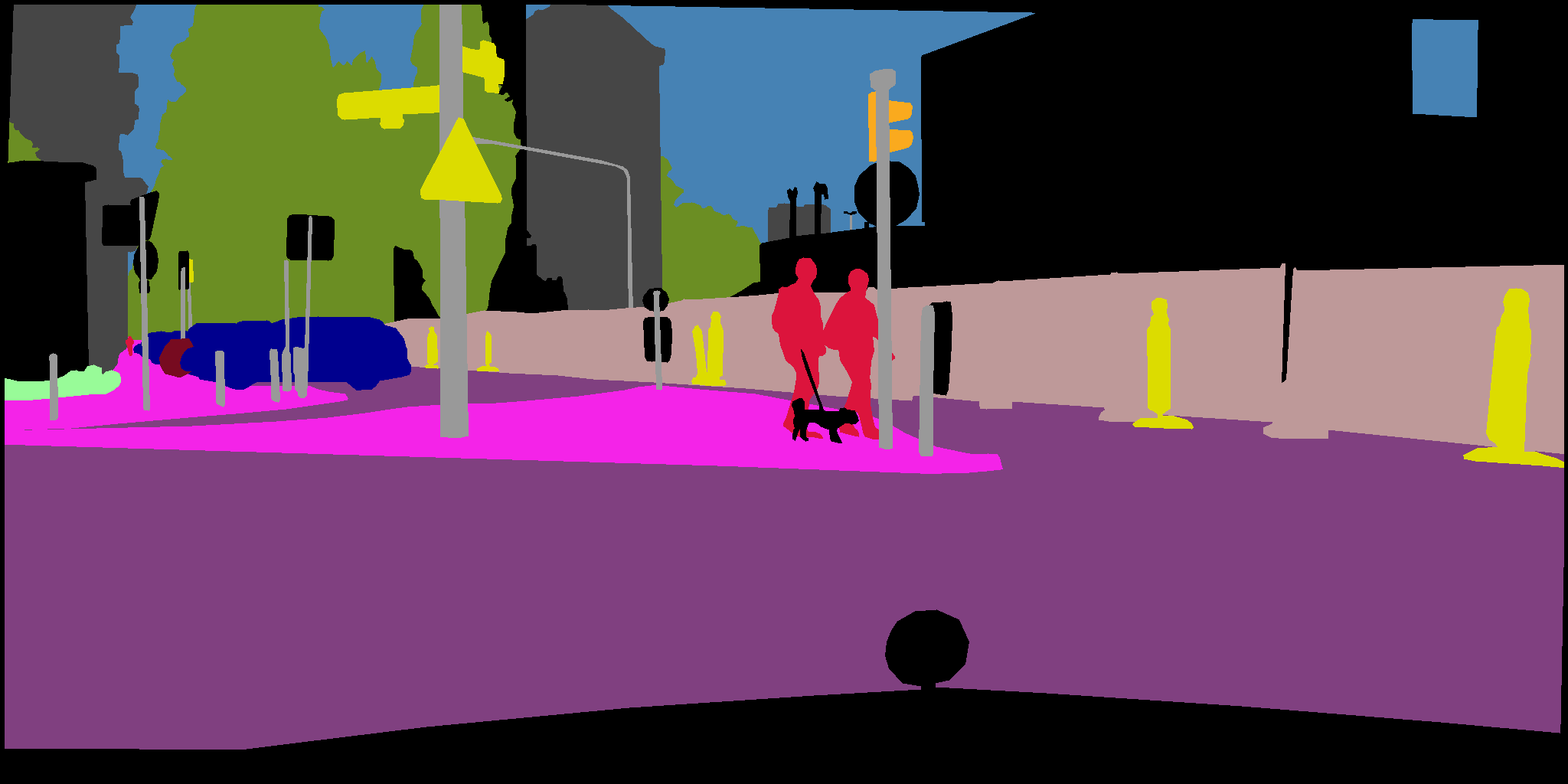}
    \end{subfigure}}
\makebox[\linewidth][c]{    \begin{subfigure}{.2\textwidth}
        \centering
        \includegraphics[width=.98\linewidth]{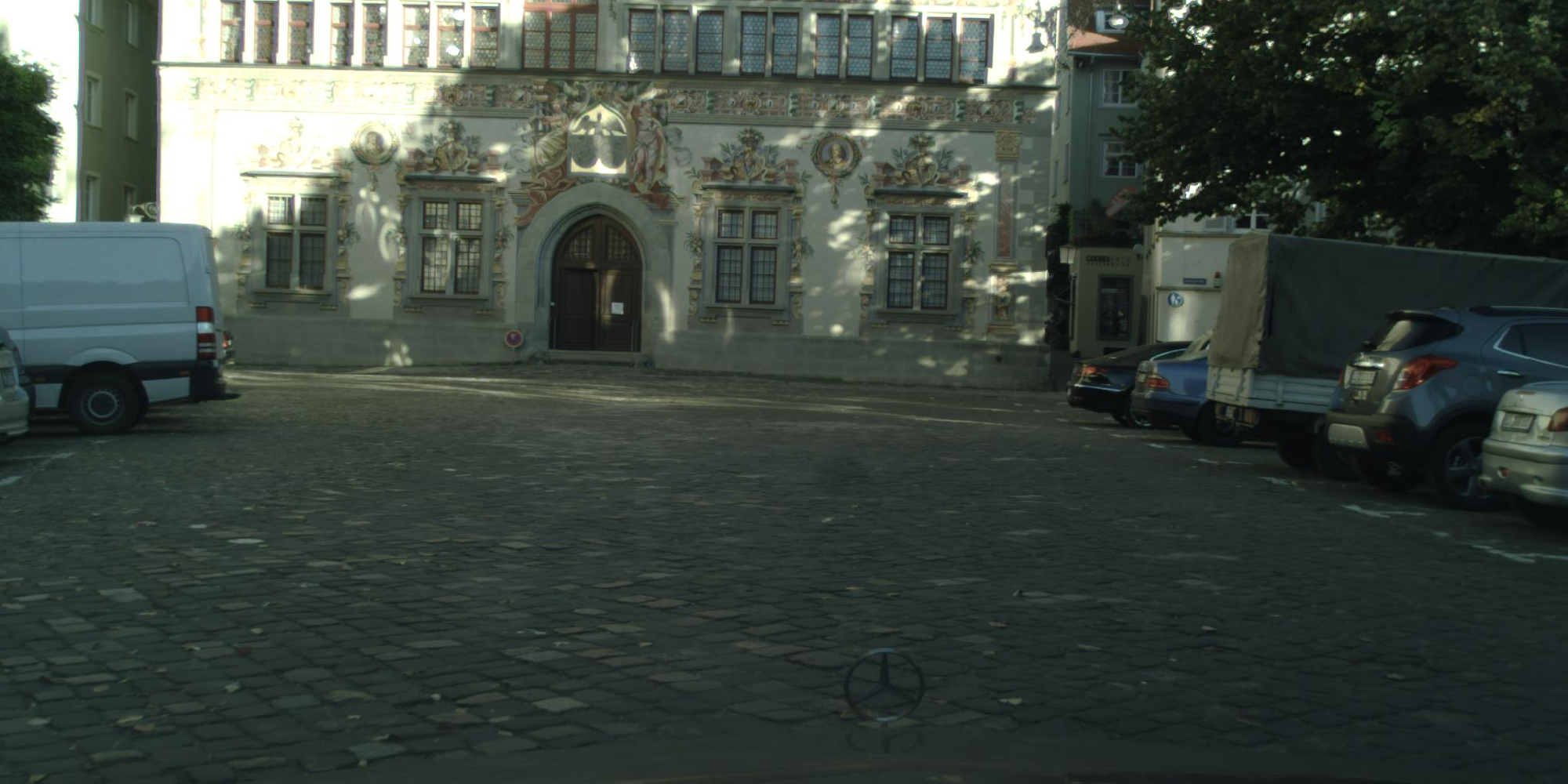}
        \caption*{Target Image}
        \label{fig:sfig1}
    \end{subfigure}    \begin{subfigure}{.2\textwidth}
        \centering
        \includegraphics[width=.98\linewidth]{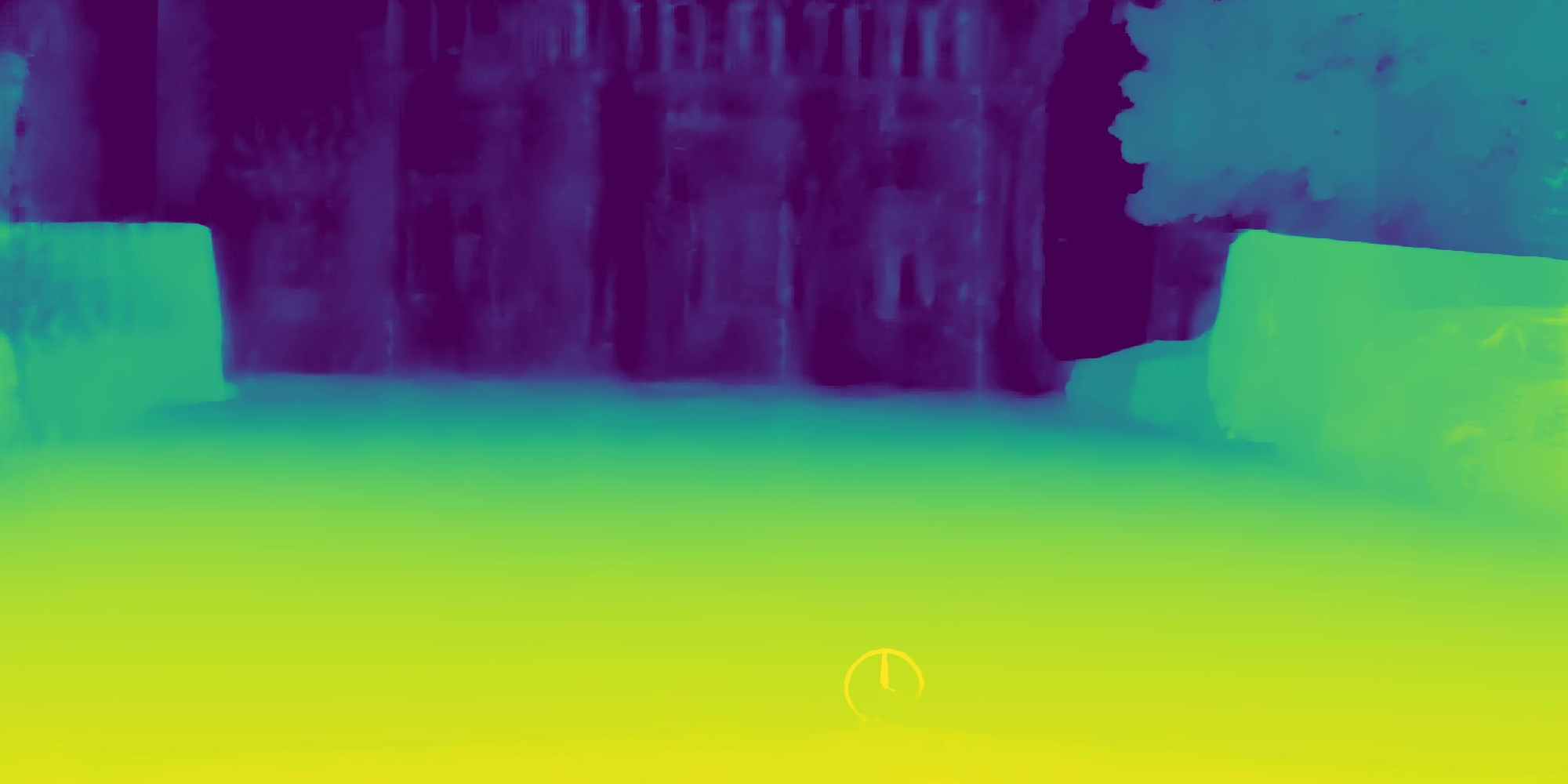}
        \caption*{Estimated Depth}
        \label{fig:sfig2}
    \end{subfigure}    \begin{subfigure}{.2\textwidth}
        \centering
        \includegraphics[width=.98\linewidth]{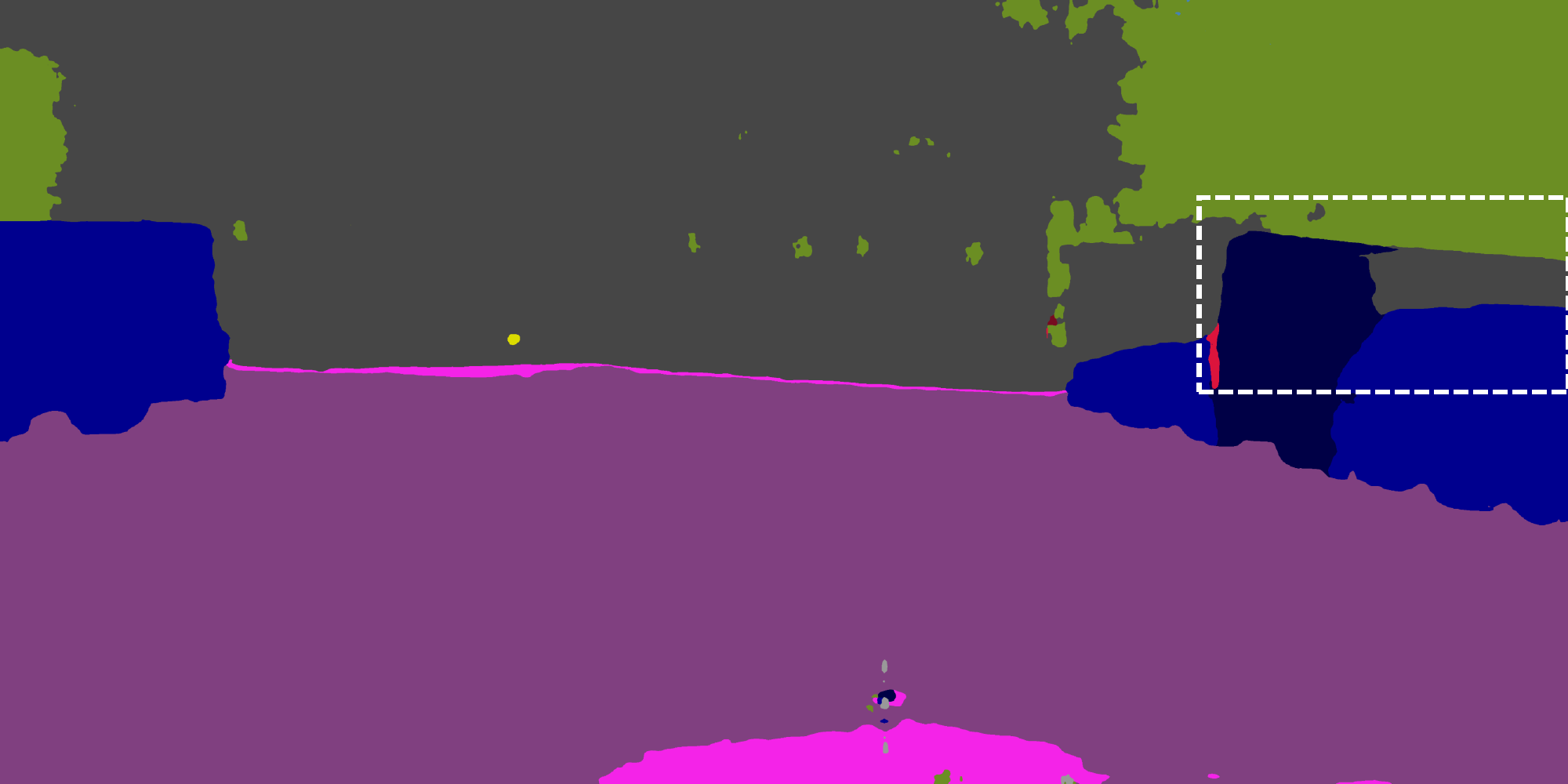}
        \caption*{MIC (HRDA)~\cite{MIC}}
        \label{fig:sfig3}
    \end{subfigure}    \begin{subfigure}{.2\textwidth}
        \centering
        \includegraphics[width=.98\linewidth]{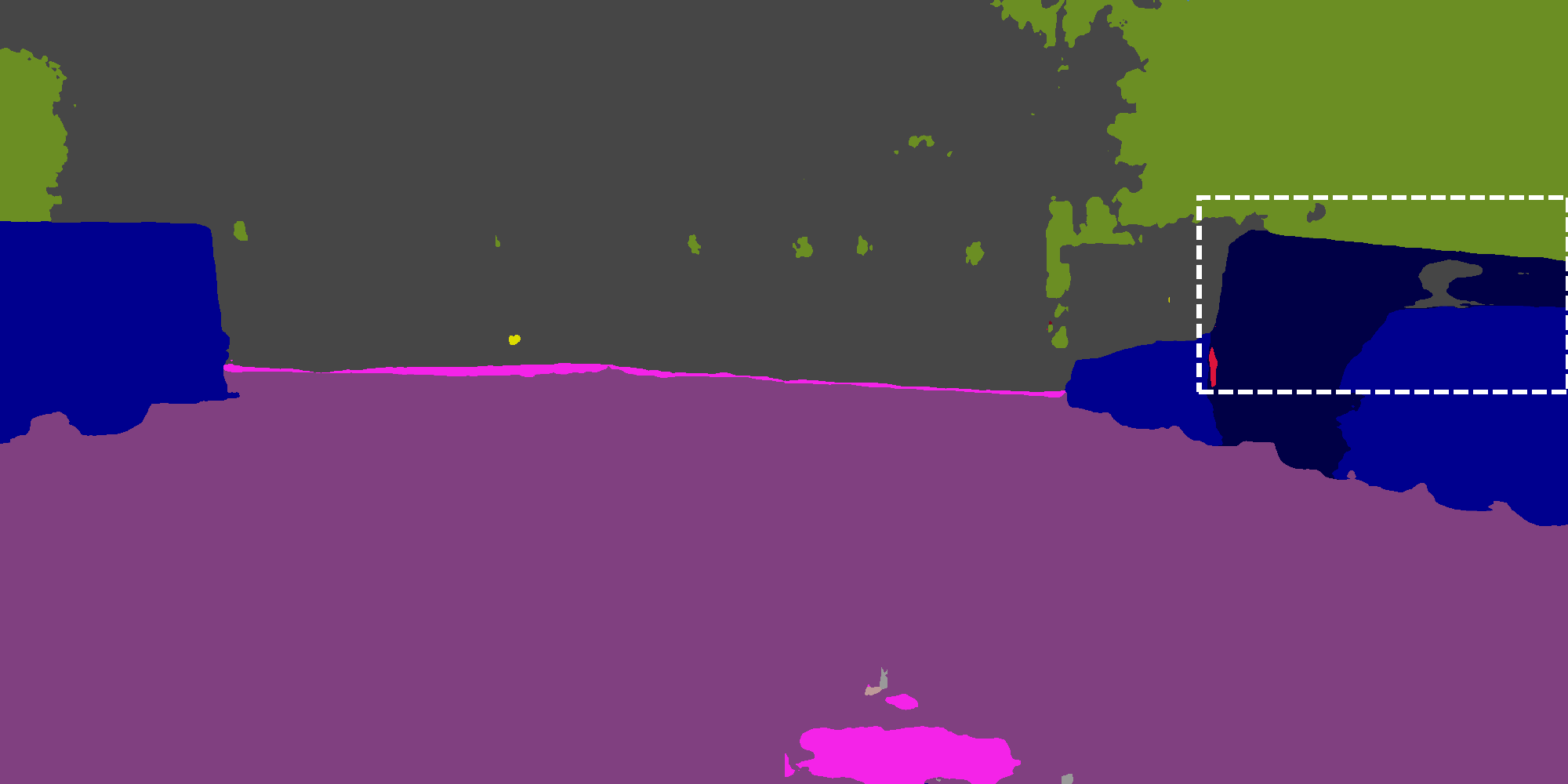}
        \caption*{\method\ (ours)}
        \label{fig:sfig4}
    \end{subfigure}    \begin{subfigure}{.2\textwidth}
        \centering
        \includegraphics[width=.98\linewidth]{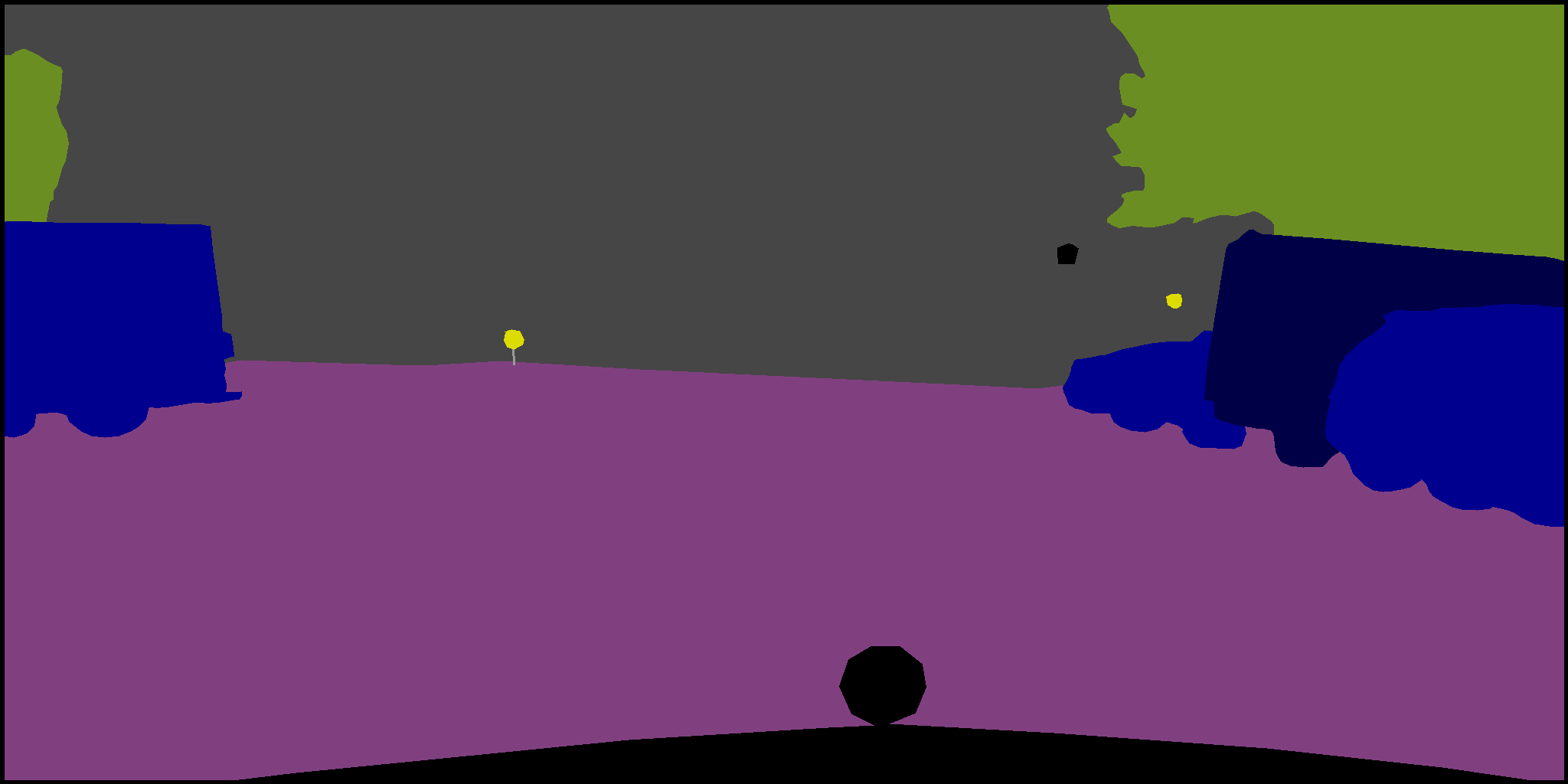}
        \caption*{Ground Truth}
        \label{fig:sfig5}
    \end{subfigure}}

\caption{\textbf{Qualitative results.} These results show the improvements of \method\ in comparison to MIC (HRDA).
We highlight improvements on thin structures, such as pole and traffic sign, as well as on larger objects like trucks, busses and fences.
In rows 1, 3, and 4, we can see that thin structures have a distinct depth profile, which helps in predicting accurate boundaries.
In rows 2, 4, and 5, we observe that the depth region for the fence, bus, and truck is smooth, improving the consistency of the predicted segmentation. 
}
\label{fig:qualitative}
\end{figure}

\tabref{main_results} further provides results on SYNTHIA$\rightarrow$Cityscapes. Also here, MICDrop achieves consistent improvements over its baselines, \ie 1.1 mIoU for DAFormer, 1.0 mIoU for HRDA, and 0.6 mIoU for MIC$_\mathit{HRDA}$. The improvements are slightly smaller than for GTA$\rightarrow$Cityscapes, which might be caused by the smaller dataset size of SYNTHIA, resulting in overfitting issues.

\PAR{Boundary Analysis.}
Tab.~\ref{tab:boundary_iou} additionally studies the boundary IoU~\cite{cheng2021boundary}. Compared to the default IoU it improves by a significantly larger margin (1.6 vs 0.7), supporting our hypothesis that MICDrop particularly improves segmentation boundaries. The class-wise boundary IoUs further demonstrate that both classes with fine structures (\eg pole or sign) and classes that are prone to oversegmentation (\eg truck and building) are improved, quantitatively supporting our motivation in Fig~\ref{fig:teaser}. 

\PAR{Qualitative Analysis.}
In~\figref{qualitative}, we showcase a qualitative comparison with the current state-of-the-art model.
We note that the estimated depth exhibits sharp discontinuities, providing strong cues for thin structures such as poles or traffic signs (\cf row 1, 3, and 4).
Moreover, these examples demonstrate how the piecewise smooth depth can help to mitigate oversegmentation of larger objects by guiding the network to predict more consistent semantic segmentation within depth contours, as can be seen for the truck, the bus, and fence (\cf row 2, 4, and 5).
Further qualitative comparisons with other methods are provided in the supplementary material.

\begin{table*}[b]
\centering
\scriptsize
\resizebox{\linewidth}{!}{
\begin{subtable}[t]{0.7\linewidth}
\centering
\begin{tabular}[t]{lccr}
    \toprule[0.2em]
    Masking Strategy & Masking RGB & Masking Depth & mIoU (\textuparrow)\\
    \toprule[0.2em]
    Baseline (w/o Depth) & \xmark & \xmark & 68.3 \tiny{\textpm 0.5} \\
    Baseline (w/ Depth) & \xmark & \xmark & 69.1 \tiny{\textpm 0.2} \\
    \midrule
    Only RGB  & \cmark & \xmark & 69.3 \tiny{\textpm 0.1} \\
    Independent & \cmark & \cmark & 69.1 \tiny{\textpm 0.6} \\
    \midrule
    Complementary (ours) & \cmark & \cmark & \textbf{70.1} \tiny{\textpm 0.1} \\
    - Different per Level & \cmark & \cmark & 69.7 \tiny{\textpm 0.1} \\
    \bottomrule[0.1em]
\end{tabular}
\caption{\textbf{Dropout strategy ablation.}}
\label{tab:dropout}
\end{subtable}
\begin{subtable}[t]{0.4\linewidth}
\centering
\begin{tabular}[t]{lr}
    \toprule[0.2em]
    Fusion Operation & mIoU (\textuparrow) \\
    \toprule[0.2em]
    Baseline (no Depth) & 68.3 \tiny{\textpm 0.5} \\
    \midrule
    Add & 69.3 \tiny{\textpm 0.4} \\
    CMX~\cite{CMX} & 68.6 \tiny{\textpm 0.3} \\
    \midrule
    Local Self-Attn & 69.7 \tiny{\textpm 0.1} \\
    Global Cross-Attn & 68.1 \tiny{\textpm 0.8} \\
    \midrule
    Local+Global (ours) & \textbf{70.1} \tiny{\textpm 0.1} \\
    \bottomrule[0.1em]
\end{tabular}
\caption{\textbf{Feature Fusion ablation.}}
\label{tab:feature_fusion}
\end{subtable}
}
\caption{\textbf{Ablation study.} We use DAFormer~\cite{DAFormer} trained on GTA as our baseline model. In (a), we study different dropout strategies. In (b), we ablate different designs to fuse RGB and depth features. Mean and std. deviation are reported over 3 seeds.
}
\label{tab:ablation}
\end{table*}

\subsection{Ablation Studies}
\label{sec:ablation}
We start the experimental validation of our design choices by ablating our dropout strategy.
After that, we compare different operations for the task of feature fusion.
For a fair comparison, we also finetune the pretrained baseline model without any changes using the same hyper-parameter described before but did not observe any performance improvements (68.3 \textpm 0.2 mIoU).

\PAR{Cross-Modal Complementary Dropout.}
The ablation study in~\tabref{dropout} explores the impact of various masking strategies.
All experiments use our proposed feature fusion module.
Adding depth information to our baseline without masking increases the mIoU from 68.3 to 69.1 on the GTA dataset, showing the promise of depth.
However, we show experimentally that depth features are not fully utilized by the decoder by testing simple masking strategies first.

When masking RGB features, the network can leverage depth information marginally better by 0.2 percentage points, indicating that feature corruption, when done right, could enhance cross-modal feature integration.
However, applying independent masking to both RGB and depth features simultaneously does not show improvements over no masking.
As evident from a significantly higher standard deviation, strategies in which the same regions in depth and RGB might be masked make the training more unstable.

Notably, using the same \emph{complementary masking across all levels} leads to a substantial gain: an increase of 1.0 mIoU over the baseline with depth (with or without independent masking) and 1.8 mIoU over the DAFormer baseline.
Furthermore, we show that true complementary masking is essential for effective learning.
For that ablation, we allow the network to recover masked features from other feature levels as we apply complementary masking independently at each feature level, resulting in a 0.4 mIoU decrease.
These findings support that complementary masking plays a crucial role in effectively leveraging depth information for semantic segmentation in UDA, as it achieves a great balance between geometric and visual scene information.

\PAR{Feature Fusion.} The fusion of depth and RGB features is the essential block in our RGB-D semantic segmentation.
In~\tabref{feature_fusion}, we compare our proposed module with different feature fusion operations.
To best utilize both modalities, we deploy our proposed complementary masking strategy across all tested feature fusion operations.
We first explore one simple fusion technique, namely feature addition.
The scores show that a naive feature fusion technique exhibits suboptimal performance in our context.
We further examine the SOTA RGB-D method CMX~\cite{CMX}, which fuses features at various encoder stages using cross-attention. However, CMX only obtains a marginal improvement of 0.3 mIoU in the UDA setting.

Turning our focus to the individual efficacy of our proposed global and local feature fusion blocks, we observed distinct outcomes.
The local self-attention block, employed independently, outperformed our naive addition baseline, indicating its effectiveness in contextual feature integration.
In contrast, the global depth-guided cross-attention block, when used alone, failed to demonstrate improvement and exhibited significant training instability, as evidenced by a high standard deviation of 0.8 in mIoU.
Analogous to the results observed with CMX, we conjecture that these findings underscore the significance of controlling the flow of local information in UDA.
However, it is crucial to note that these blocks were designed to \emph{complement} each other.
When combined, their synergy becomes clear, validating our hypothesis that both local and global attention mechanisms are indispensable for optimal performance.
This combination led to a notable improvement of additional 0.4 mIoU over local self-attention, achieving an overall gain of 1.8 mIoU over the baseline~\cite{DAFormer}. In summary, our fusion module effectively harnesses both \emph{global and local cues}, significantly enhancing the overall effectiveness of our RGB and depth feature fusion task.

\section{Conclusion}
\label{sec:conc}
We present a novel complementary dropout method specifically tailored for UDA.
Coupled with our cross-modal fusion module that combines RGB and depth features, our approach consistently improves various recent UDA methods, achieving state-of-the-art results.
In particular, on both GTA and SYNTHIA, \method\ achieves a boost of 0.7 to 1.8 mIoU, depending on the method used for encoding RGB features.
Thus, \method\ demonstrates the effectiveness of utilizing depth in UDA without the need for retraining existing encoders, achieved by adopting a many-to-one prediction framework rather than traditional multi-task learning or auxiliary predictions.
The plugin design of \method\ is intended to facilitate ease of integration into future domain-adaptive semantic segmentation methods.
We hope that our simple but effective approach inspires further research into leveraging complementary cues in UDA.

%
%
\bibliographystyle{splncs04}
\bibliography{main}

\renewcommand{\thesection}{\Alph{section}}
\renewcommand{\thetable}{S\arabic{table}}
\renewcommand{\thefigure}{S\arabic{figure}}

\Crefname{figure}{\textbf{Figure.}}{\textbf{Figures.}}

\clearpage
\noindent\textbf{\Large Supplementary Material}

\setcounter{section}{0}
\setcounter{table}{0}
\setcounter{figure}{0}

\section{Overview}
In~\cref{sec:supp_arch}, we provide further insight into our design decisions on the encoder architecture and overall training strategy.
Furthermore, we investigate the impact of various hyperparameters for masked feature training in~\cref{sec:supp_mask_parameters}.
In~\cref{sec:limitations}, we provide a discussion on the potential limitations of our method.
Finally, we supplement our experimental evaluation in \cref{sec:supp_qual_eval} with additional qualitative results obtained from various UDA methods. All of our experiment results are reported over three different random seeds.   

\section{Architecture}
\label{sec:supp_arch}

\begin{figure}
    \centering
    \includegraphics[width=0.7\linewidth]{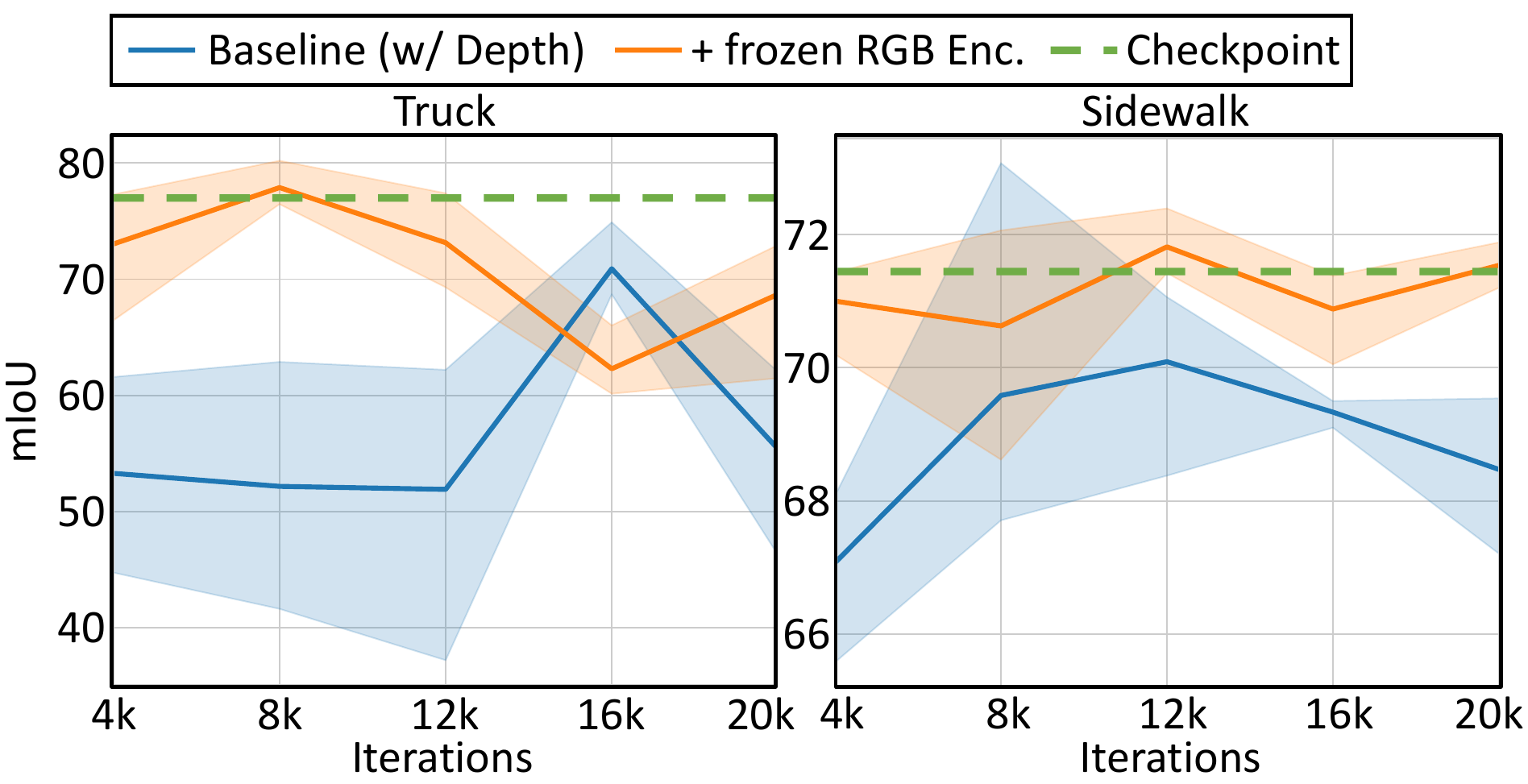}
    \caption{\textbf{Classwise performance.} This figure highlights not only improved average performance but also a reduction of strong deviations in classwise performances when using a frozen backbone. The dotted checkpoint line indicates the model's performance at its initialization with pretrained weights.}
    \label{fig:encoder_classwise}
\end{figure}

\PAR{Frozen Pretrained RGB Encoder.}
We observe that it is not necessary to further train a pretrained RGB encoder, as evidenced in~\cref{tab:frozen_encoder}.
In fact, we find that additional training tends to destabilize the encoder. 
As presented in~\cref{fig:encoder_classwise}, this effect is particularly noticeable for large structures such as \textit{truck} or \textit{sidewalk}.
Freezing the RGB encoder parameters prevents representation drift, allowing us to use the depth information effectively. 
Furthermore, freezing the RGB encoder considerably reduces the resource requirements for training, facilitating reproduction by other researchers.
This is reflected in the higher throughput during training, lower VRAM utilization, and a lower number of GPUs used.

\section{Mask Parameters}
\label{sec:supp_mask_parameters}
We use the following designations to introduce depth features as an additional baseline.
We refer to the model without depth features as the RGB baseline. 
Conversely, the depth baseline denotes the model that incorporates depth features without the use of complementary masking.
Performance-wise, the RGB baseline achieved 68.3 mIoU, while the depth baseline obtained 69.1 mIoU.

\begin{table}
\footnotesize
\centering
\resizebox{0.8\columnwidth}{!}{ \begin{tabular}[b]{lcccc}
    \toprule[0.2em]
    & \multicolumn{3}{c}{Resources} & Metric \\
    Architecture & Throughput & Mem. per GPU &  \#GPUs & mIoU (\textuparrow)\\
    \toprule[0.2em]
    Baseline (w/ Depth) & 0.47 it/s & 11.09 GB & 2 & 68.7 \tiny{\textpm 0.3}\\
    + freeze RGB Enc. & 0.70 it/s & 8.24 GB & 1 & 69.1 \tiny{\textpm 0.2} \\
    \bottomrule[0.1em]
\end{tabular}
}
\caption{\textbf{Effect of freezing the RGB encoder.} The tables highlight the benefits gained from freezing the RGB encoder. This process notably decreases resource usage while also yielding slight performance improvements.}
\label{tab:frozen_encoder}
\end{table}

\subsection{Image Batch}
\label{sec:batch}

We use images from the source domain $\textbf{X}_s$ and from the target domain $\textbf{X}_t$ in each training iteration. 
The batch of source images is designated as the \emph{source batch}.
Building on the findings of ~\cite{DACS}, several subsequent methods~\cite{DAFormer, HRDA, SePiCo, chen2023pipa} have adopted cross-domain class mix augmentation.
Following this approach, we adopt a similar augmentation pipeline, resulting in a class mix image batch, hereafter referred to as \emph{mix batch}.
With the introduction of our masking strategy, we use \emph{full batch} to refer to batches containing unmasked features from both the \emph{mix batch} and \emph{source batch}.
Using~\cref{table:batches}, we discuss the influence of using masked features of individual image batches separately and in combination with the \emph{full batch}.

\begin{table}
\begin{center}
\footnotesize
\begin{tabular}[b]{lcccr}
    \toprule[0.2em]
    & Full Feat. & Mask Src Feat. & Mask Mix Feat. & mIoU (\textuparrow)\\
    \toprule[0.2em]
    1 & \cmark & \xmark & \xmark & 69.1 \tiny{\textpm 0.2} \\
    2 & \xmark & \cmark & \xmark & 66.4 \tiny{\textpm 0.1} \\
    3 & \xmark & \xmark & \cmark & 67.8 \tiny{\textpm 0.2} \\
    \midrule
    4 & \cmark & \cmark & \xmark & 69.0 \tiny{\textpm 0.3} \\
    \rowcolor{ours} 5 & \cmark & \xmark & \cmark  & \textbf{70.1 \tiny{\textpm 0.1}} \\
    \midrule
    6 & \xmark & \cmark & \cmark & 68.8 \tiny{\textpm 0.2} \\
    7 & \cmark & \cmark & \cmark & 69.9 \tiny{\textpm 0.3} \\
    \bottomrule[0.1em]
\end{tabular}
\centering
    \caption{\textbf{Detailed analysis of the choice of image batches used for loss computation}. The chosen composition of image batches used in the paper is highlighted in \colorbox{ours}{green}.}
    \label{table:batches}
\end{center}
\end{table}

\PAR{Full Features.}
In row 1, using the \emph{full batch}, we demonstrate the depth baseline. Here, the unmasked features are fused with our presented method.
In this context, we can observe an improvement of +0.8 mIoU compared to the RGB baseline.
An analysis of Rows 1, 4, 5, and 7 reveals that employing the \emph{full batch} is essential for stabilizing performance.

\PAR{Masked Source Features.}
Masking the feature maps aims at learning redundancies across modalities.
As shown in row 2, it is evident that simply retraining on source images alone is unable to recover the RGB baseline performance, achieving the lowest performance of all combinations with an overall decrease of -1.9 mIoU.
This result highlights the significant distribution gap between inference and training. Additionally, the lack of target domain samples causes the model to overfit on the source domain.
However, when combined with the \emph{mix batch} in row 6, we can see that the combination of both batches beats the performance of the individual components but is not sufficient to close the performance gap to the \emph{full batch} on its own.
Notably, in row 4, the addition of the \emph{full batch} results in an improvement of +0.6 mIoU. As this performance is lower than solely training on the \emph{full batch}, row 1,  we suspect that the \emph{source batch} hinders further improvements.

\PAR{Masked Mix Features.}
As observed earlier, we show in row 3 that solely optimizing on the masked image batch is not sufficient to obtain the RGB baseline performance. 
However, in combination with the \emph{full features}, this significantly outperforms the RGB baseline performance with an overall improvement of +1.8 mIoU.
Further adding the masked \emph{source batch} reduces the overall performance slightly yet still maintains a significant improvement of +1.7 mIoU.
Thus, we argue that for optimal improvement, mainly the masked \emph{mix batch} and the \emph{full batch} are necessary.

\subsection{Patch Masking Ratio}
\label{sec:ratio}

An important hyperparameter for the masking strategy is the ratio between the complementary RGB and depth features.
In our experiments, we investigate the effect of fixed ratios for the entire training and a scheduled ratio, which linearly adjusts the ratio during training. 

\begin{figure}[t]
    \centering
\includegraphics[width=0.5\linewidth]{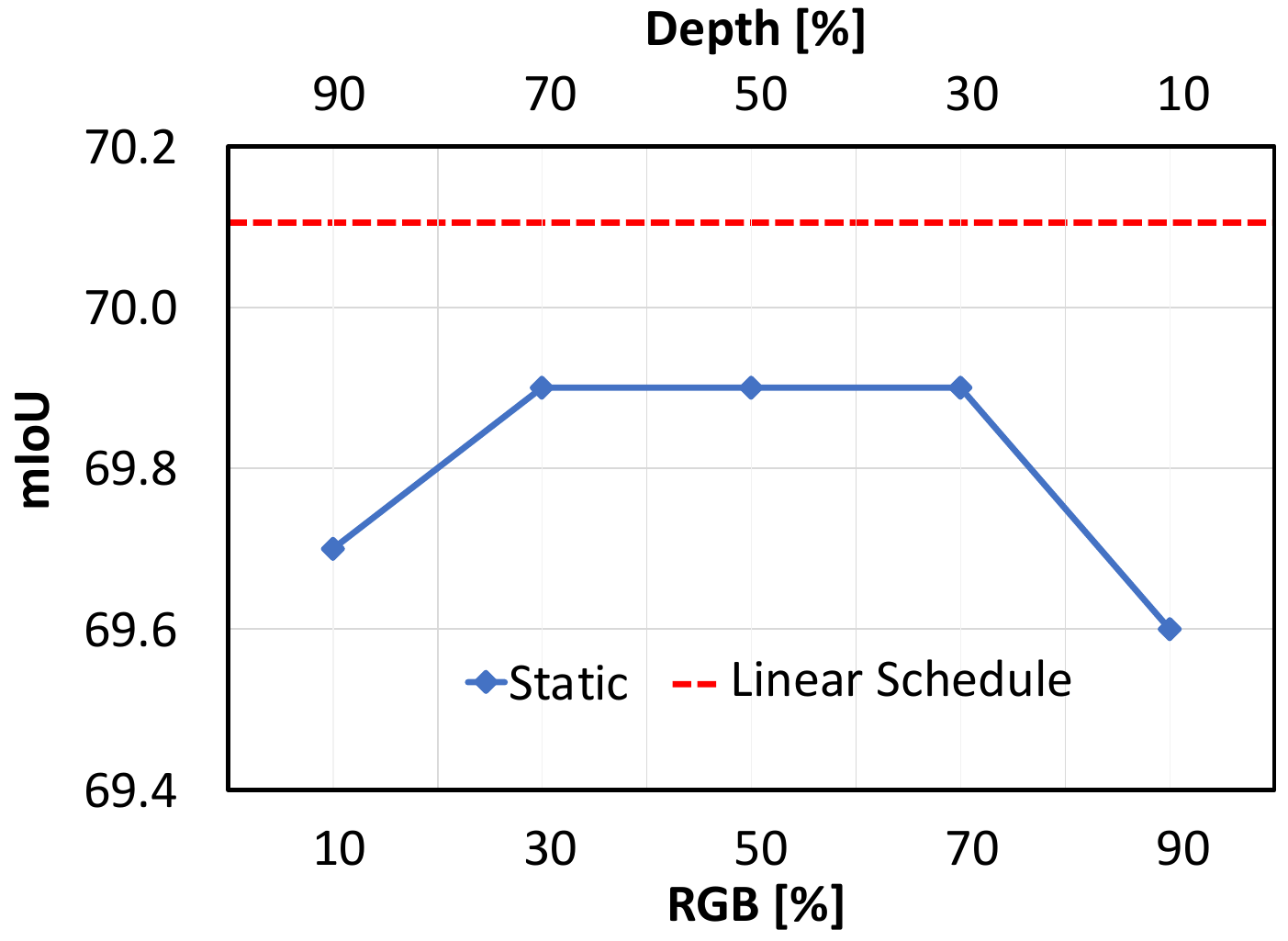}
    \caption{\textbf{Effect of Patch Masking Ratio.} Here, the \textcolor{blue}{blue} line represents a consistent ratio between the different modalities. Conversely, the \textcolor{red}{red} line illustrates a gradual adjustment.}
    \label{fig:patch_ratio}
\end{figure}

\PAR{Fixed Ratio.}
For fixed ratios, as depicted in blue in~\cref{fig:patch_ratio}, it is evident that only ratios that substantially reduce the proportion of a single modality yield inferior results across experiments.
Overall, we find that our method consistently yields significant improvements over the depth baseline across a range of modality ratios. Note that even inferior choices improve performance over the depth baseline.
\PAR{Linear Schedule.}
Beyond fixed modality ratios, we also explore the use of a linear schedule.
This approach, represented in red in~\cref{fig:patch_ratio}, obtains the overall best performance, outperforming other experiments across all evaluated ratios.

\subsection{Patch Size}
\label{sec:patch_size}
This section investigates the effect of different patch sizes for masking feature maps, as shown in~\cref{tab:patch_size_}. 
Here, the patch sizes are specified in relation to the input image size and downsampled to the same scale for the respective feature map. 
Note that the same mask is used for all levels of the hierarchical encoder.
Our observations reveal that larger image patches and, thus, larger regions with exclusively one modality result in the strongest performance. 
These findings support our hypothesis, in which we argue that the larger image patches hinder the ability to reconstruct information using only RGB features. This obstruction necessitates the use of depth features, thus leading to improved performance.

\begin{table}
\begin{center}
\footnotesize
\begin{tabular}[t]{llr}
    \toprule[0.2em]
    Patch Size & mIoU (\textuparrow) \\
    \toprule[0.2em]
    16$^\ddagger$ & 69.4 \tiny\textpm{0.1}\\
    32 & 69.7 \tiny{\textpm{0.3}} \\
    \rowcolor{ours} 64 & 70.1 \tiny{\textpm{0.1}} \\
    128 & 69.9 \tiny{\textpm 0.2} \\
    \bottomrule[0.1em]
\end{tabular}

\centering
    \caption{\textbf{Effect of various patch sizes.} $^\ddagger$ denotes the special case, where the masked patch size is smaller than the corresponding bottleneck feature layer stride. In this case, masking for the bottleneck features is omitted, as consistent masking across scales is not possible. We highlight the patch size we use for our experiments in \colorbox{ours}{green}.}
    \label{tab:patch_size_}
\end{center}
\end{table}

\section{Potential Limitations}
\label{sec:limitations}
\begin{figure}[t]
    \begin{subfigure}{.24\columnwidth}
        \centering
        \caption*{MIC (HRDA)\cite{MIC}}
    \includegraphics[width=\textwidth]{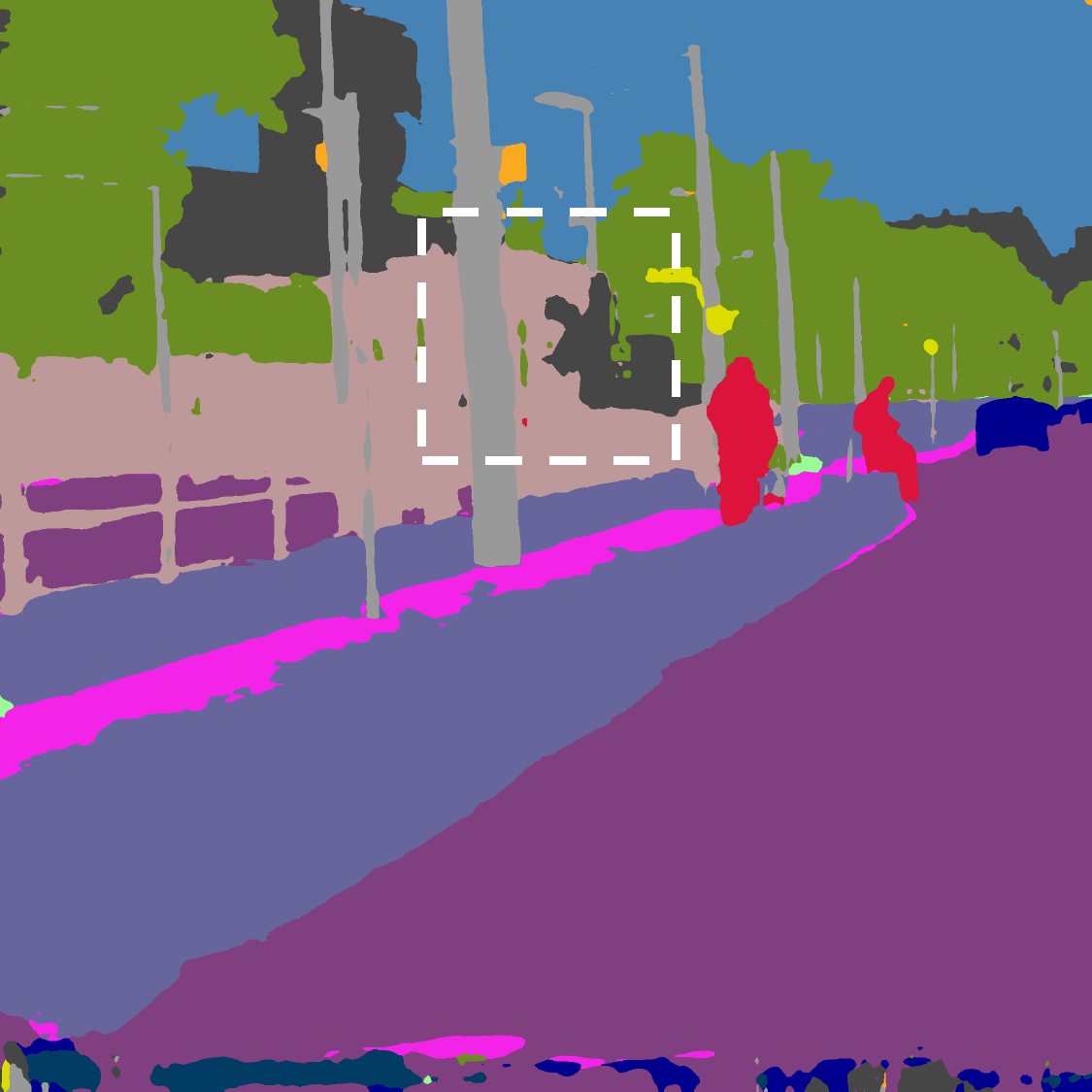}
    \end{subfigure}    
    \begin{subfigure}{.24\columnwidth}
        \centering
        \caption*{\method\ (Ours)}        \includegraphics[width=\textwidth]{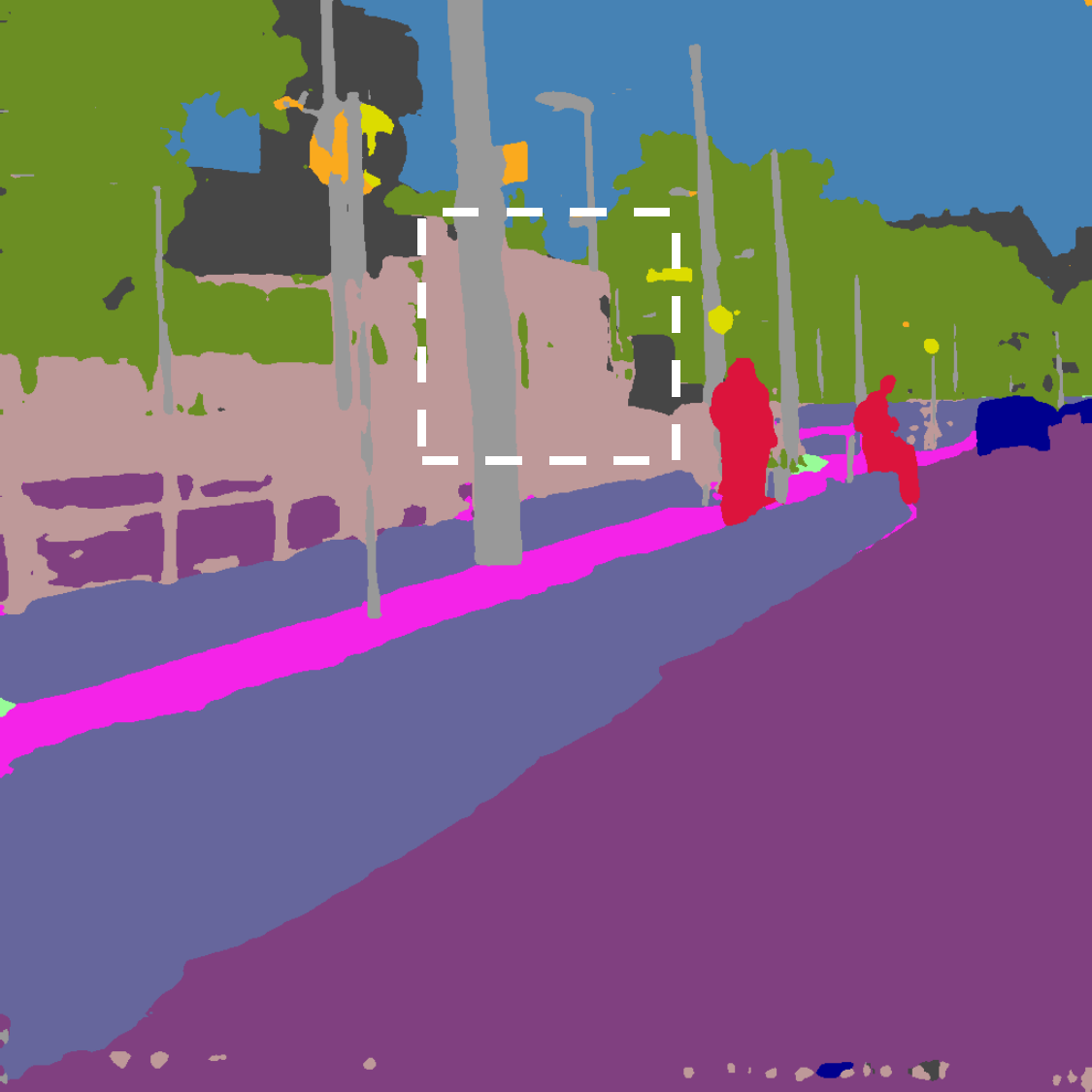}
    \end{subfigure}
    \begin{subfigure}{.24\columnwidth}
        \centering
        \caption*{Target Image}
    \includegraphics[width=\textwidth]{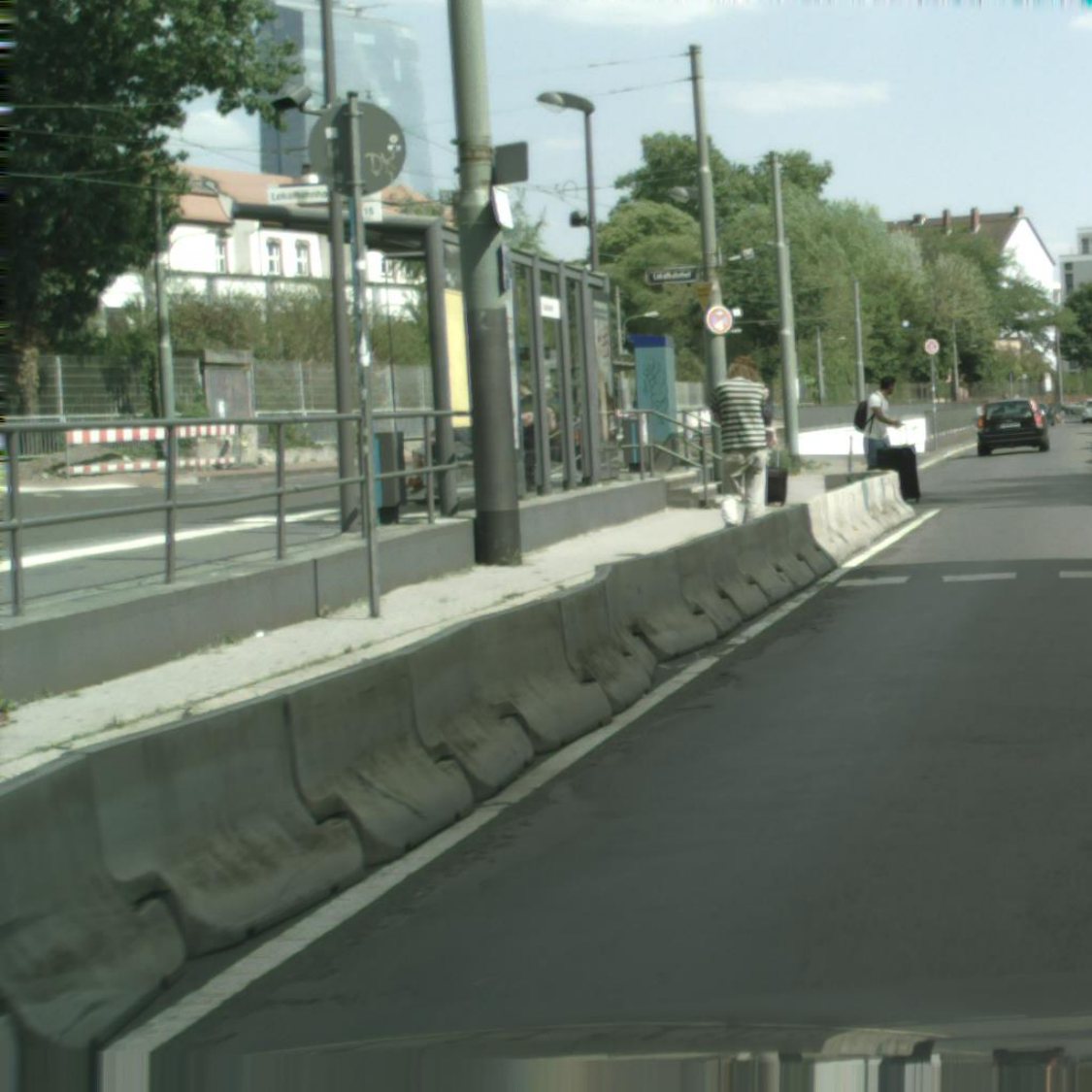}
    \end{subfigure}    
    \begin{subfigure}{.24\columnwidth}
        \centering
        \caption*{Ground Truth}
        \includegraphics[width=\textwidth]{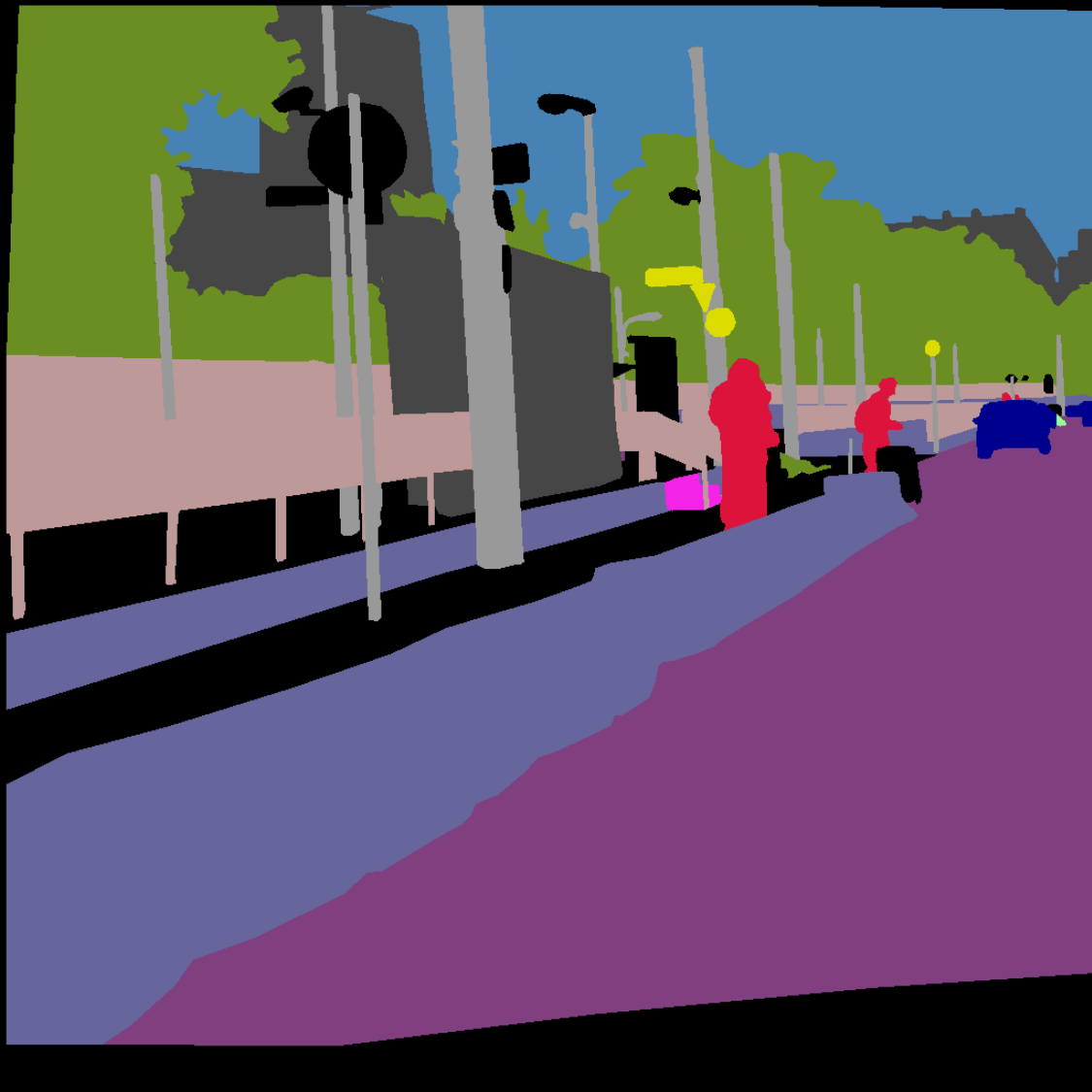}
    \end{subfigure}\centering
\caption{\textbf{Incorrect label propagation}. The baseline method struggles to accurately classify \textit{building}. Incorporating depth information further exacerbates this issue. Driven by the uniform depth, the inaccurate prediction is propagated to the entire object.}
\label{fig:label_propagation}
\end{figure}

\PAR{Hyperparameters.}
Our proposed method includes several hyperparameters, such as masking patch size, masking ratio, and image domain batch. In~\cref{fig:patch_ratio}, \cref{table:batches}, and~\cref{tab:patch_size_}, we explore the effect of these hyperparameters. We notice that moderate deviations from the selected hyperparameters continue to demonstrate strong improvements and are, therefore, reasonably robust. Nevertheless, there these hyperparameters might not be optimal for a custom dataset and might require additional tuning.

\PAR{Label propagation.}
A notable challenge in current state-of-the-art UDA methods is their limited capability of accurately identifying segmentation boundaries resulting in oversegmentation of objects with ambiguous appearance. To address this challenge, MICDrop employs geometric information \ie depth as cue to propagate semantic information. While we show that this enhances segmentation performance significantly, we notice that it can also be disadvantageous in some cases. As shown in ~\cref{fig:label_propagation}, we see an example in which the majority of the \textit{building} class is predicted erroneously as \textit{fence} by the pretrained RGB encoder. Using geometric cues, these problems of the pretrained RGB network are inherited and, in this case, amplified. 
However, this case rarely occurs such that MICDrop's advantages of mitigated oversegmentation and segmentation of fine structures quantitatively outweigh this occasional failure mode.

\section{Qualitative Evaluation}
\label{sec:supp_qual_eval}
\begin{figure*}
\makebox[\linewidth][c]{    \begin{subfigure}{\linewidth}
        \centering
        \label{fig:palette}
        \includegraphics[width=\linewidth]{figures/color_palette.png}
    \end{subfigure}}
\makebox[\linewidth][c]{    \begin{subfigure}{.2\textwidth}
        \centering
        \includegraphics[width=.98\linewidth]{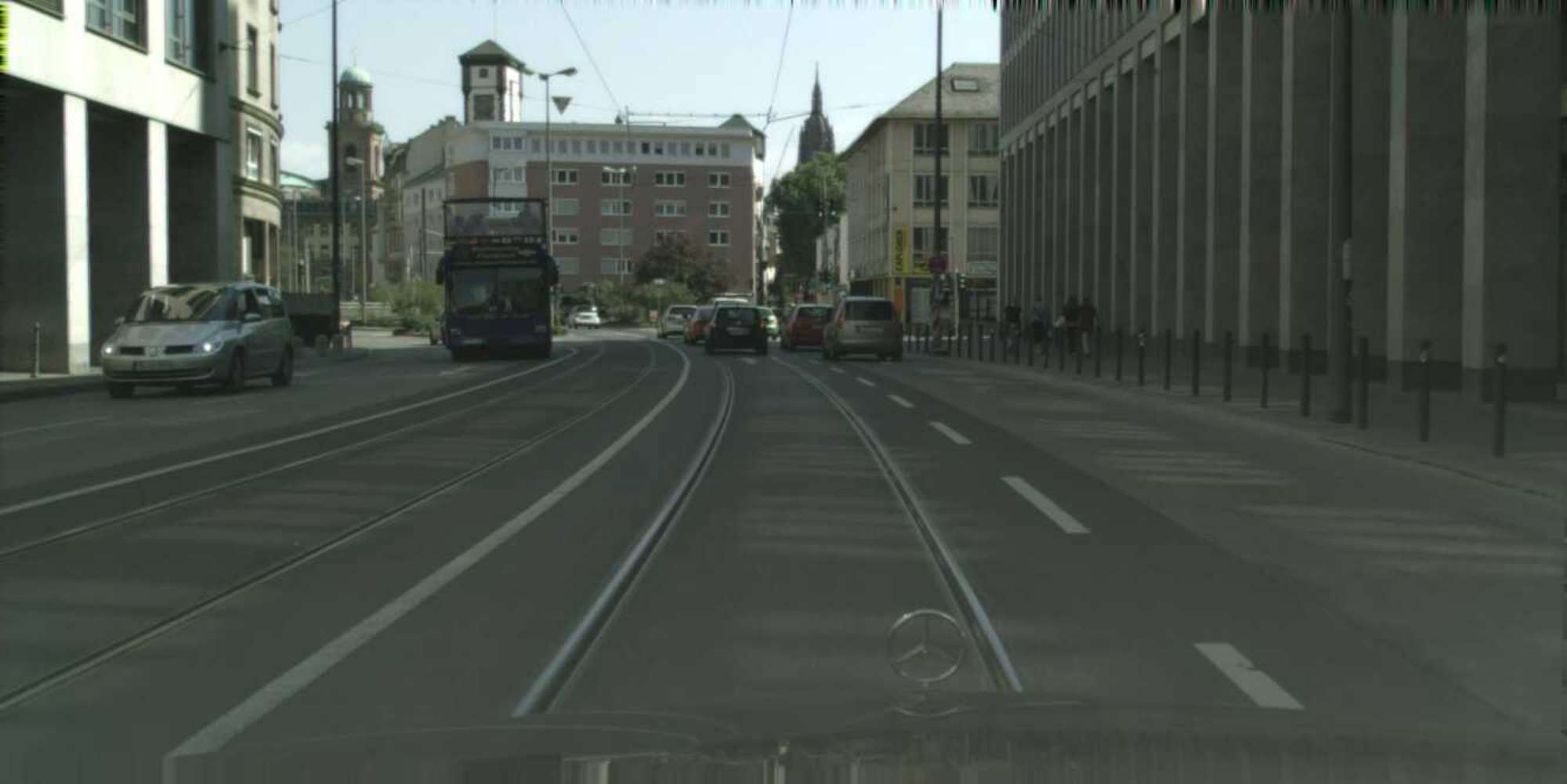}
    \end{subfigure}    \begin{subfigure}{.2\textwidth}
        \centering
        \includegraphics[width=.98\linewidth]{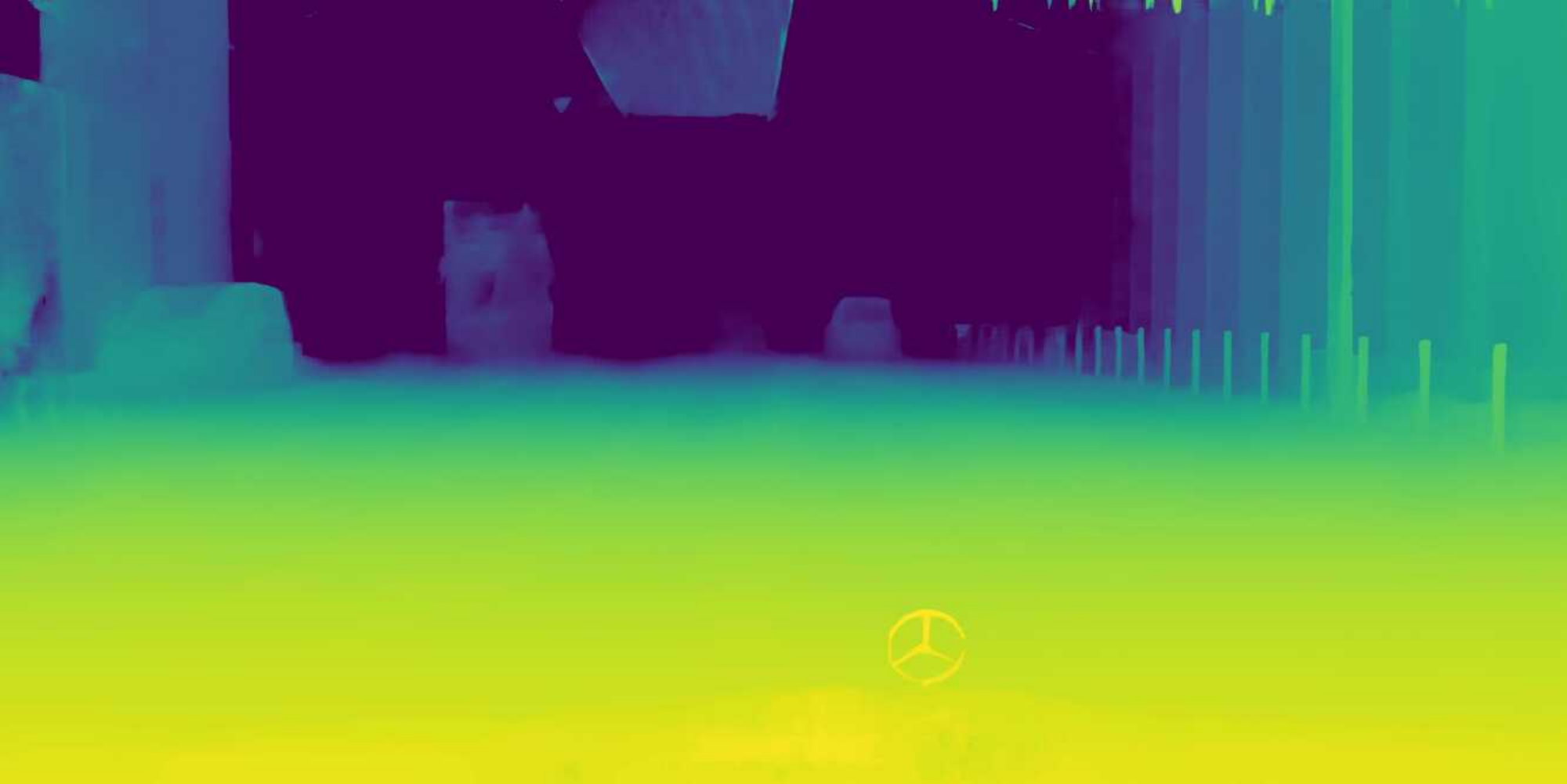}
    \end{subfigure}    \begin{subfigure}{.2\textwidth}
        \centering
        \includegraphics[width=.98\linewidth]{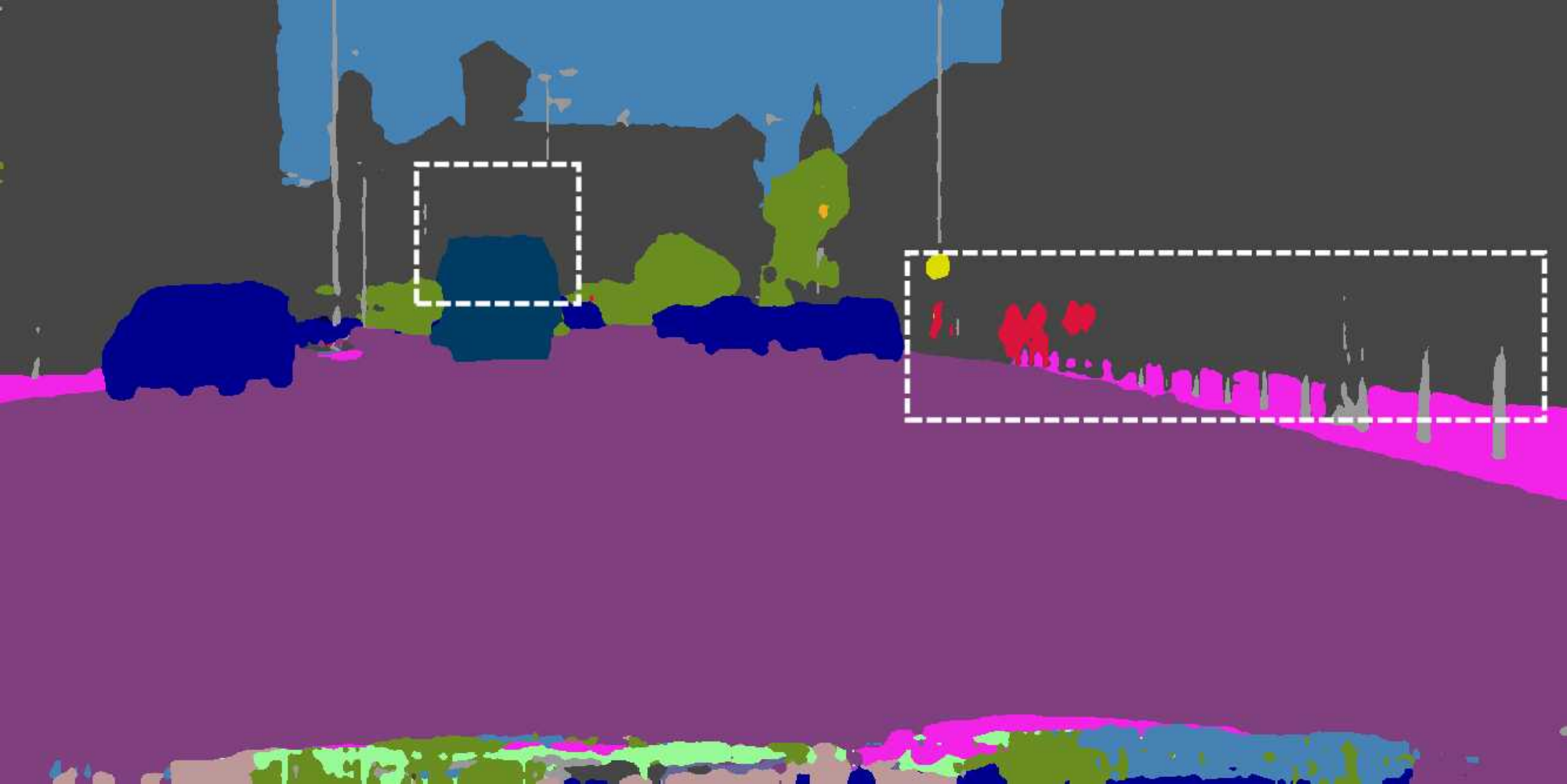}
    \end{subfigure}    \begin{subfigure}{.2\textwidth}
        \centering
        \includegraphics[width=.98\linewidth]{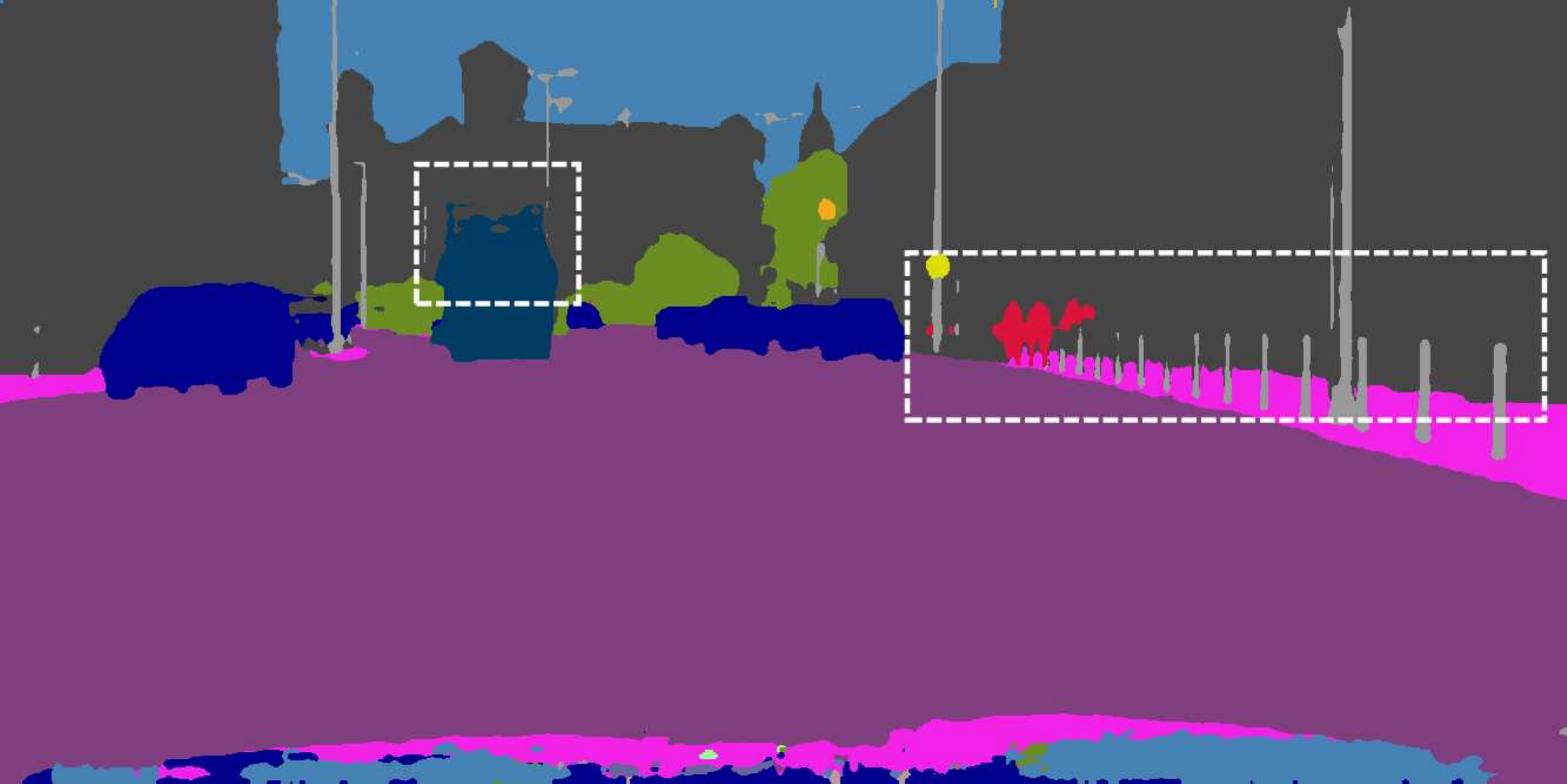}
    \end{subfigure}    \begin{subfigure}{.2\textwidth}
        \centering
        \includegraphics[width=.98\linewidth]{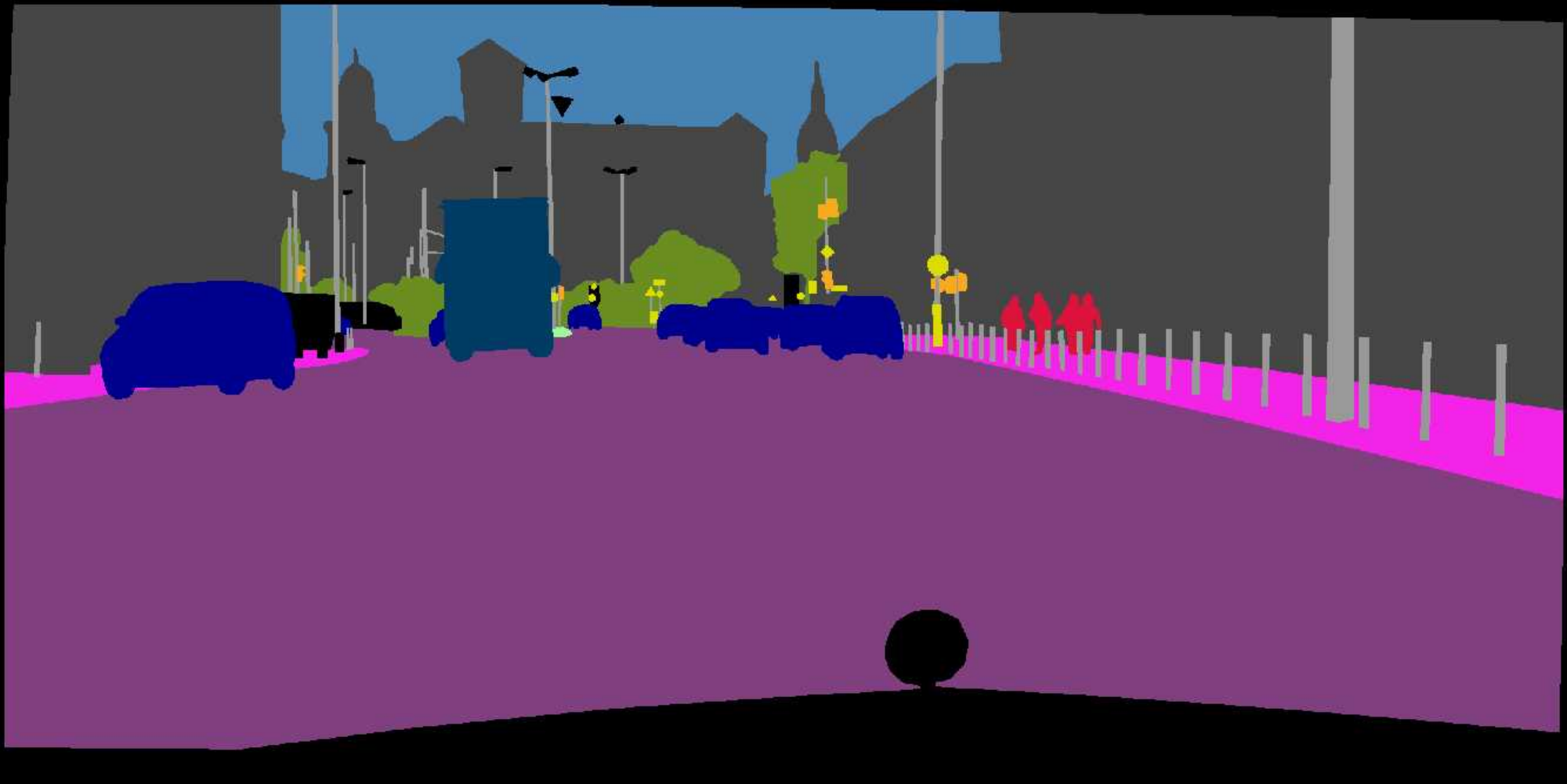}
    \end{subfigure}}
\makebox[\linewidth][c]{    \begin{subfigure}{.2\textwidth}
        \centering
        \includegraphics[width=.98\linewidth]{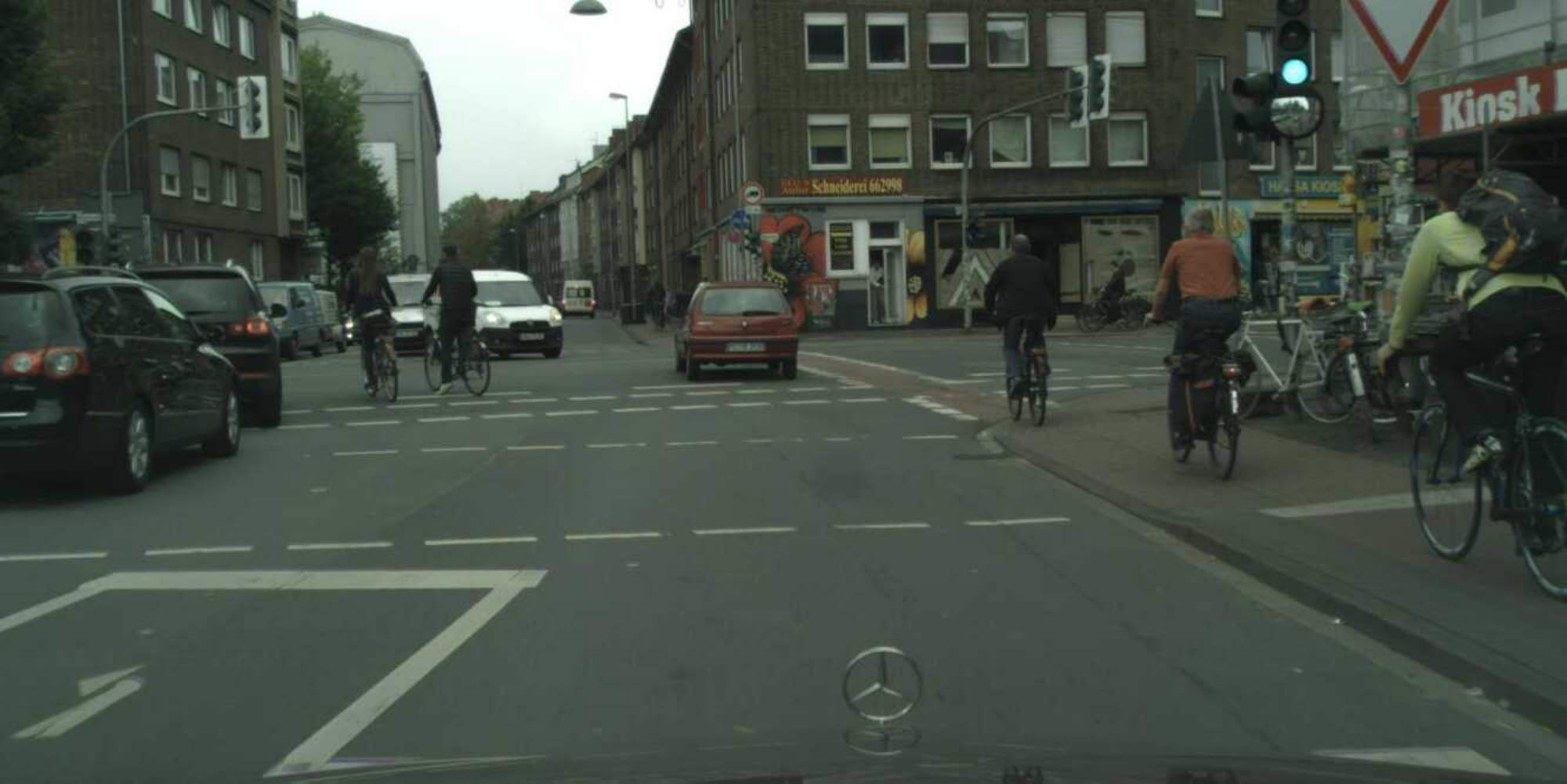}
    \end{subfigure}    \begin{subfigure}{.2\textwidth}
        \centering
        \includegraphics[width=.98\linewidth]{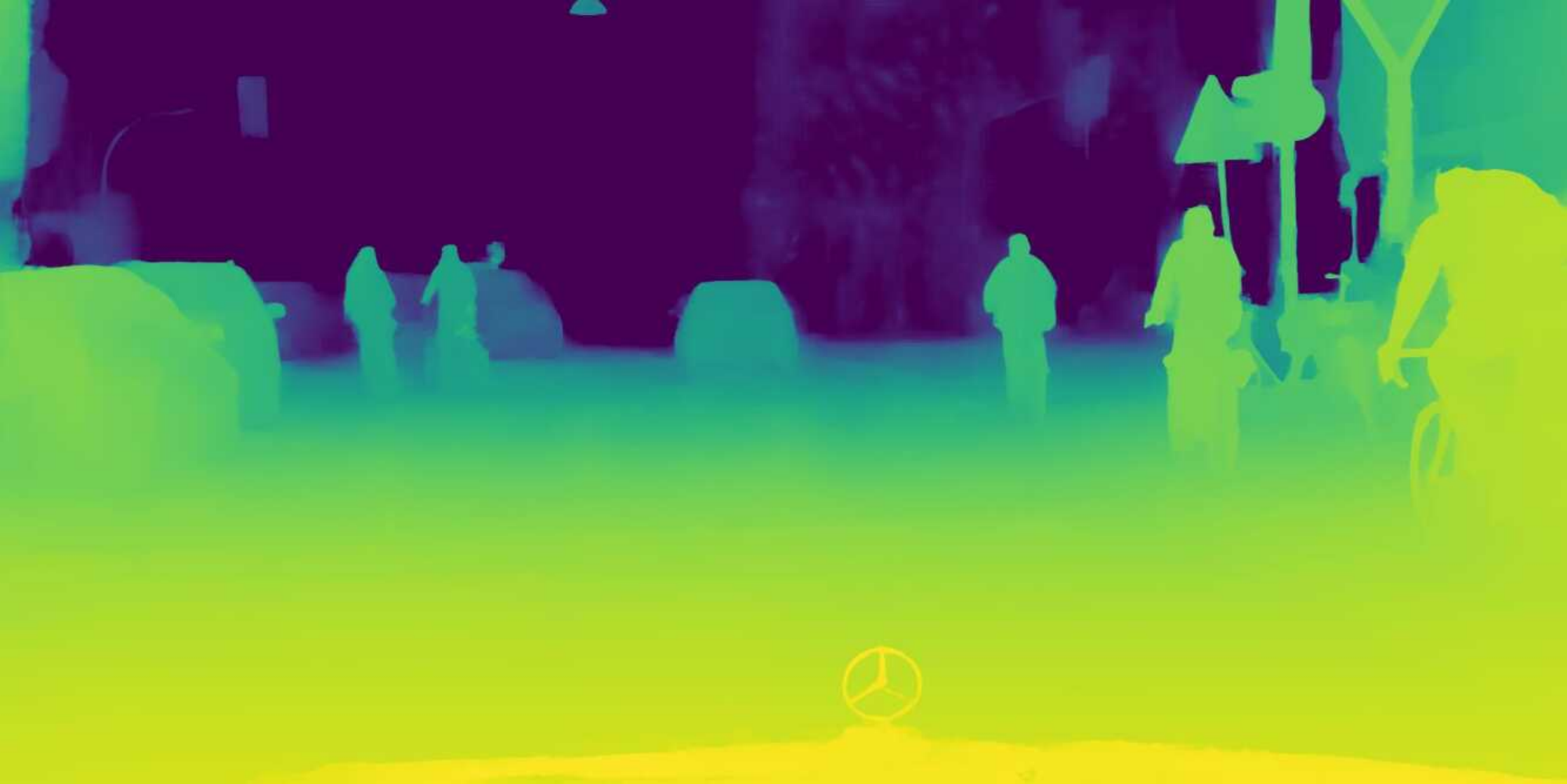}
    \end{subfigure}    \begin{subfigure}{.2\textwidth}
        \centering
        \includegraphics[width=.98\linewidth]{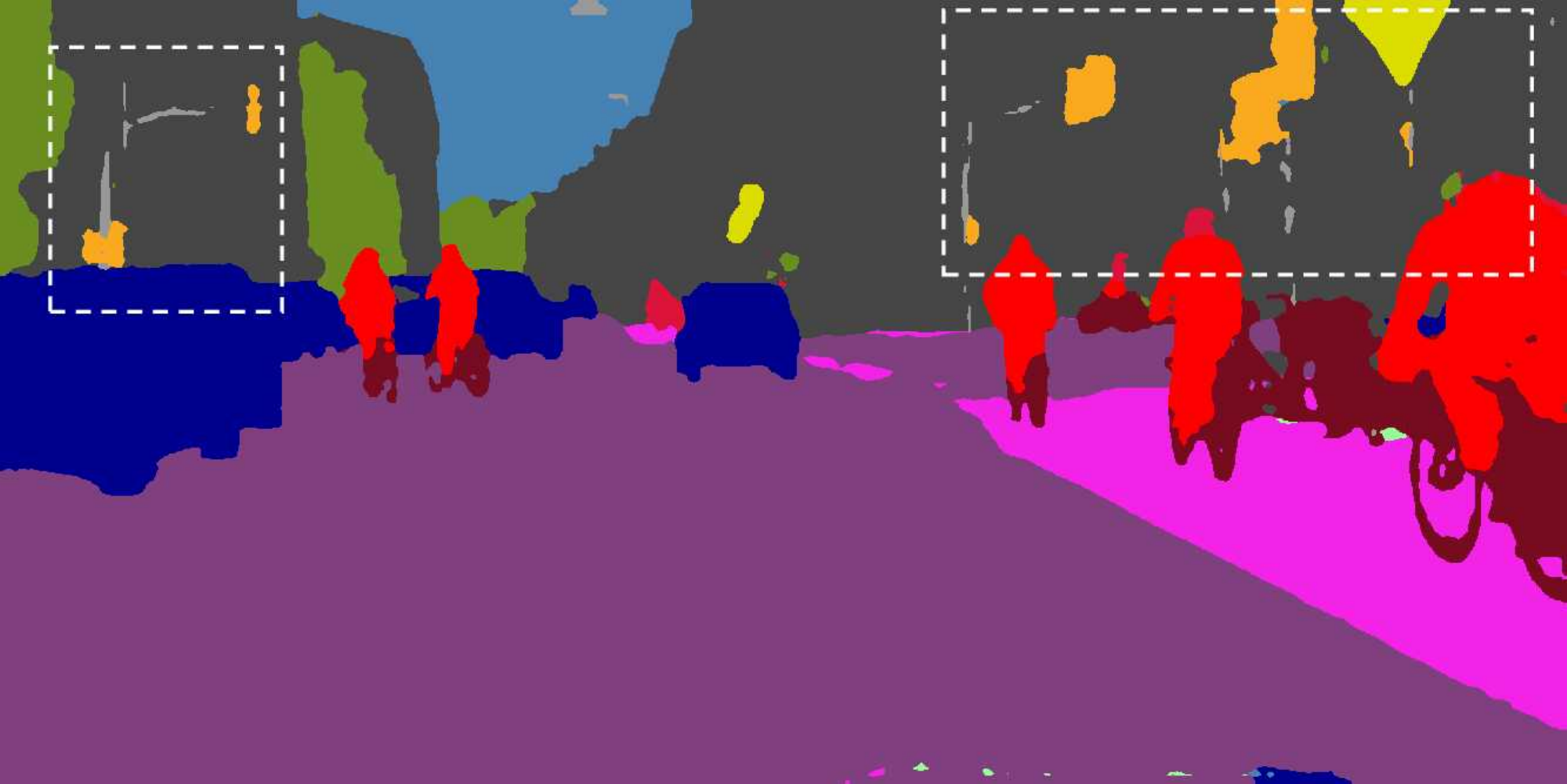}
    \end{subfigure}    \begin{subfigure}{.2\textwidth}
        \centering
        \includegraphics[width=.98\linewidth]{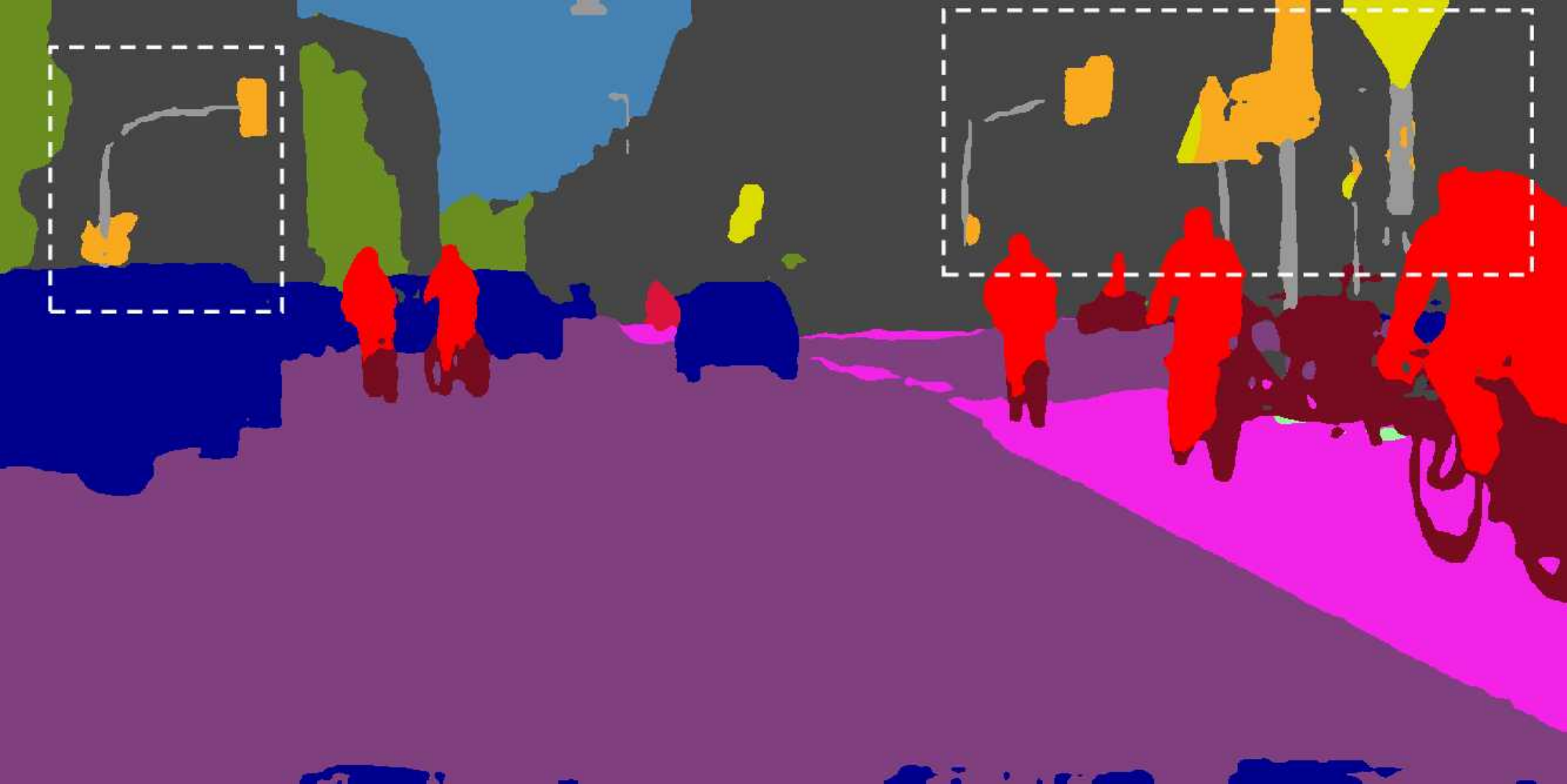}
    \end{subfigure}    \begin{subfigure}{.2\textwidth}
        \centering
        \includegraphics[width=.98\linewidth]{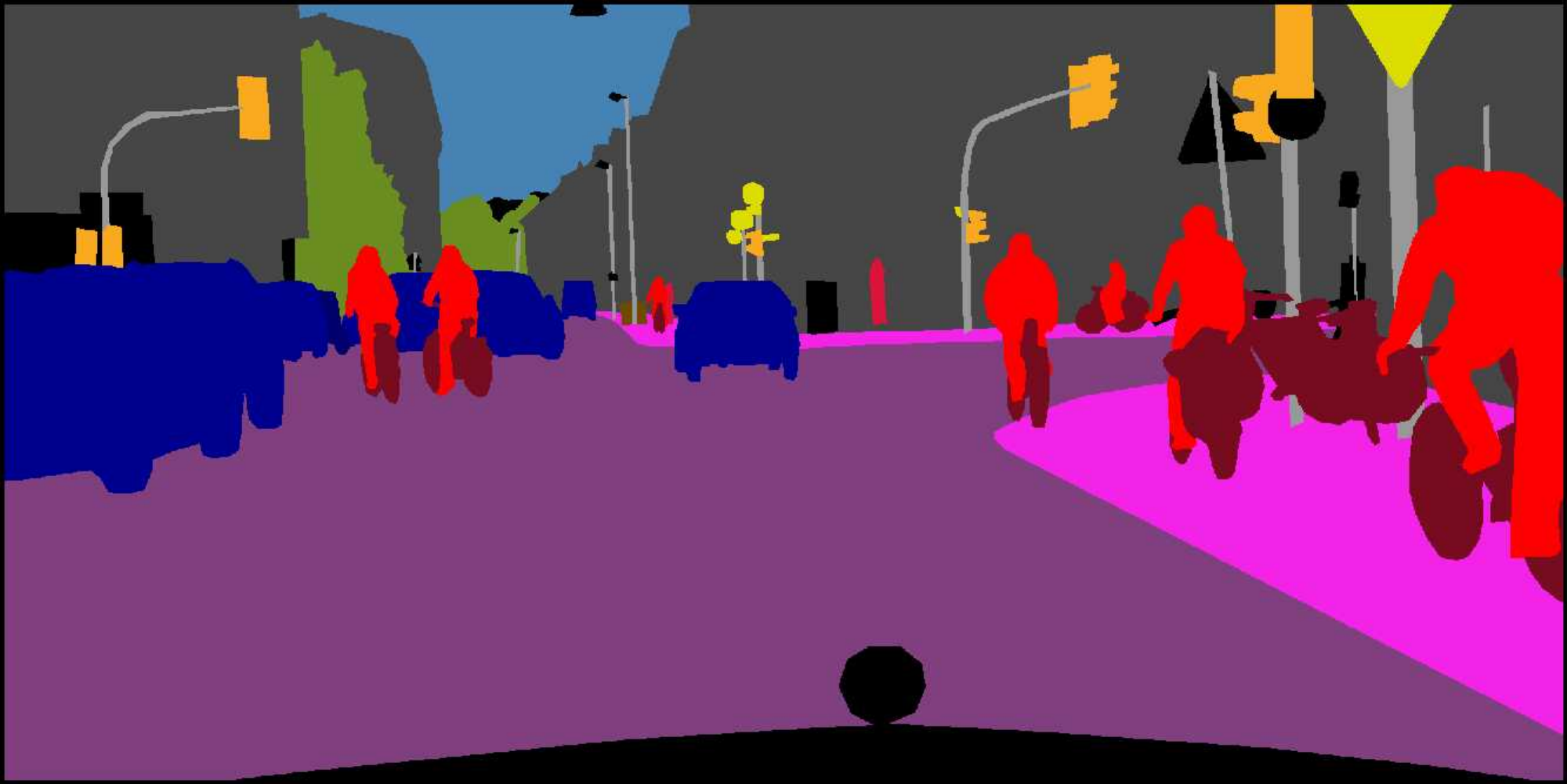}
    \end{subfigure}}
\makebox[\linewidth][c]{    \begin{subfigure}{.2\textwidth}
        \centering
        \includegraphics[width=.98\linewidth]{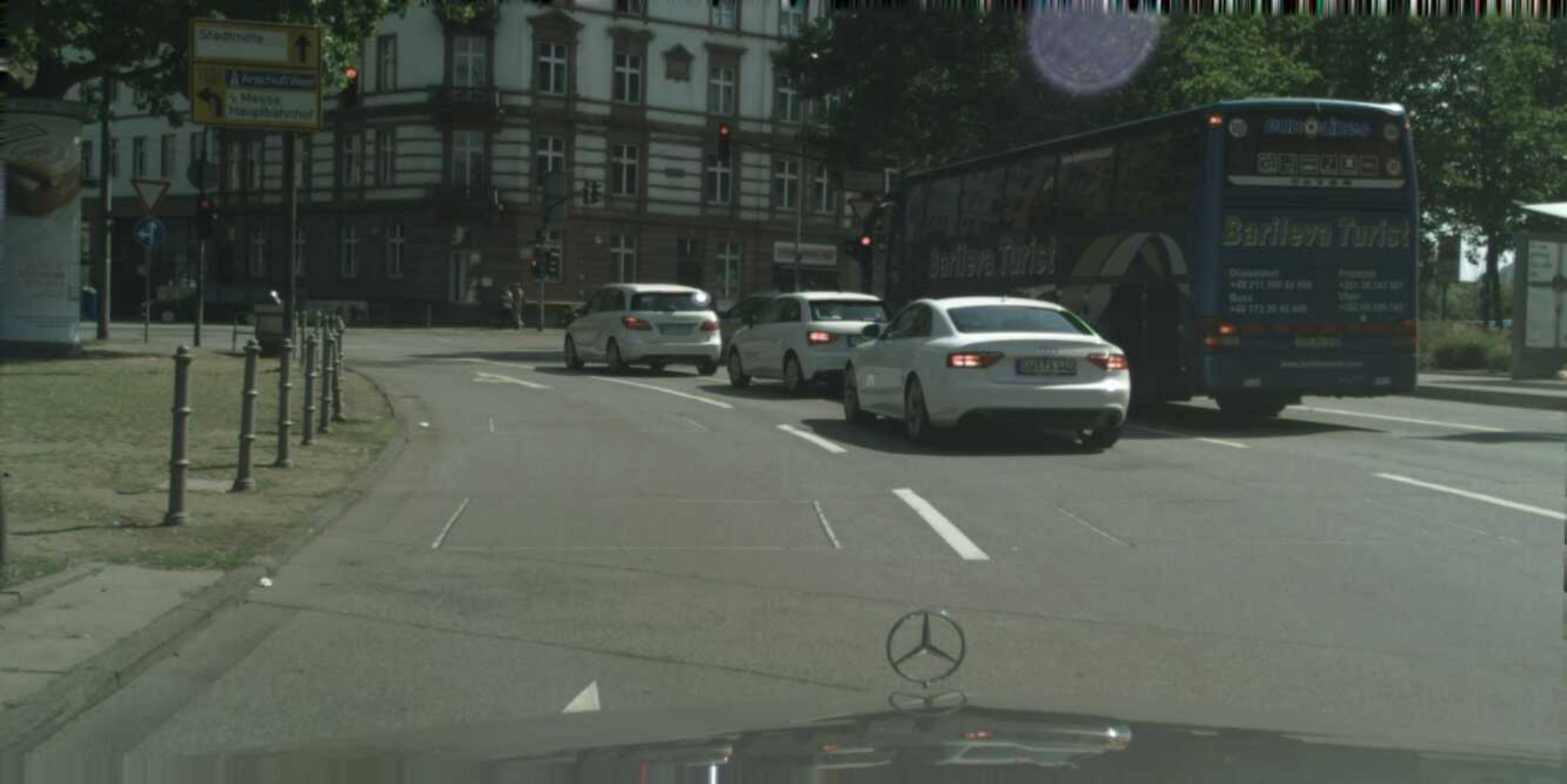}
    \end{subfigure}    \begin{subfigure}{.2\textwidth}
        \centering
        \includegraphics[width=.98\linewidth]{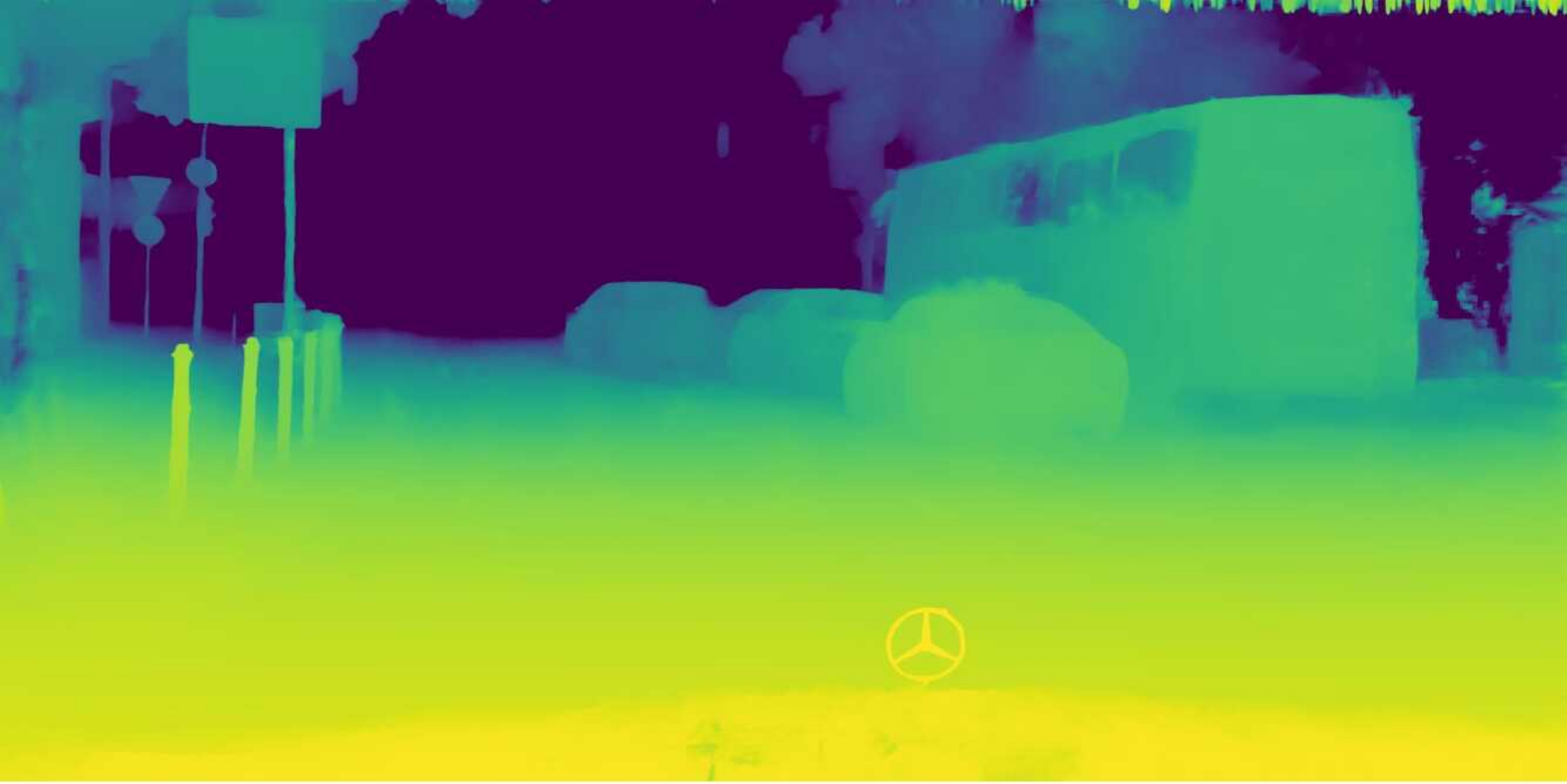}
    \end{subfigure}    \begin{subfigure}{.2\textwidth}
        \centering
        \includegraphics[width=.98\linewidth]{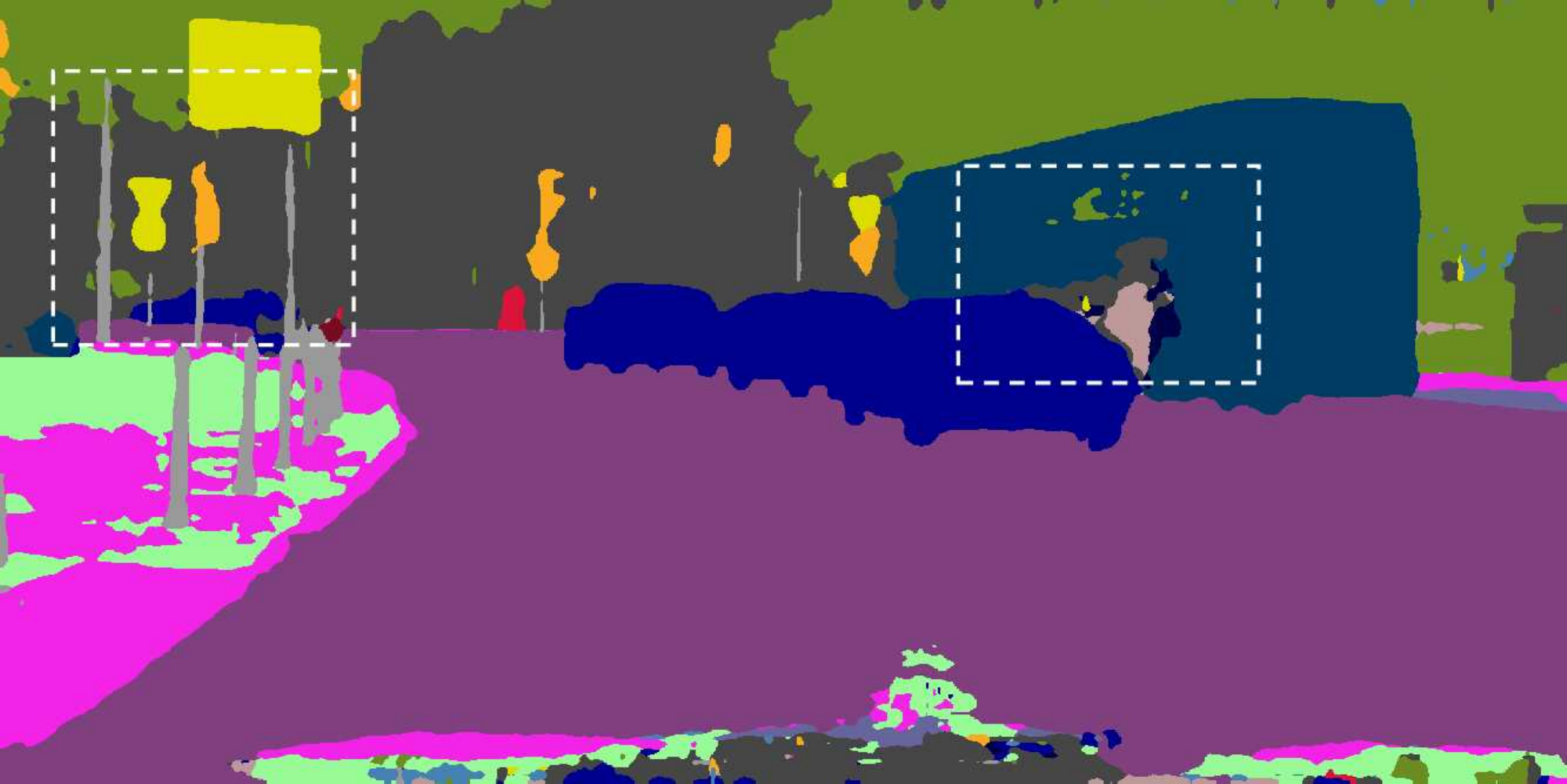}
    \end{subfigure}    \begin{subfigure}{.2\textwidth}
        \centering
        \includegraphics[width=.98\linewidth]{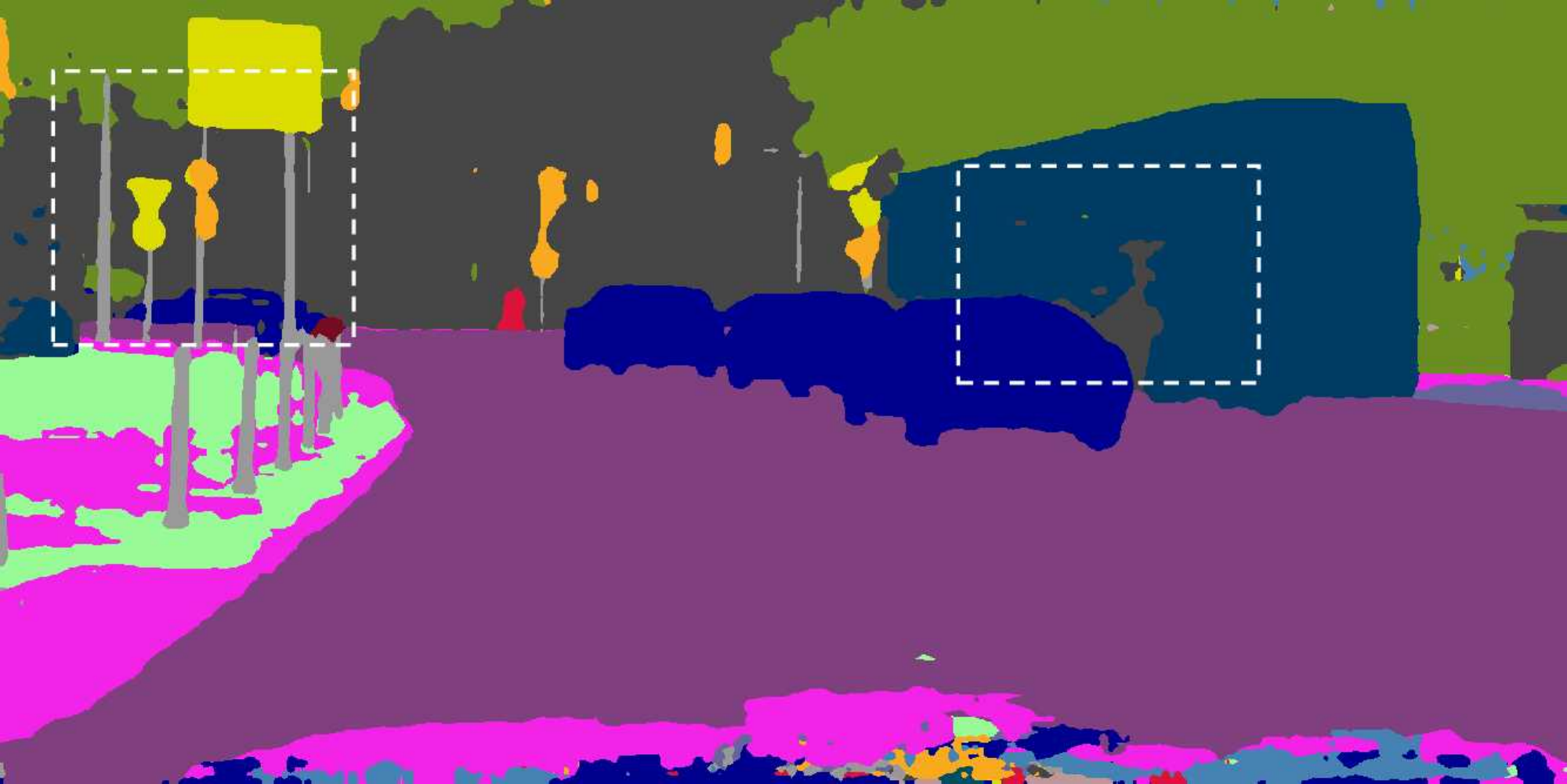}
    \end{subfigure}    \begin{subfigure}{.2\textwidth}
        \centering
        \includegraphics[width=.98\linewidth]{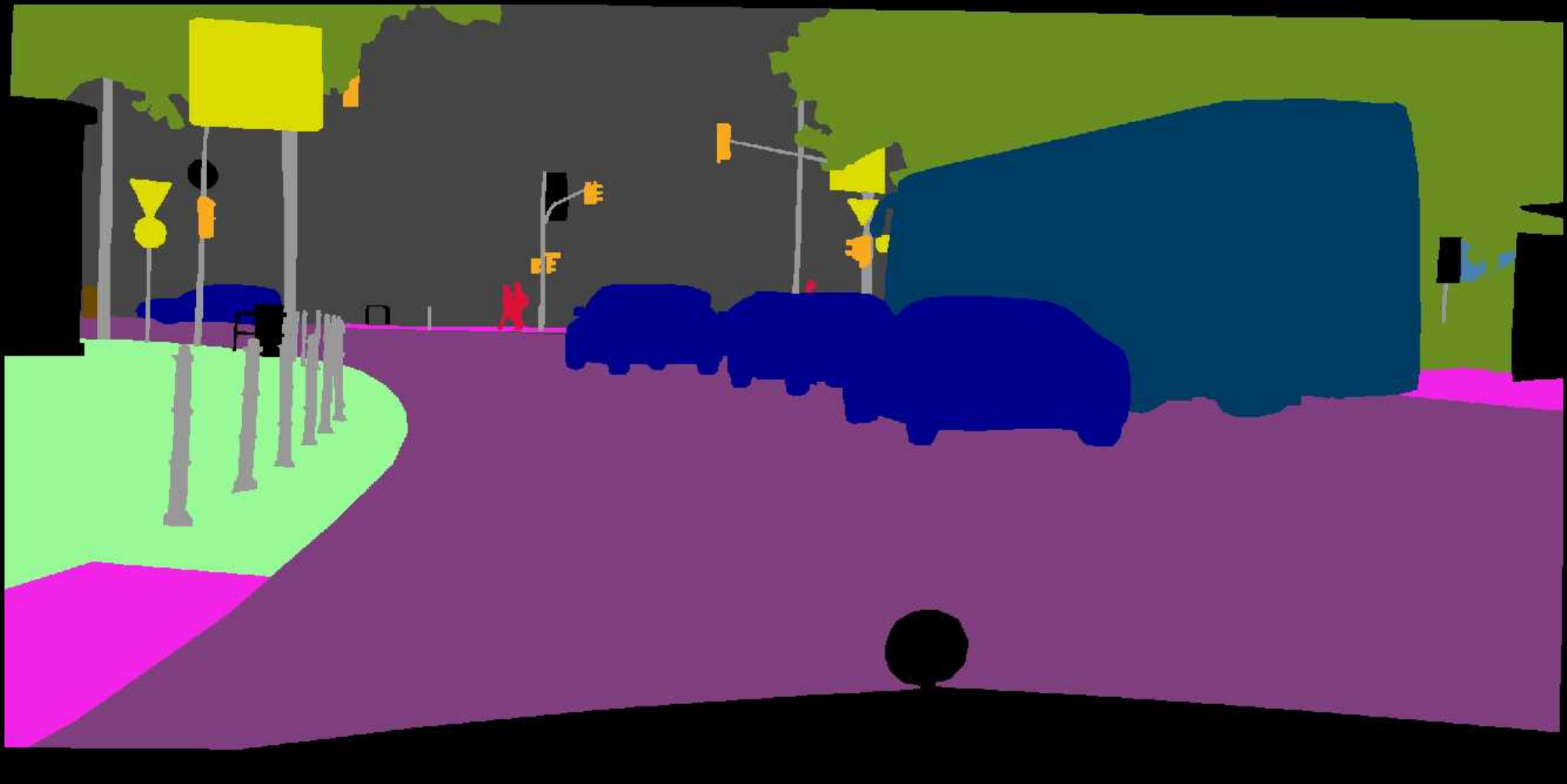}
    \end{subfigure}}
\makebox[\linewidth][c]{    \begin{subfigure}{.2\textwidth}
        \centering
        \includegraphics[width=.98\linewidth]{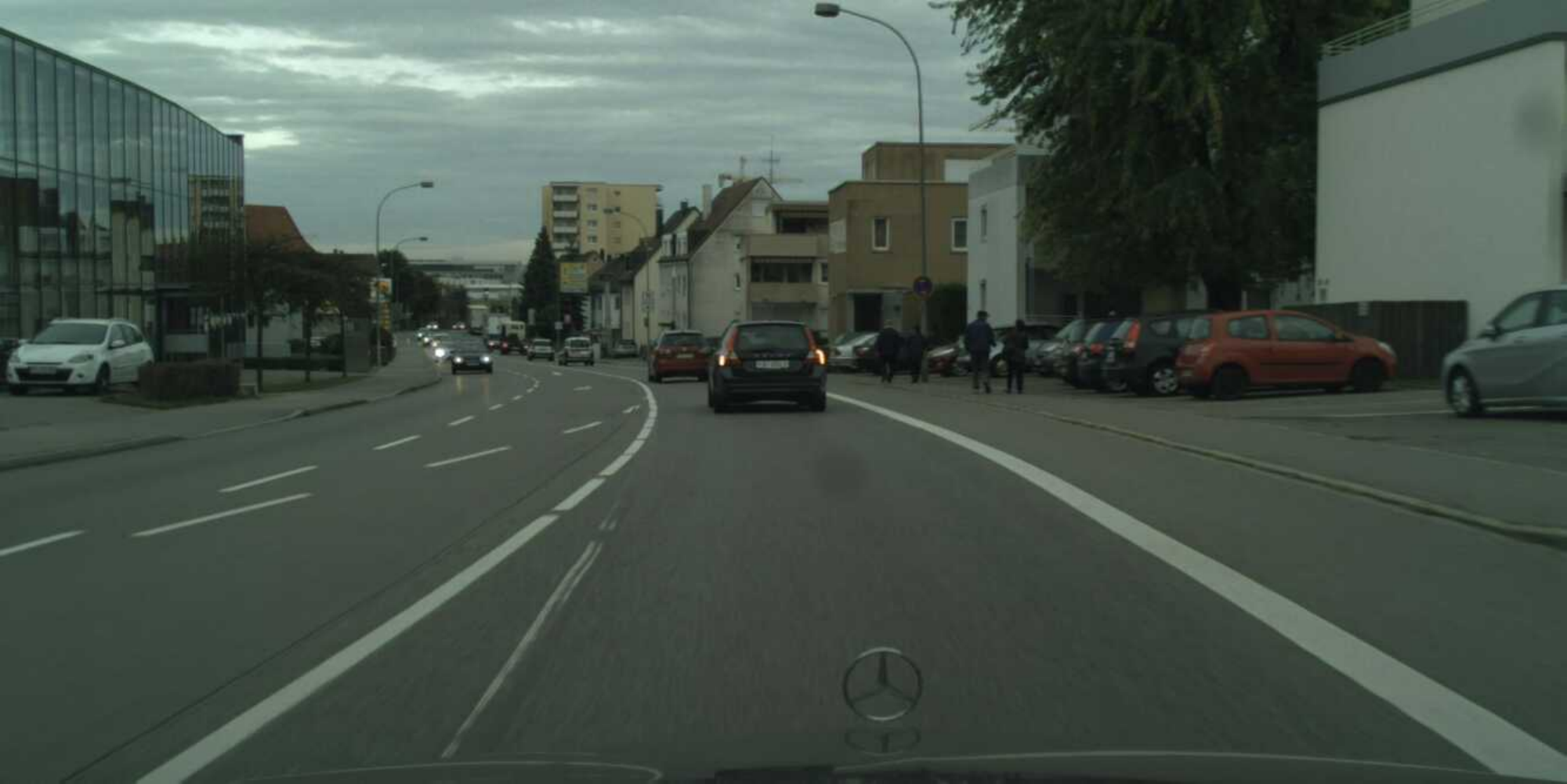}
    \end{subfigure}    \begin{subfigure}{.2\textwidth}
        \centering
        \includegraphics[width=.98\linewidth]{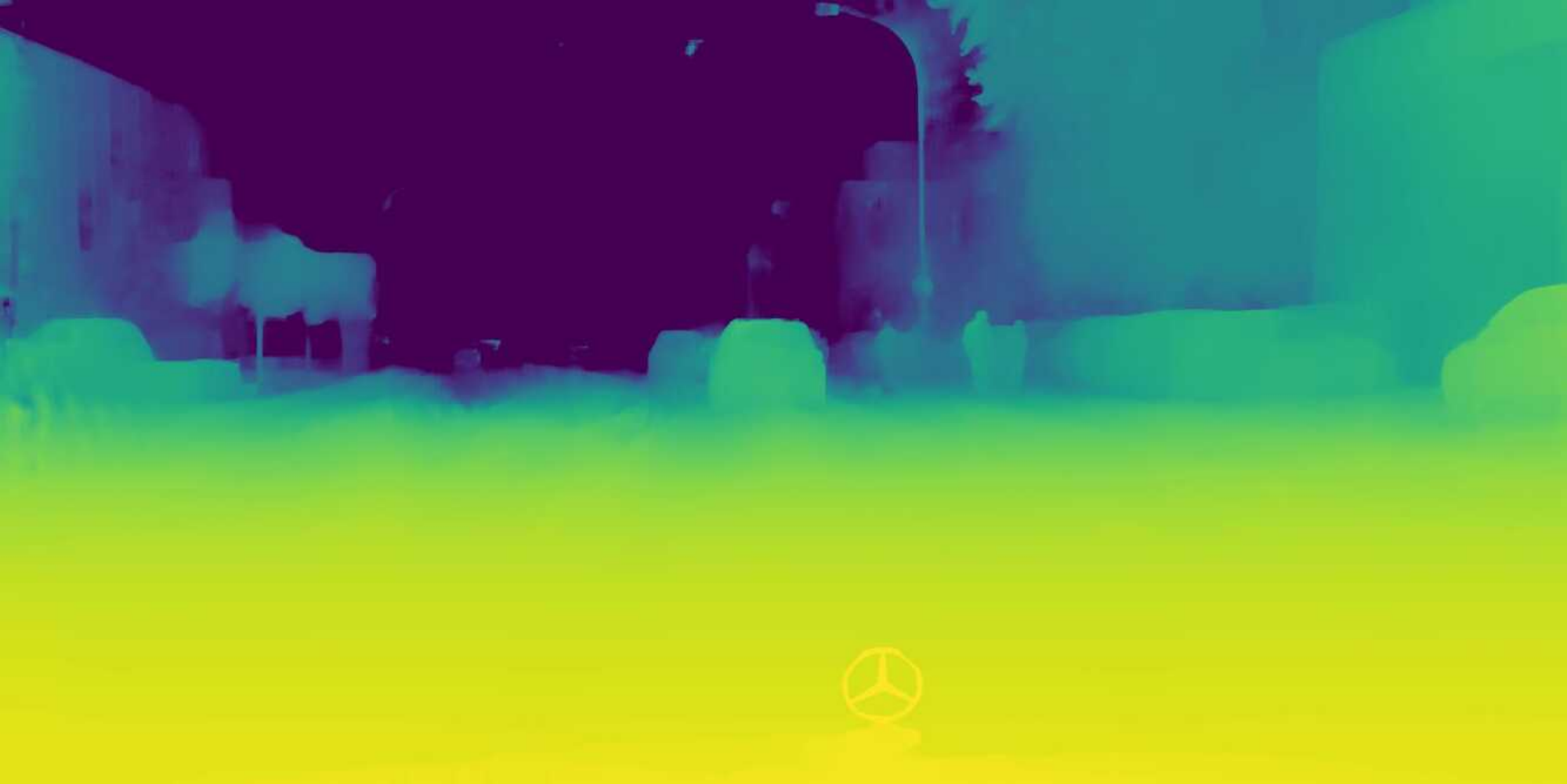}
    \end{subfigure}    \begin{subfigure}{.2\textwidth}
        \centering
        \includegraphics[width=.98\linewidth]{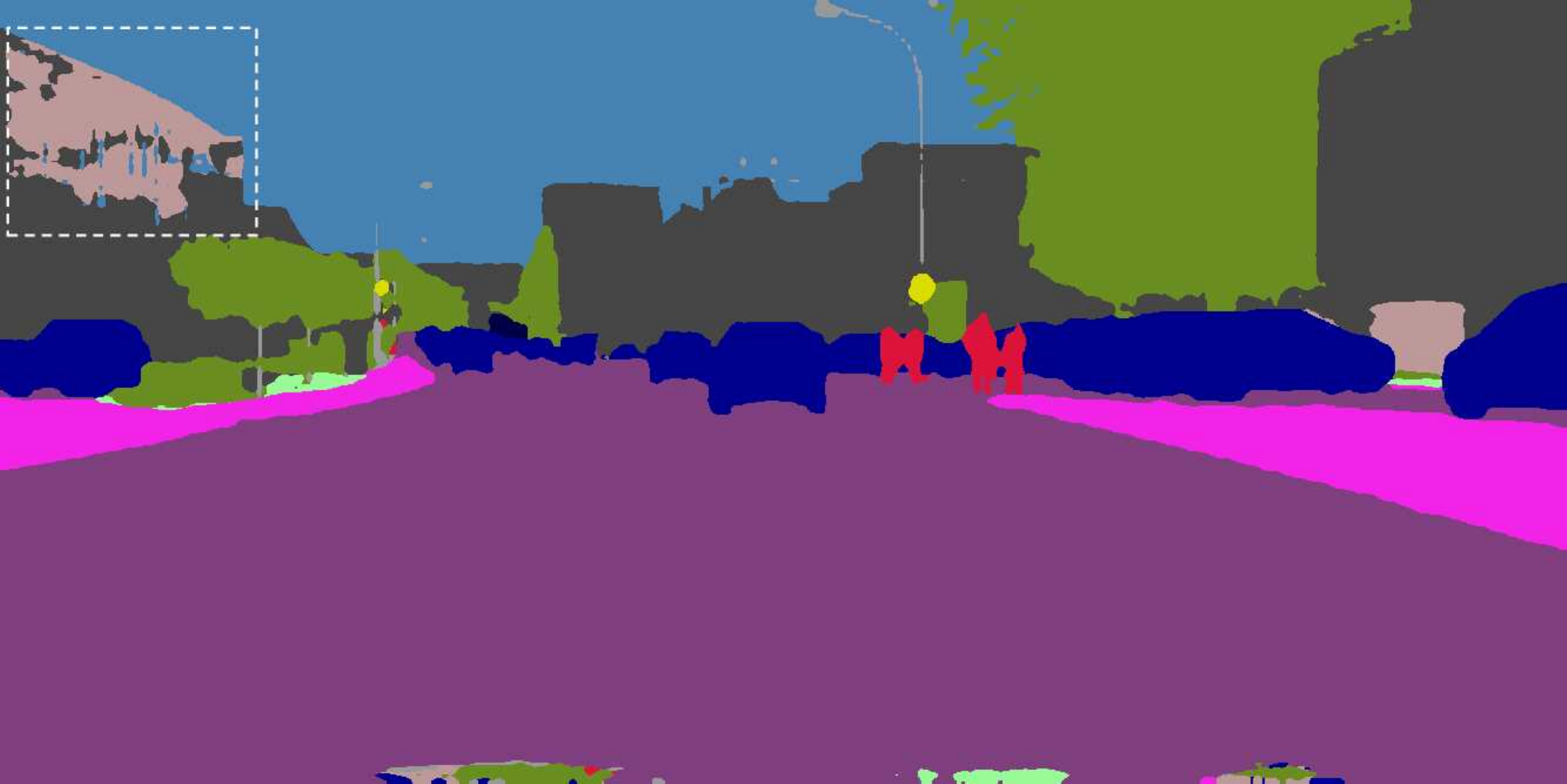}
    \end{subfigure}    \begin{subfigure}{.2\textwidth}
        \centering
        \includegraphics[width=.98\linewidth]{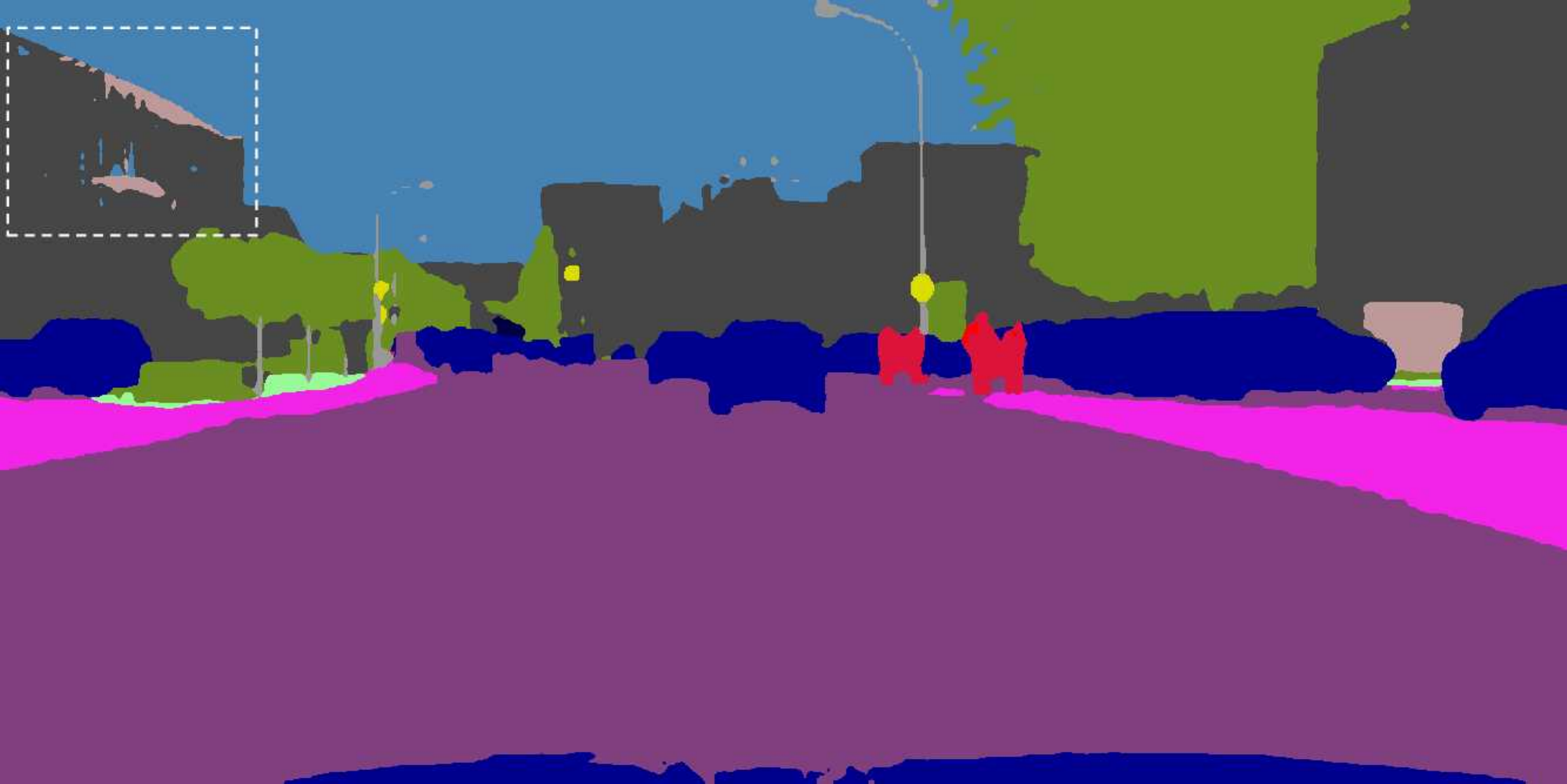}
    \end{subfigure}    \begin{subfigure}{.2\textwidth}
        \centering
        \includegraphics[width=.98\linewidth]{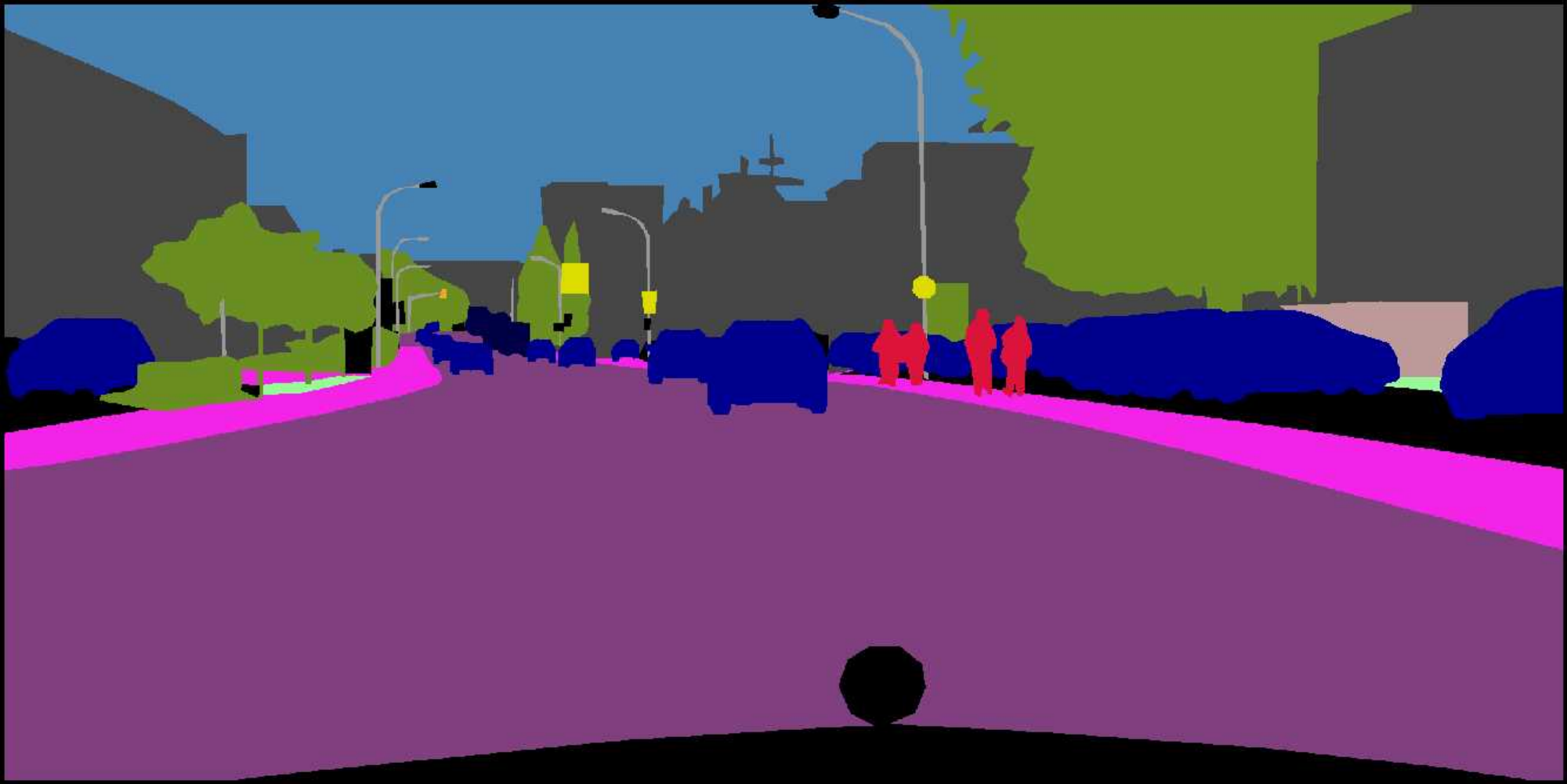}
    \end{subfigure}}
\makebox[\linewidth][c]{    \begin{subfigure}{.2\textwidth}
        \centering
        \includegraphics[width=.98\linewidth]{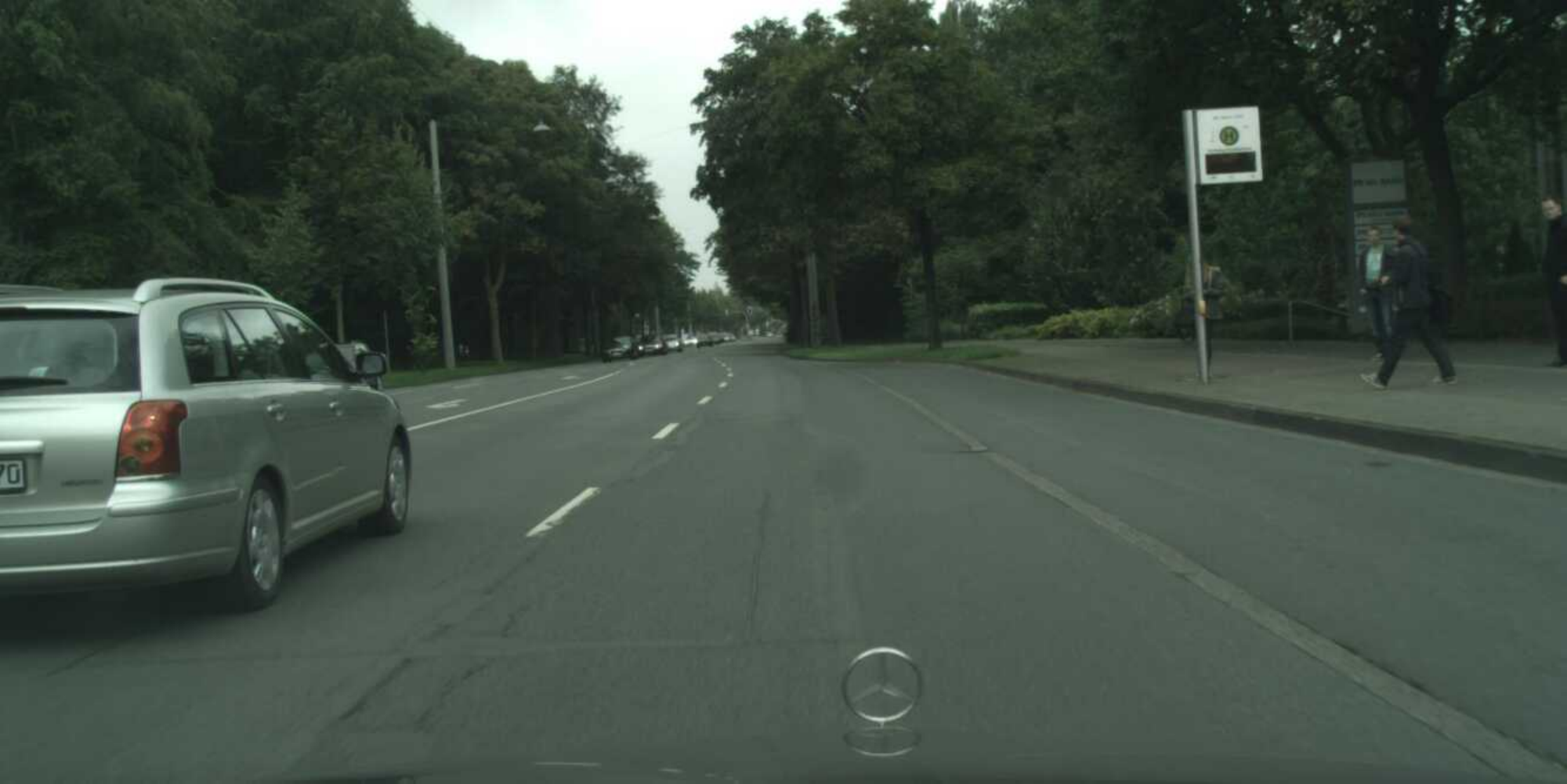}
    \end{subfigure}    \begin{subfigure}{.2\textwidth}
        \centering
        \includegraphics[width=.98\linewidth]{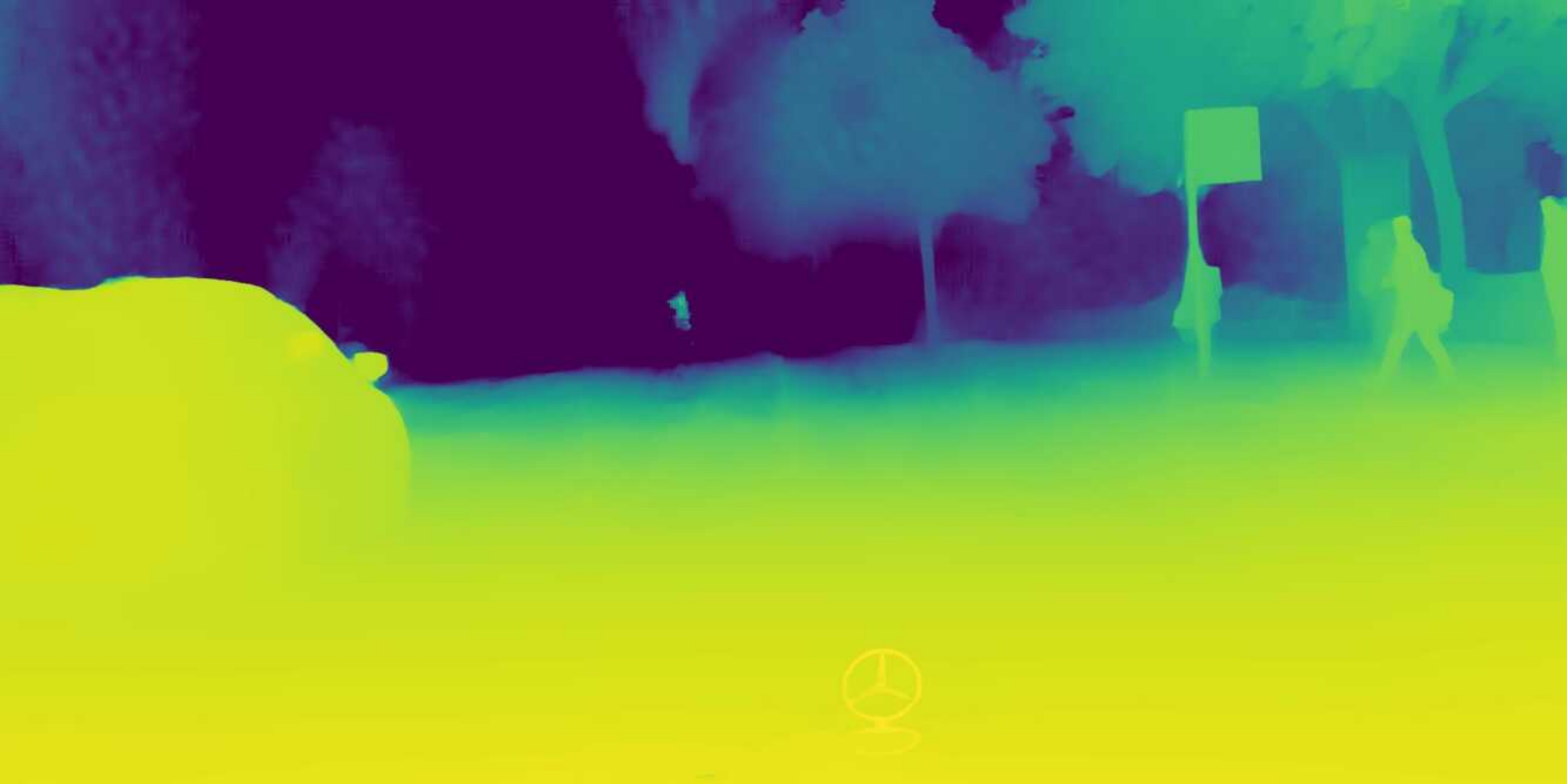}
    \end{subfigure}    \begin{subfigure}{.2\textwidth}
        \centering
        \includegraphics[width=.98\linewidth]{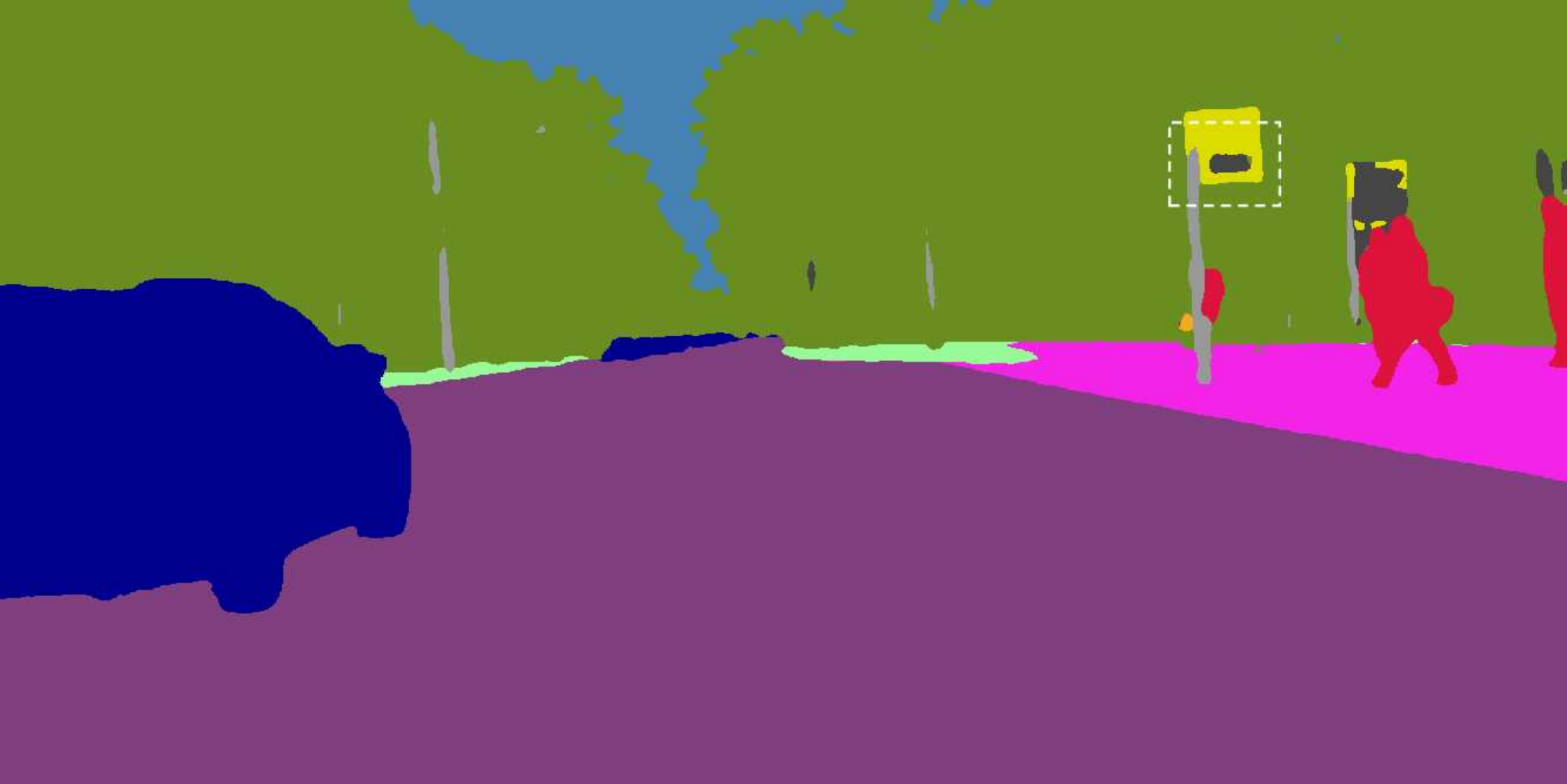}
    \end{subfigure}    \begin{subfigure}{.2\textwidth}
        \centering
        \includegraphics[width=.98\linewidth]{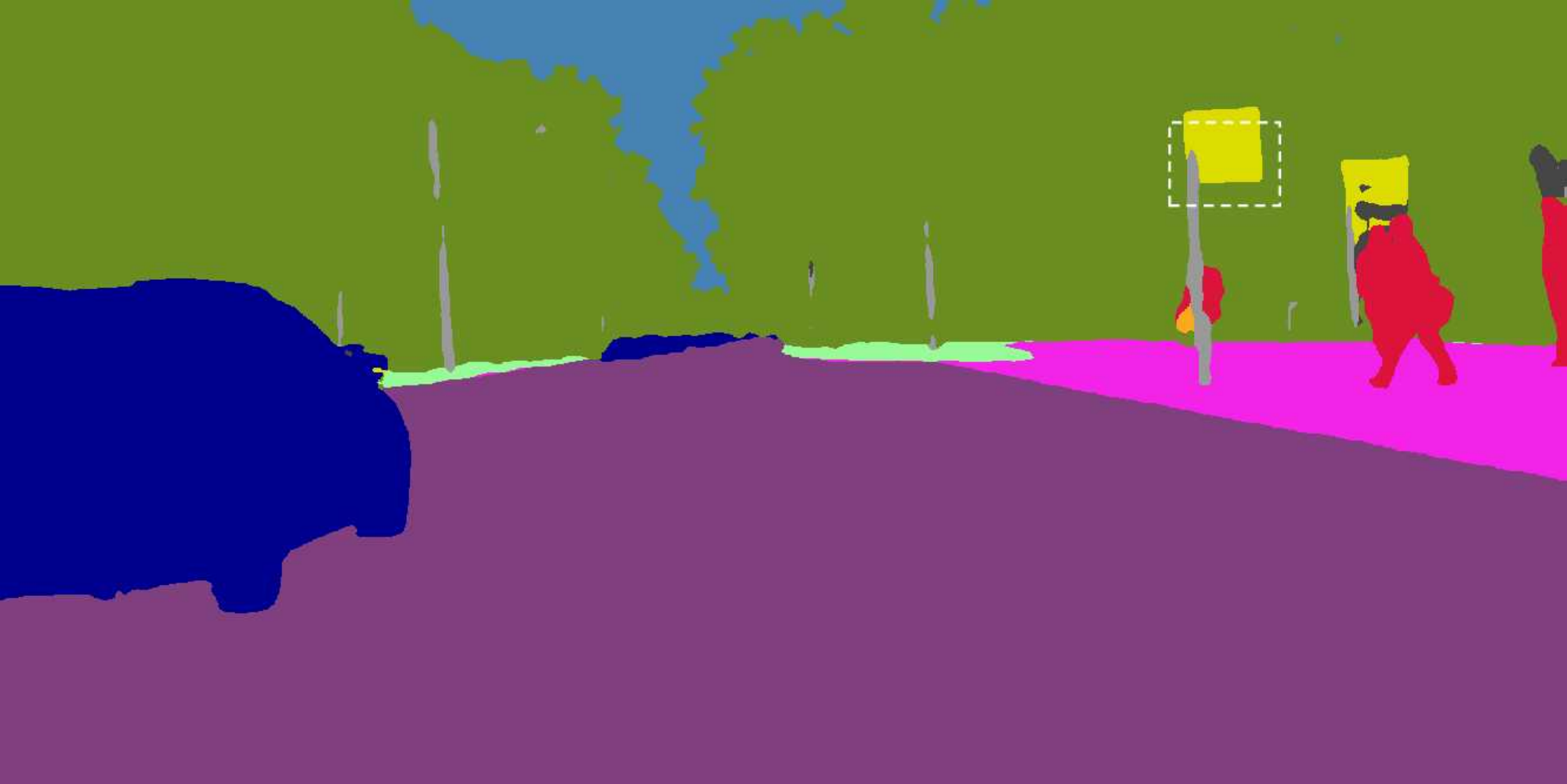}
    \end{subfigure}    \begin{subfigure}{.2\textwidth}
        \centering
        \includegraphics[width=.98\linewidth]{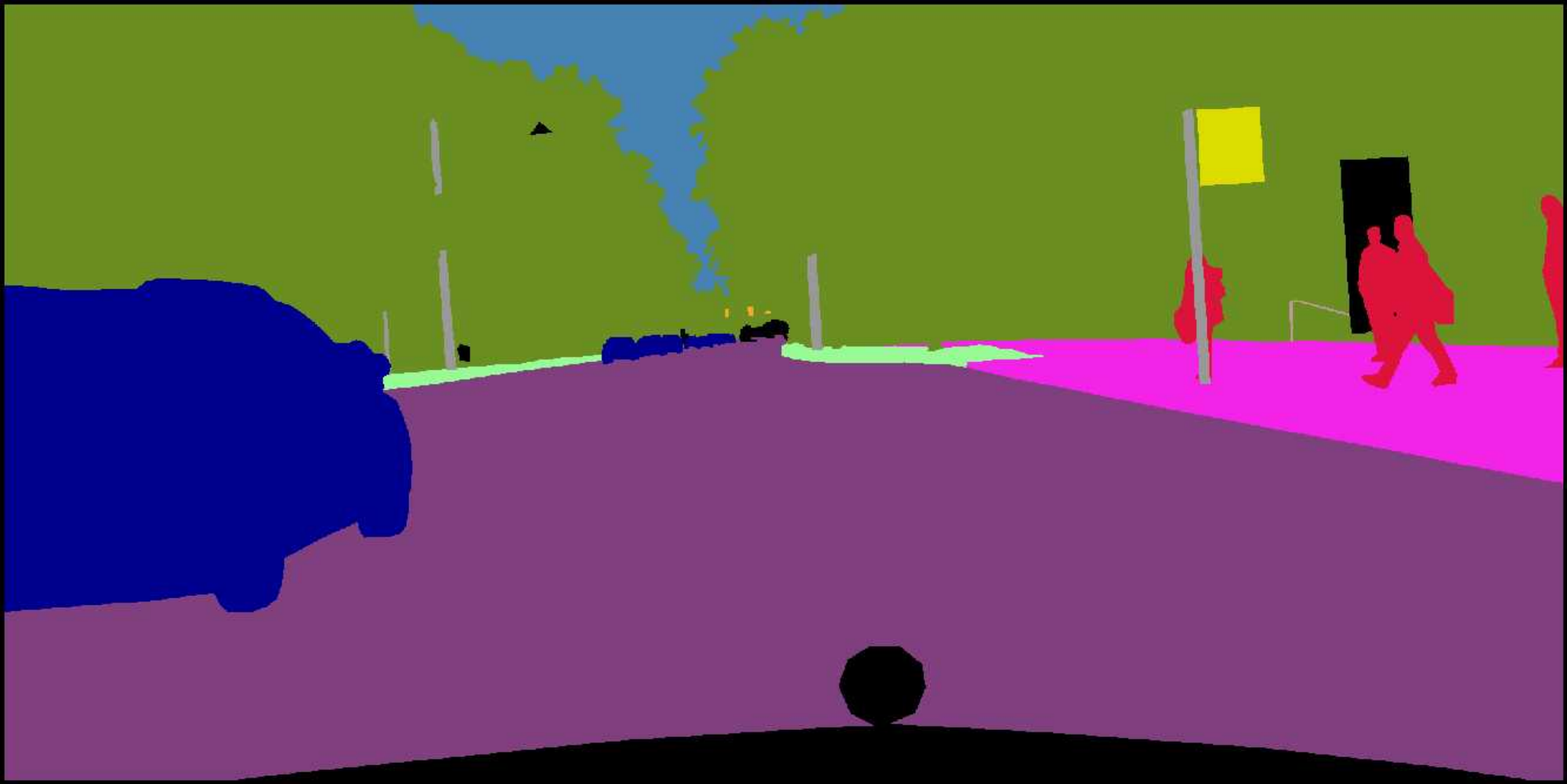}
    \end{subfigure}}
\makebox[\linewidth][c]{    \begin{subfigure}{.2\textwidth}
        \centering
        \includegraphics[width=.98\linewidth]{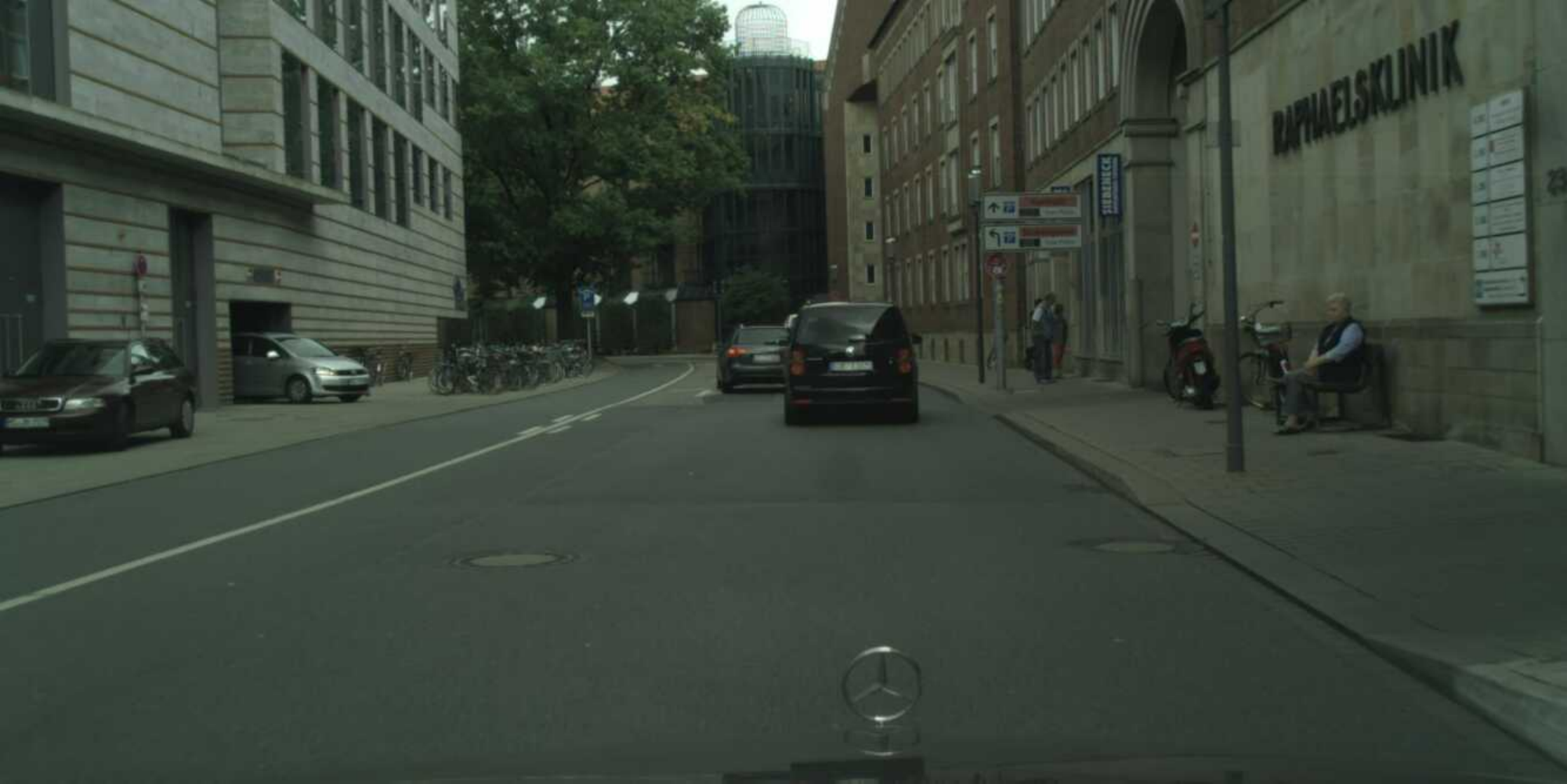}
    \end{subfigure}    \begin{subfigure}{.2\textwidth}
        \centering
        \includegraphics[width=.98\linewidth]{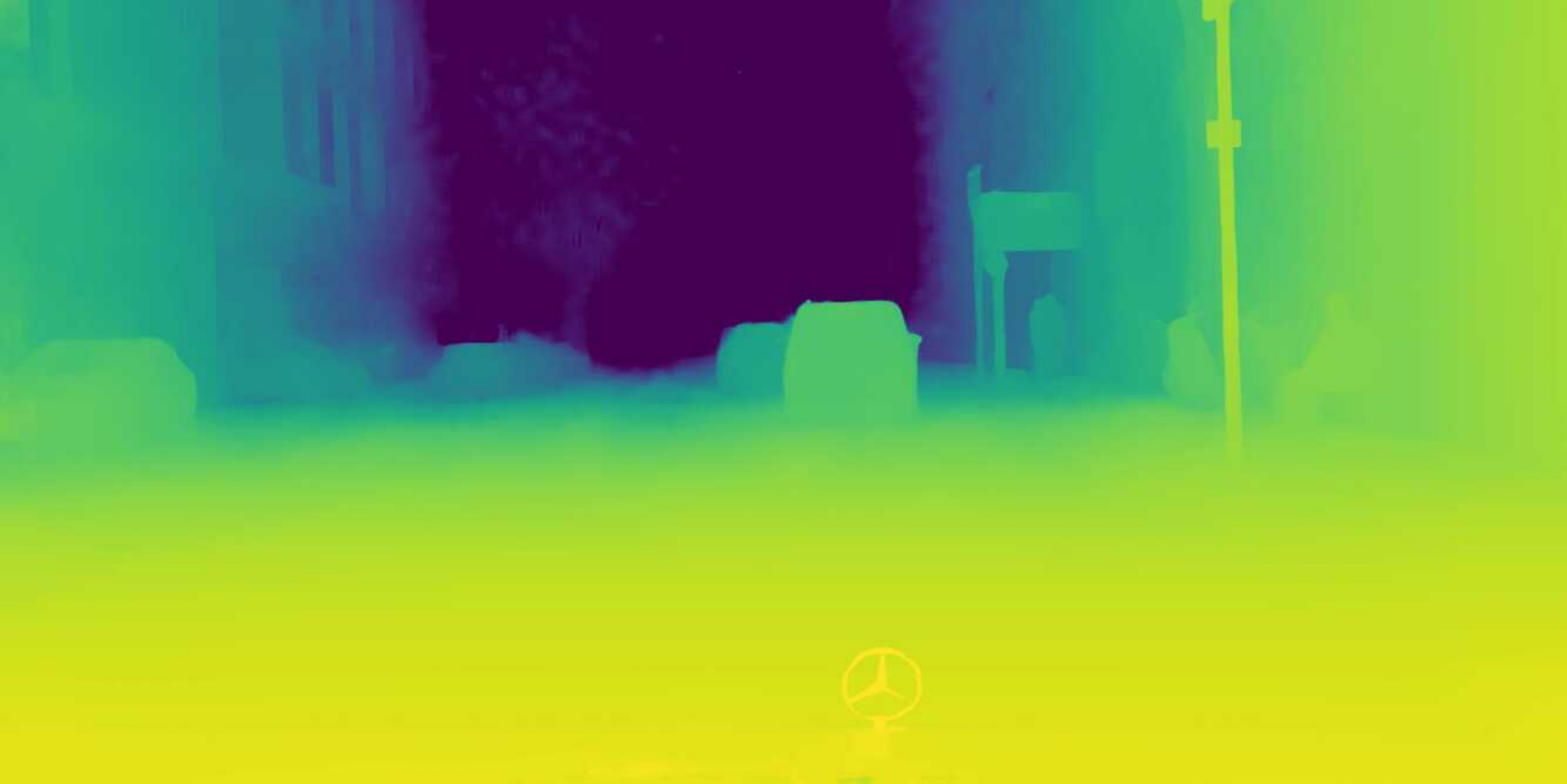}
    \end{subfigure}    \begin{subfigure}{.2\textwidth}
        \centering
        \includegraphics[width=.98\linewidth]{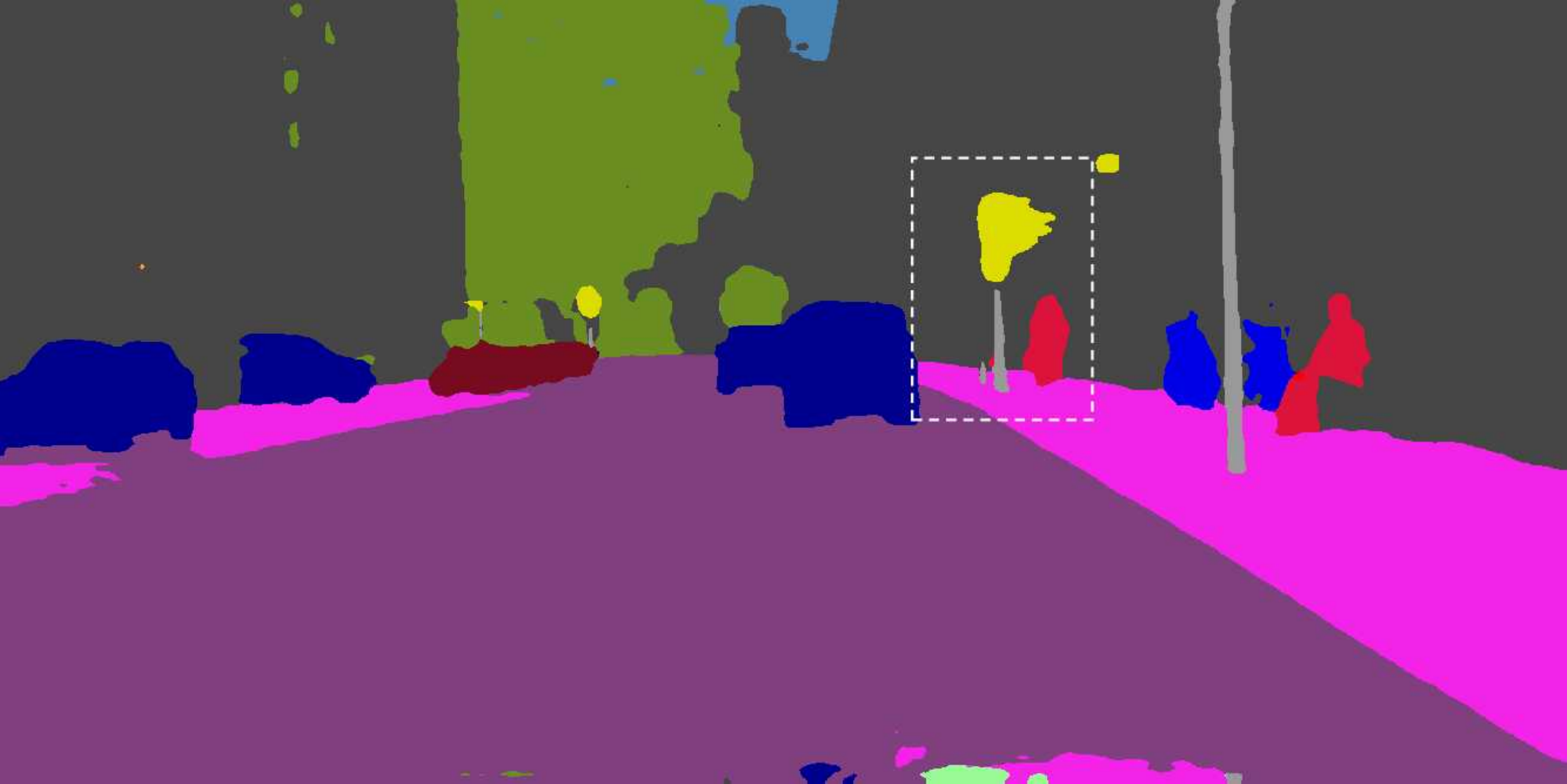}
    \end{subfigure}    \begin{subfigure}{.2\textwidth}
        \centering
        \includegraphics[width=.98\linewidth]{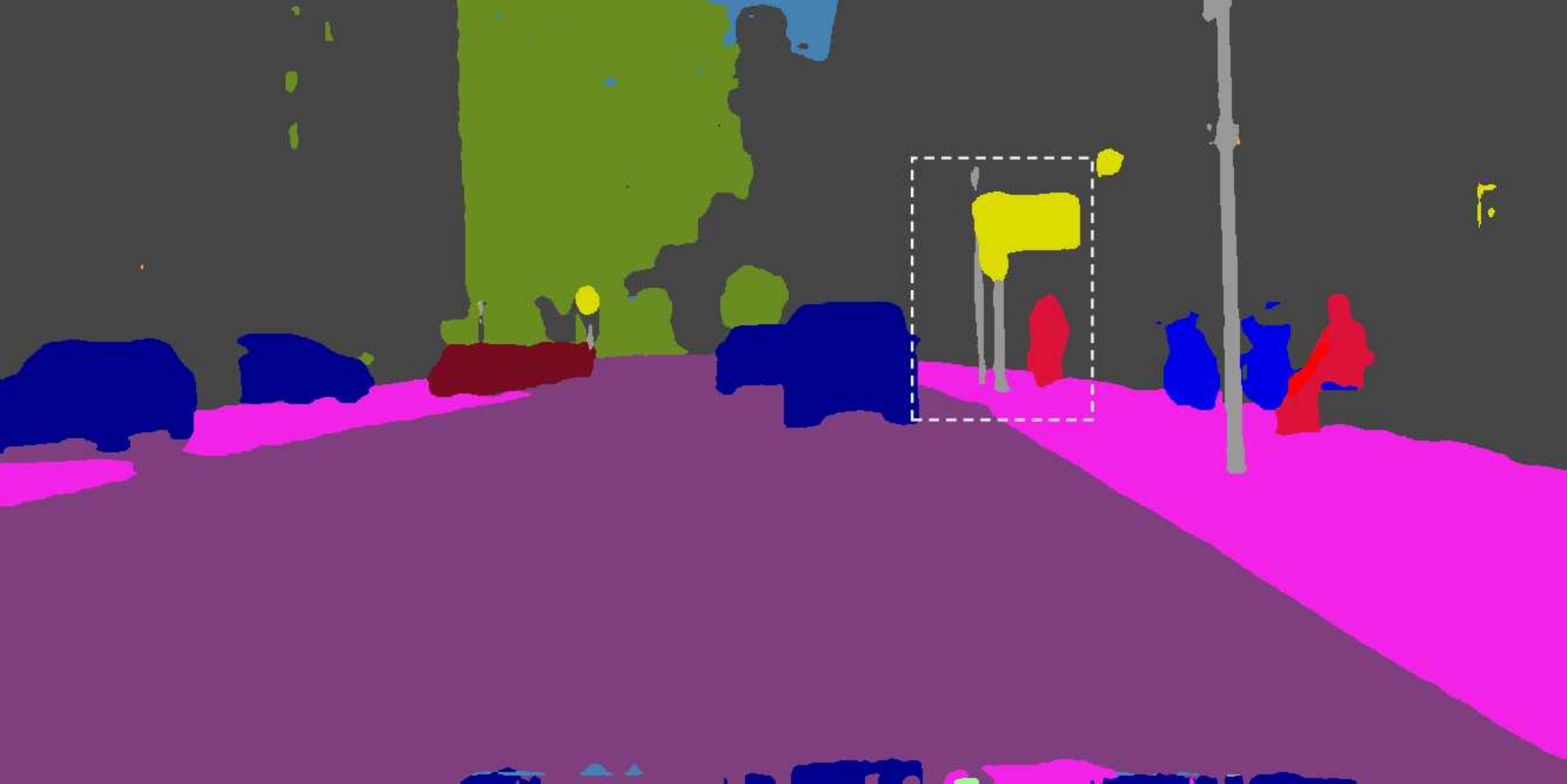}
    \end{subfigure}    \begin{subfigure}{.2\textwidth}
        \centering
        \includegraphics[width=.98\linewidth]{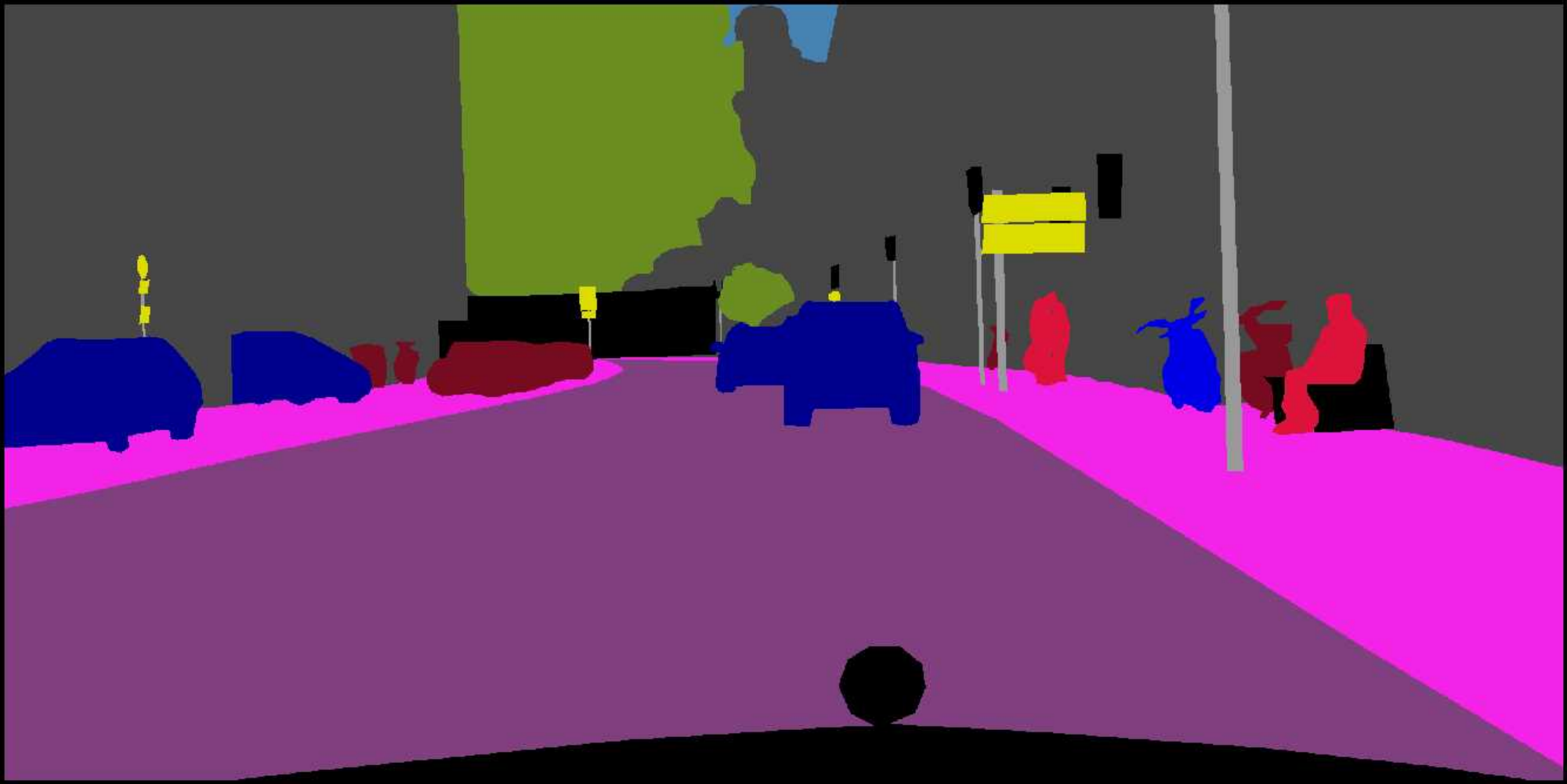}
    \end{subfigure}}
\makebox[\linewidth][c]{    \begin{subfigure}{.2\textwidth}
        \centering
        \includegraphics[width=.98\linewidth]{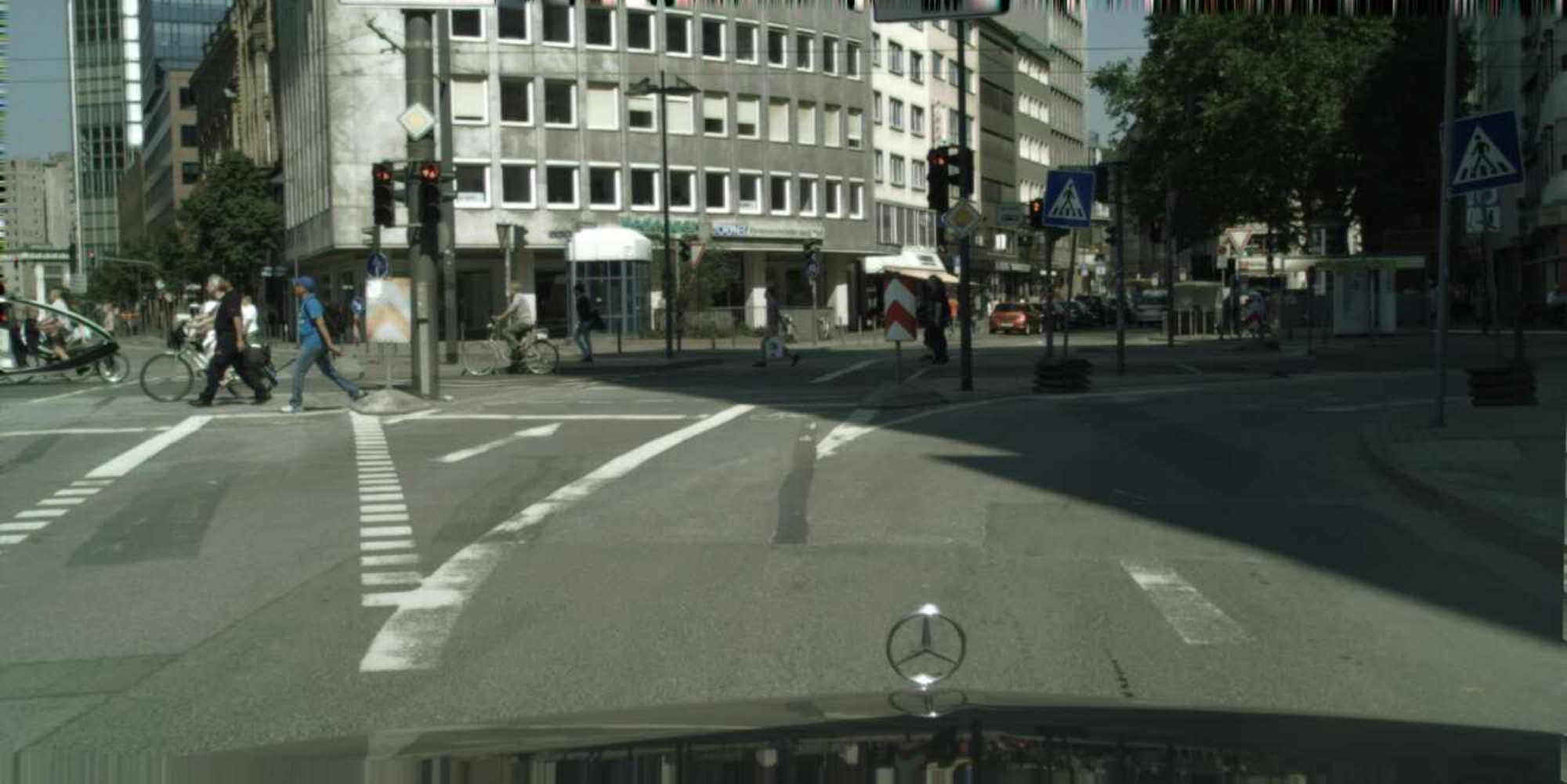}
        \caption*{Target Image}
    \end{subfigure}    \begin{subfigure}{.2\textwidth}
        \centering
        \includegraphics[width=.98\linewidth]{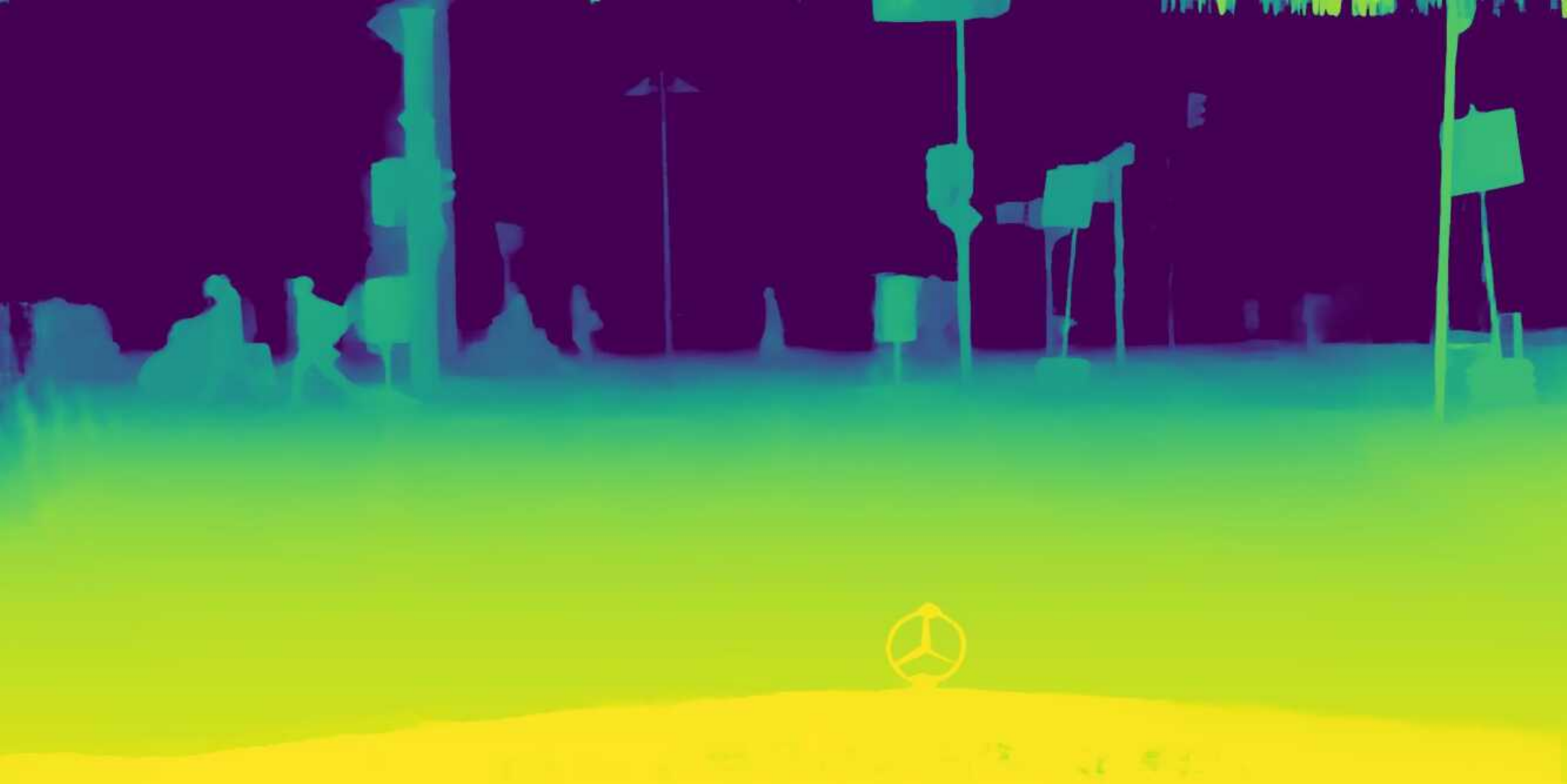}
        \caption*{Estimated Depth}
    \end{subfigure}    \begin{subfigure}{.2\textwidth}
        \centering
        \includegraphics[width=.98\linewidth]{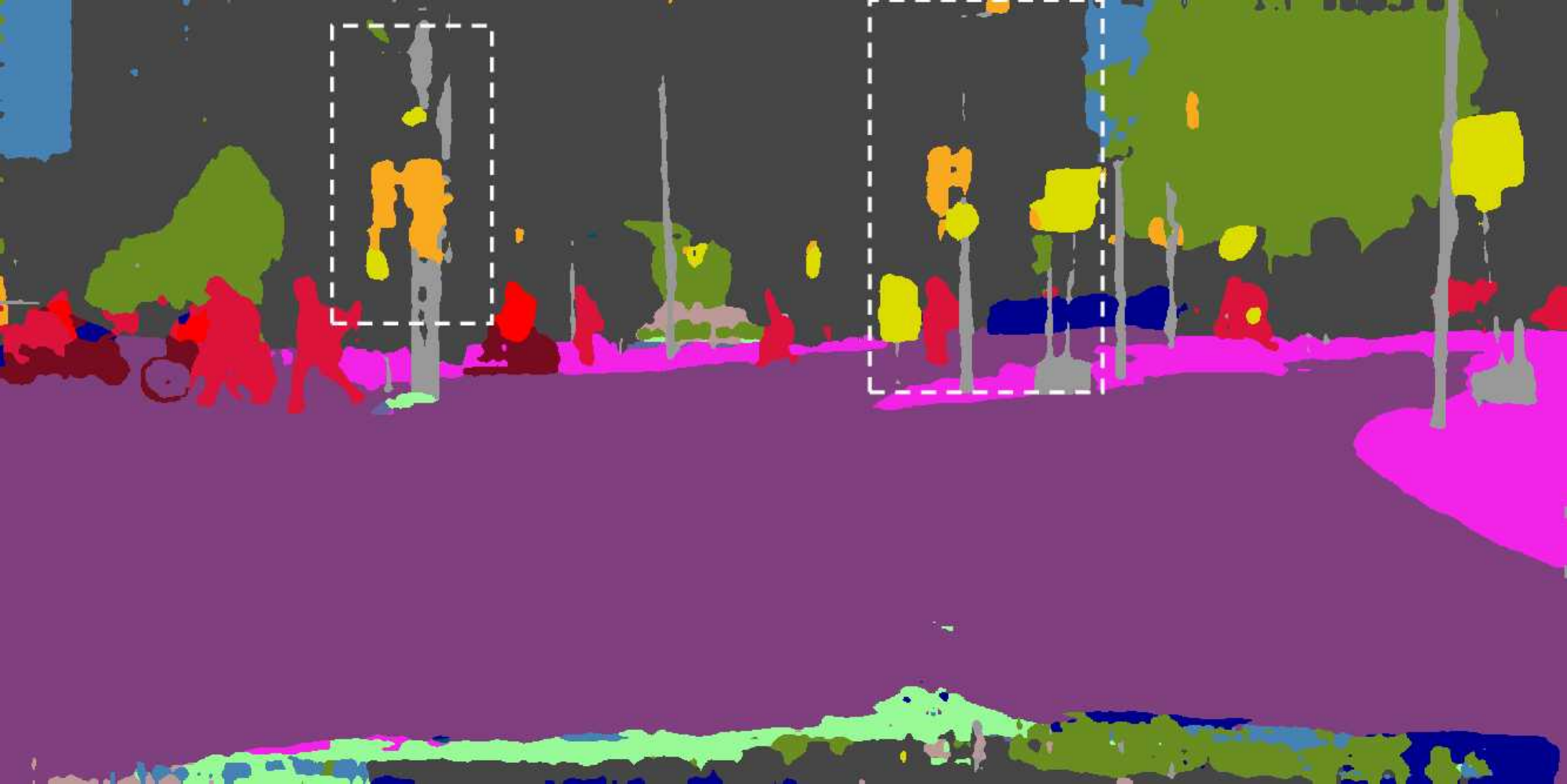}
        \caption*{DAFormer~\cite{DAFormer}}
    \end{subfigure}    \begin{subfigure}{.2\textwidth}
        \centering
        \includegraphics[width=.98\linewidth]{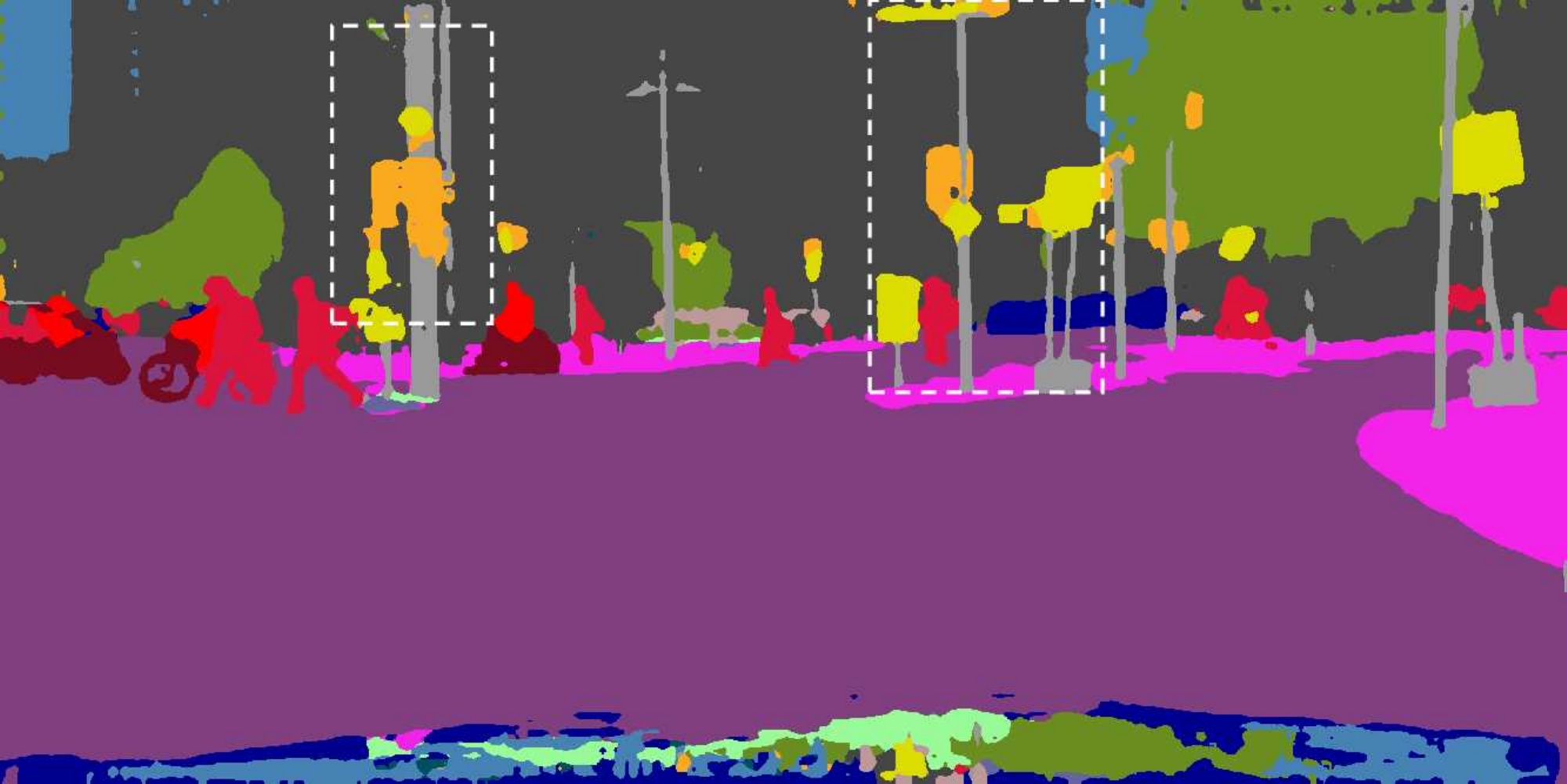}
        \caption*{\method\ (ours)}
    \end{subfigure}    \begin{subfigure}{.2\textwidth}
        \centering
        \includegraphics[width=.98\linewidth]{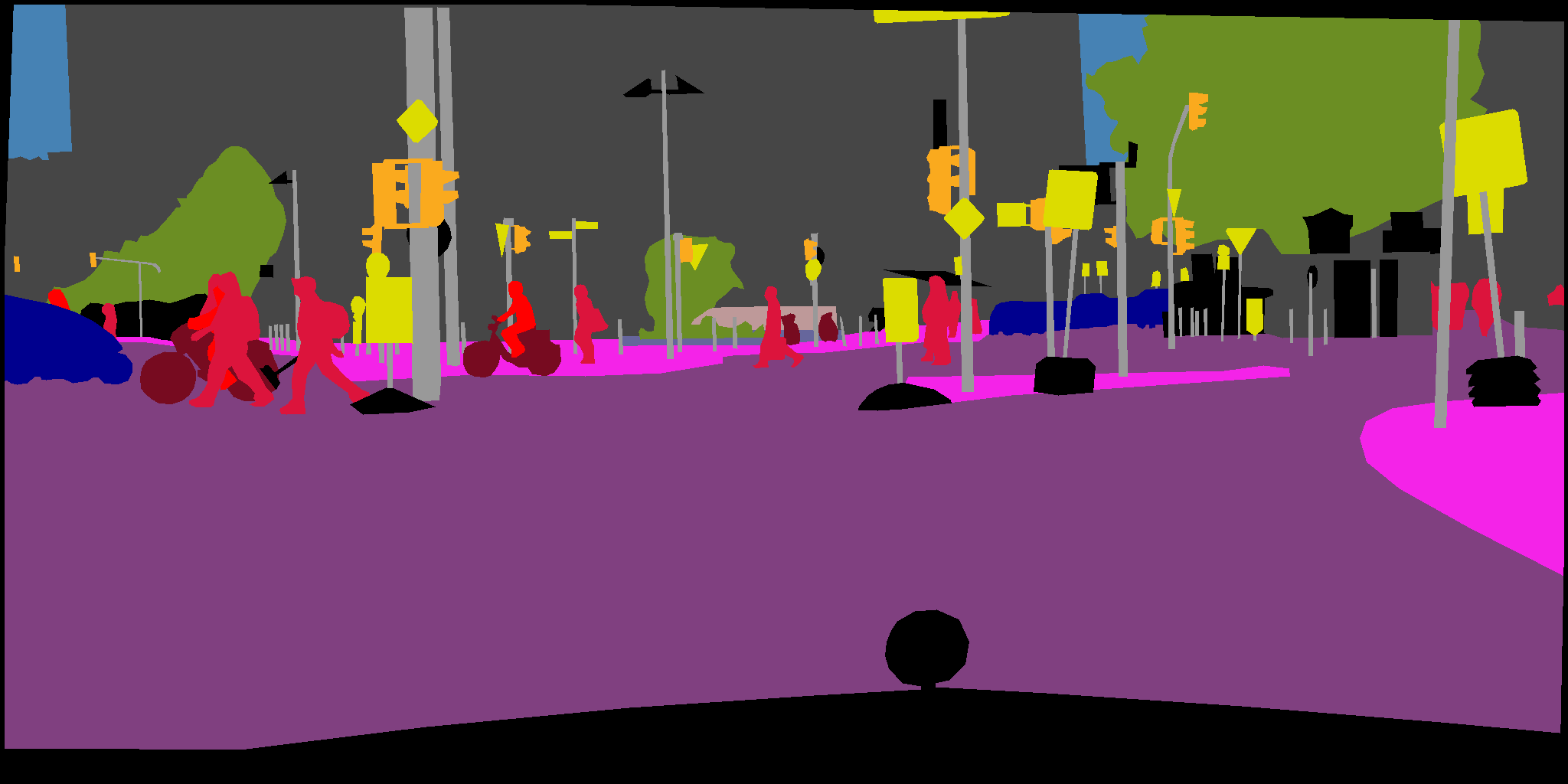}
        \caption*{Ground Truth}
    \end{subfigure}}

\centering

\caption{\textbf{Qualitative results using DAFormer~\cite{DAFormer}.}
In rows 1, 3, and 4, depth cues correct mislabeled patches as part of larger structures.    
Rows 5 and 6 obtain more accurate \textit{traffic sign} segmentation with consistent depth and depth discontinuities, respectively.
In rows 1, 2, and 7, correct segmentation of thin \textit{poles} is enabled through depth information.
}
\label{fig:qualitative_supp_daf}
\end{figure*}

\begin{figure*}
\makebox[\linewidth][c]{    \begin{subfigure}{\linewidth}
        \centering
        \label{fig:palette}
        \includegraphics[width=\linewidth]{figures/color_palette.png}
    \end{subfigure}}
\makebox[\linewidth][c]{
    \begin{subfigure}{.2\textwidth}
        \centering
        \includegraphics[width=.98\linewidth]{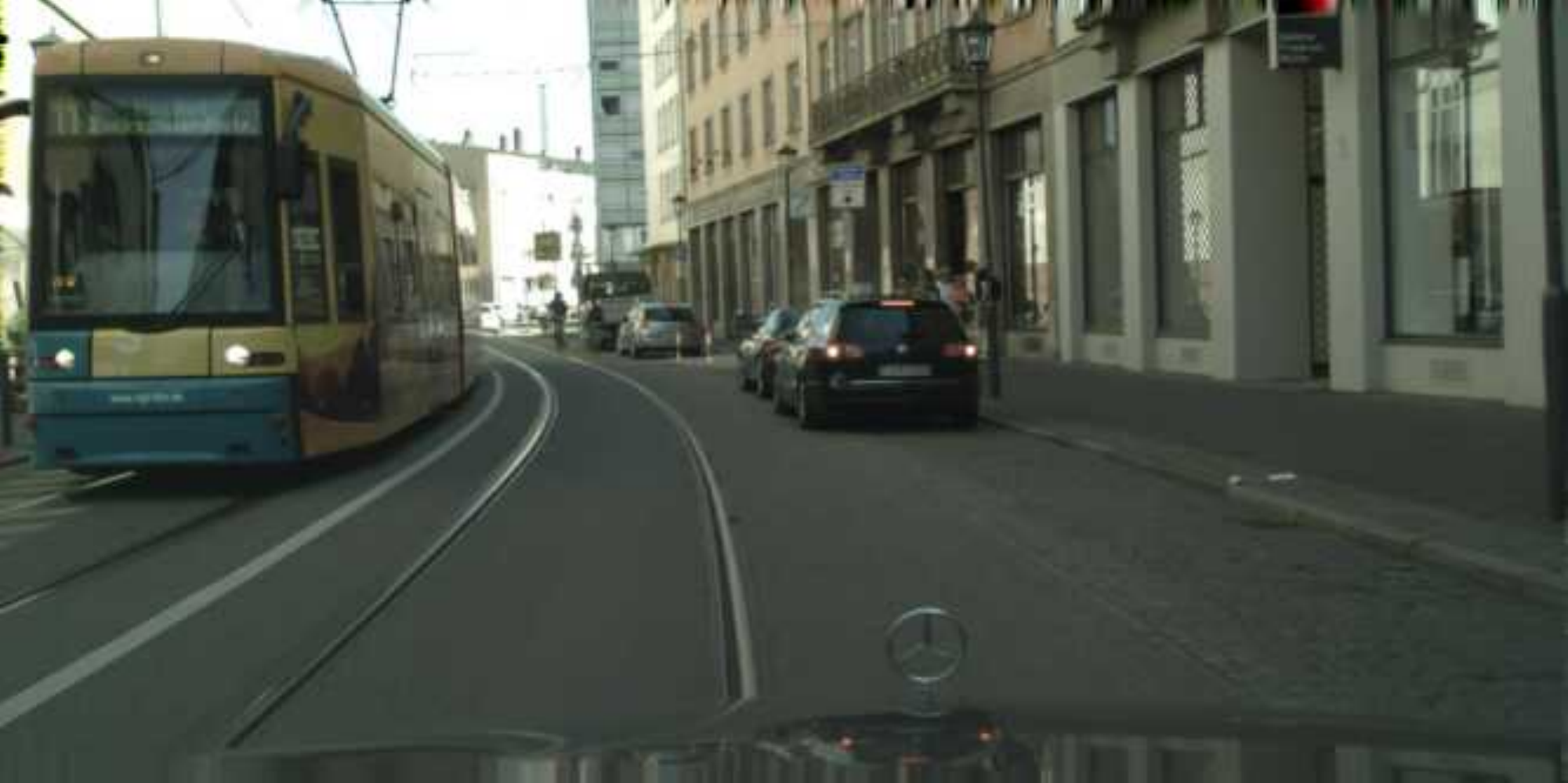}
    \end{subfigure}    \begin{subfigure}{.2\textwidth}
        \centering
        \includegraphics[width=.98\linewidth]{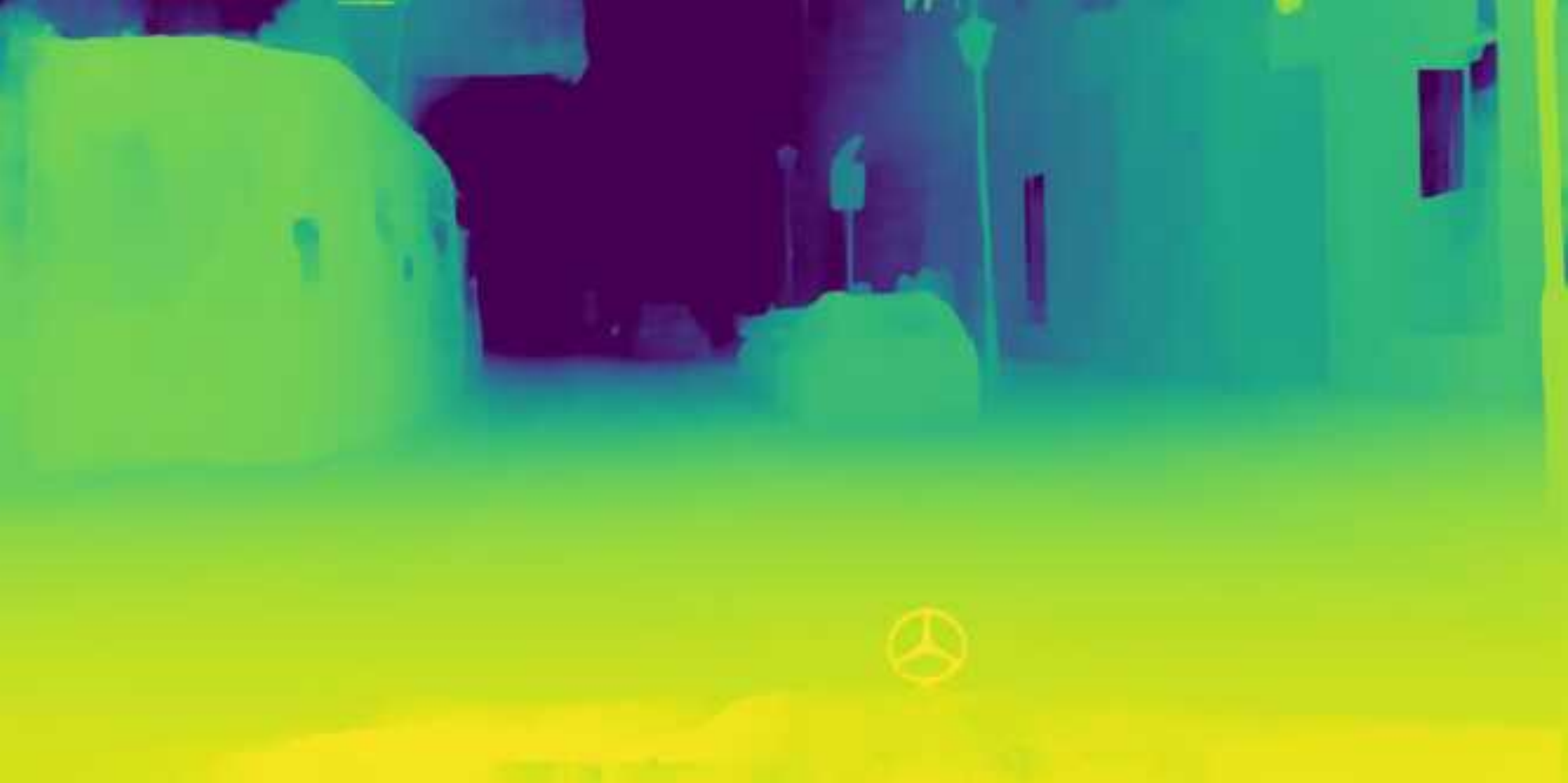}
    \end{subfigure}
    \begin{subfigure}{.2\textwidth}
        \centering
        \includegraphics[width=.98\linewidth]{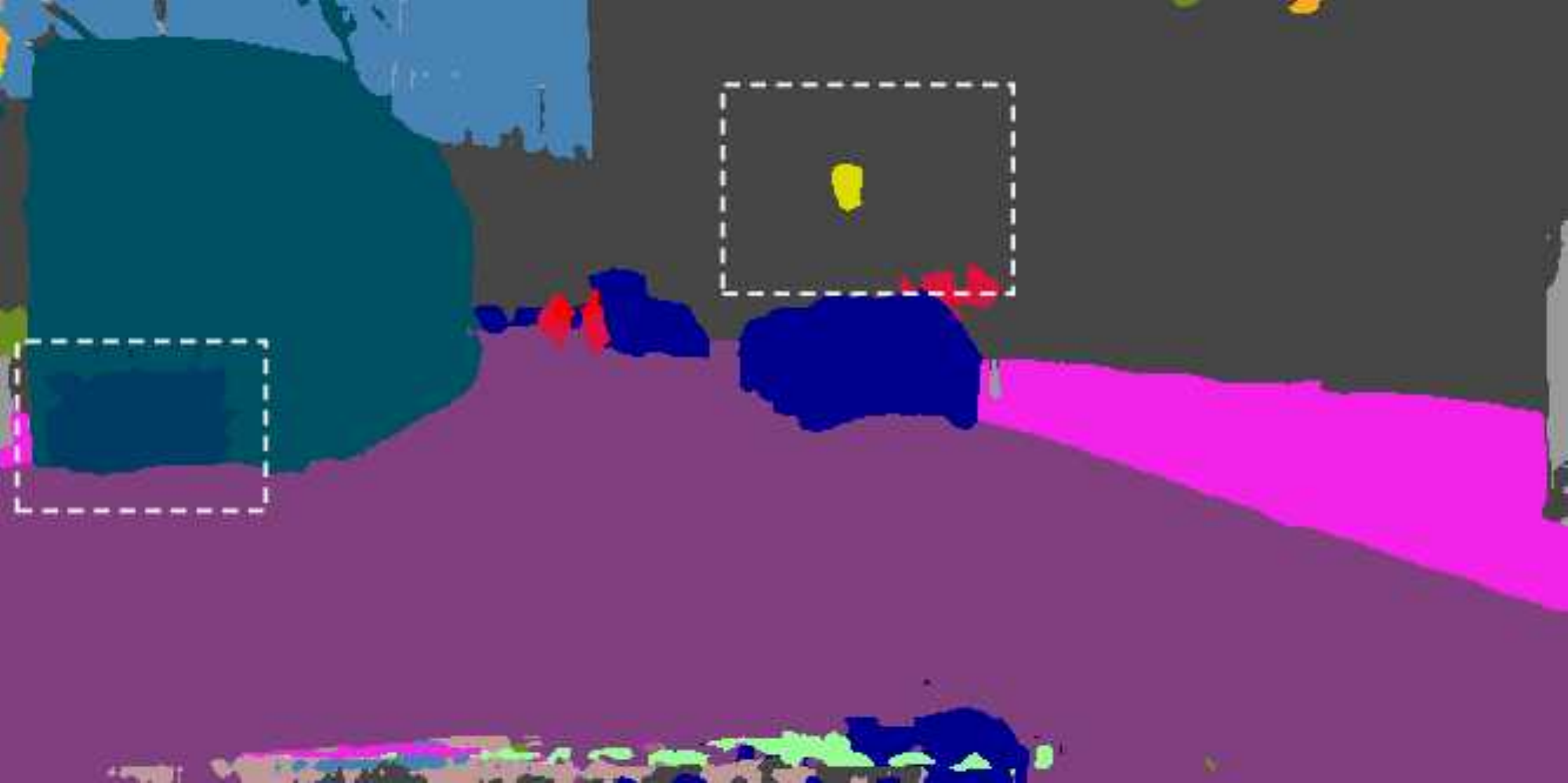}
    \end{subfigure}    \begin{subfigure}{.2\textwidth}
        \centering
        \includegraphics[width=.98\linewidth]{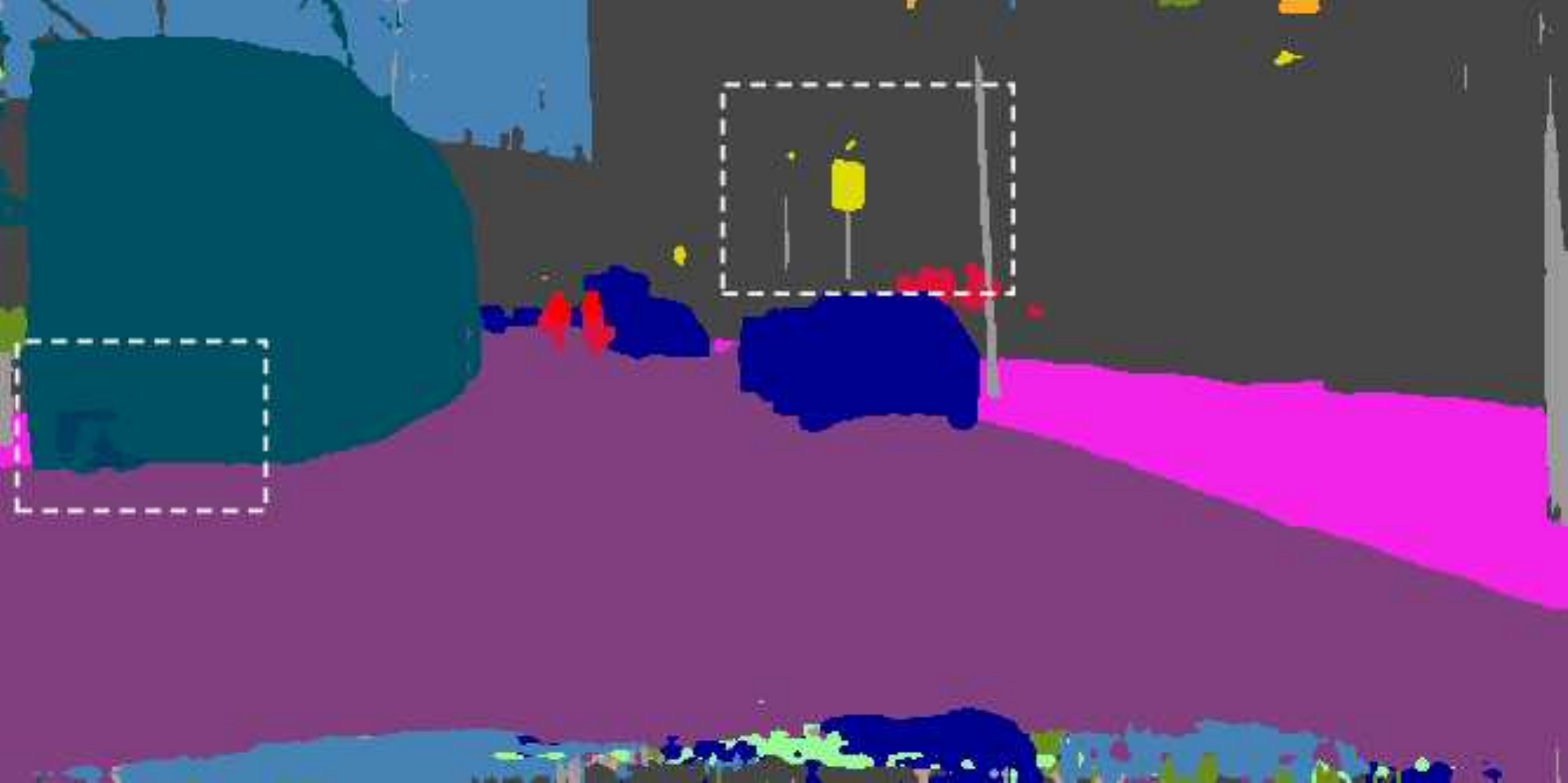}
    \end{subfigure}    \begin{subfigure}{.2\textwidth}
        \centering
        \includegraphics[width=.98\linewidth]{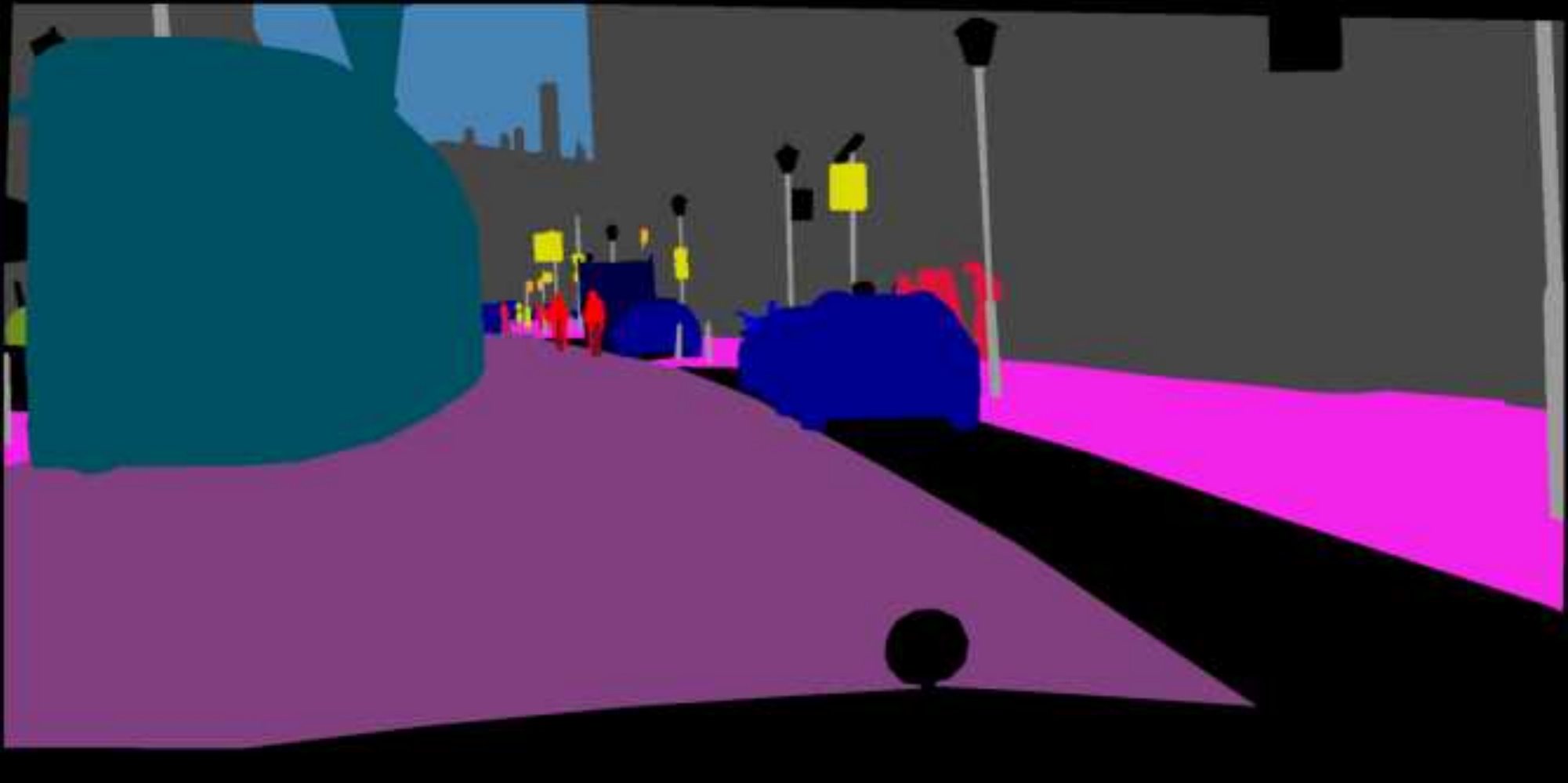}
    \end{subfigure}}

\makebox[\linewidth][c]{    \begin{subfigure}{.2\textwidth}
        \centering
        \includegraphics[width=.98\linewidth]{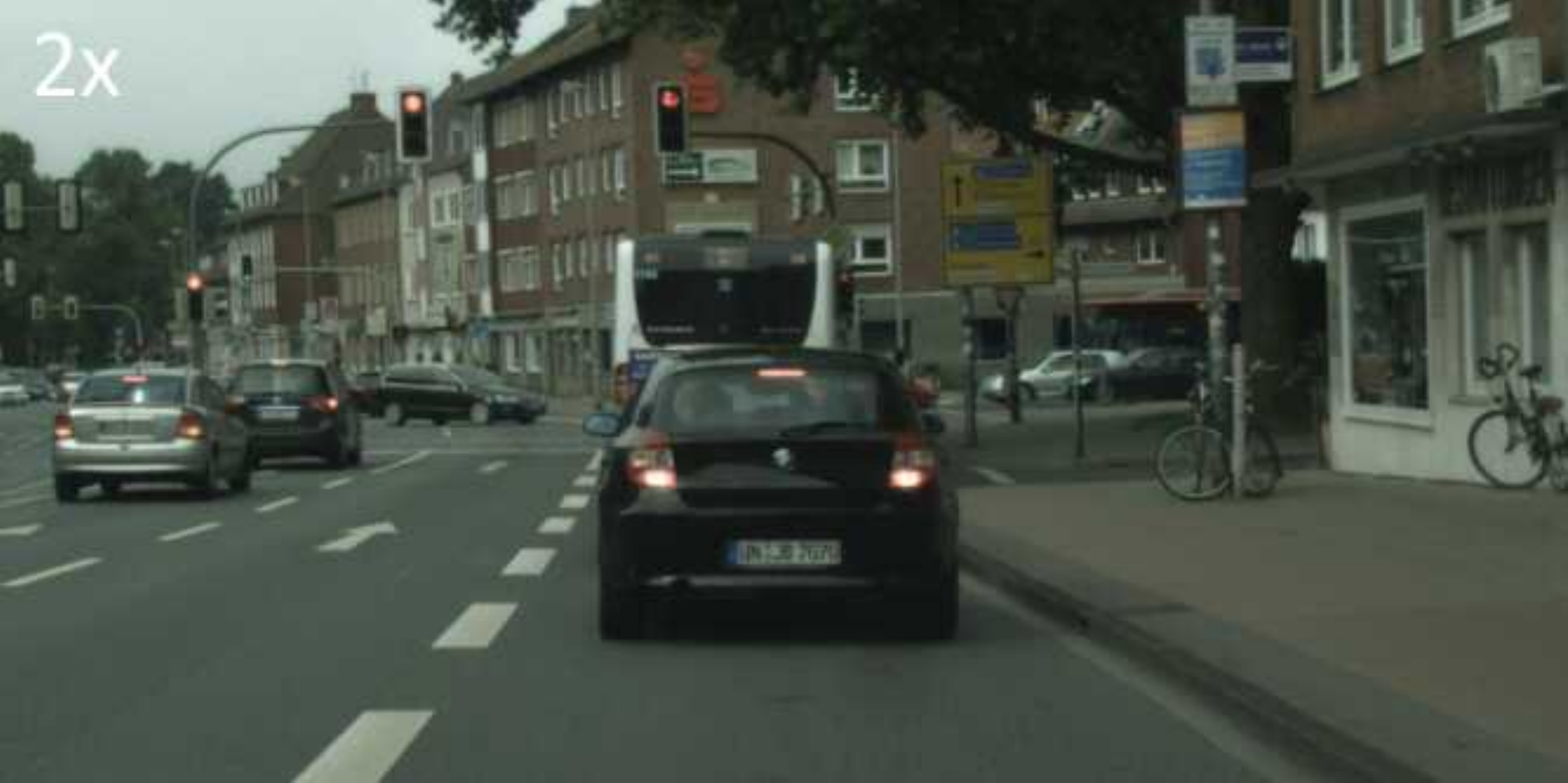}
    \end{subfigure}    \begin{subfigure}{.2\textwidth}
        \centering
        \includegraphics[width=.98\linewidth]{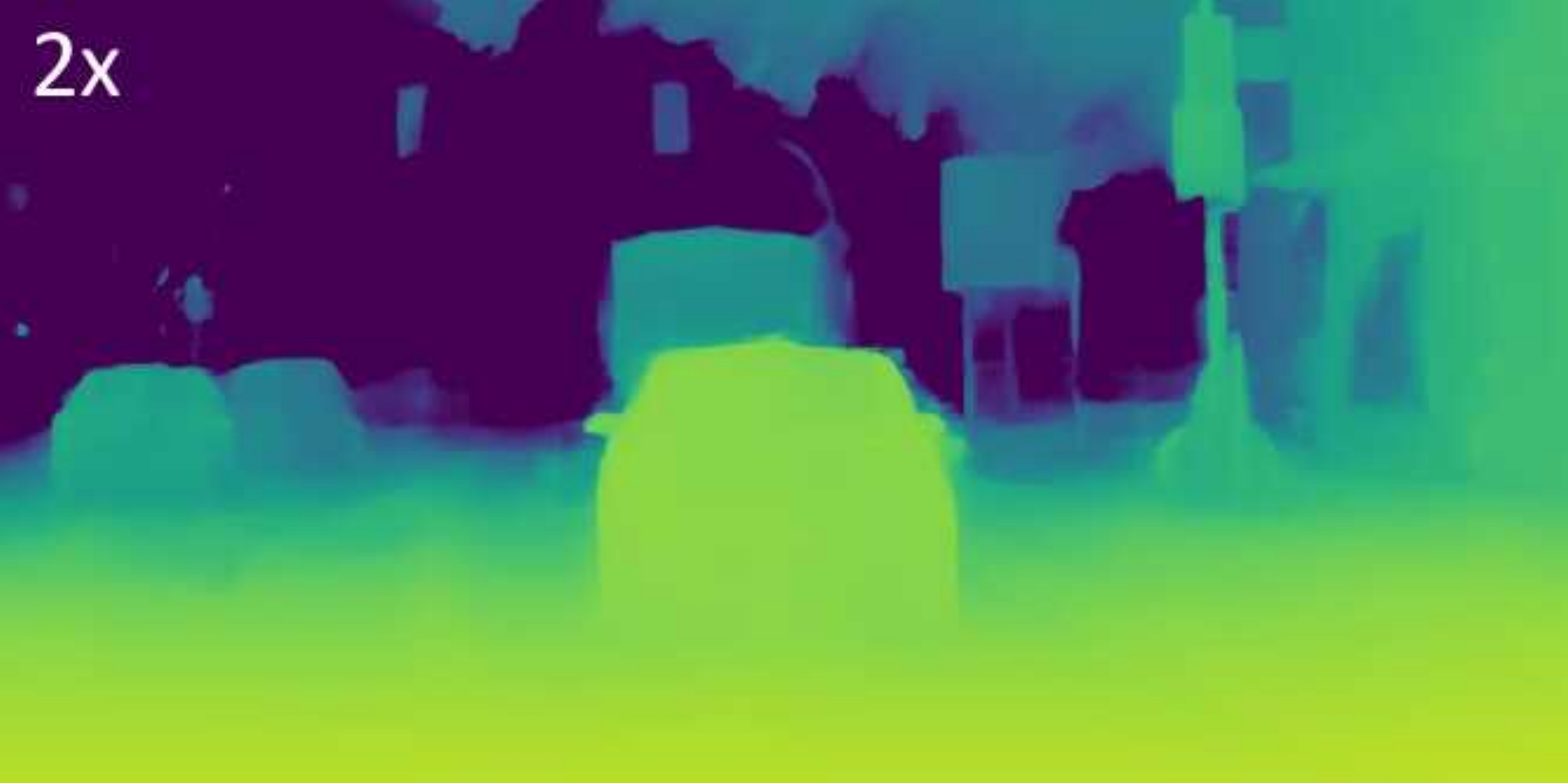}
    \end{subfigure}    \begin{subfigure}{.2\textwidth}
        \centering
        \includegraphics[width=.98\linewidth]{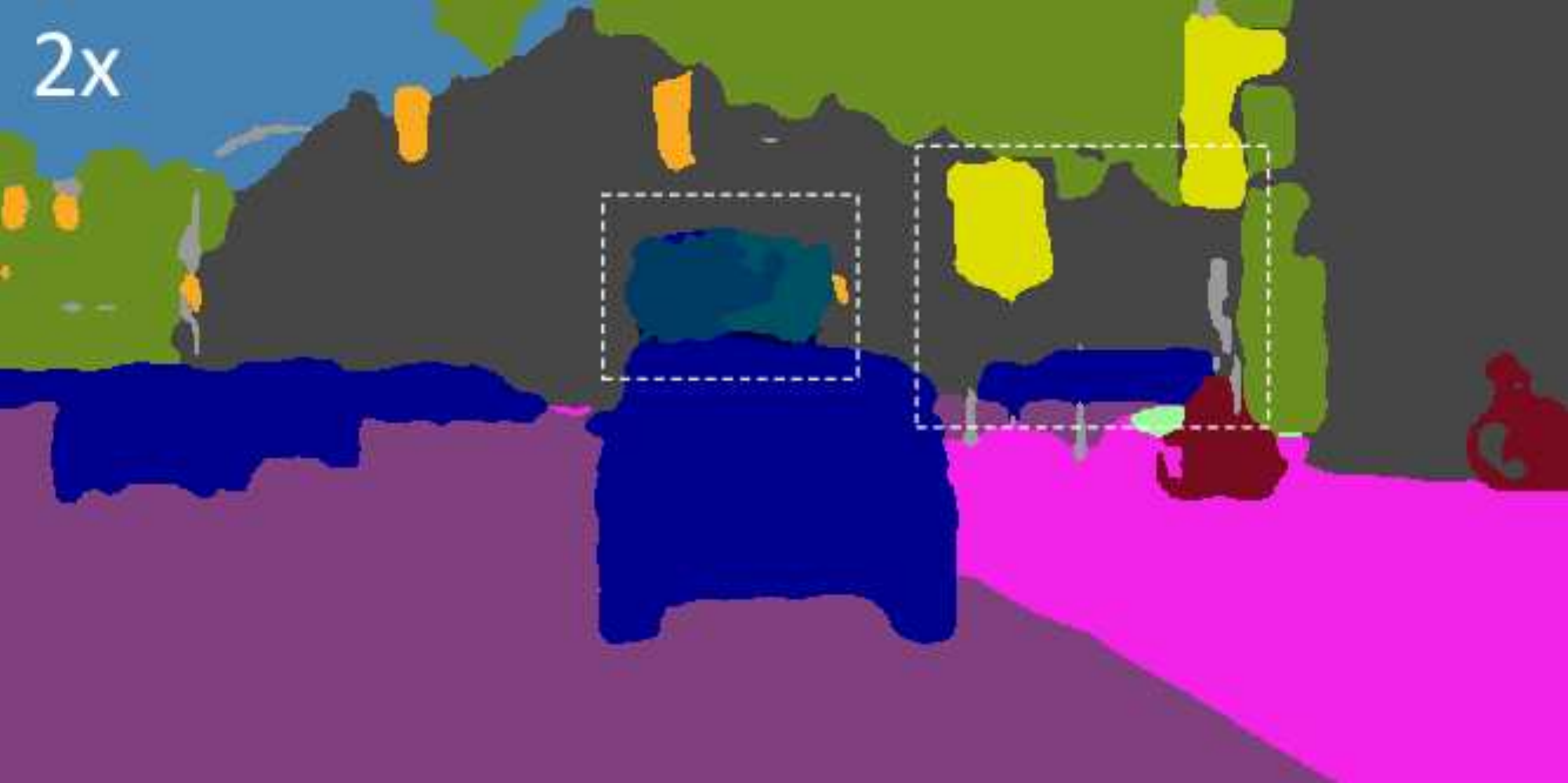}
    \end{subfigure}    \begin{subfigure}{.2\textwidth}
        \centering
        \includegraphics[width=.98\linewidth]{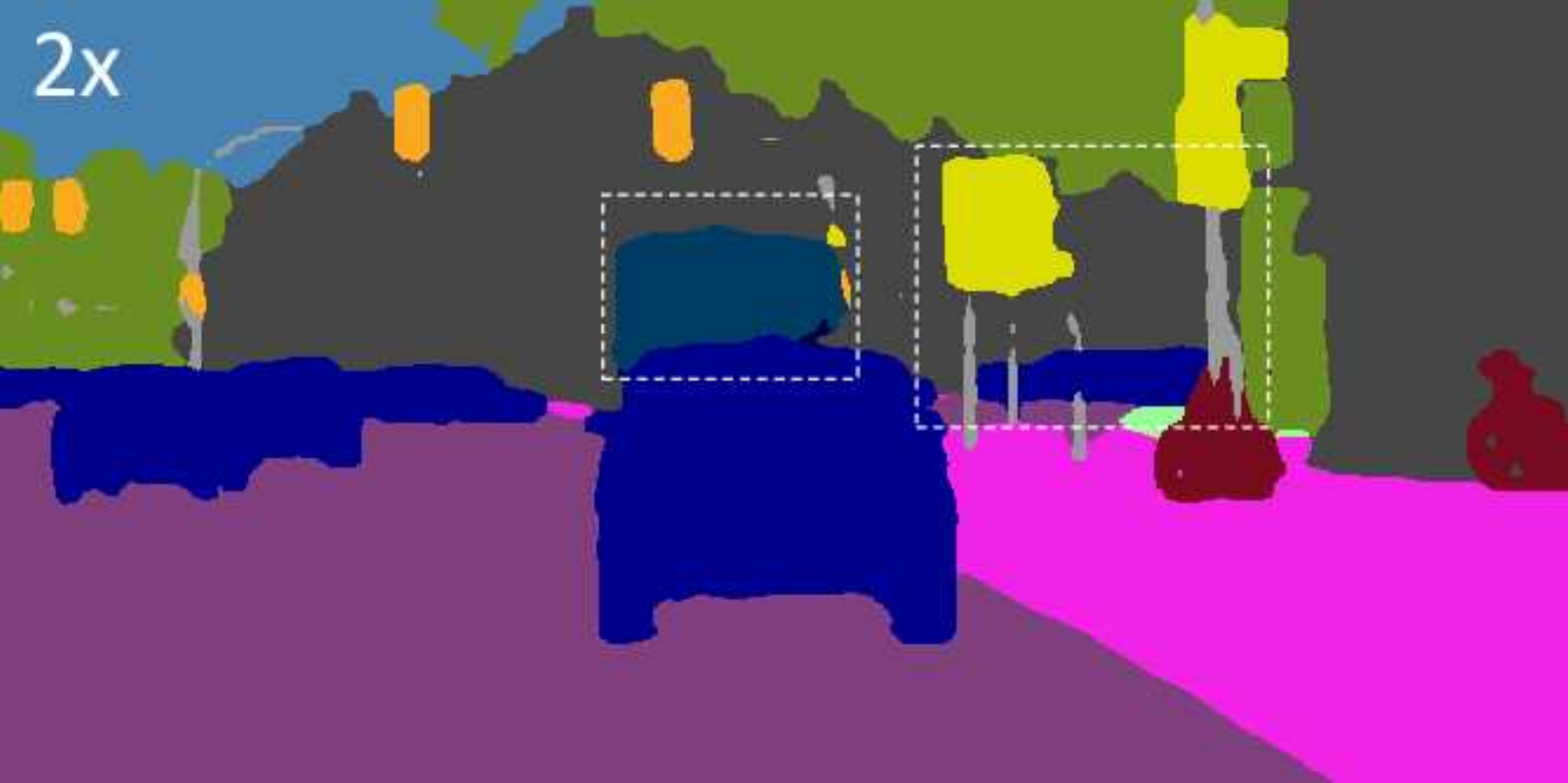}
    \end{subfigure}    \begin{subfigure}{.2\textwidth}
        \centering
        \includegraphics[width=.98\linewidth]{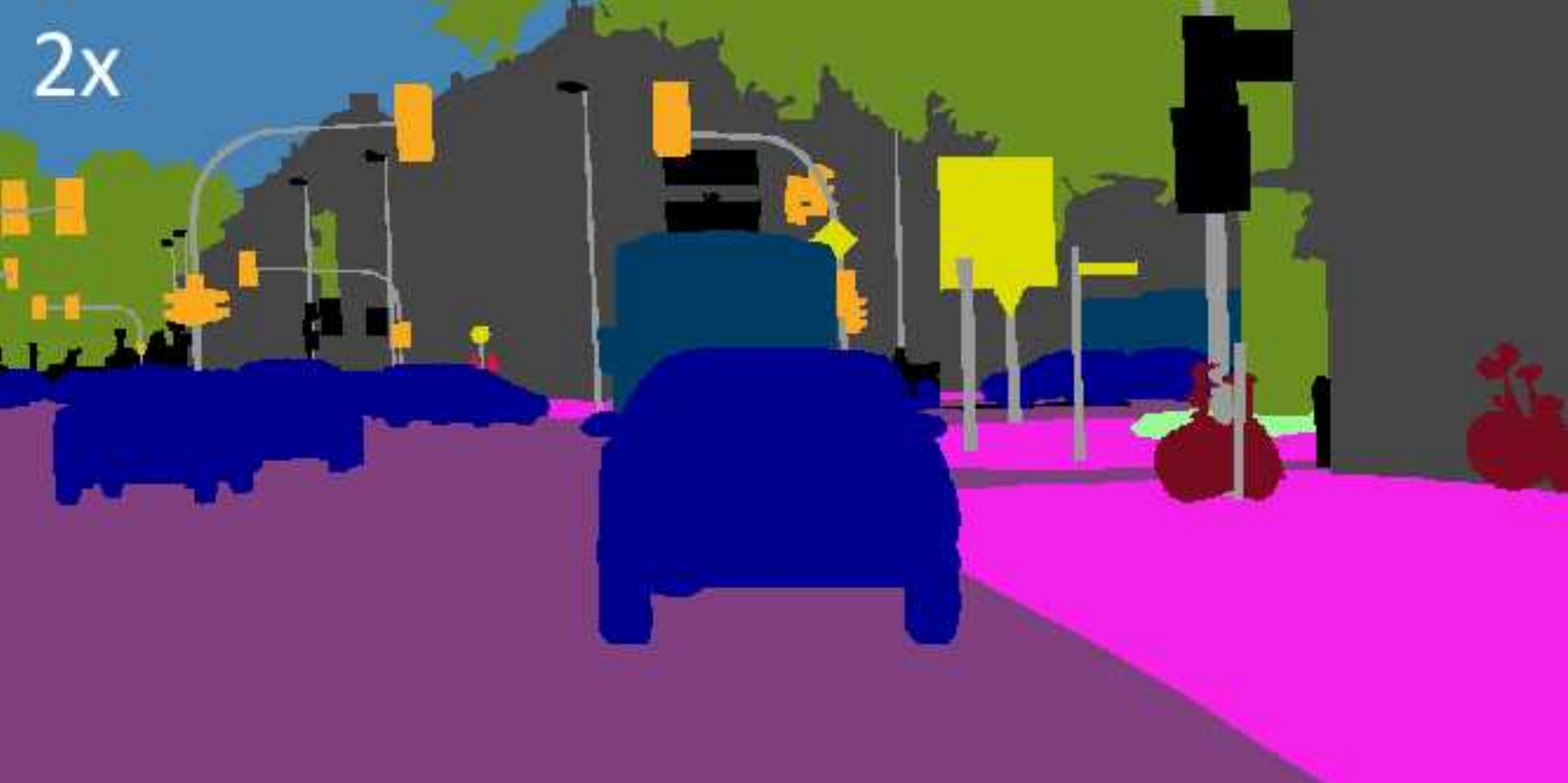}
    \end{subfigure}}
\makebox[\linewidth][c]{    \begin{subfigure}{.2\textwidth}
        \centering
        \includegraphics[width=.98\linewidth]{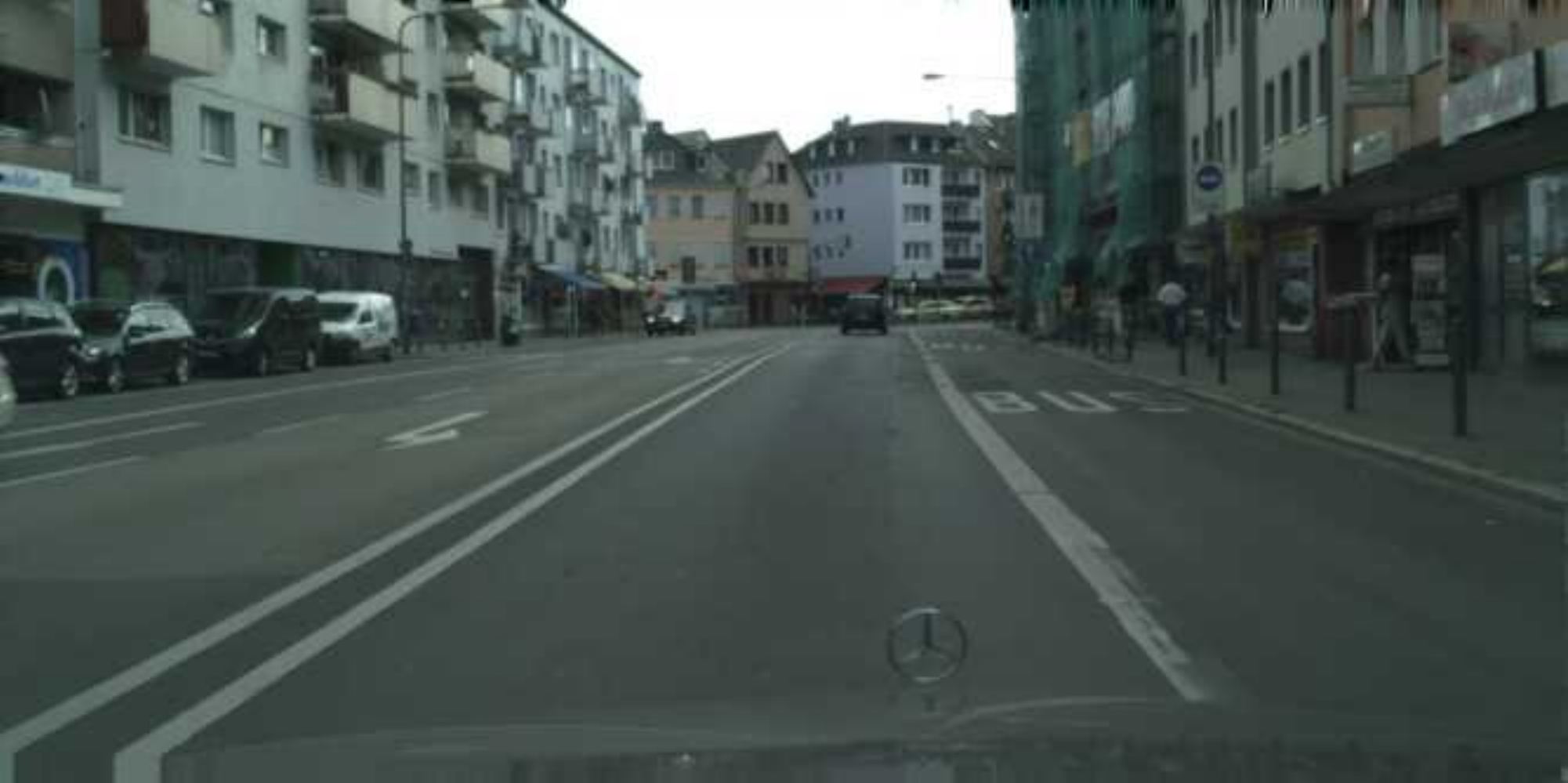}
    \end{subfigure}    \begin{subfigure}{.2\textwidth}
        \centering
        \includegraphics[width=.98\linewidth]{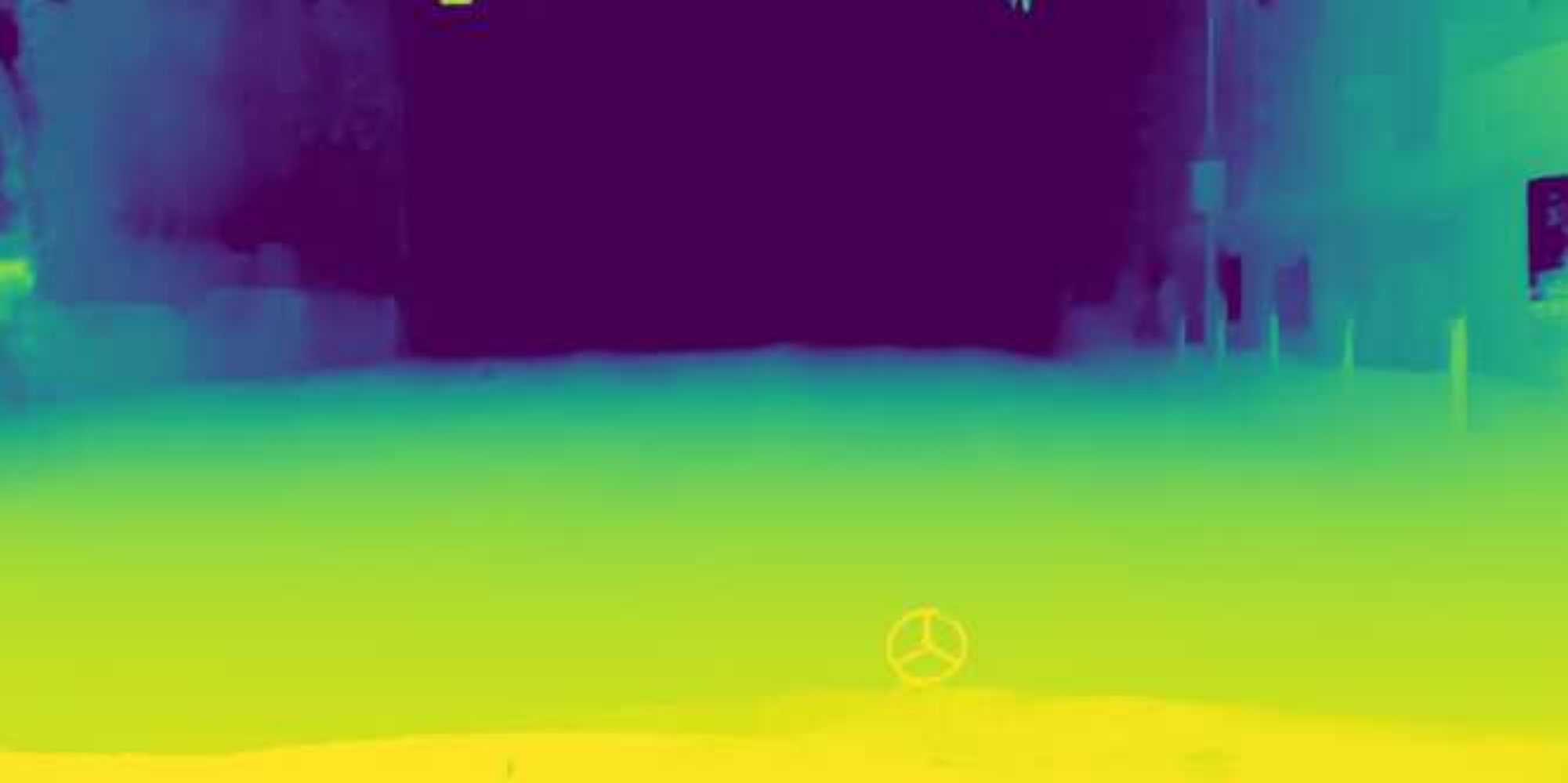}
    \end{subfigure}    \begin{subfigure}{.2\textwidth}
        \centering
        \includegraphics[width=.98\linewidth]{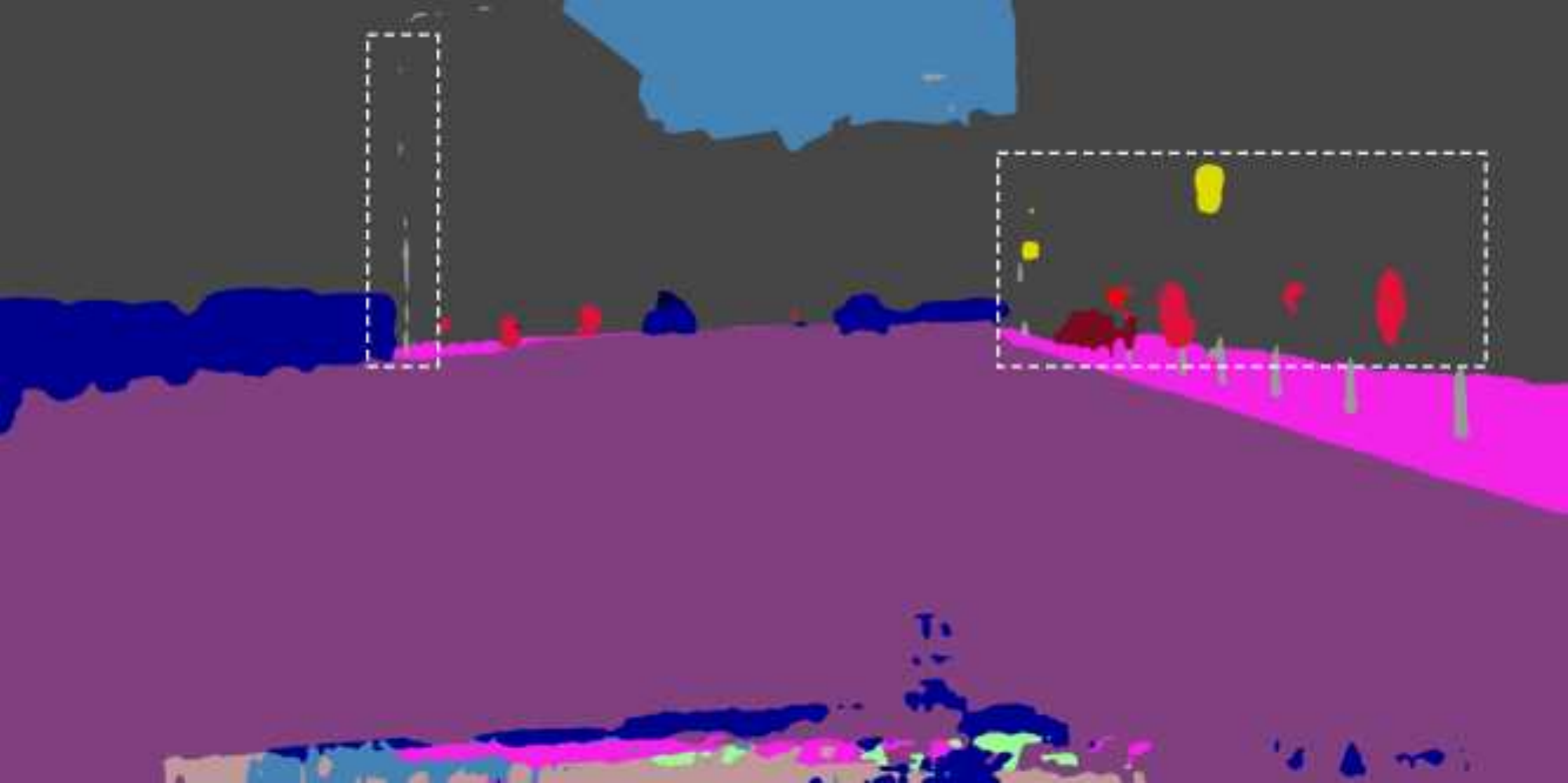}
    \end{subfigure}    \begin{subfigure}{.2\textwidth}
        \centering
        \includegraphics[width=.98\linewidth]{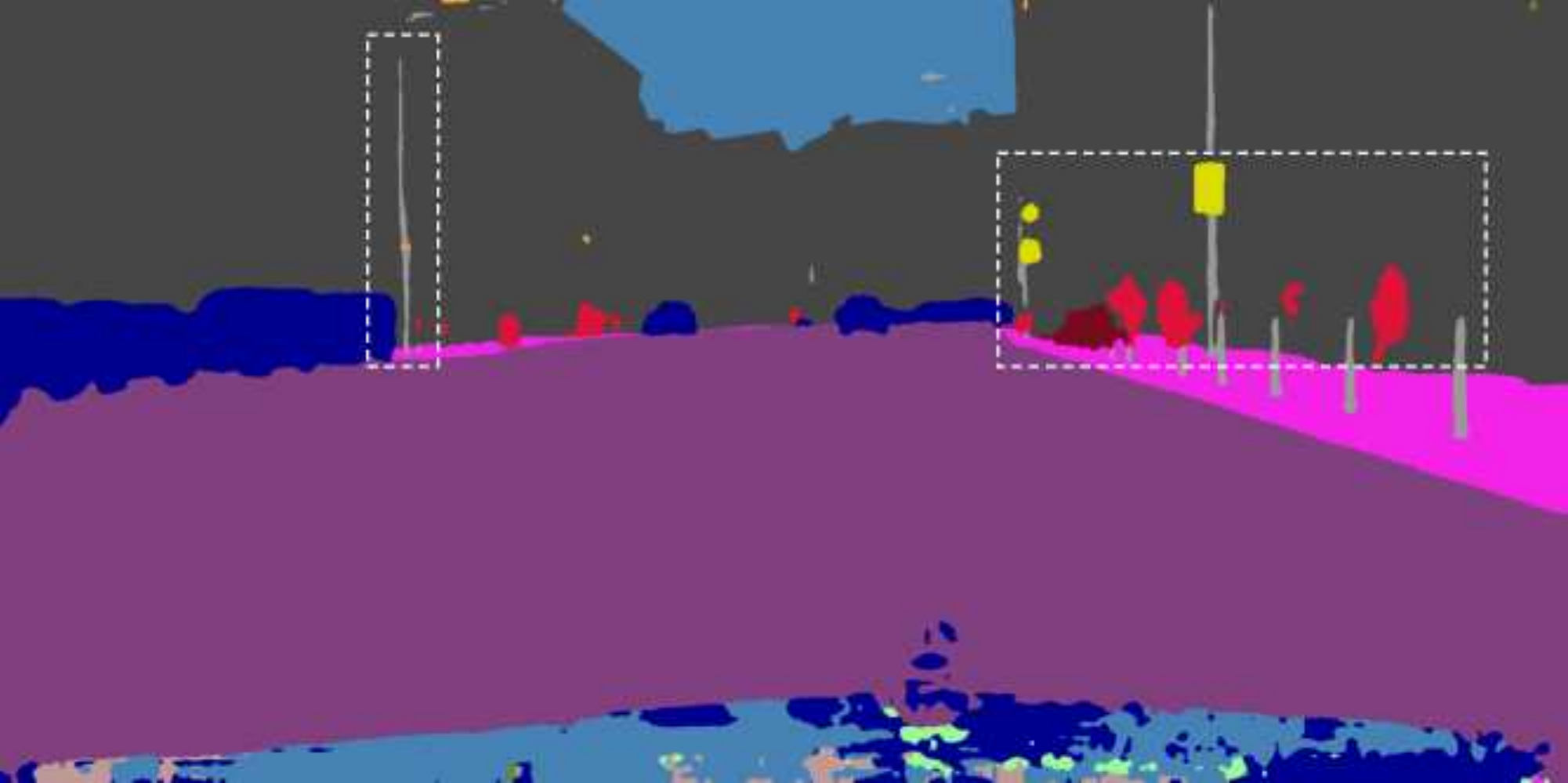}
    \end{subfigure}    \begin{subfigure}{.2\textwidth}
        \centering
        \includegraphics[width=.98\linewidth]{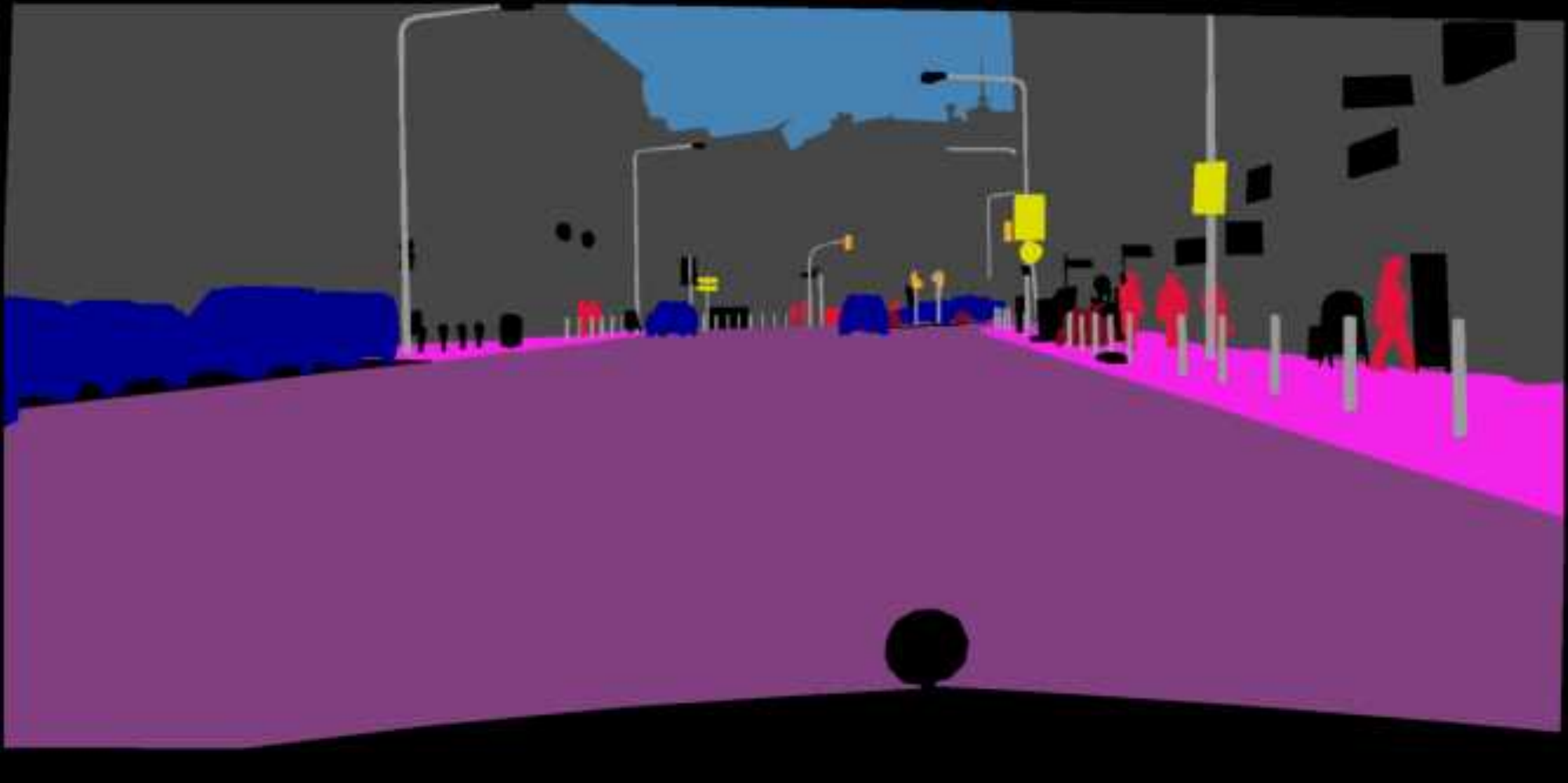}
    \end{subfigure}}
\makebox[\linewidth][c]{    \begin{subfigure}{.2\textwidth}
        \centering
        \includegraphics[width=.98\linewidth]{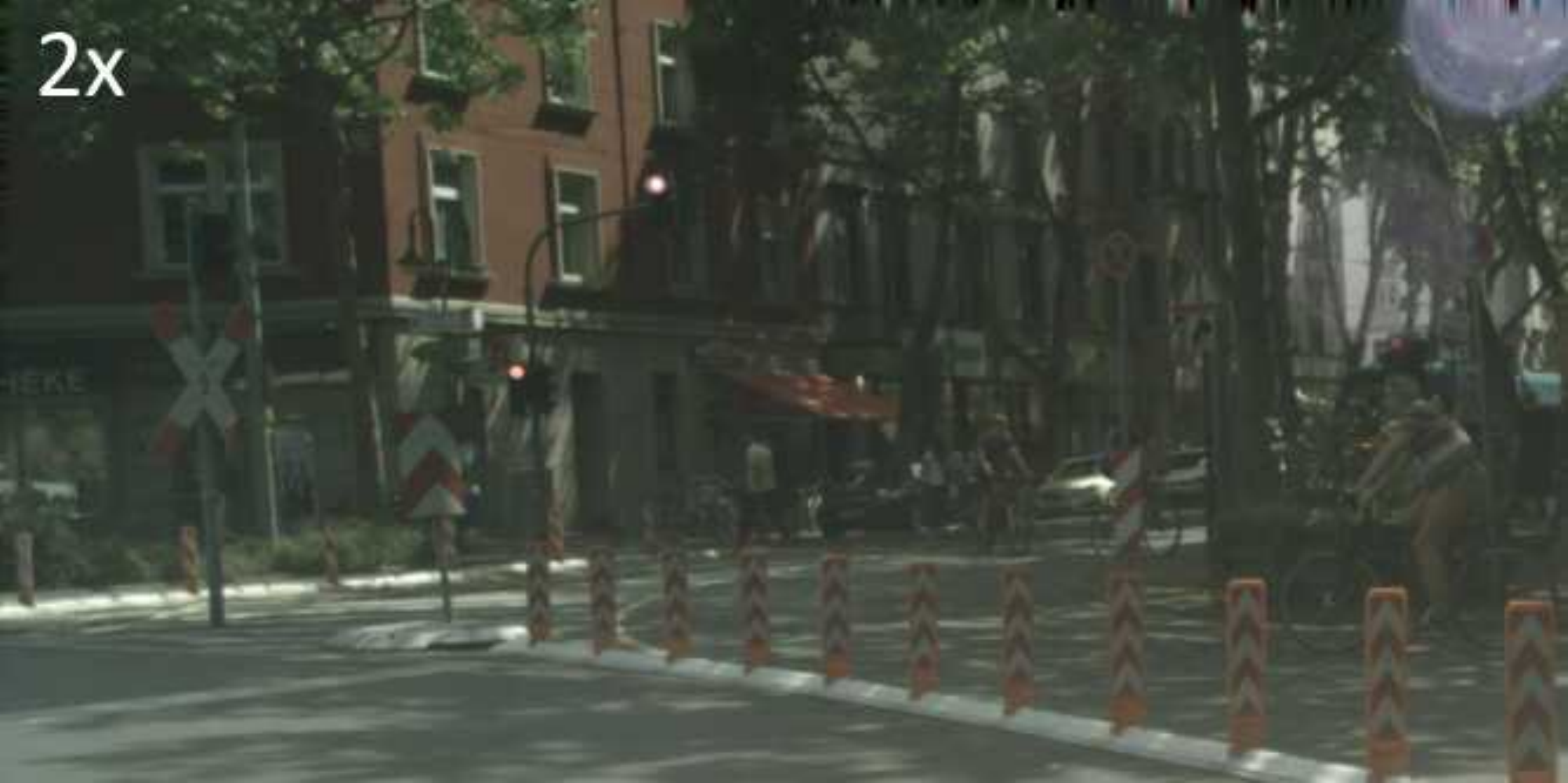}
    \end{subfigure}    \begin{subfigure}{.2\textwidth}
        \centering
        \includegraphics[width=.98\linewidth]{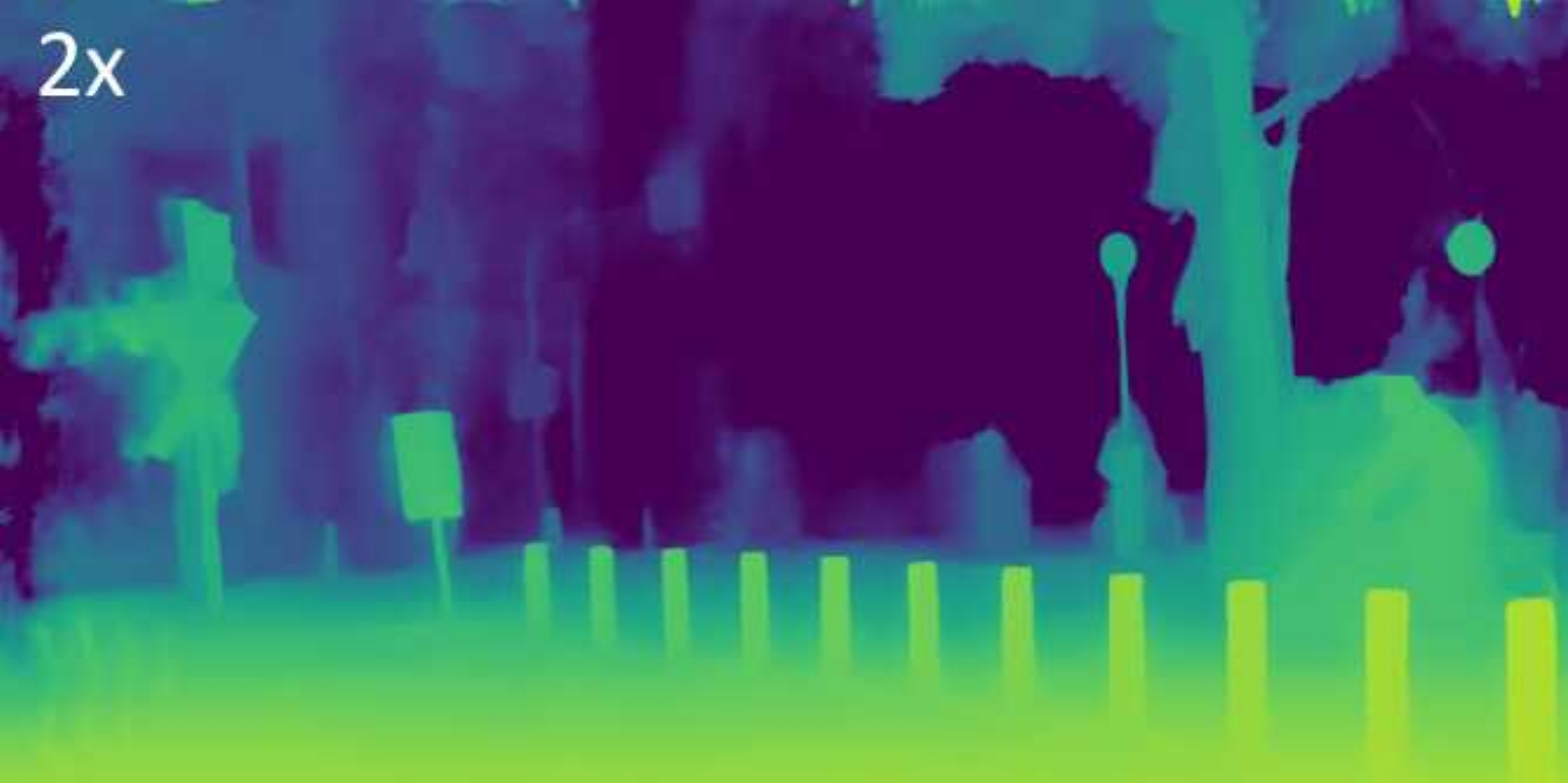}
    \end{subfigure}    \begin{subfigure}{.2\textwidth}
        \centering
        \includegraphics[width=.98\linewidth]{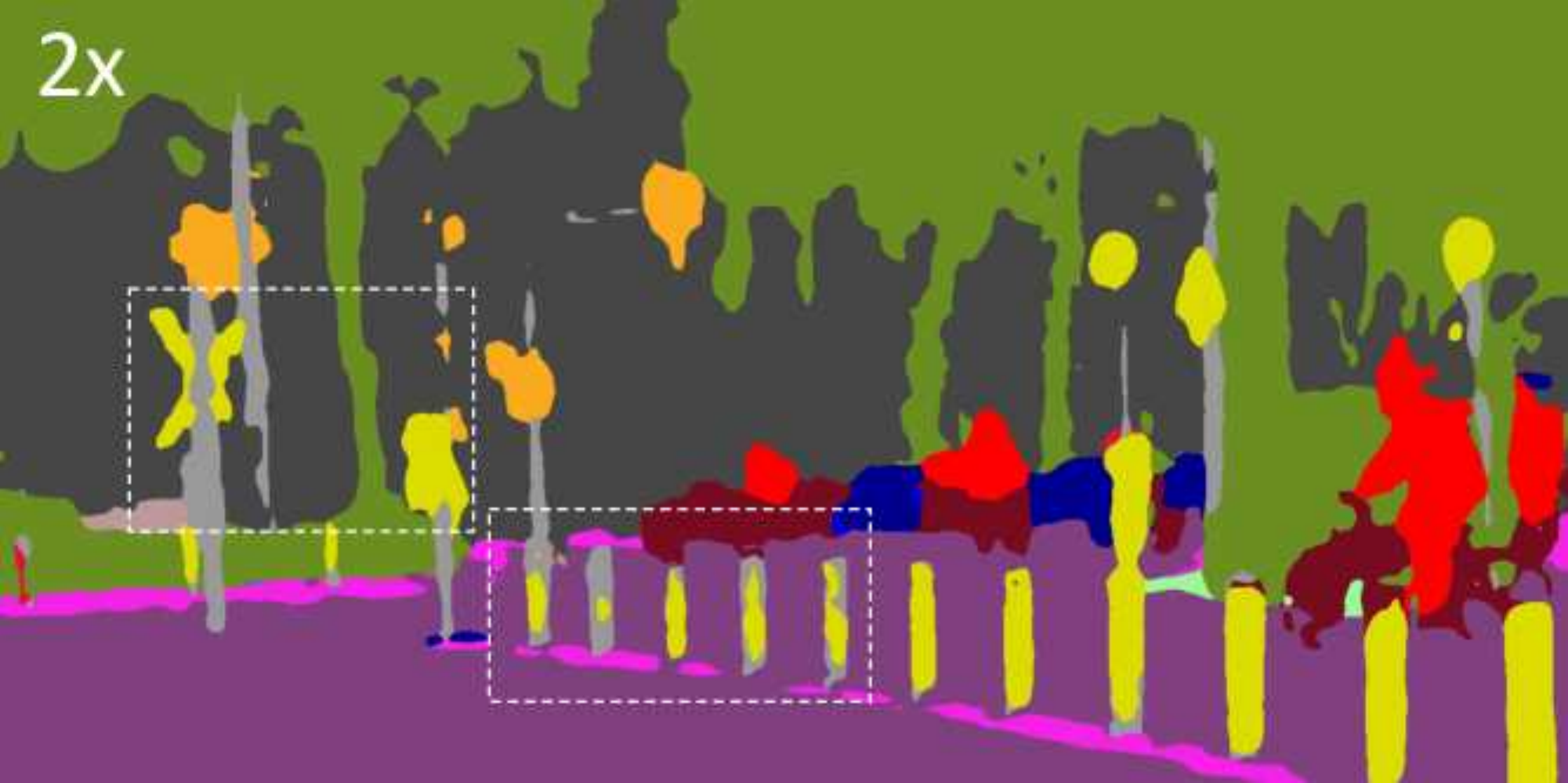}
    \end{subfigure}    \begin{subfigure}{.2\textwidth}
        \centering
        \includegraphics[width=.98\linewidth]{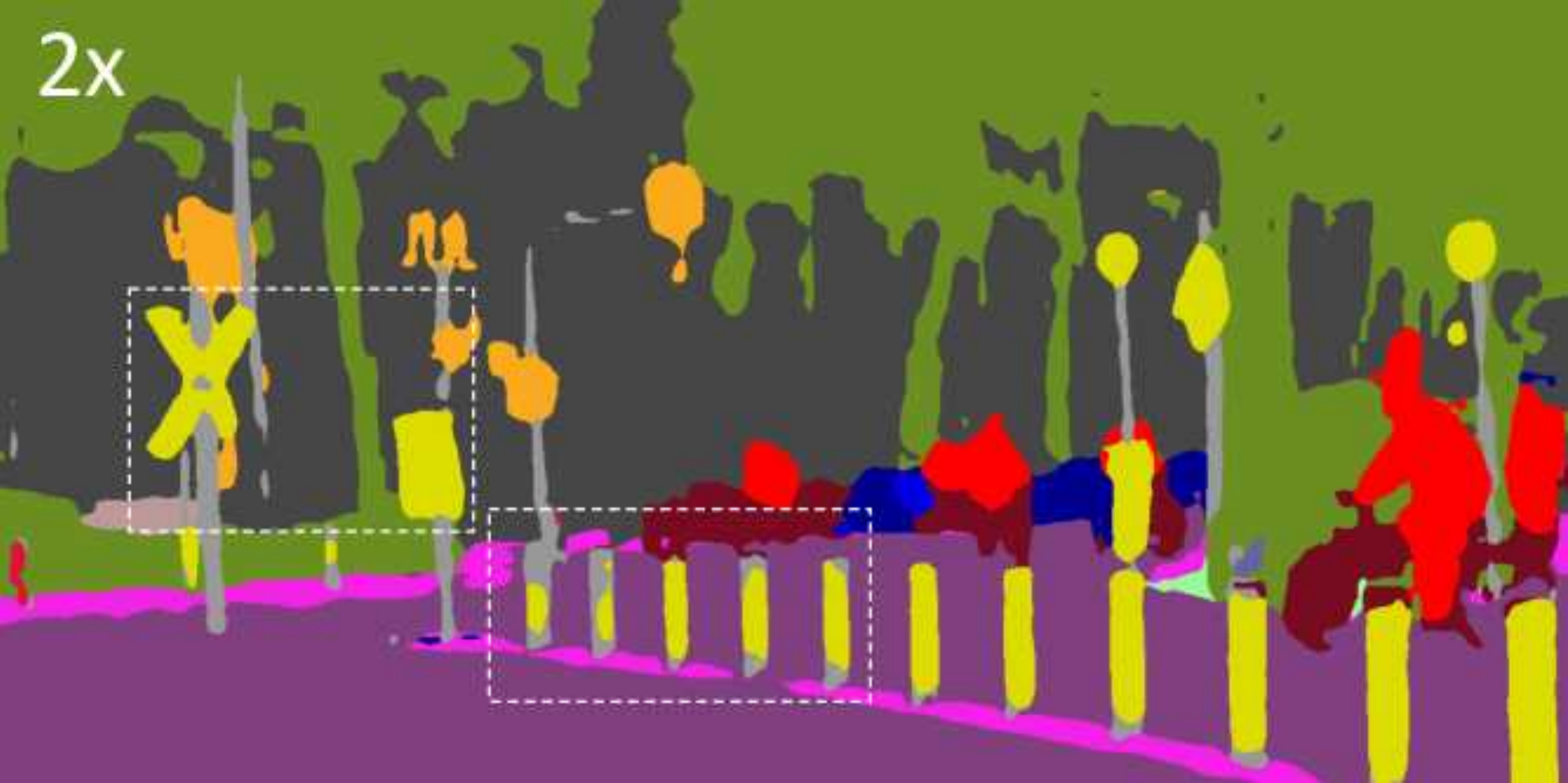}
    \end{subfigure}    \begin{subfigure}{.2\textwidth}
        \centering
        \includegraphics[width=.98\linewidth]{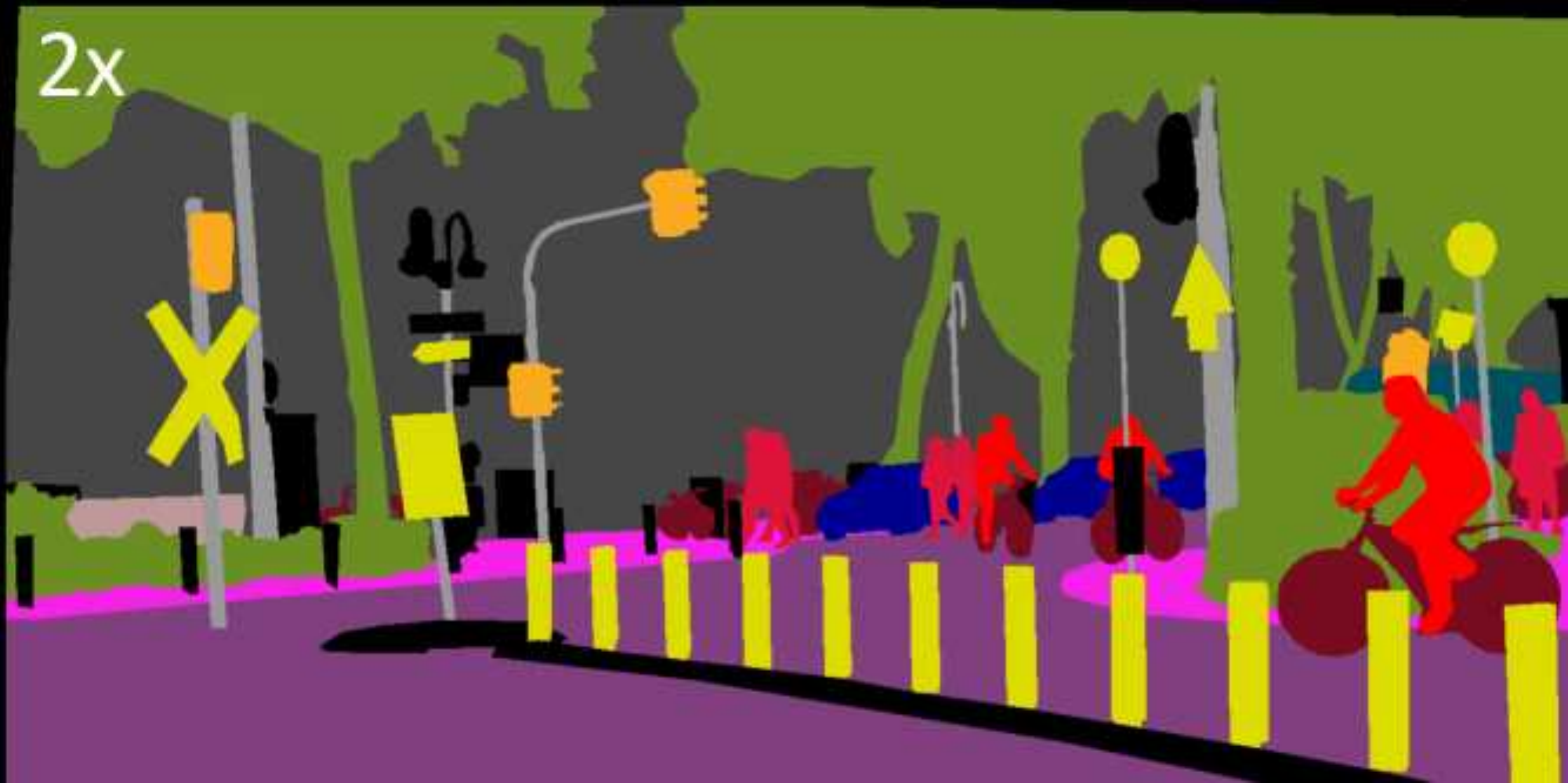}
    \end{subfigure}}
\makebox[\linewidth][c]{    \begin{subfigure}{.2\textwidth}
        \centering
        \includegraphics[width=.98\linewidth]{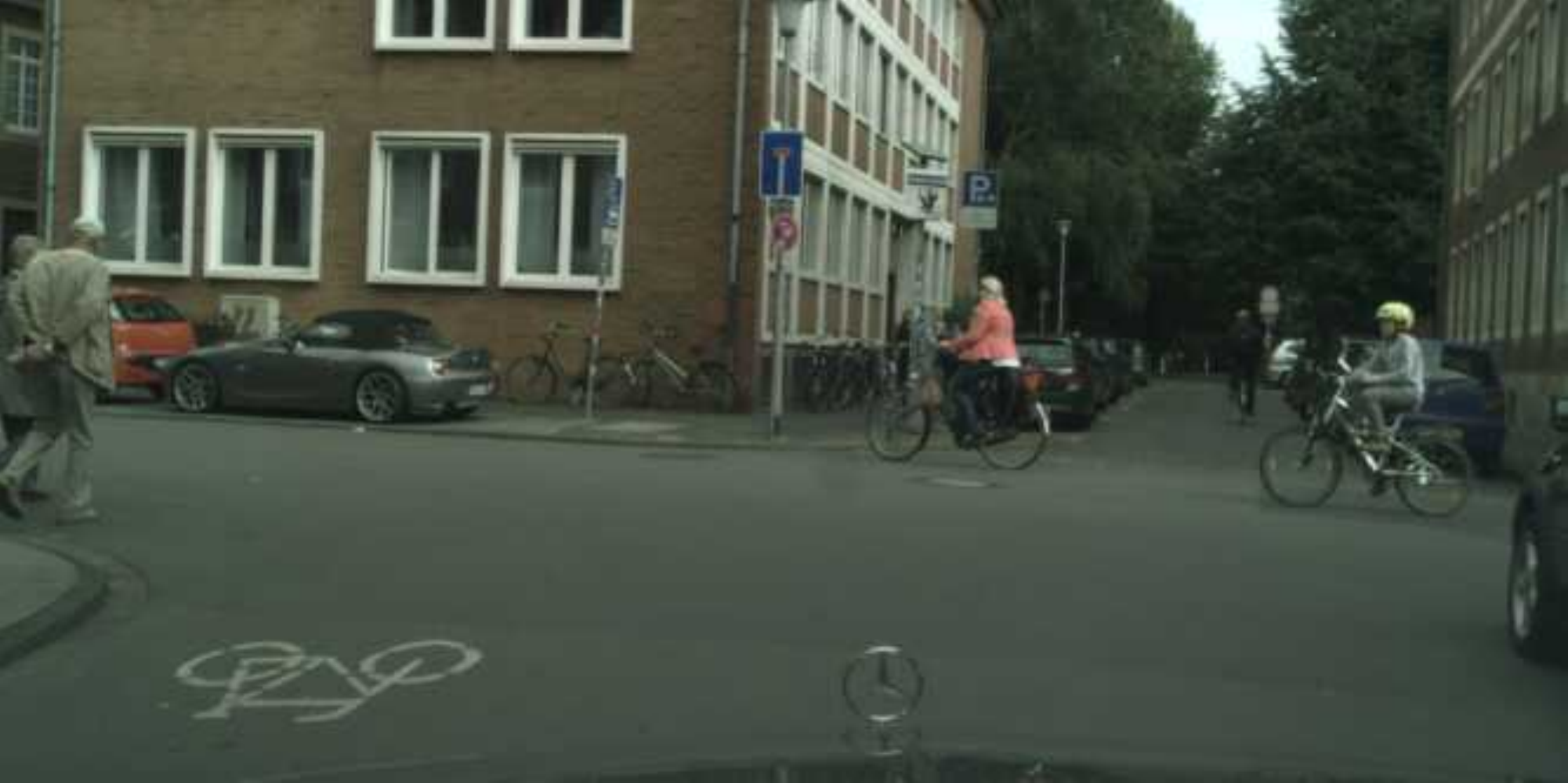}
    \end{subfigure}    \begin{subfigure}{.2\textwidth}
        \centering
        \includegraphics[width=.98\linewidth]{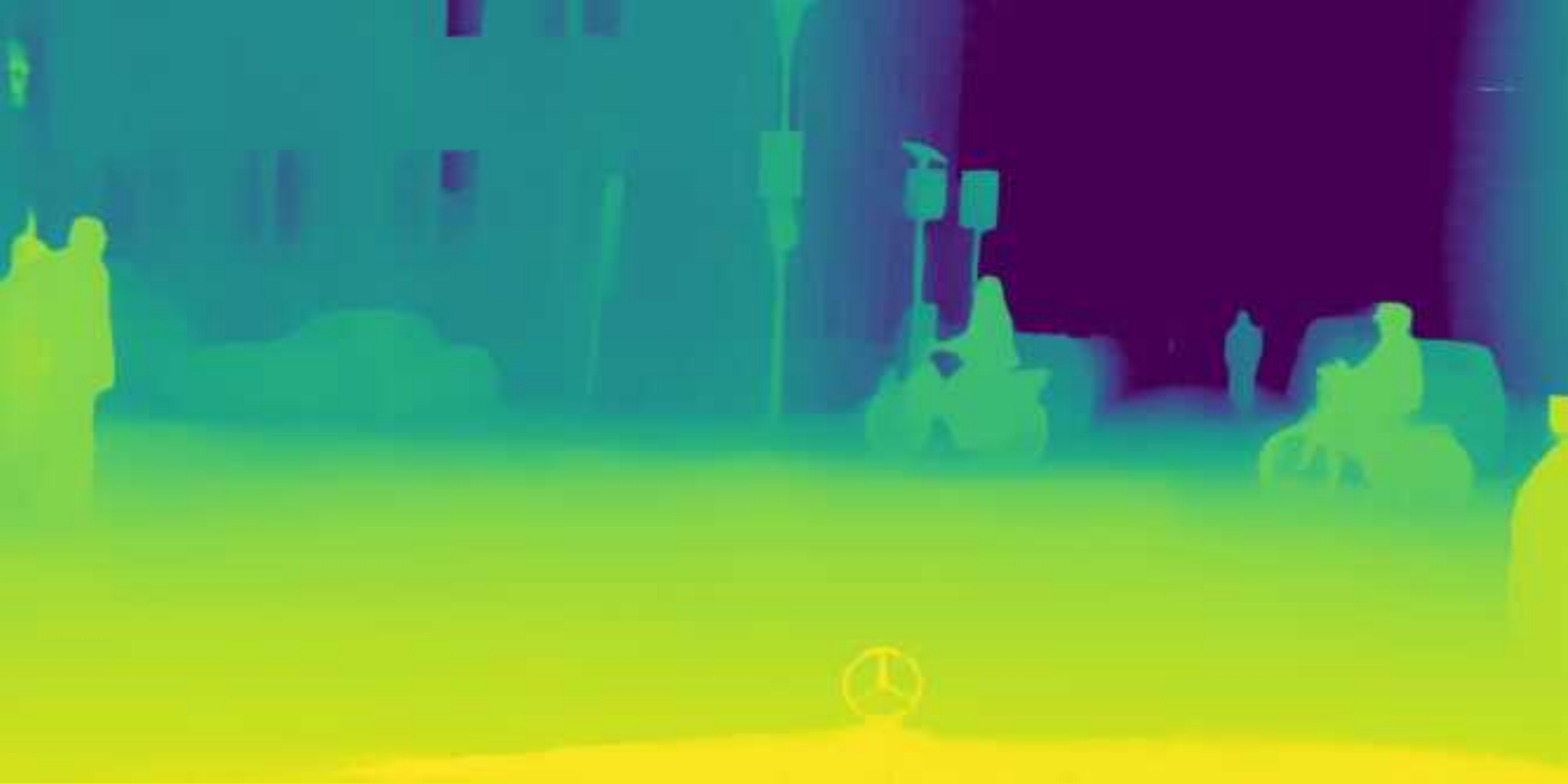}
    \end{subfigure}    \begin{subfigure}{.2\textwidth}
        \centering
        \includegraphics[width=.98\linewidth]{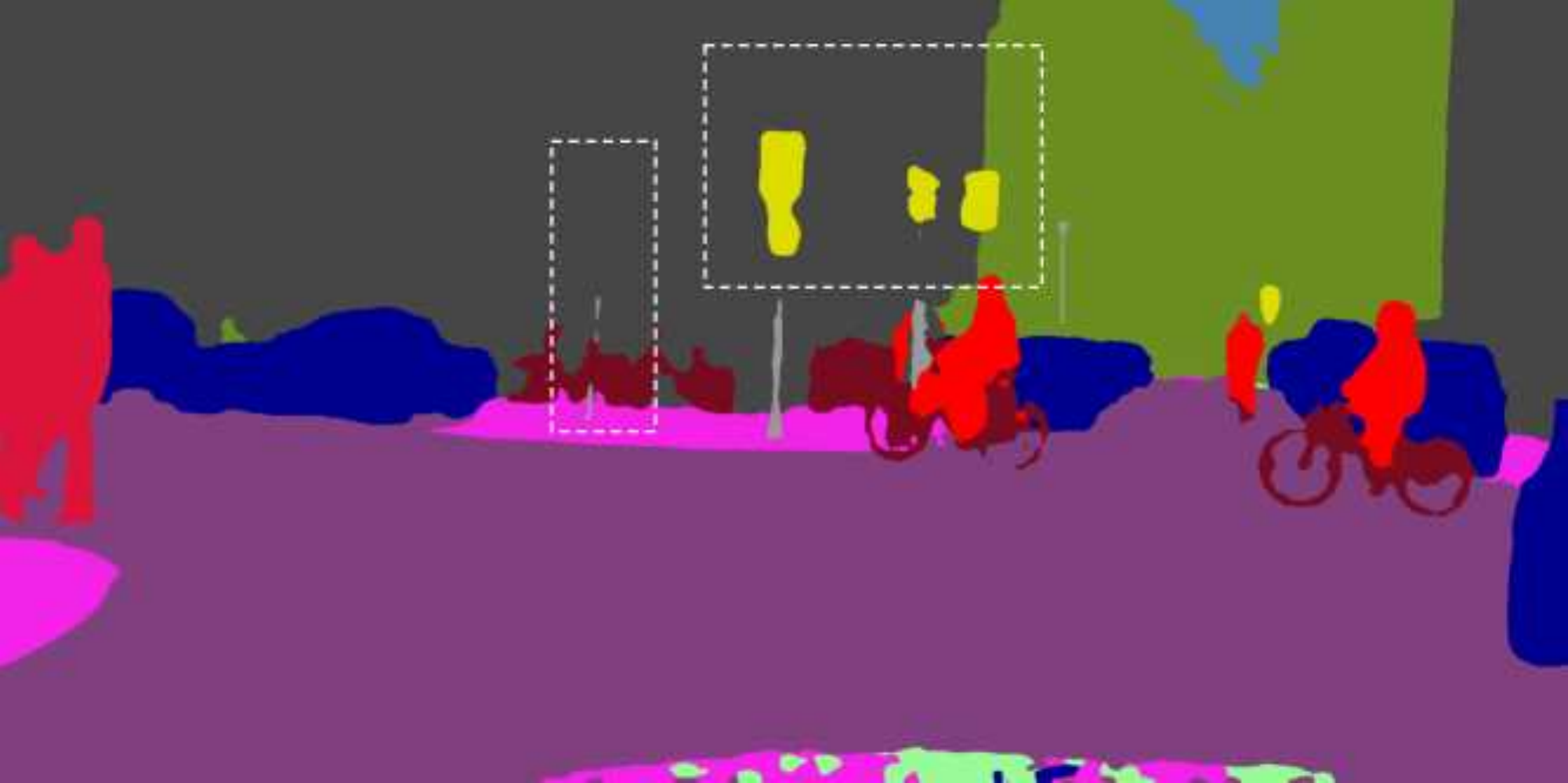}
    \end{subfigure}    \begin{subfigure}{.2\textwidth}
        \centering
        \includegraphics[width=.98\linewidth]{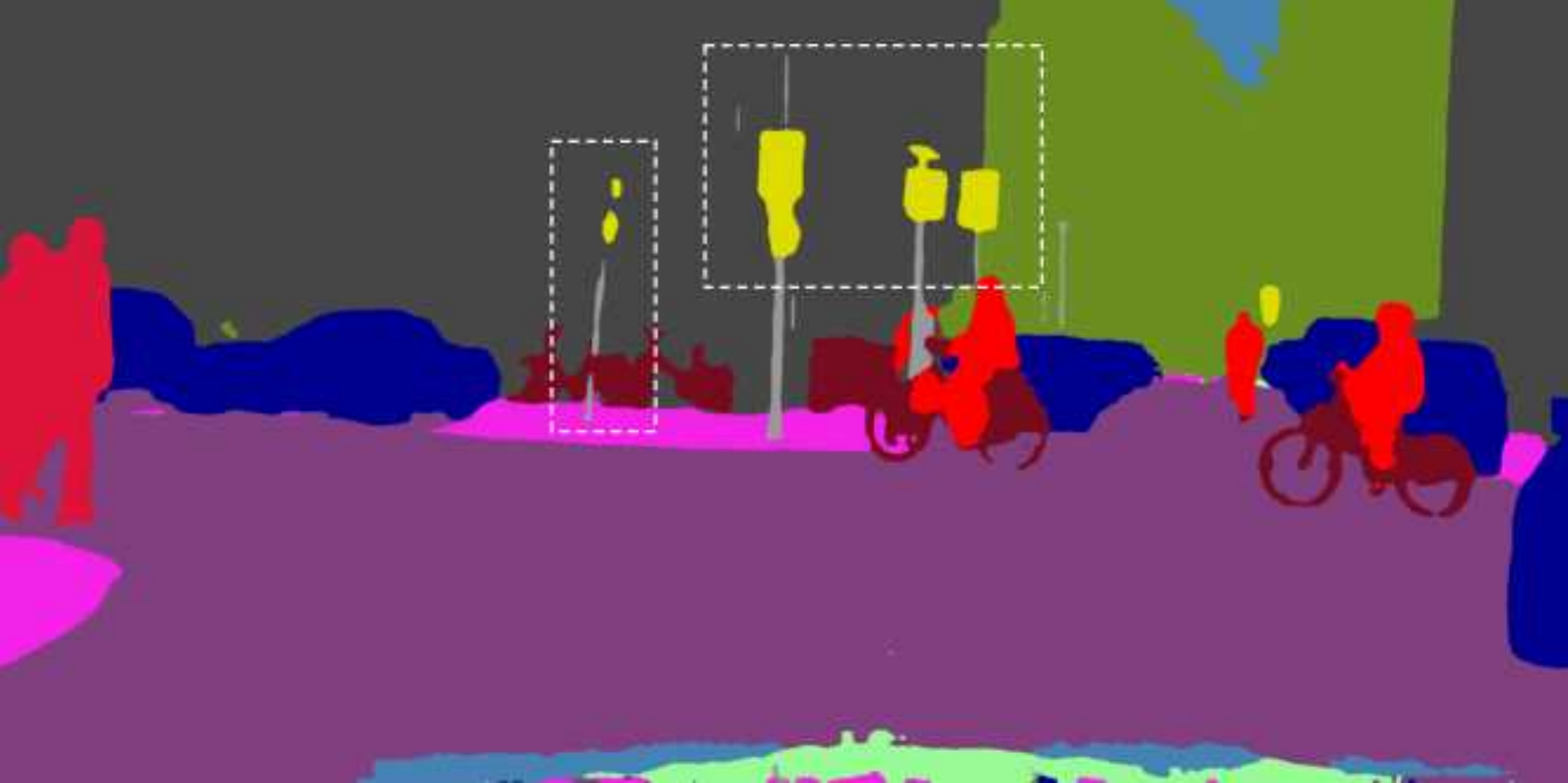}
    \end{subfigure}    \begin{subfigure}{.2\textwidth}
        \centering
        \includegraphics[width=.98\linewidth]{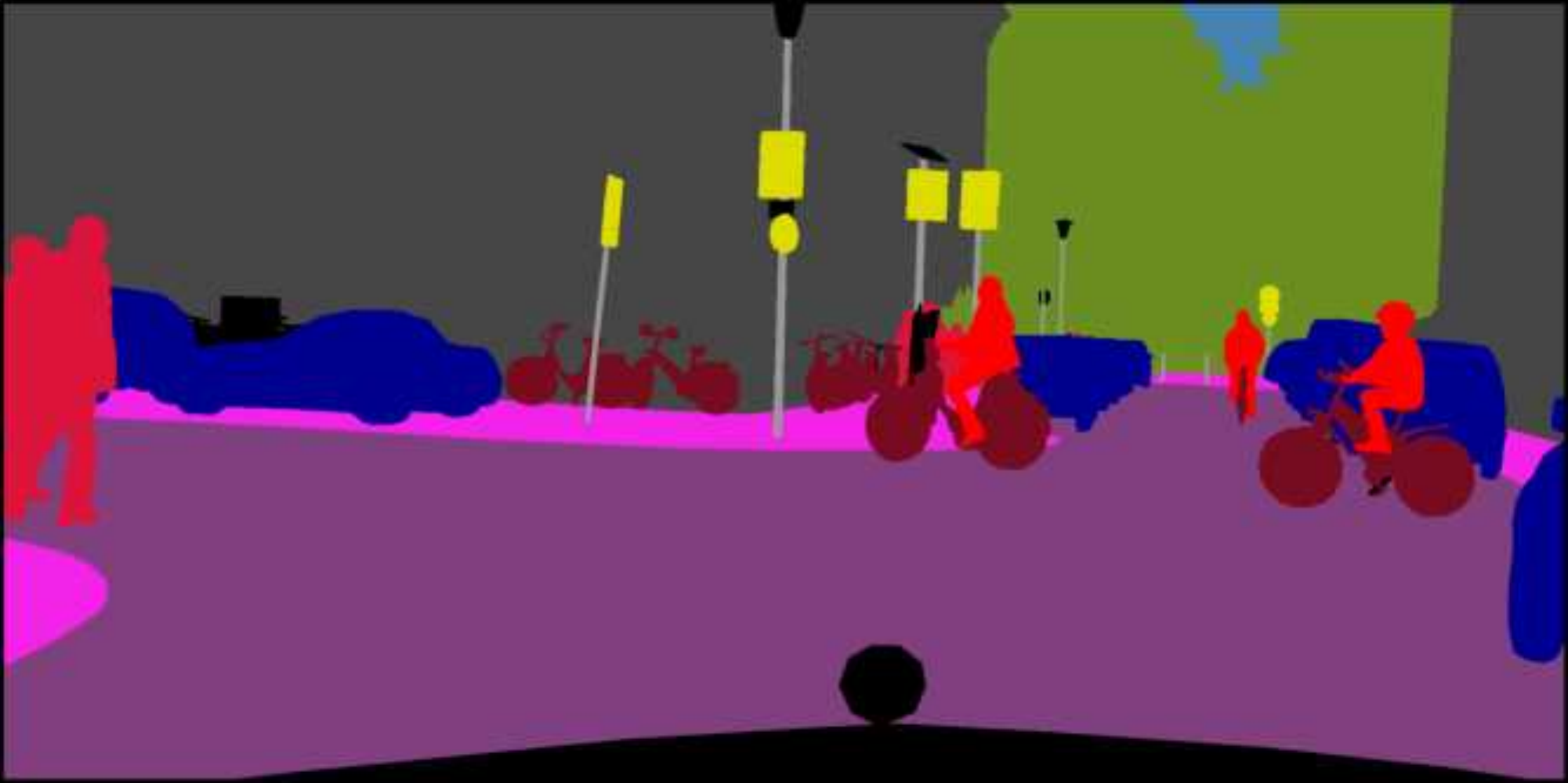}
    \end{subfigure}}
\makebox[\linewidth][c]{    \begin{subfigure}{.2\textwidth}
        \centering
        \includegraphics[width=.98\linewidth]{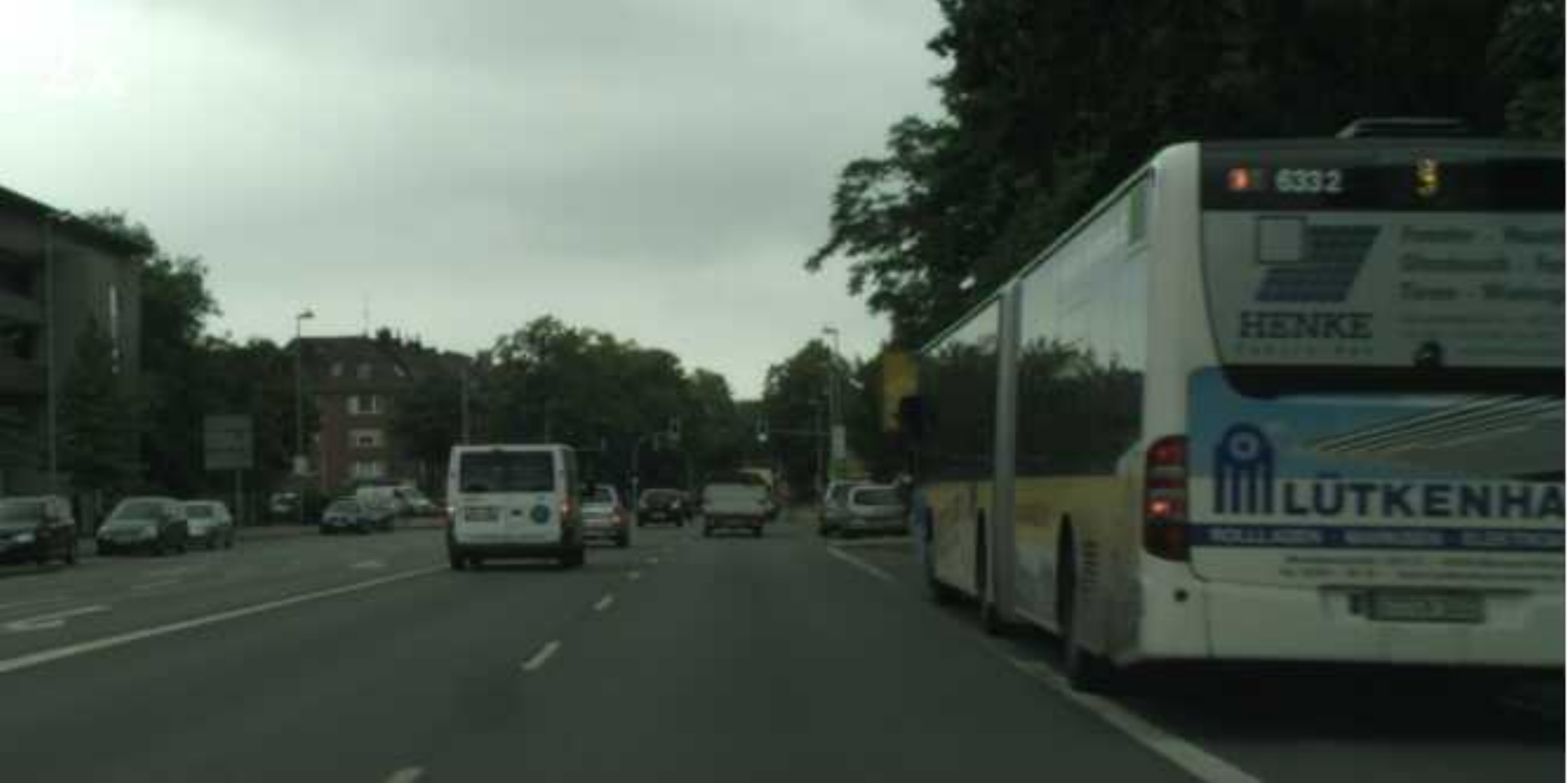}
    \end{subfigure}    \begin{subfigure}{.2\textwidth}
        \centering
        \includegraphics[width=.98\linewidth]{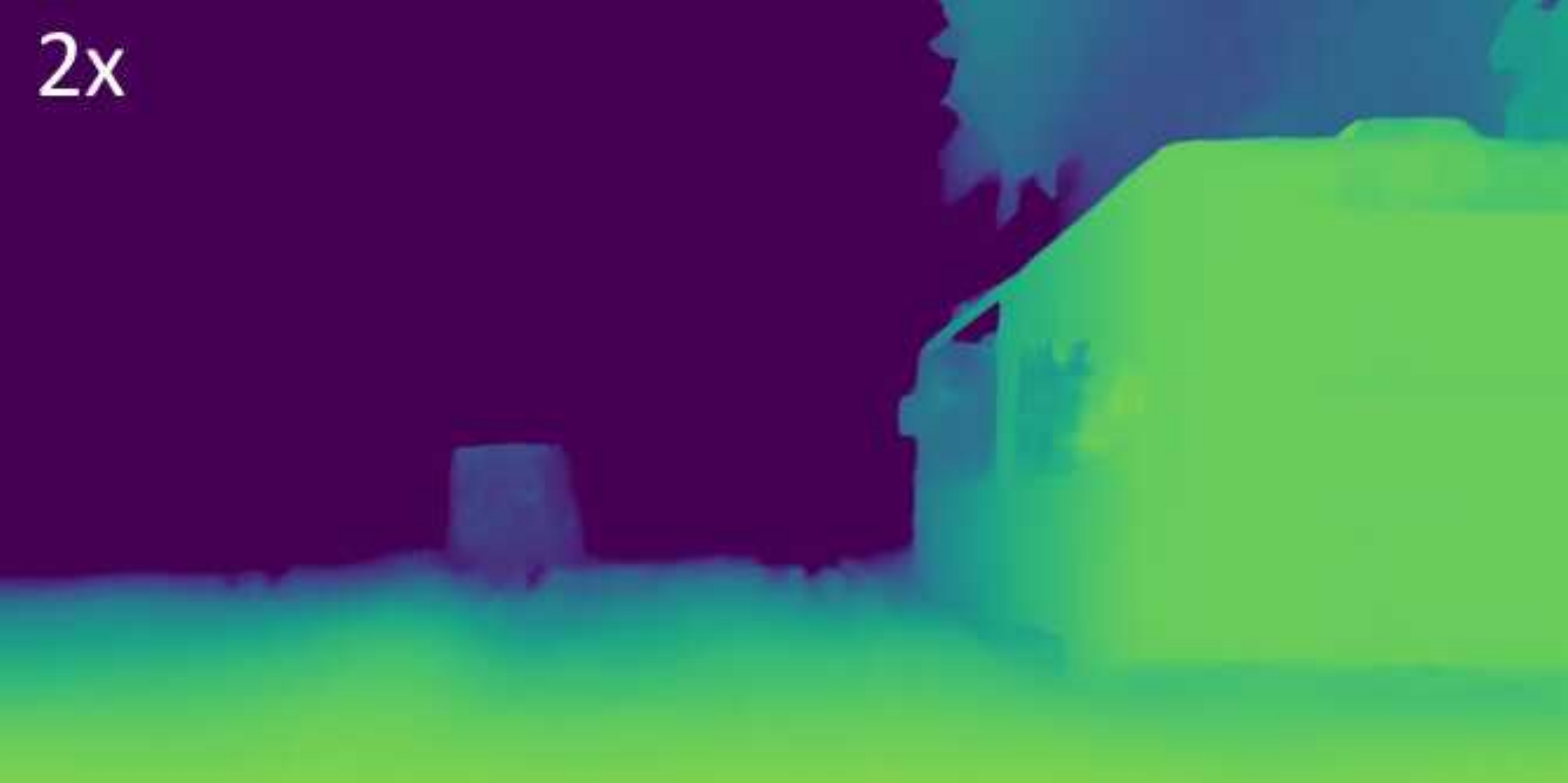}
    \end{subfigure}    \begin{subfigure}{.2\textwidth}
        \centering
        \includegraphics[width=.98\linewidth]{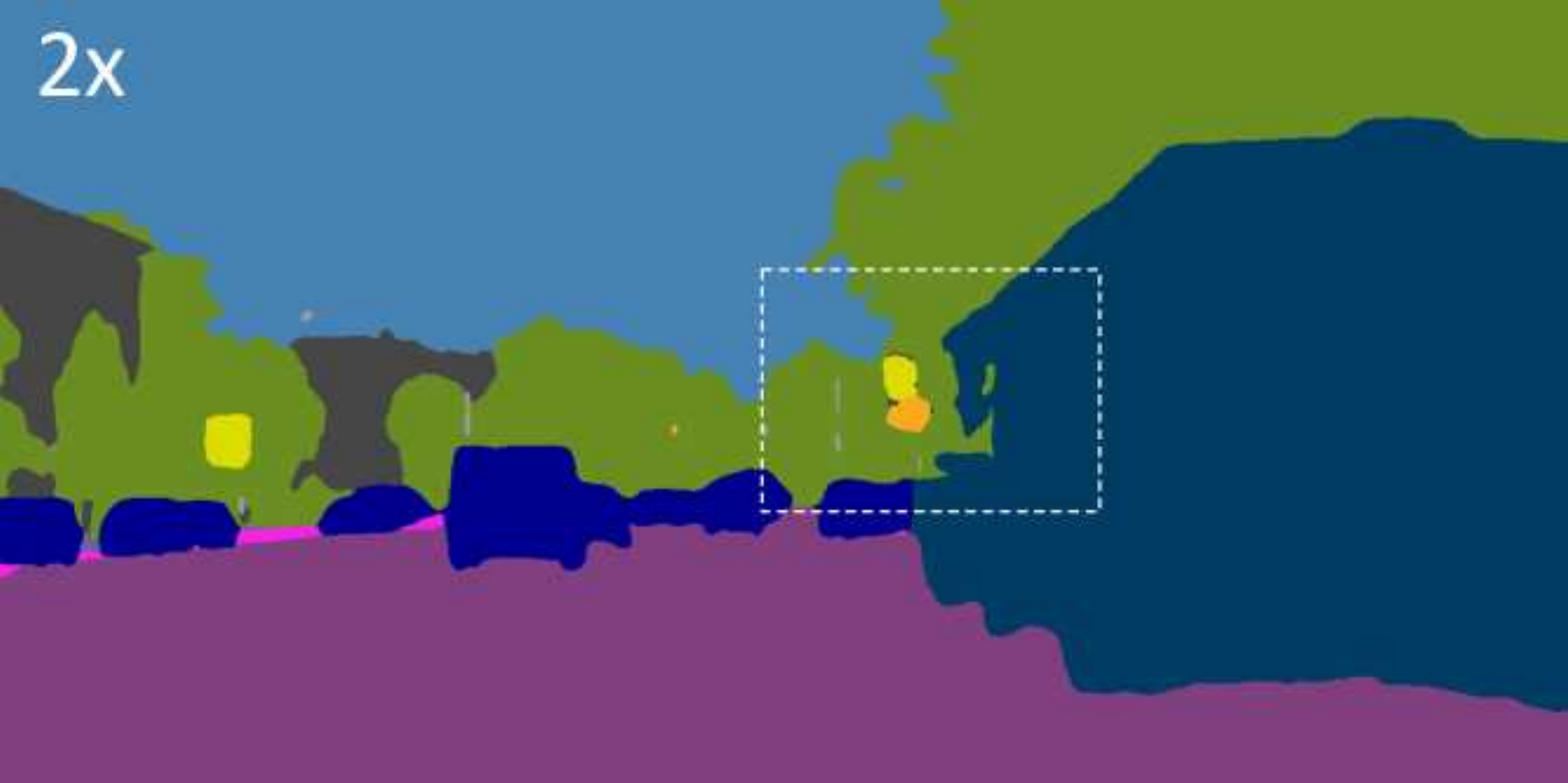}
    \end{subfigure}    \begin{subfigure}{.2\textwidth}
        \centering
        \includegraphics[width=.98\linewidth]{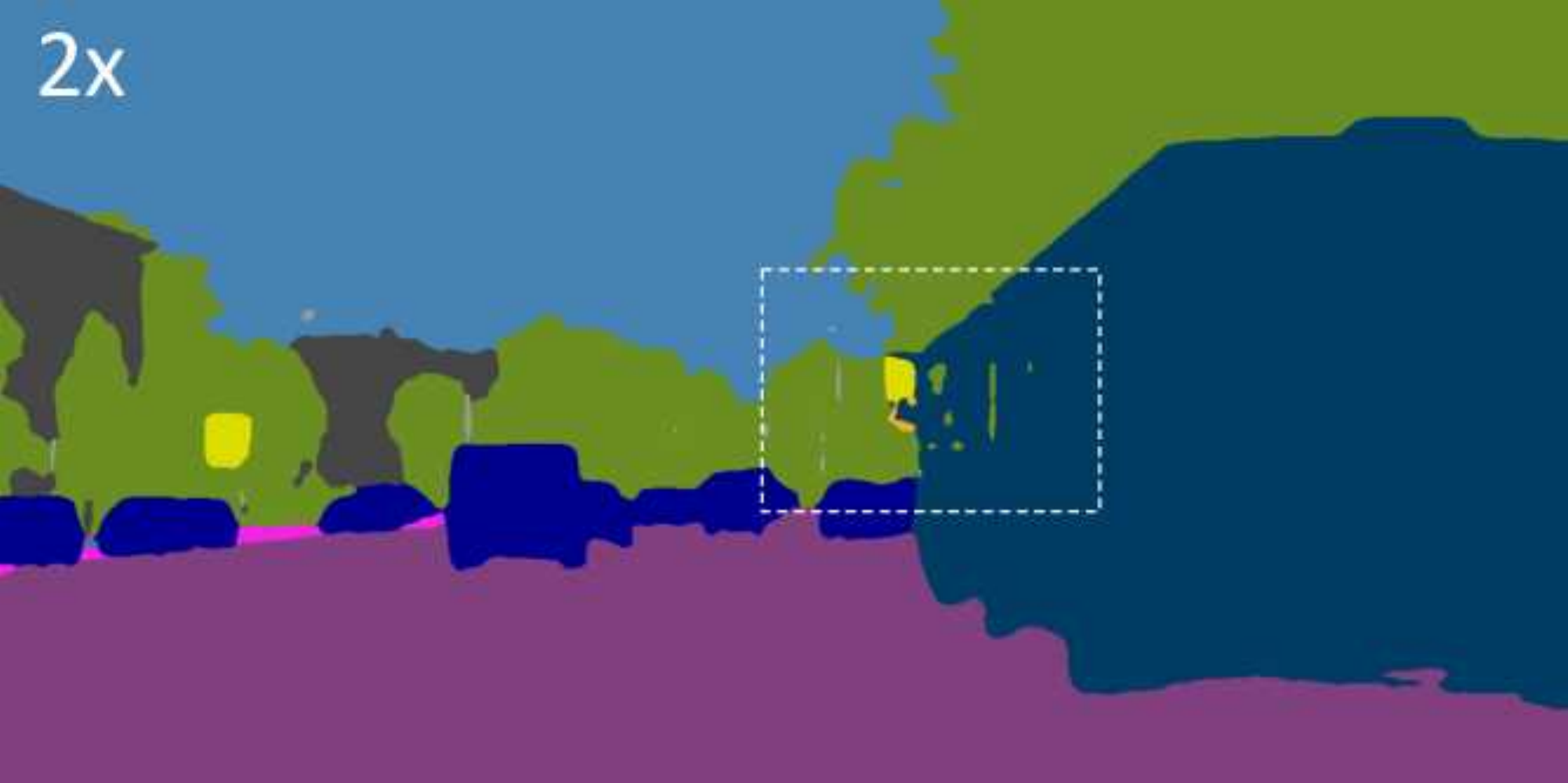}
    \end{subfigure}    \begin{subfigure}{.2\textwidth}
        \centering
        \includegraphics[width=.98\linewidth]{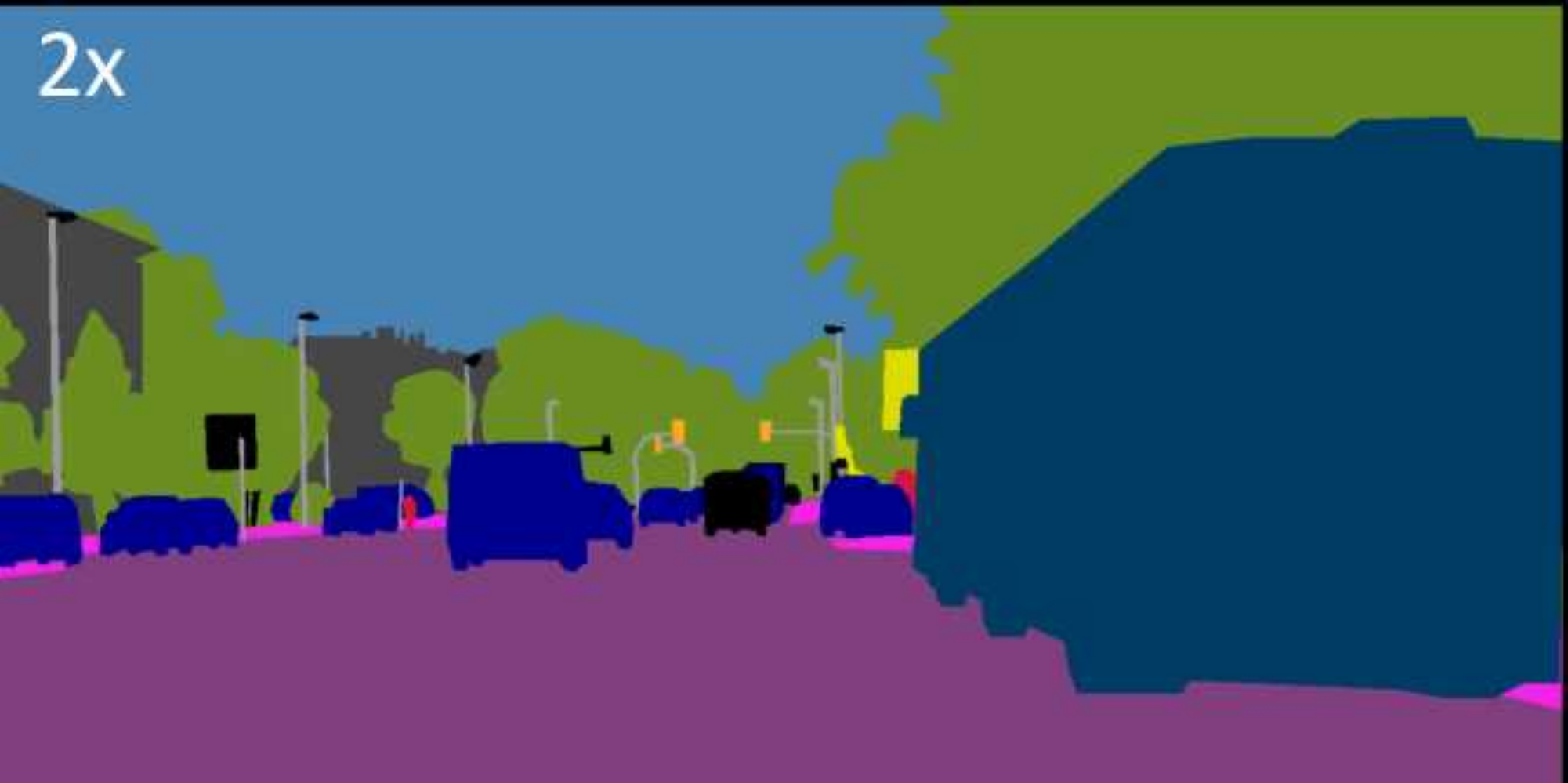}
    \end{subfigure}}
\makebox[\linewidth][c]{    \begin{subfigure}{.2\textwidth}
        \centering
        \includegraphics[width=.98\linewidth]{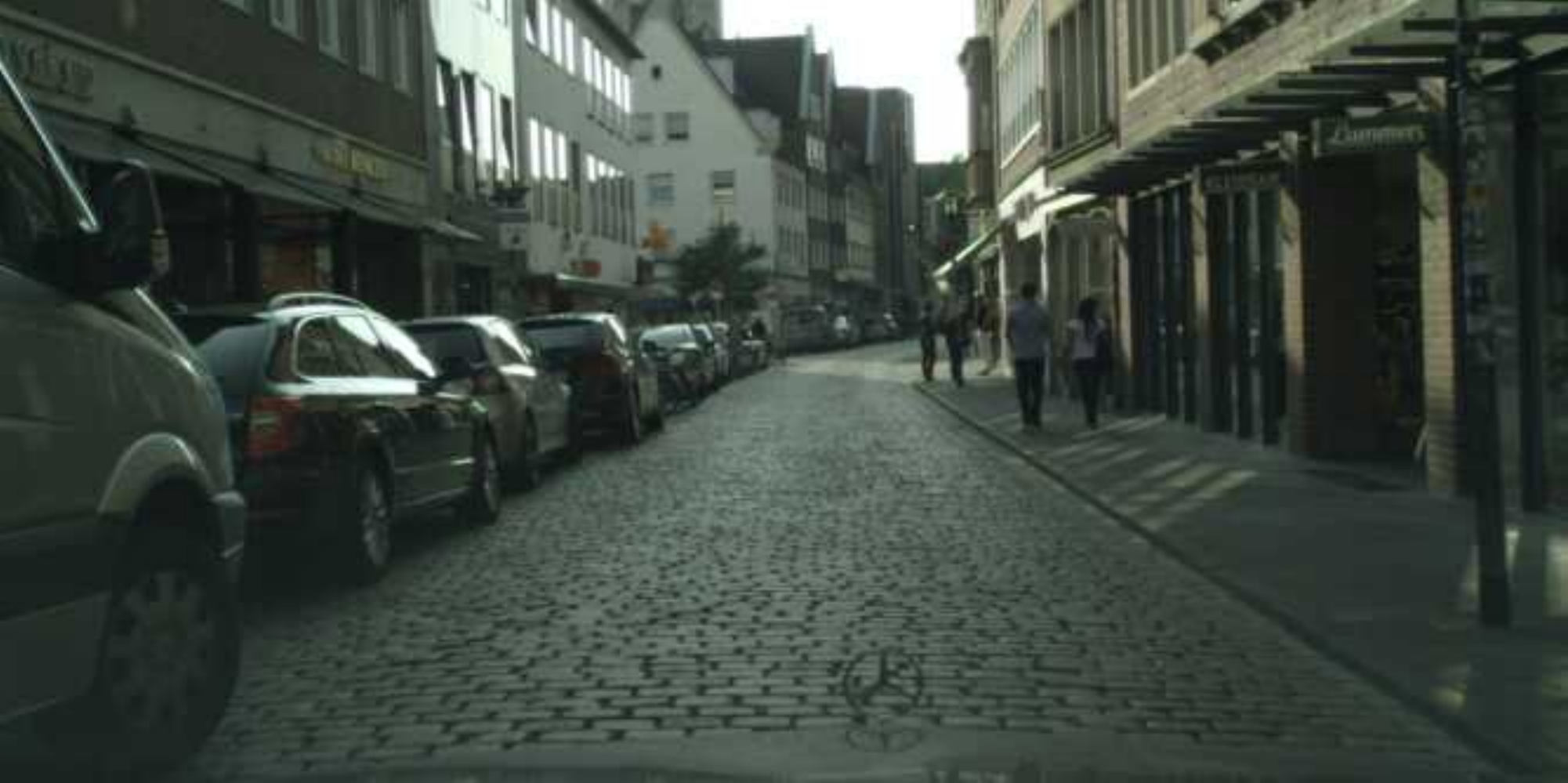}
        \caption*{Target Image}
    \end{subfigure}    \begin{subfigure}{.2\textwidth}
        \centering
        \includegraphics[width=.98\linewidth]        {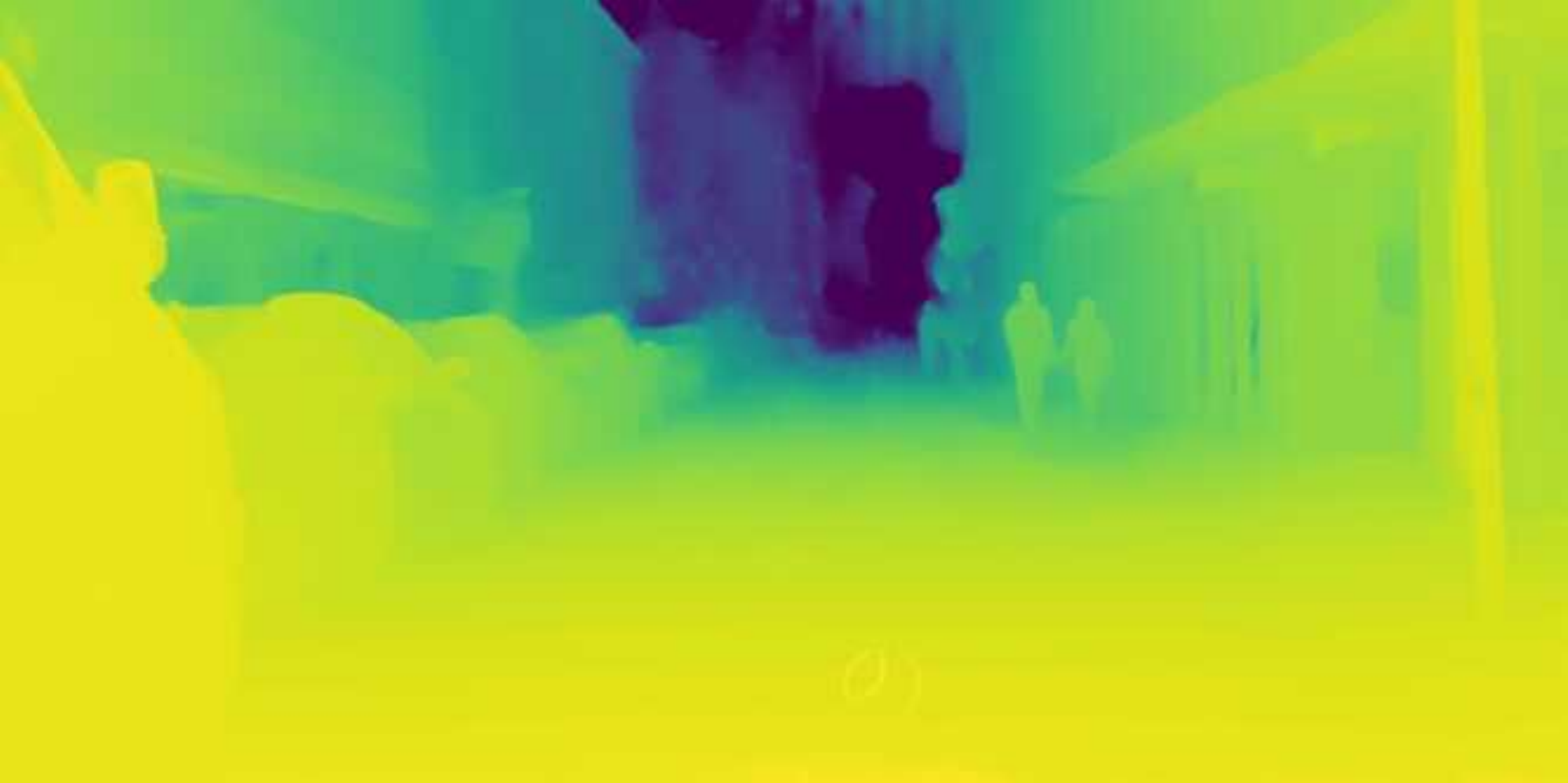}
        \caption*{Estimated Depth}
    \end{subfigure}    \begin{subfigure}{.2\textwidth}
        \centering
        \includegraphics[width=.98\linewidth]{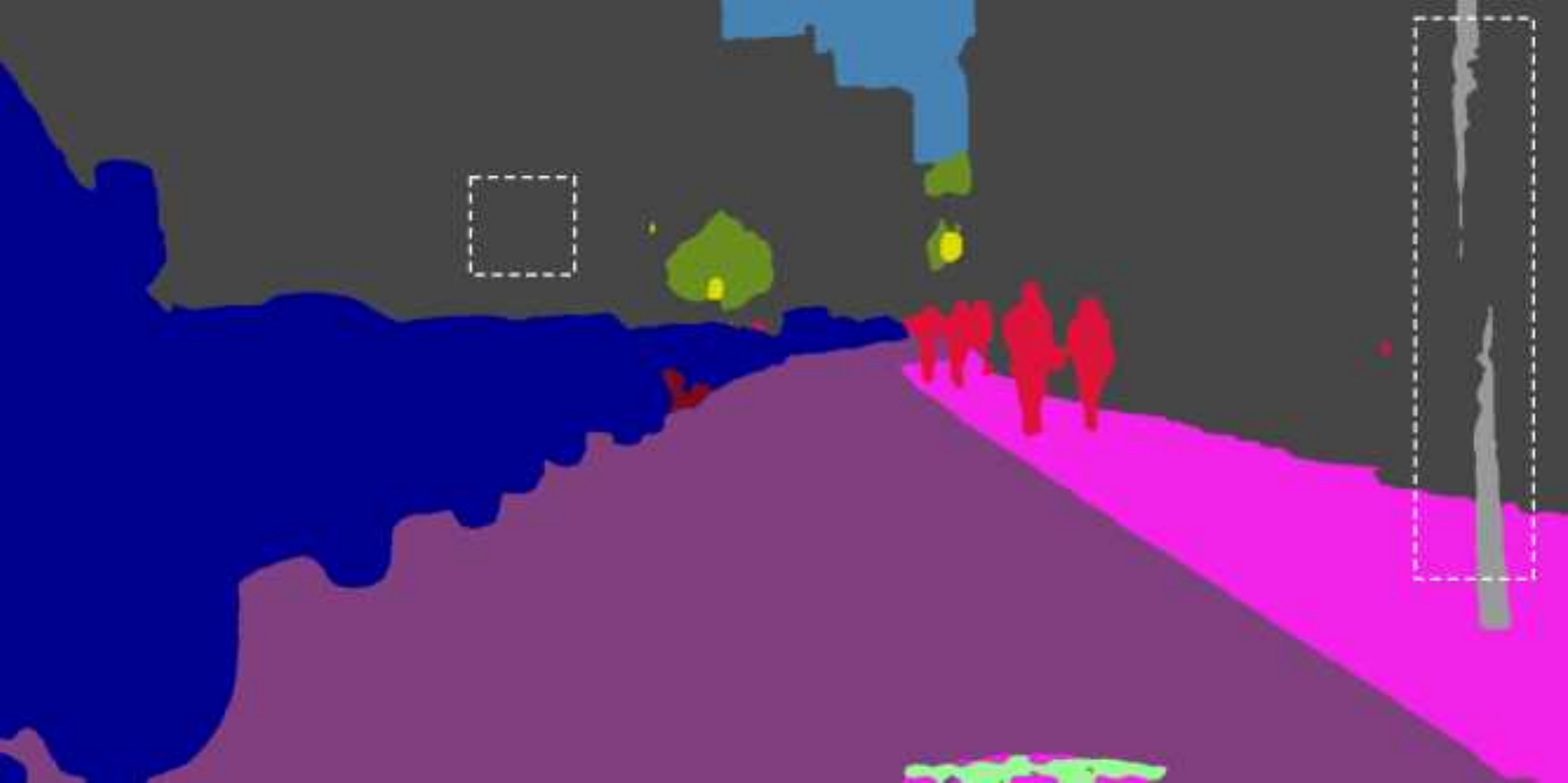}
        \caption*{MIC(DAFormer)~\cite{MIC}}
    \end{subfigure}    \begin{subfigure}{.2\textwidth}
        \centering
        \includegraphics[width=.98\linewidth]{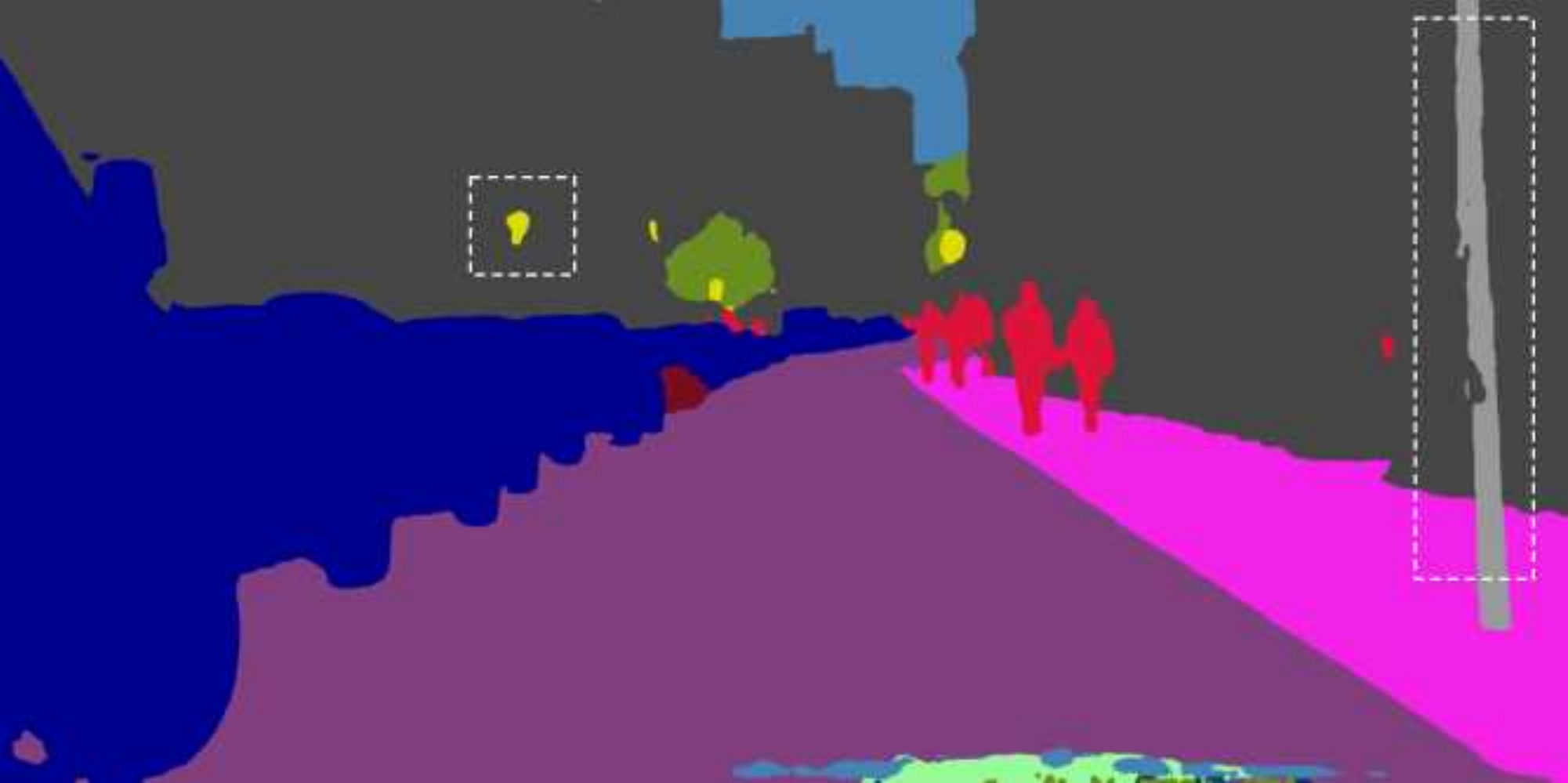}
        \caption*{\method\ (ours)}
    \end{subfigure}    \begin{subfigure}{.2\textwidth}
        \centering
        \includegraphics[width=.98\linewidth]{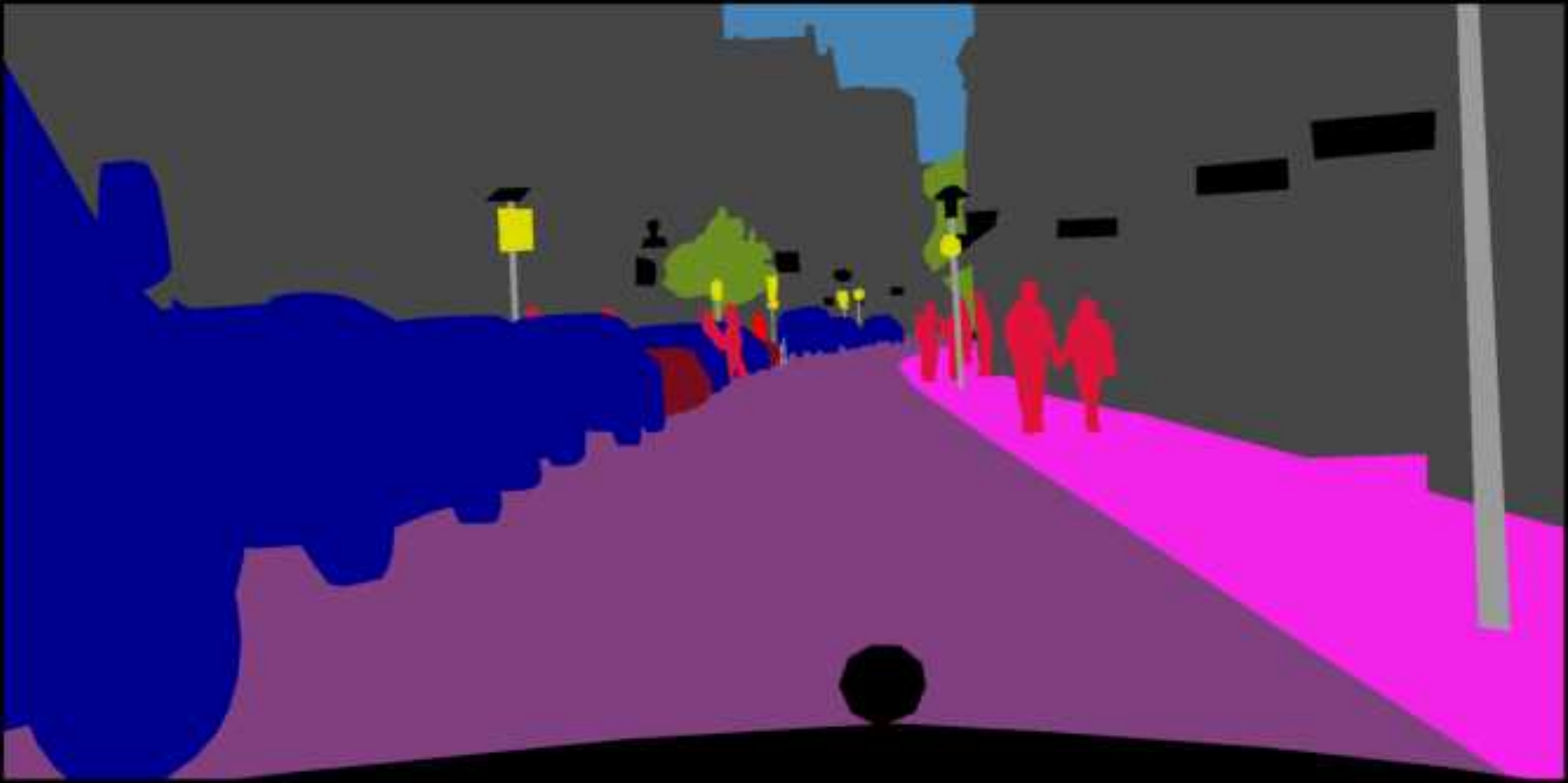}
        \caption*{Ground Truth}
    \end{subfigure}}

\centering

\caption{\textbf{Qualitative results using MIC (DAFormer)~\cite{MIC}.}
In rows 1, 2, and 6, depth cues correct mislabeled patches as part of larger structures.
Row 4 obtains more accurate \textit{traffic sign} segmentation with consistent depth.
In rows 3, 5, and 7, correct segmentation of thin \textit{poles} is enabled through depth information.
}
\label{fig:qualitative_supp_mic_daf}
\end{figure*}

\begin{figure*}
\makebox[\linewidth][c]{    \begin{subfigure}{\linewidth}
        \centering
        \label{fig:palette}
        \includegraphics[width=\linewidth]{figures/color_palette.png}
    \end{subfigure}}
\makebox[\linewidth][c]{    \begin{subfigure}{.2\textwidth}
        \centering
        \includegraphics[width=.98\linewidth]{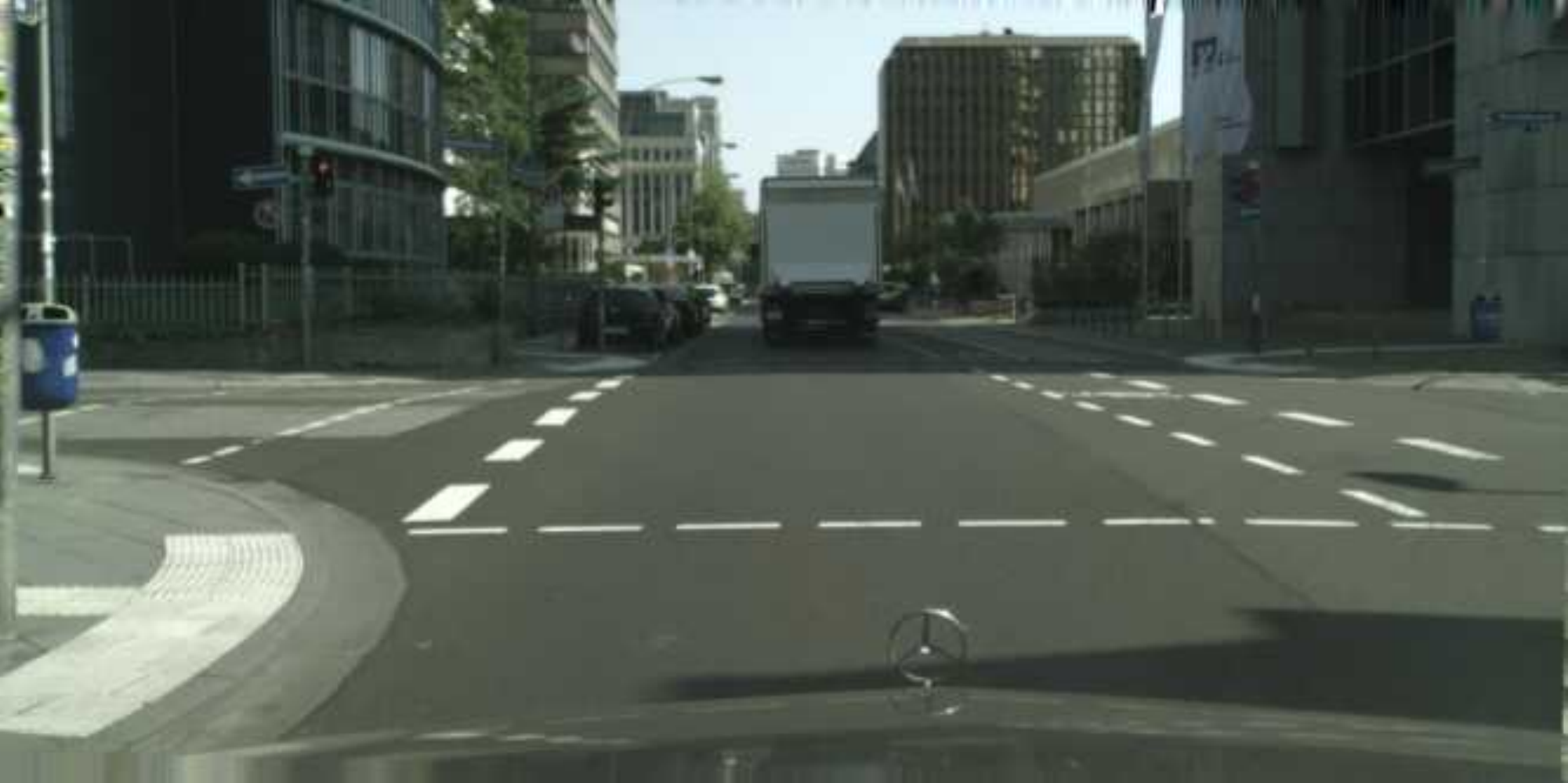}
    \end{subfigure}    \begin{subfigure}{.2\textwidth}
        \centering
        \includegraphics[width=.98\linewidth]{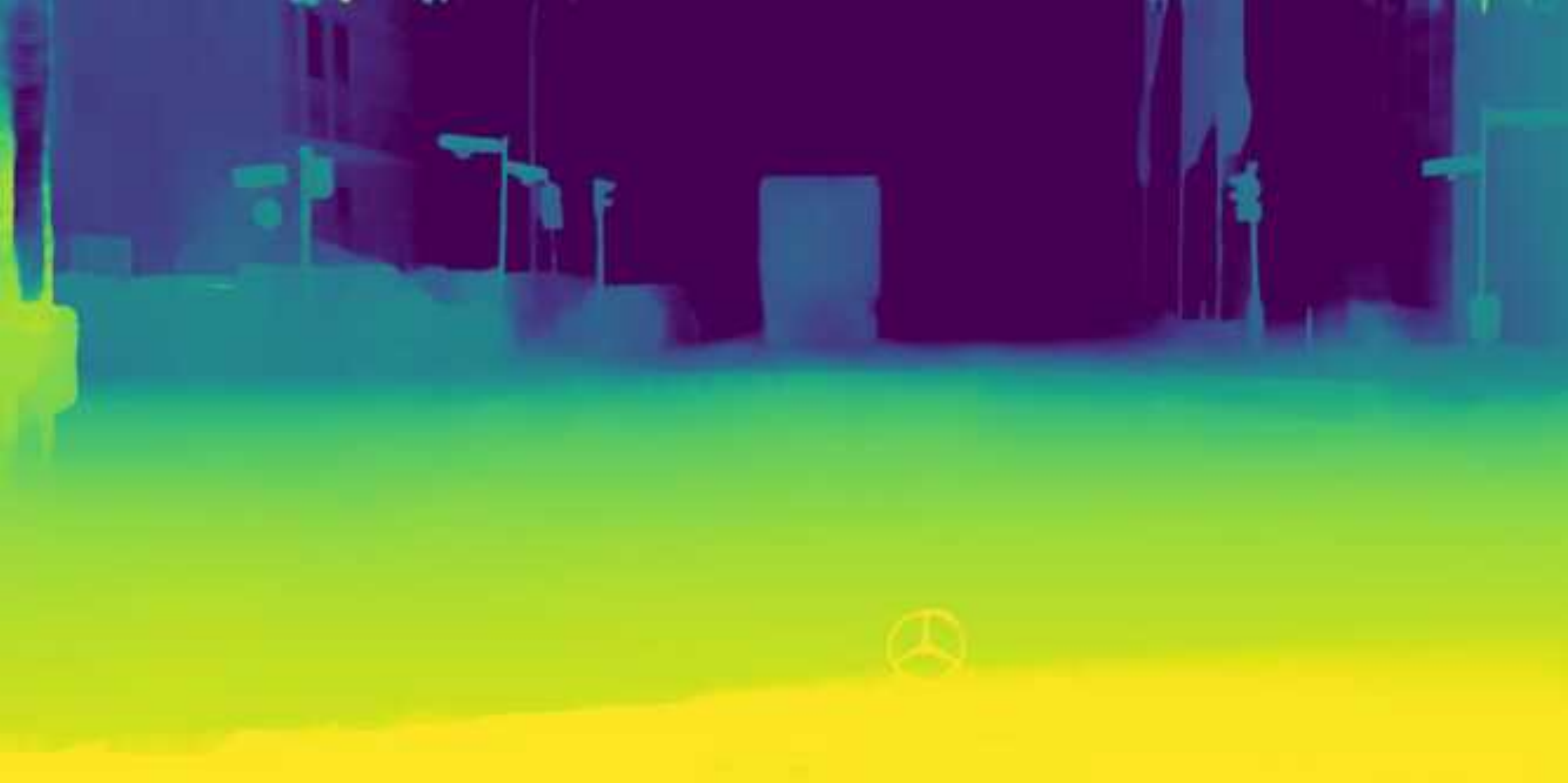}
    \end{subfigure}    \begin{subfigure}{.2\textwidth}
        \centering
        \includegraphics[width=.98\linewidth]{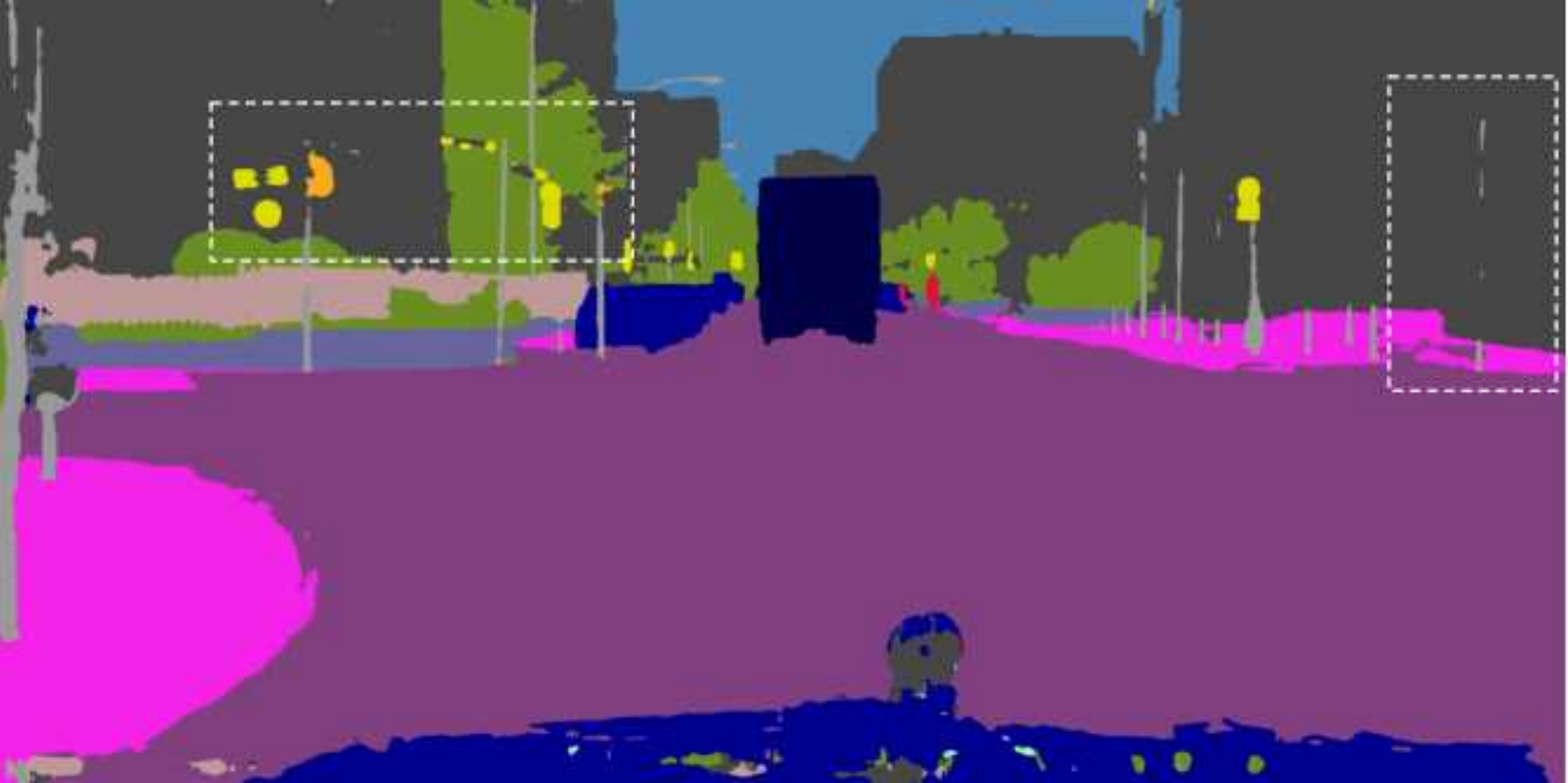}
    \end{subfigure}    \begin{subfigure}{.2\textwidth}
        \centering
        \includegraphics[width=.98\linewidth]{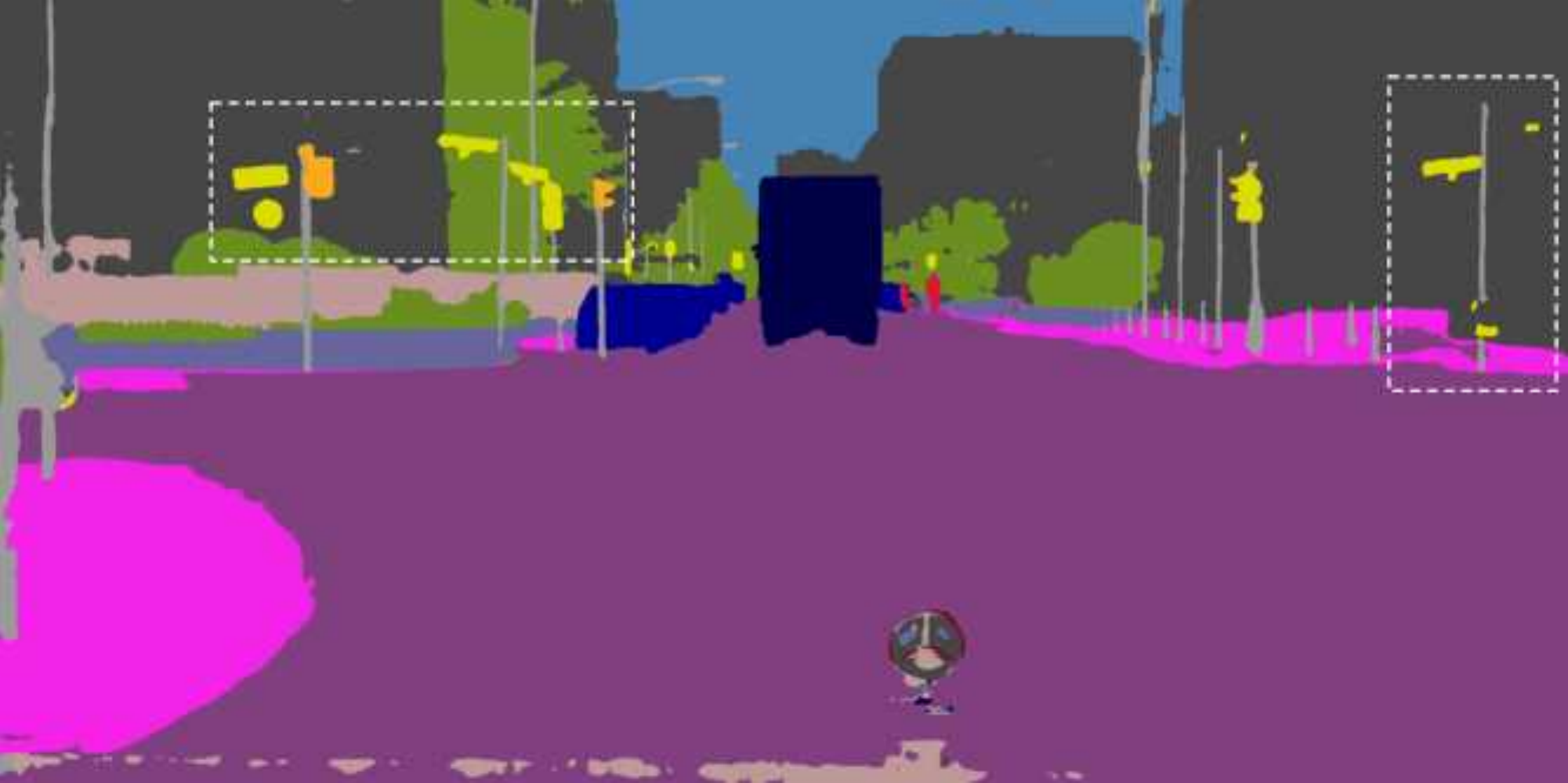}
    \end{subfigure}    \begin{subfigure}{.2\textwidth}
        \centering
        \includegraphics[width=.98\linewidth]{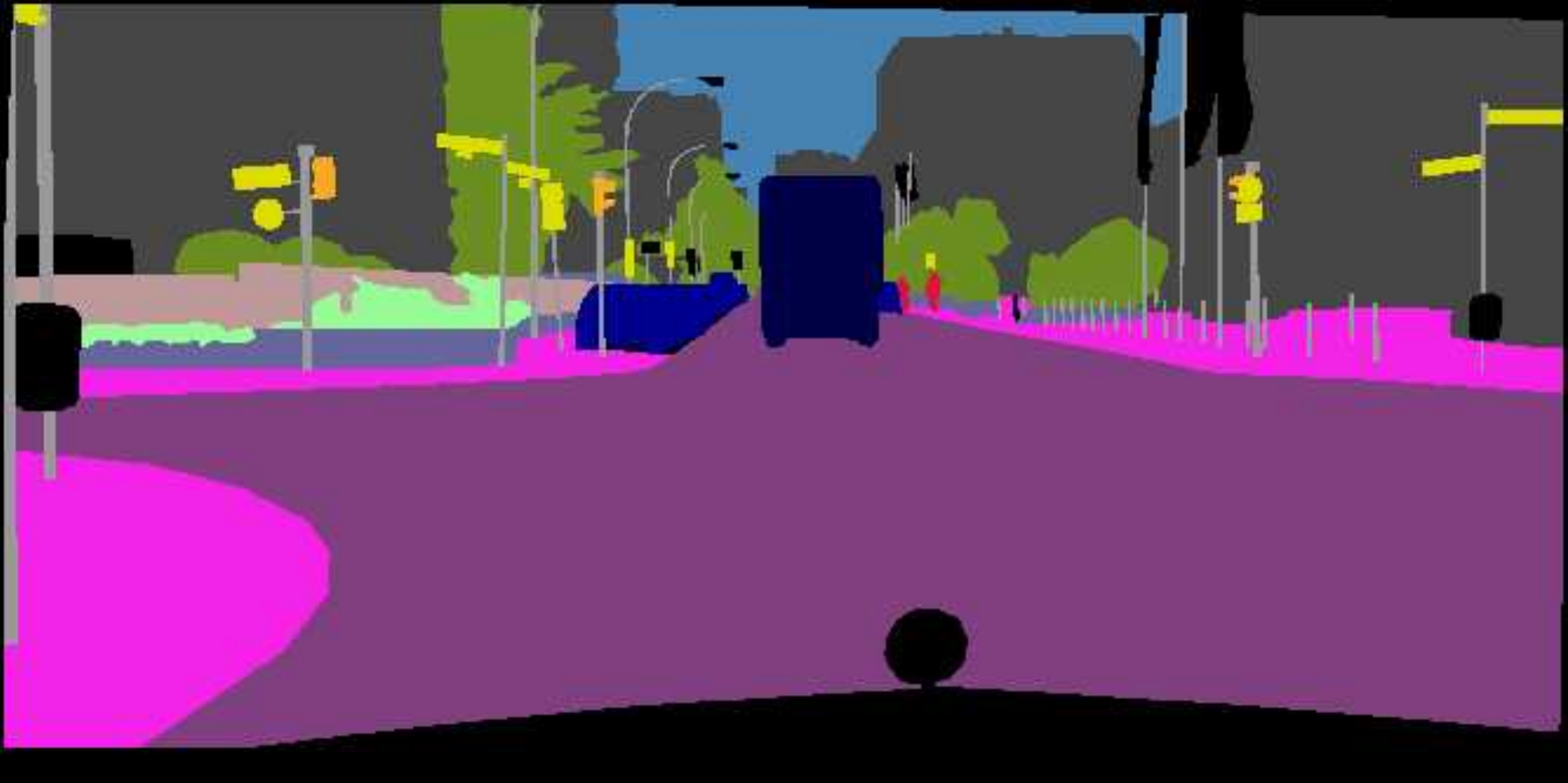}
    \end{subfigure}}
\makebox[\linewidth][c]{    \begin{subfigure}{.2\textwidth}
        \centering
        \includegraphics[width=.98\linewidth]{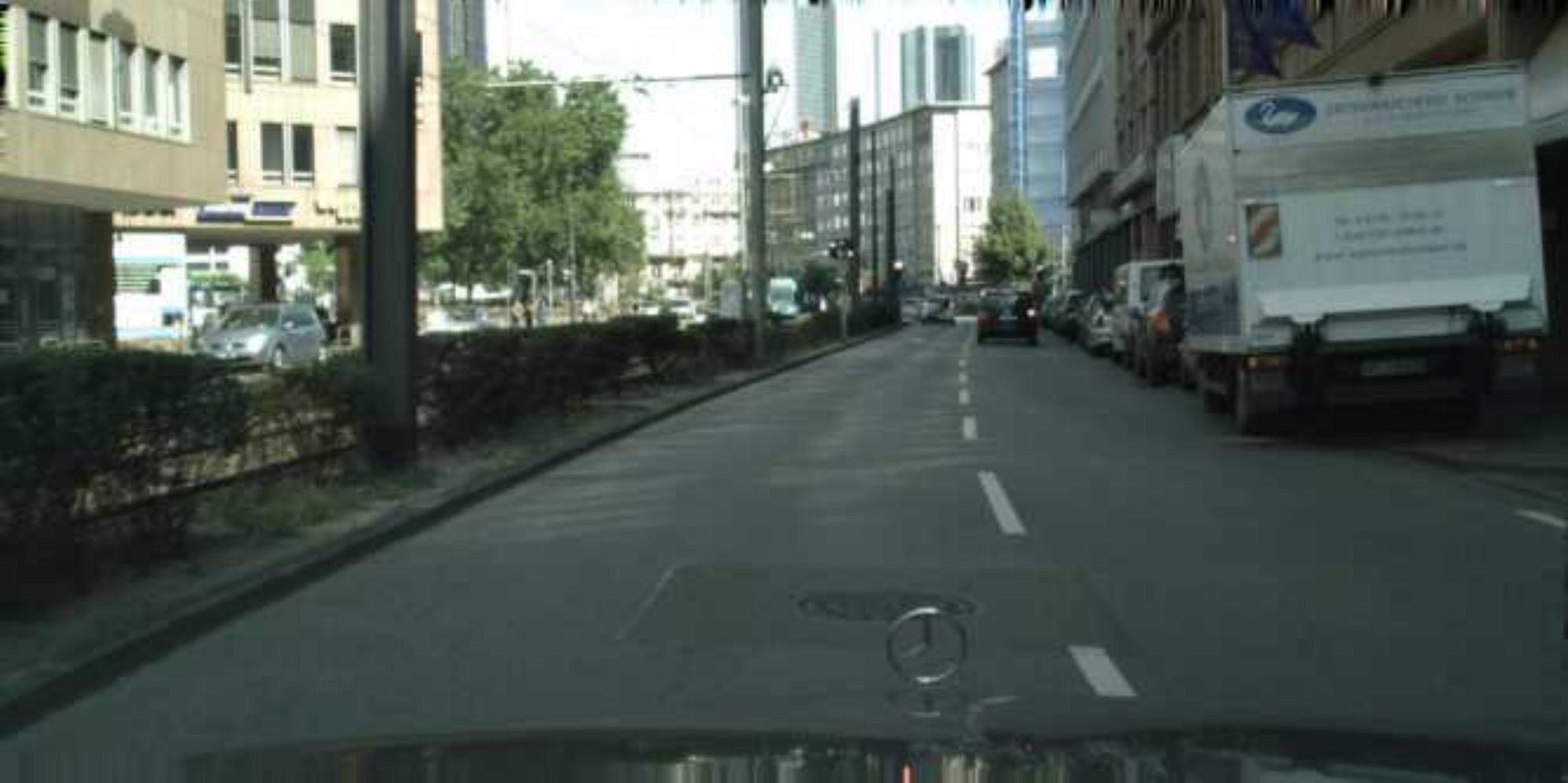}
    \end{subfigure}    \begin{subfigure}{.2\textwidth}
        \centering
        \includegraphics[width=.98\linewidth]{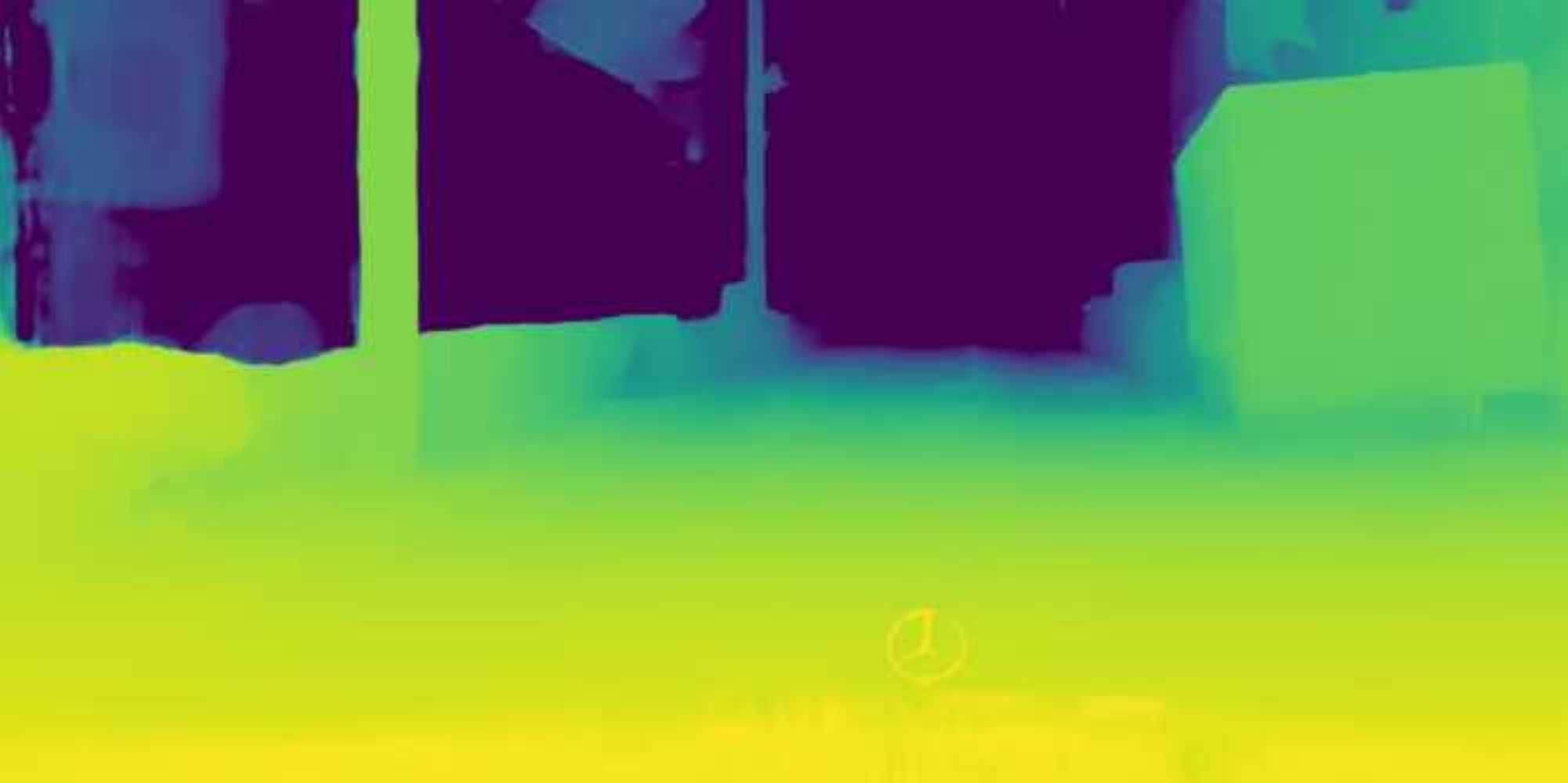}
    \end{subfigure}    \begin{subfigure}{.2\textwidth}
        \centering
        \includegraphics[width=.98\linewidth]{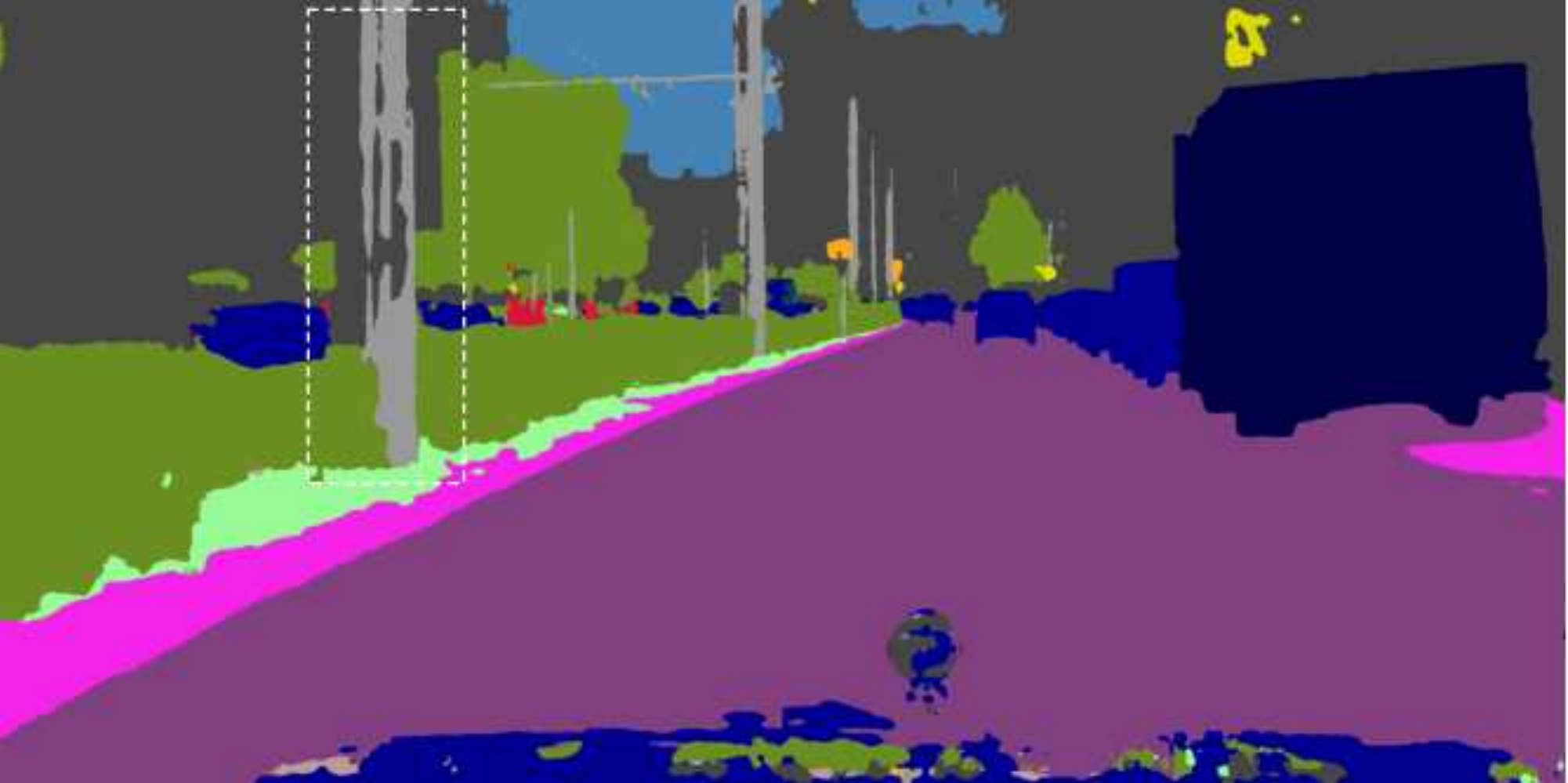}
    \end{subfigure}    \begin{subfigure}{.2\textwidth}
        \centering
        \includegraphics[width=.98\linewidth]{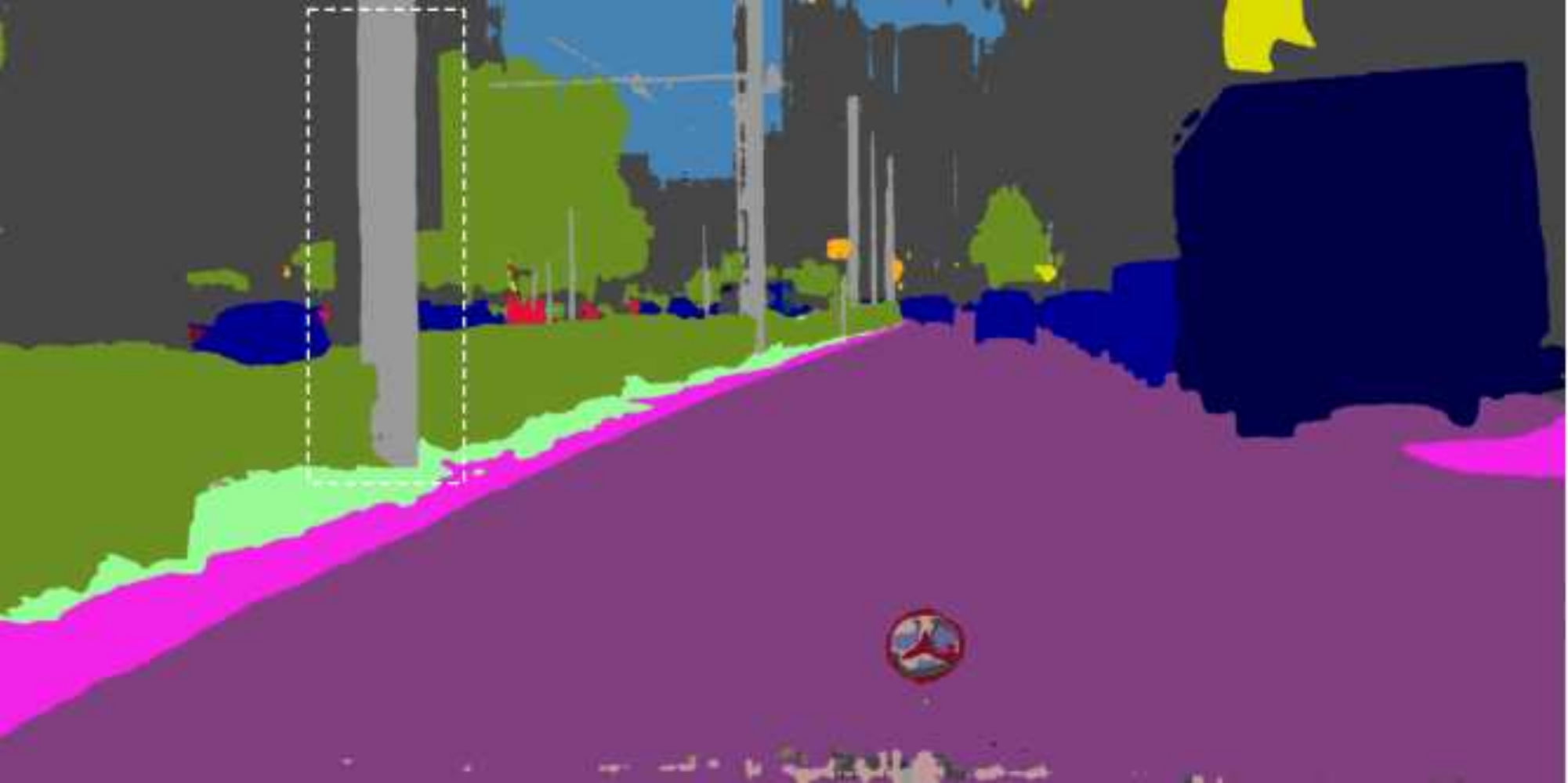}
    \end{subfigure}    \begin{subfigure}{.2\textwidth}
        \centering
        \includegraphics[width=.98\linewidth]{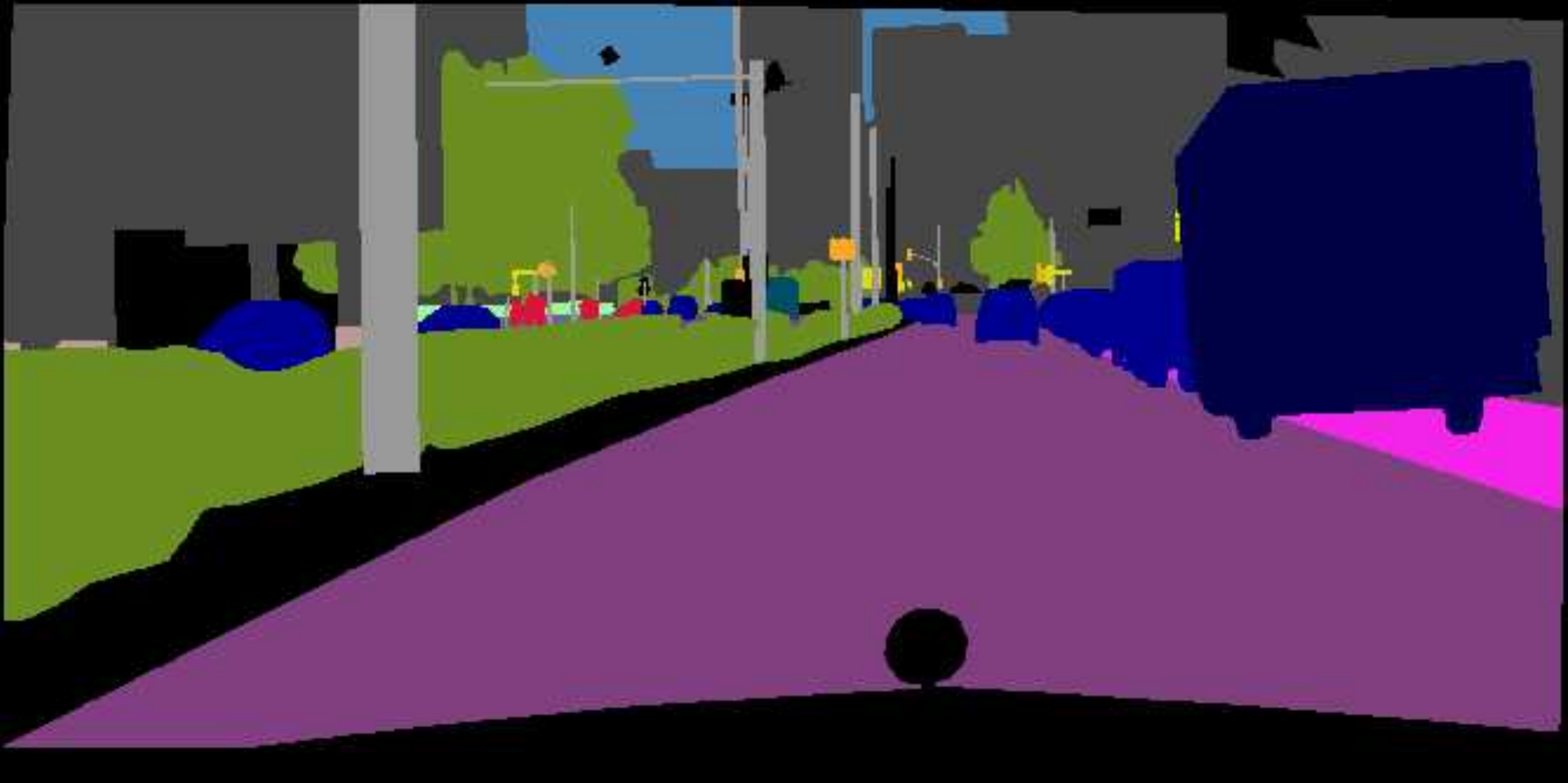}
    \end{subfigure}}
\makebox[\linewidth][c]{    \begin{subfigure}{.2\textwidth}
        \centering
        \includegraphics[width=.98\linewidth]{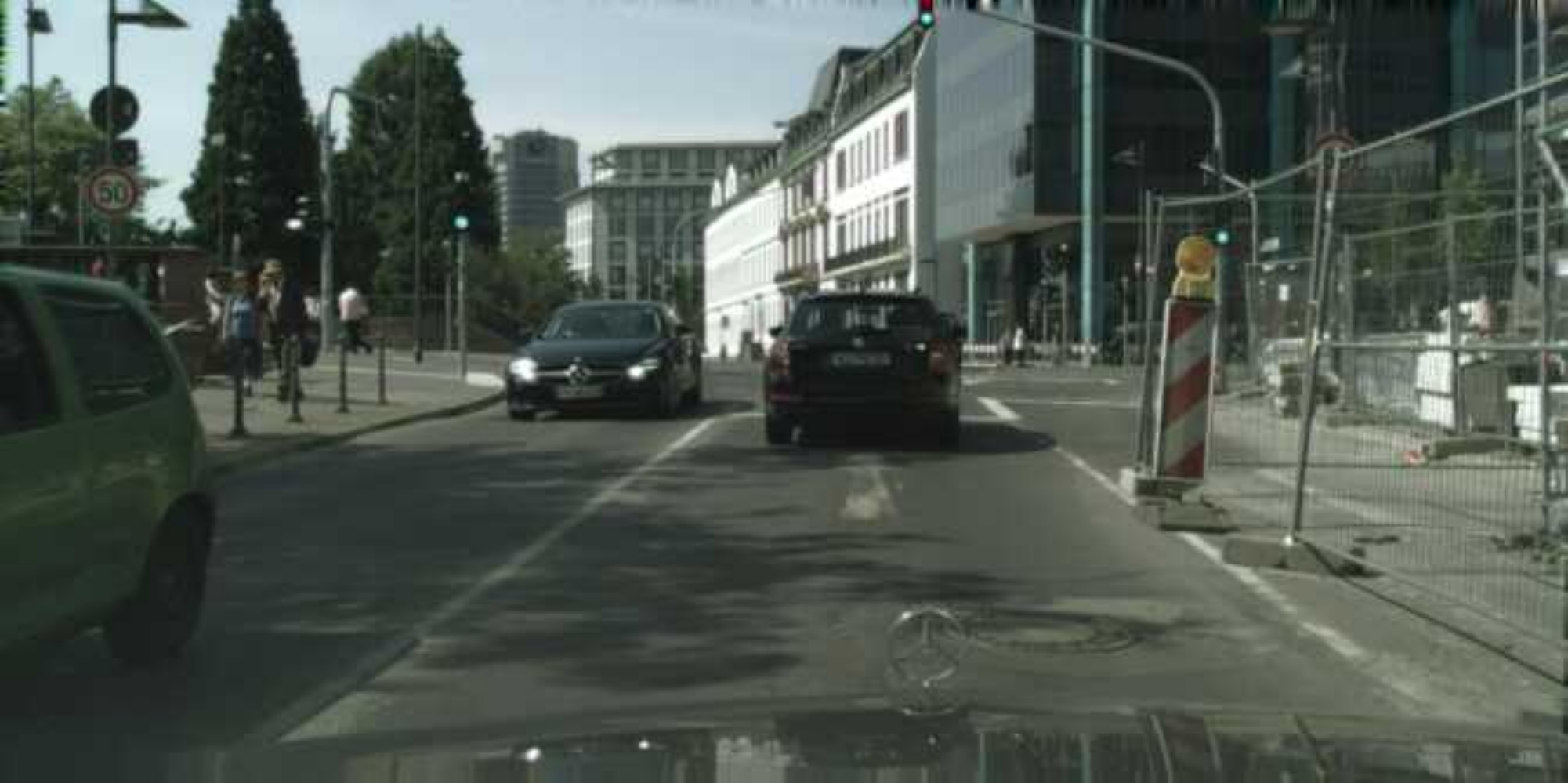}
    \end{subfigure}    \begin{subfigure}{.2\textwidth}
        \centering
        \includegraphics[width=.98\linewidth]{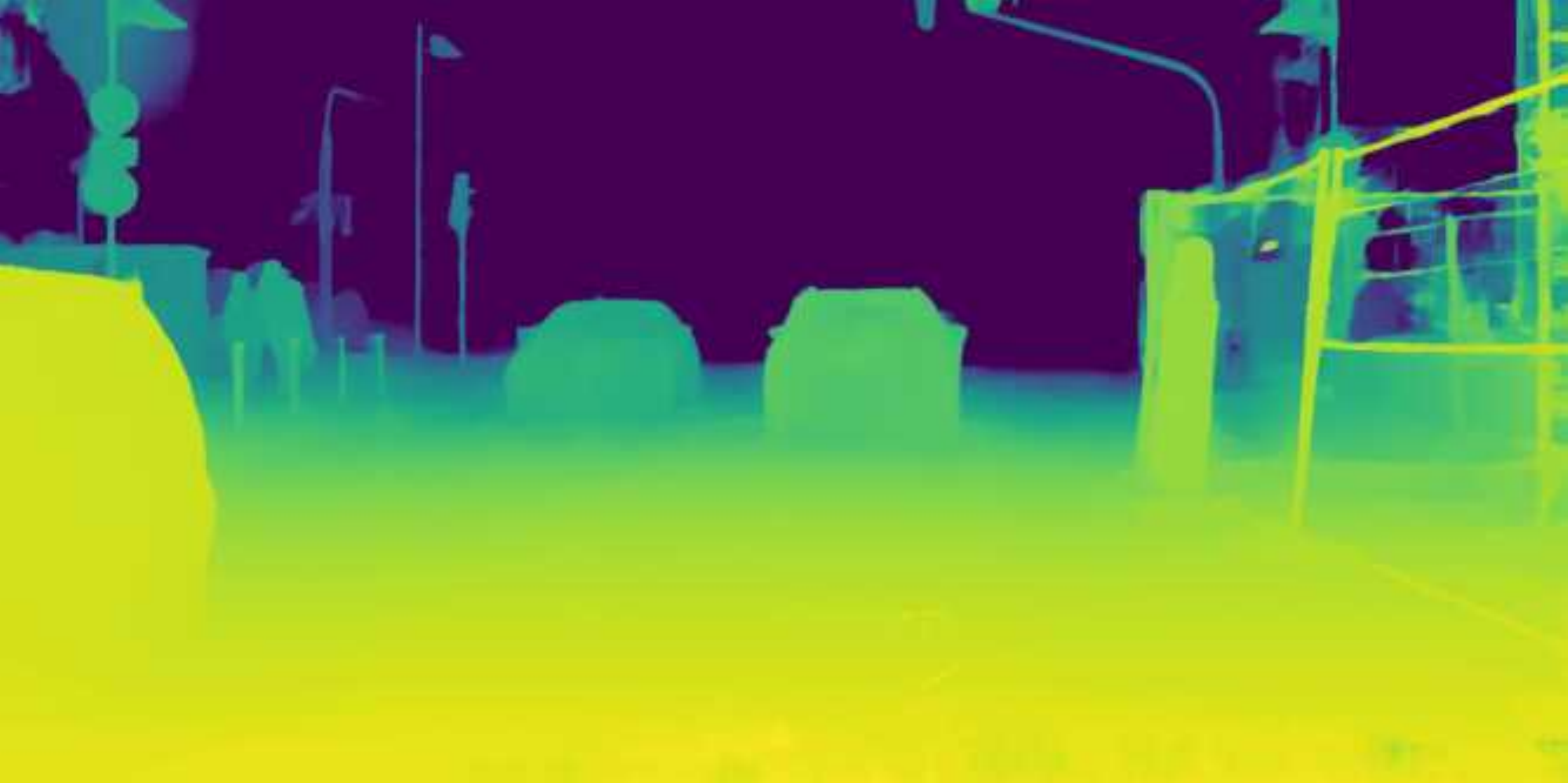}
    \end{subfigure}    \begin{subfigure}{.2\textwidth}
        \centering
        \includegraphics[width=.98\linewidth]{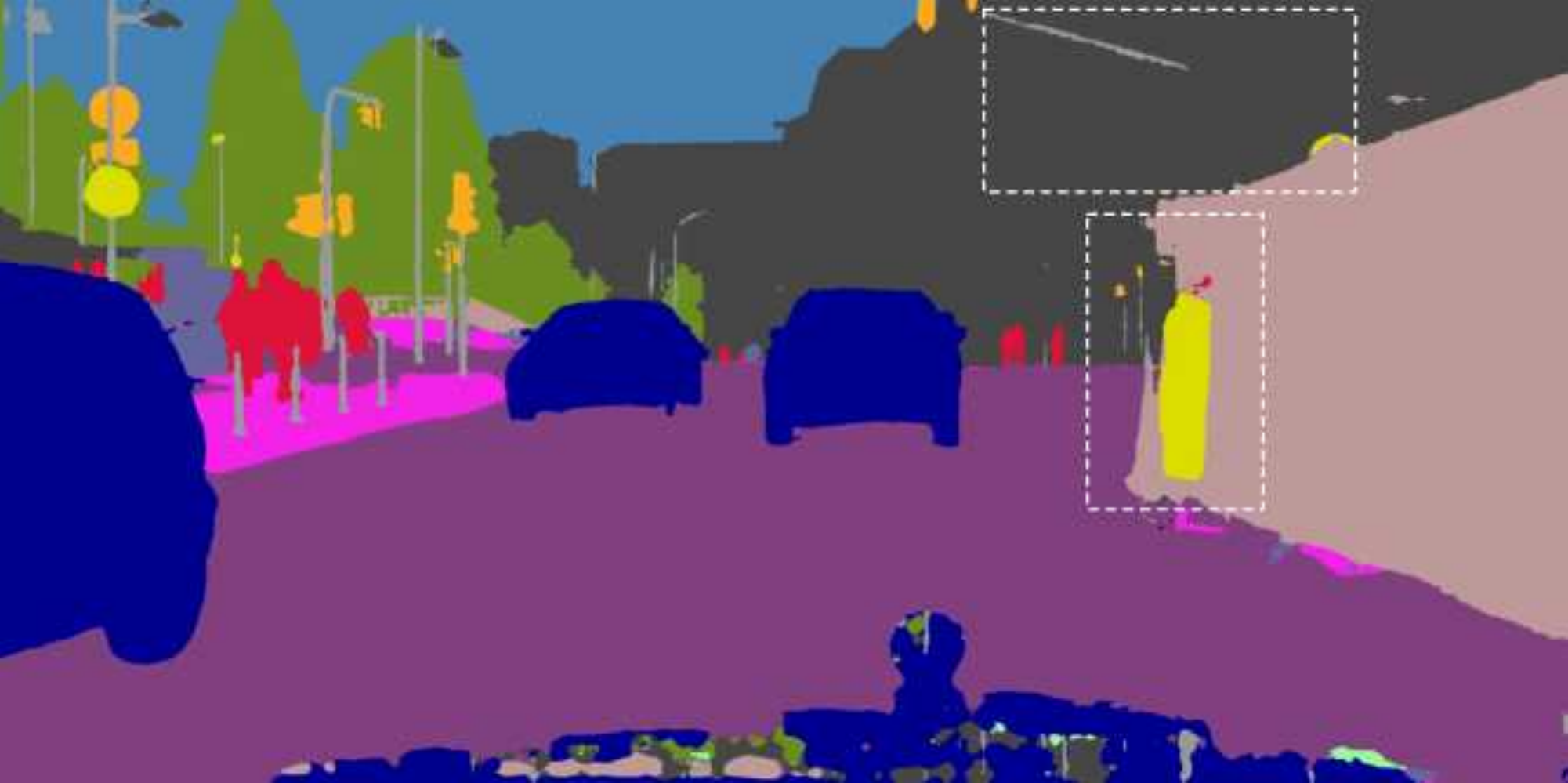}
    \end{subfigure}    \begin{subfigure}{.2\textwidth}
        \centering
        \includegraphics[width=.98\linewidth]{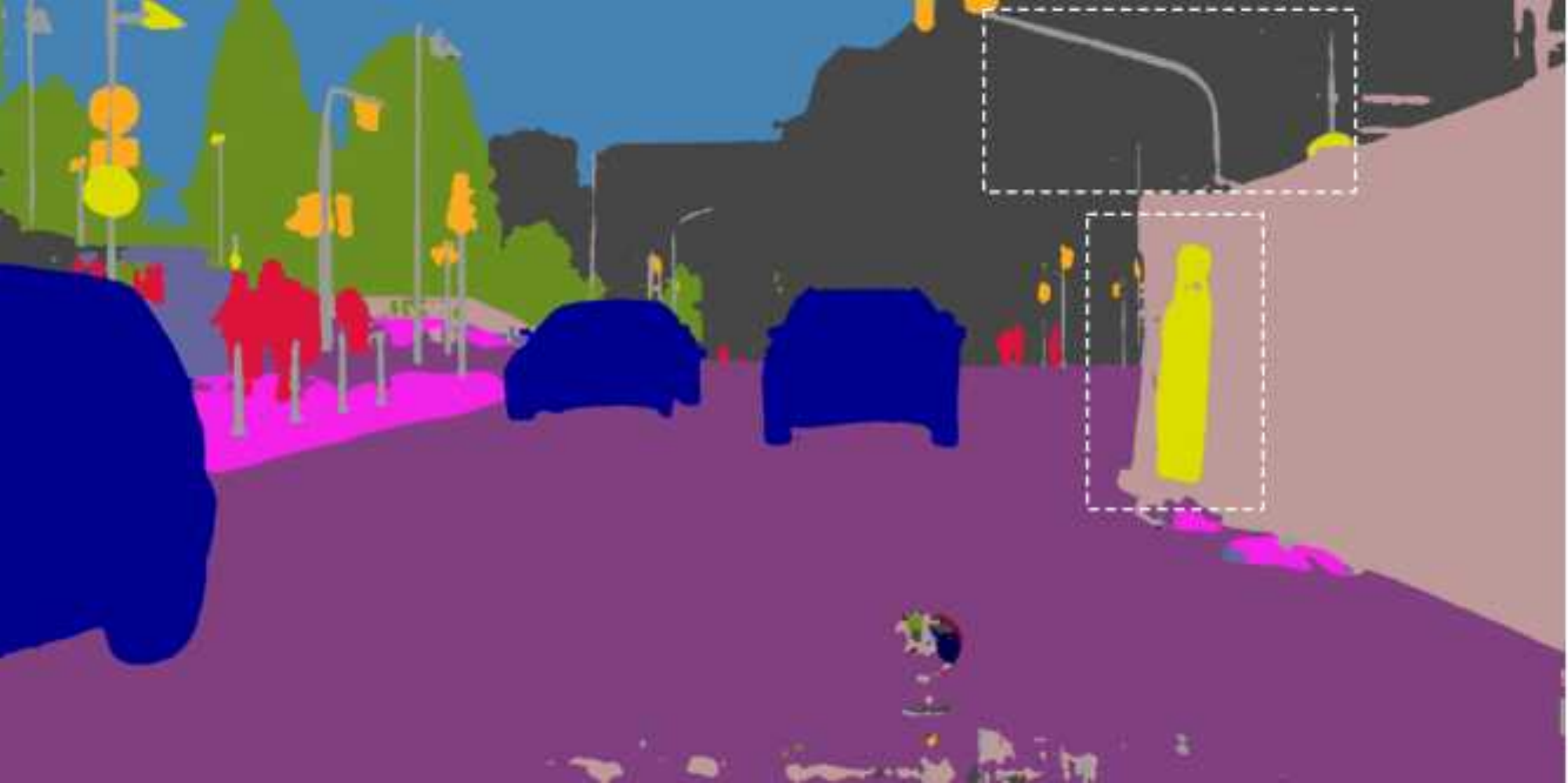}
    \end{subfigure}    \begin{subfigure}{.2\textwidth}
        \centering
        \includegraphics[width=.98\linewidth]{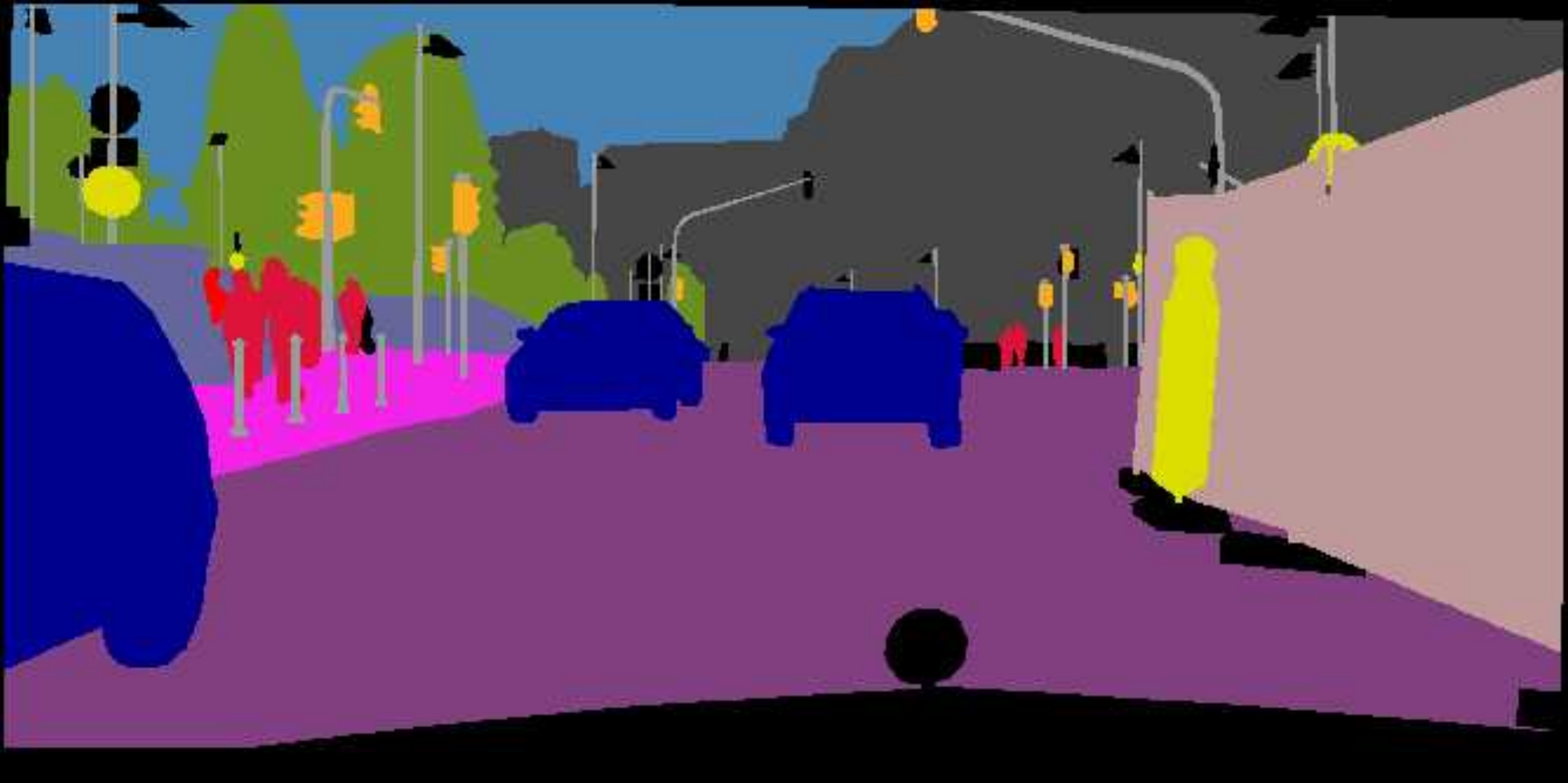}
    \end{subfigure}}
\makebox[\linewidth][c]{    \begin{subfigure}{.2\textwidth}
        \centering
        \includegraphics[width=.98\linewidth]{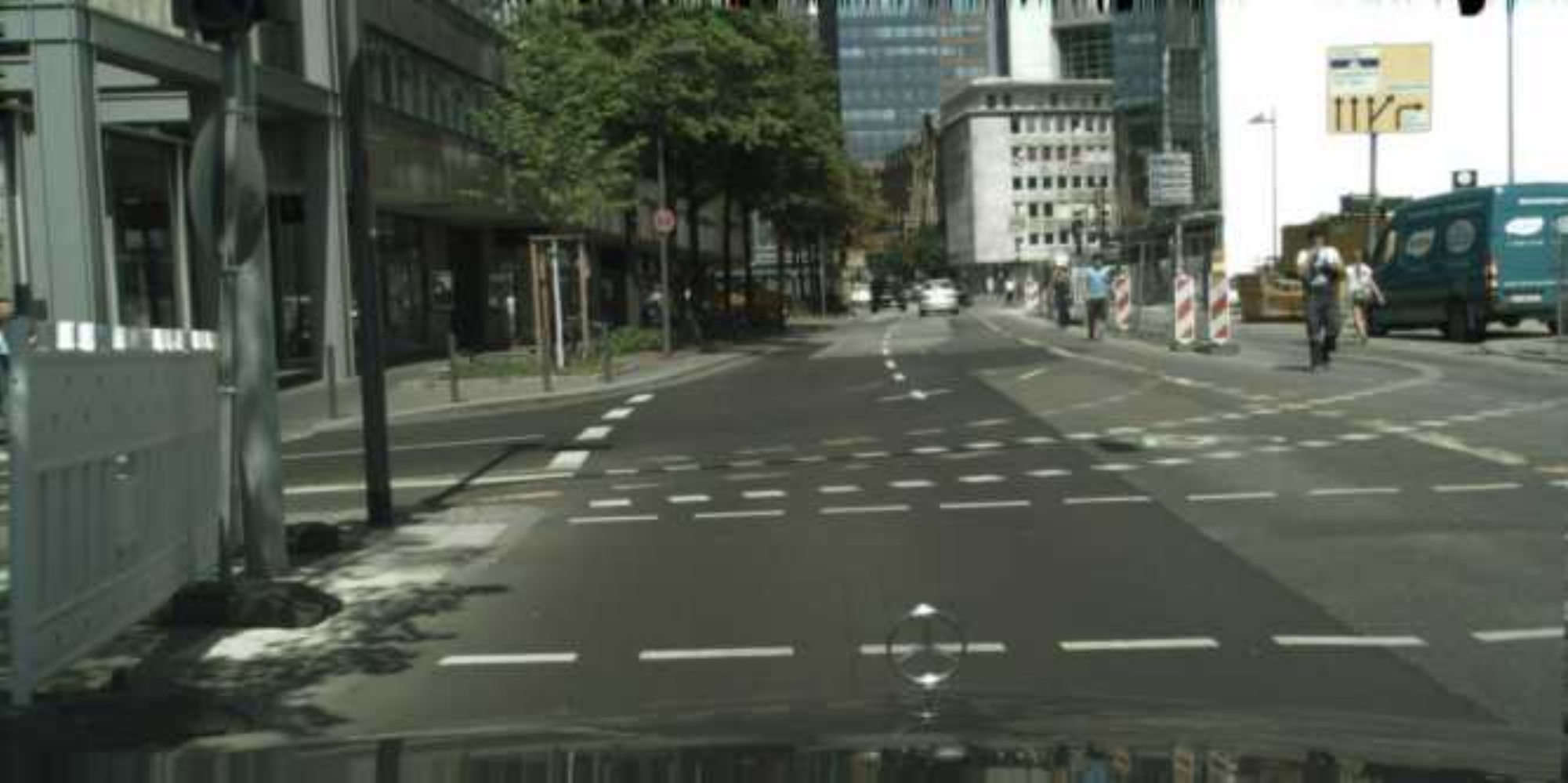}
    \end{subfigure}    \begin{subfigure}{.2\textwidth}
        \centering
        \includegraphics[width=.98\linewidth]{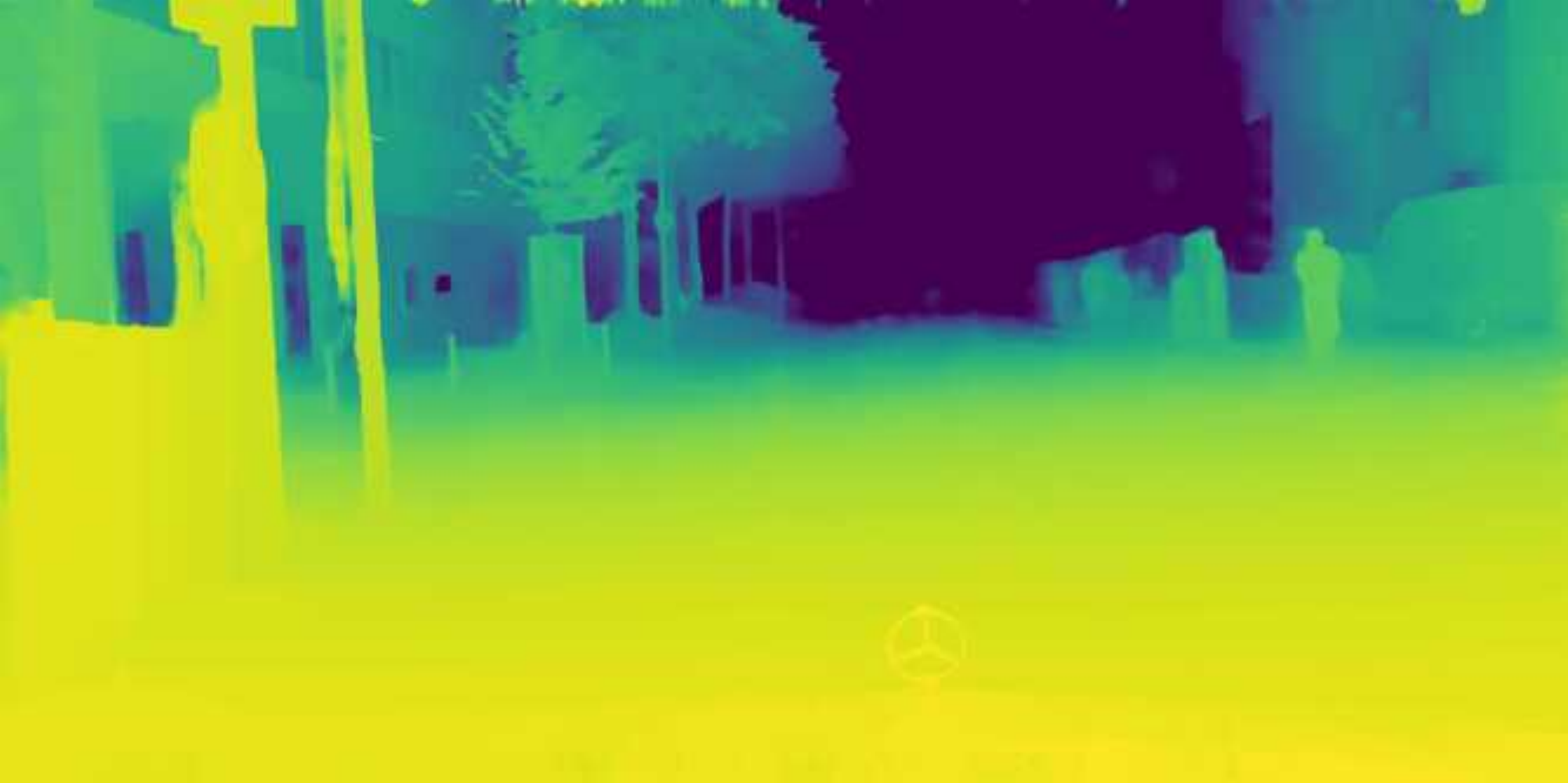}
    \end{subfigure}    \begin{subfigure}{.2\textwidth}
        \centering
        \includegraphics[width=.98\linewidth]{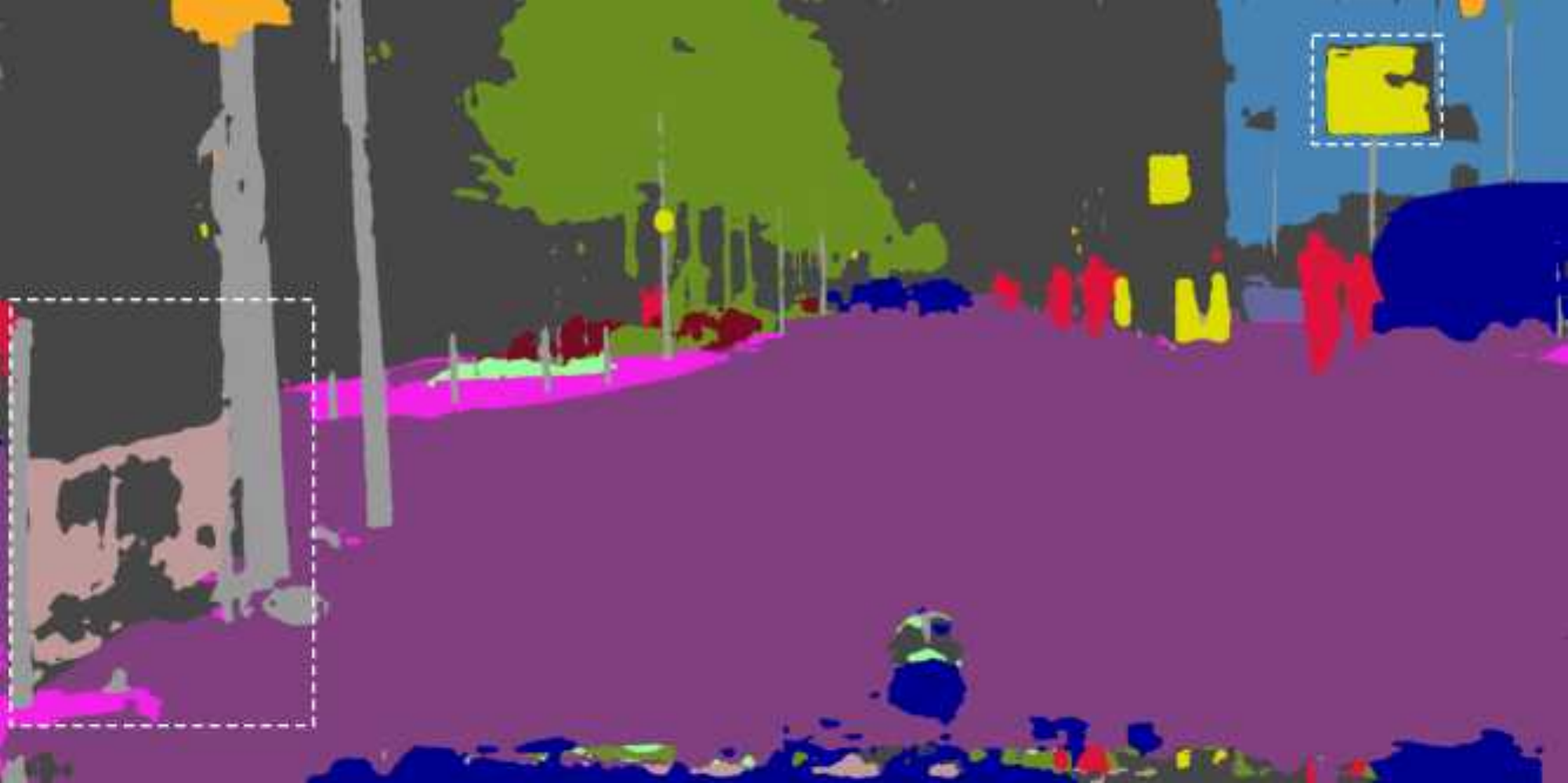}
    \end{subfigure}    \begin{subfigure}{.2\textwidth}
        \centering
        \includegraphics[width=.98\linewidth]{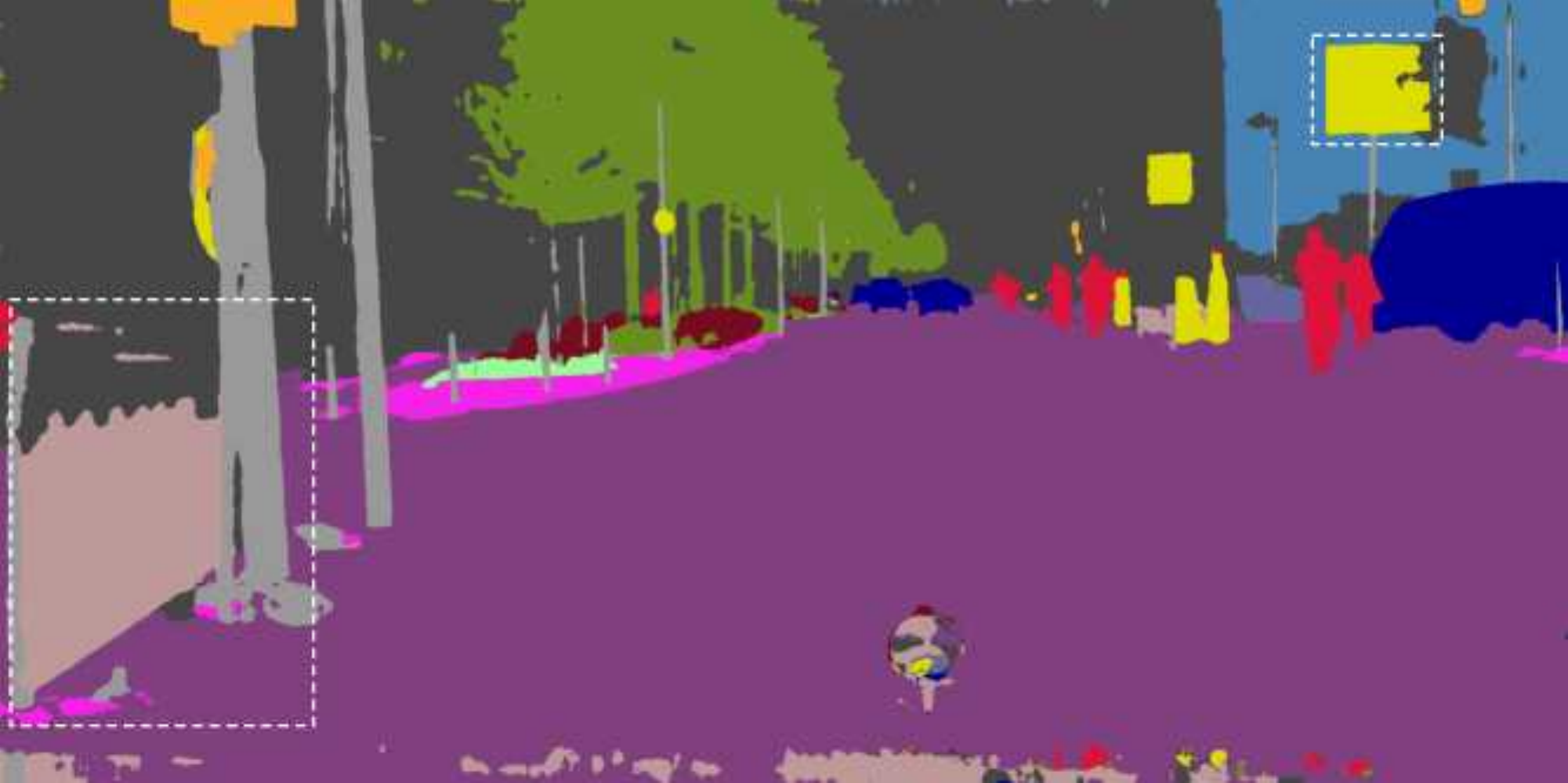}
    \end{subfigure}    \begin{subfigure}{.2\textwidth}
        \centering
        \includegraphics[width=.98\linewidth]{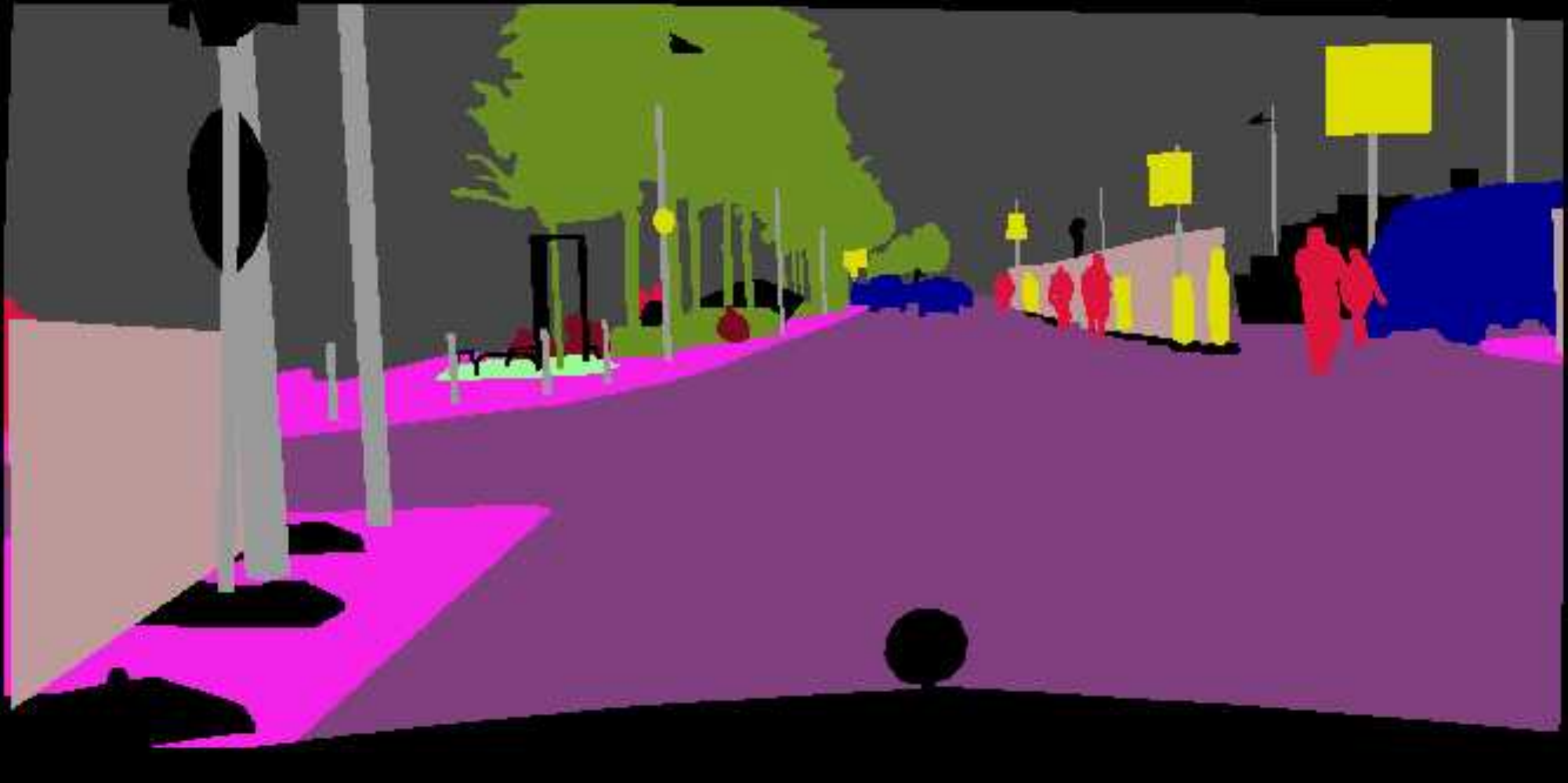}
    \end{subfigure}}
\makebox[\linewidth][c]{    \begin{subfigure}{.2\textwidth}
        \centering
        \includegraphics[width=.98\linewidth]{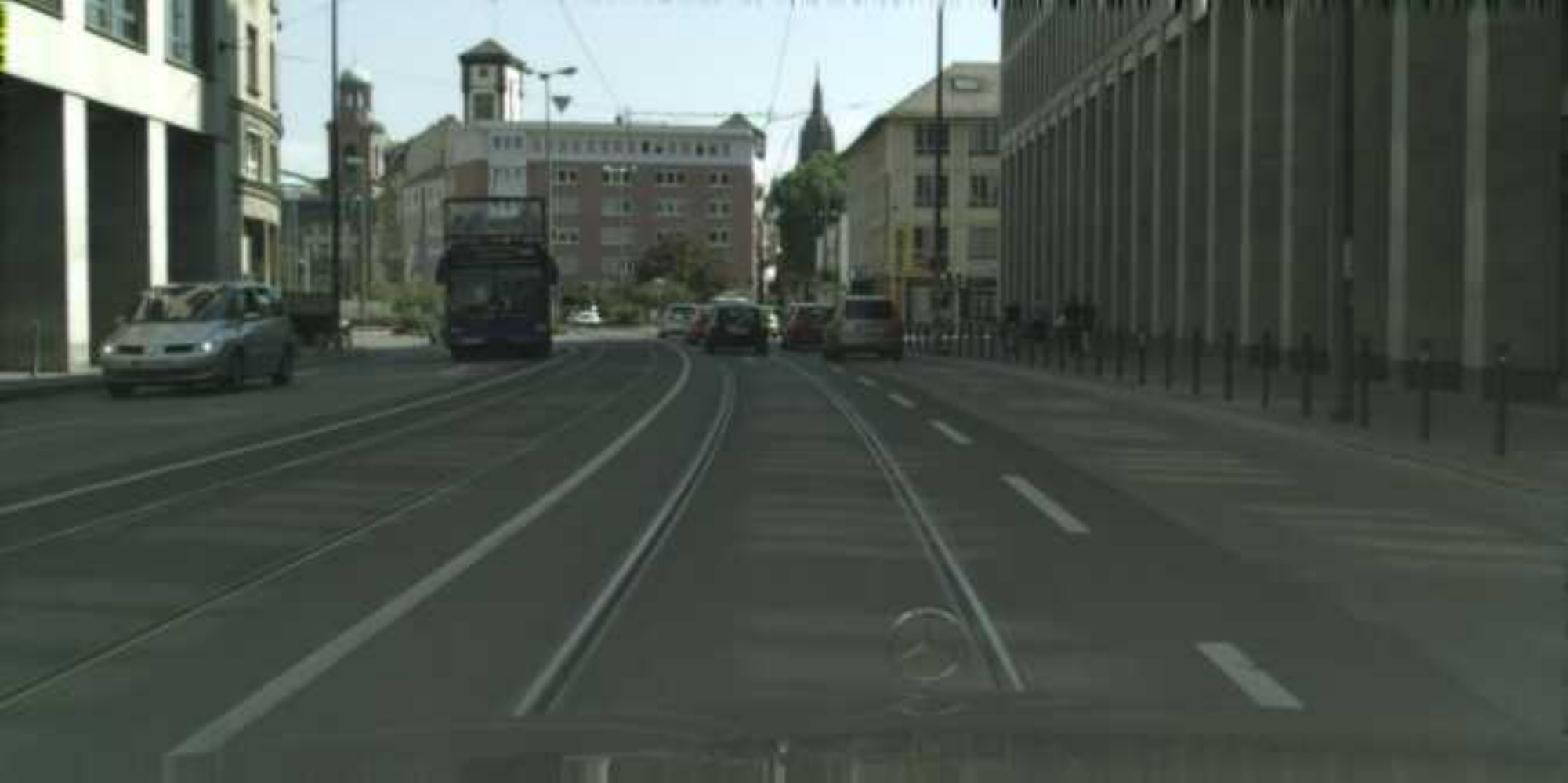}
    \end{subfigure}    \begin{subfigure}{.2\textwidth}
        \centering
        \includegraphics[width=.98\linewidth]{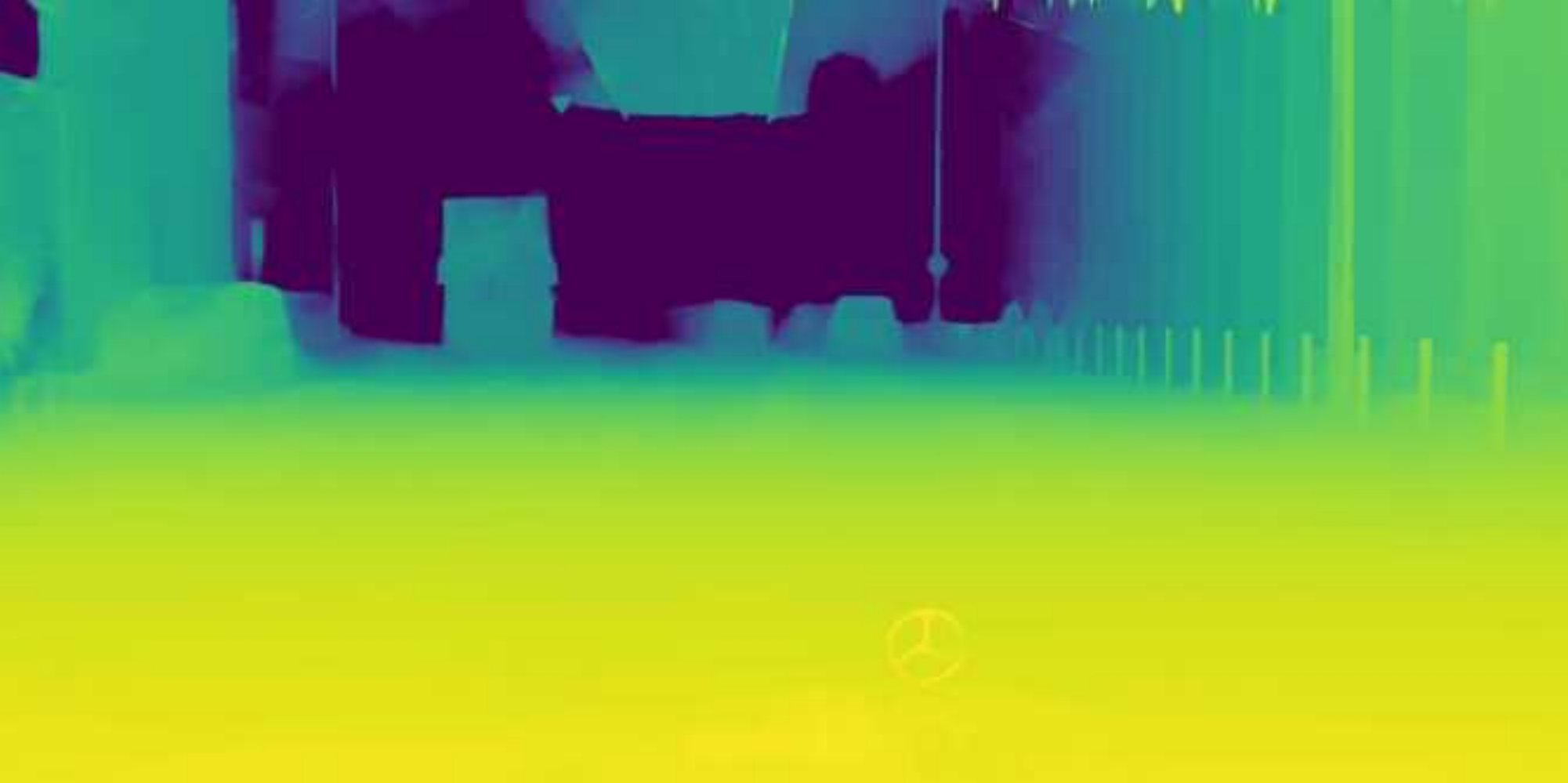}
    \end{subfigure}    \begin{subfigure}{.2\textwidth}
        \centering
        \includegraphics[width=.98\linewidth]{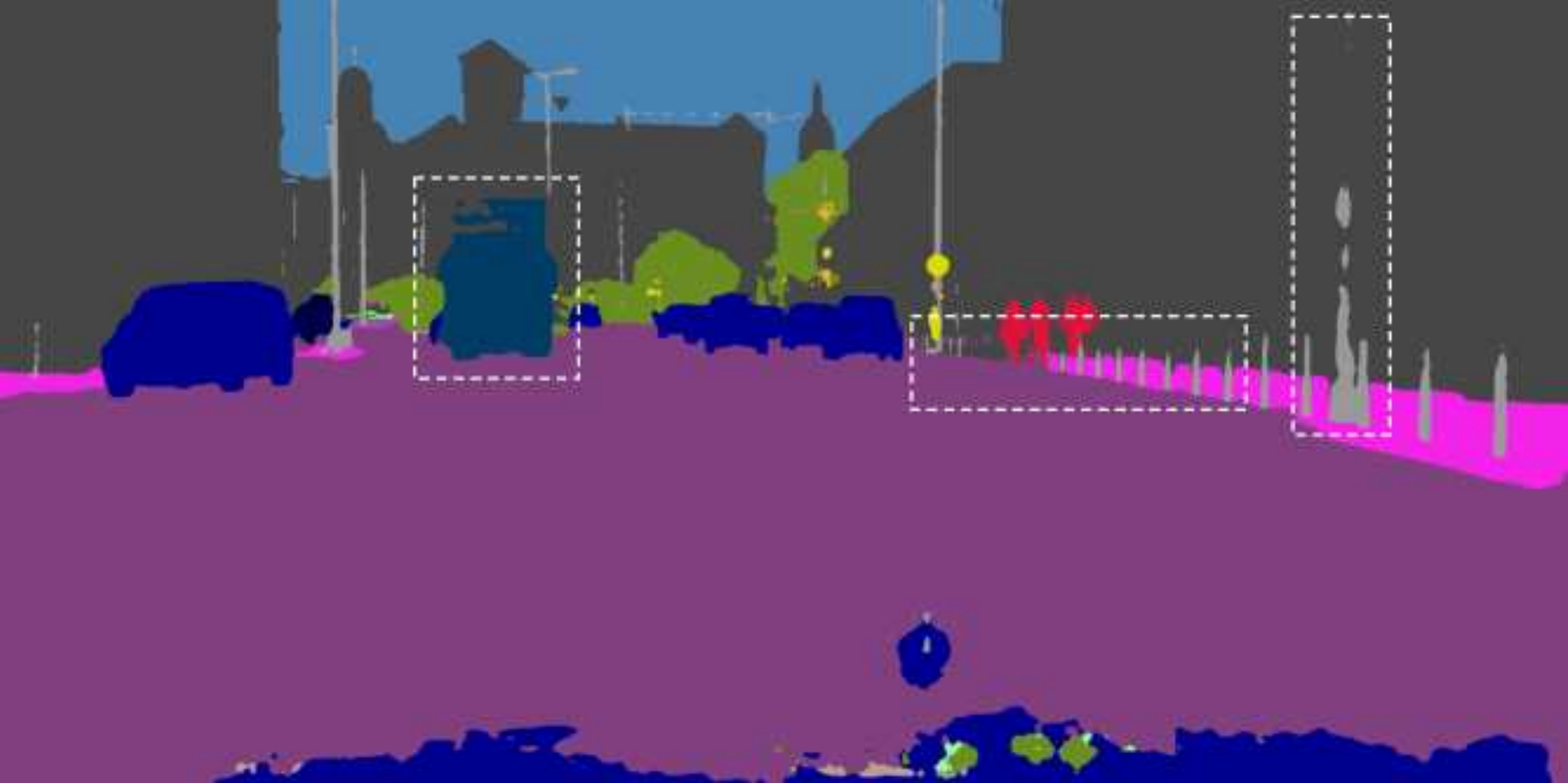}
    \end{subfigure}    \begin{subfigure}{.2\textwidth}
        \centering
        \includegraphics[width=.98\linewidth]{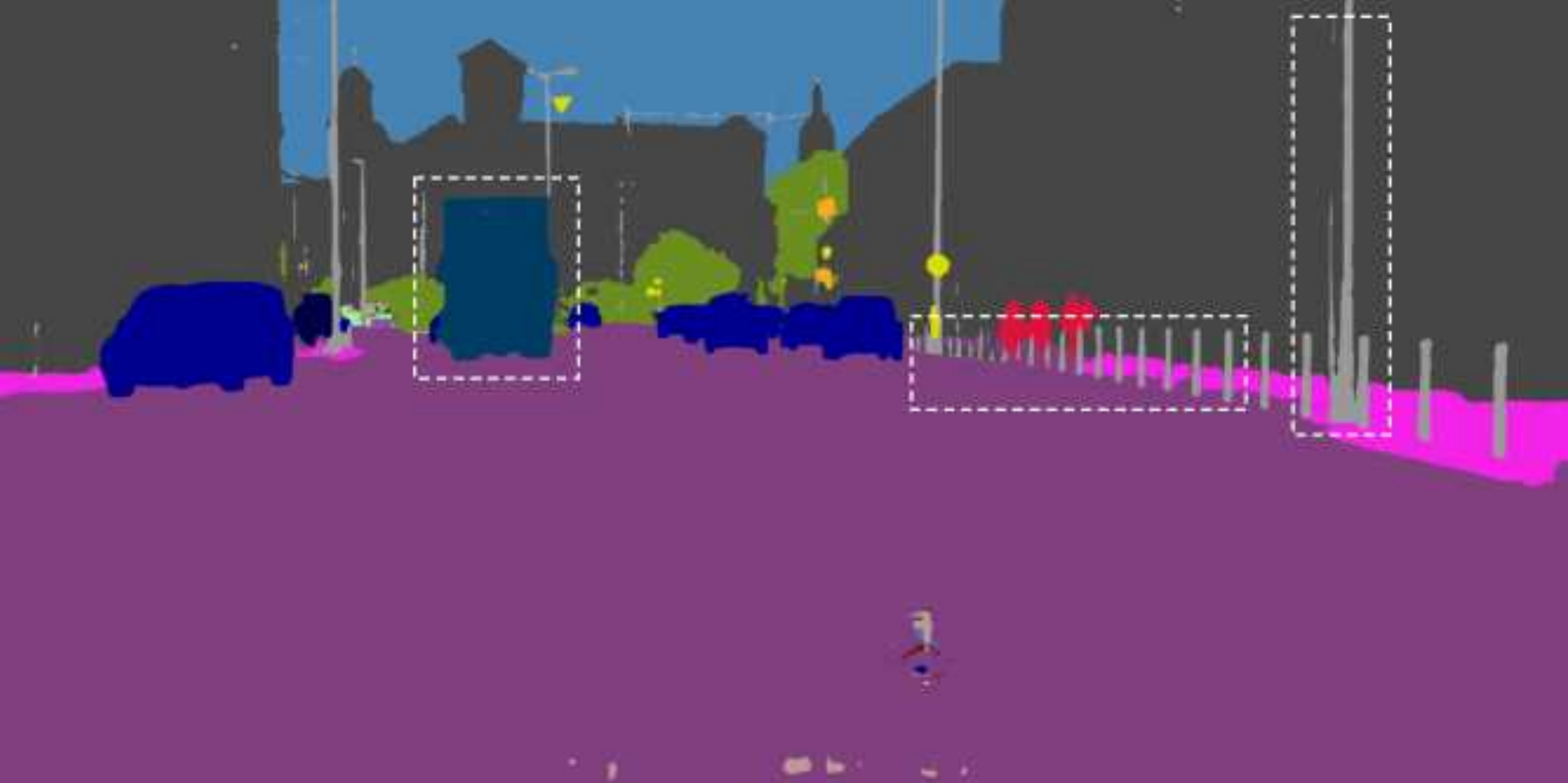}
    \end{subfigure}    \begin{subfigure}{.2\textwidth}
        \centering
        \includegraphics[width=.98\linewidth]{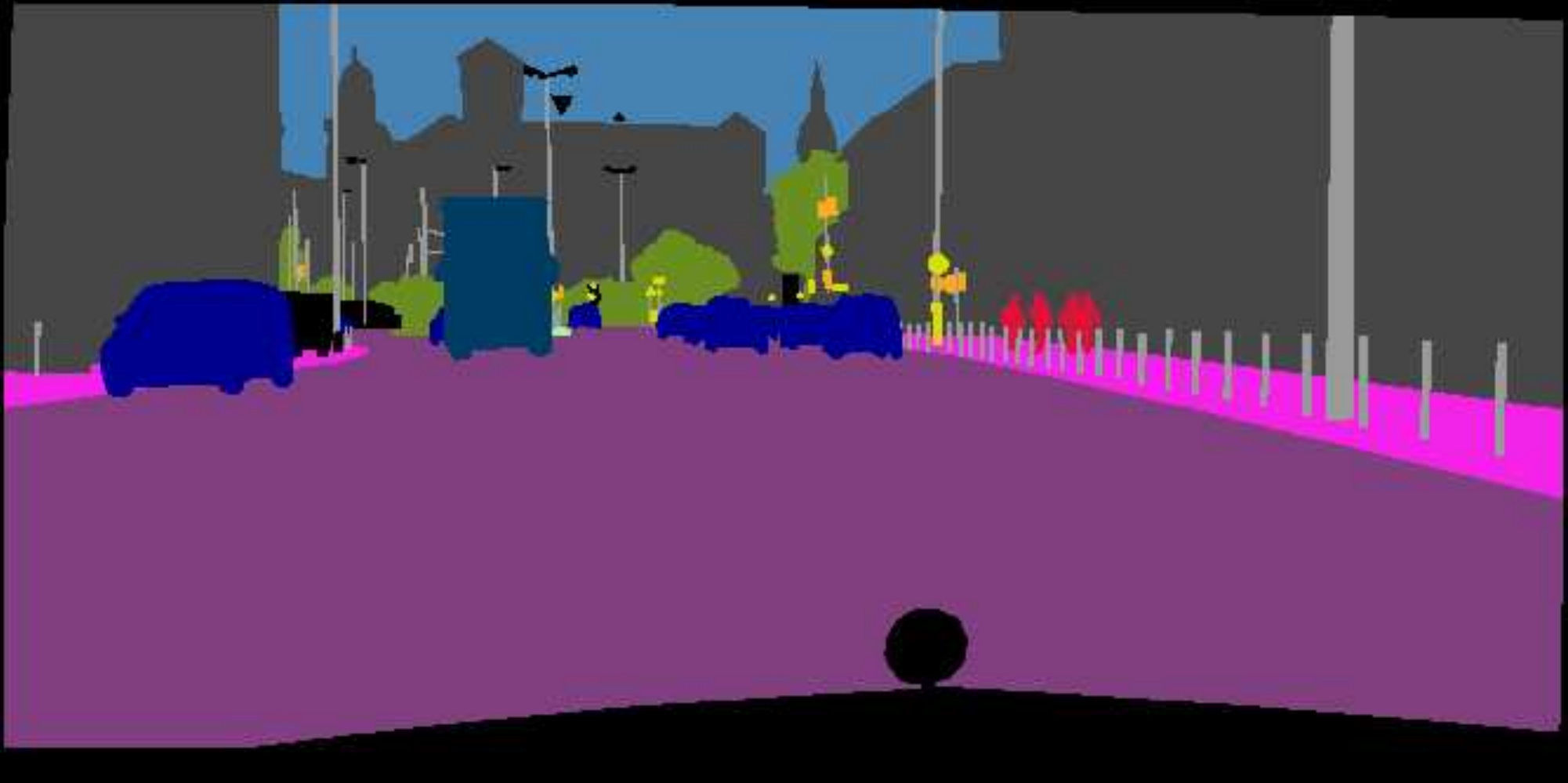}
    \end{subfigure}}
\makebox[\linewidth][c]{    \begin{subfigure}{.2\textwidth}
        \centering
        \includegraphics[width=.98\linewidth]{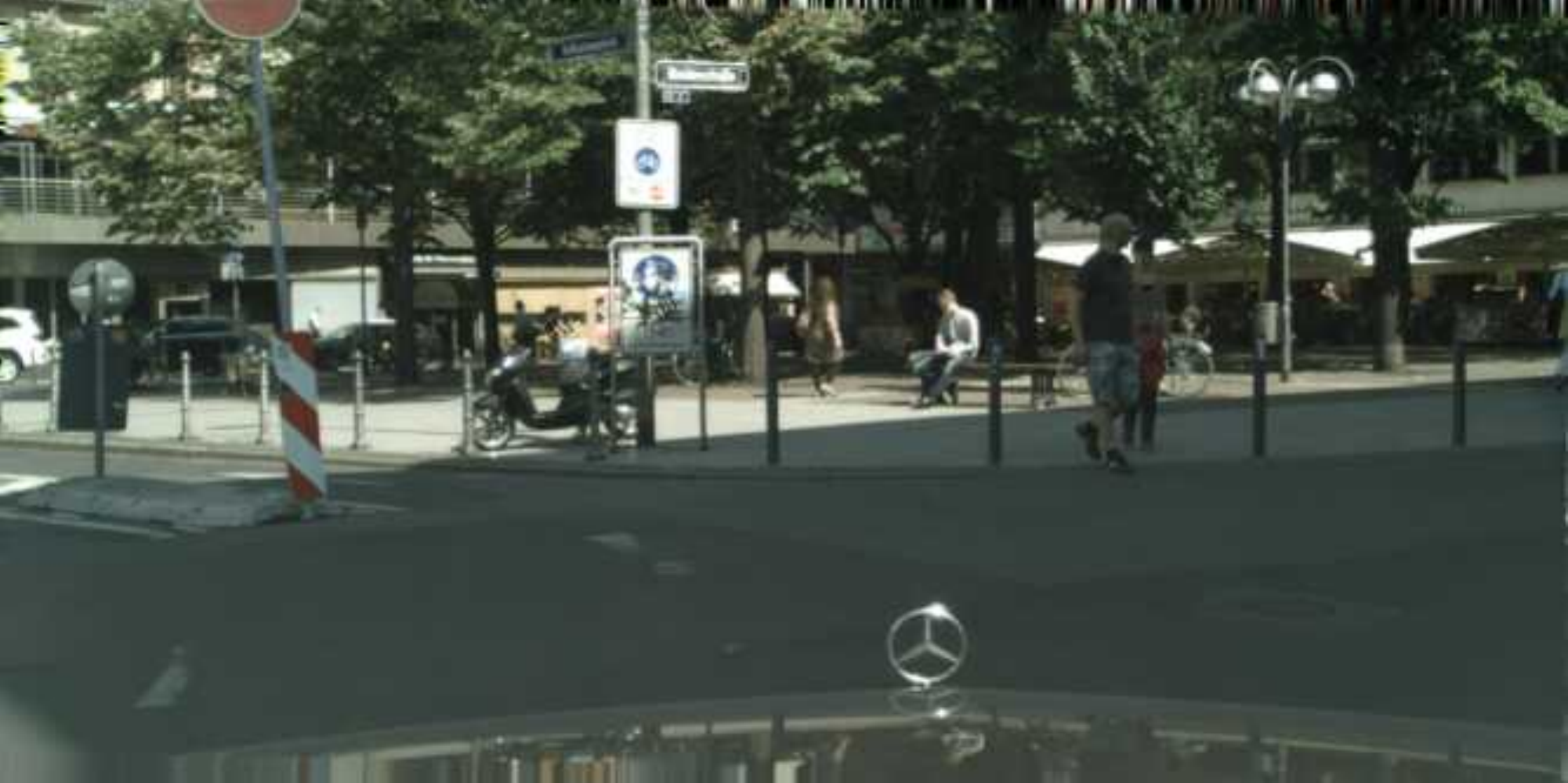}
        \caption*{Target Image}
    \end{subfigure}    \begin{subfigure}{.2\textwidth}
        \centering
        \includegraphics[width=.98\linewidth]{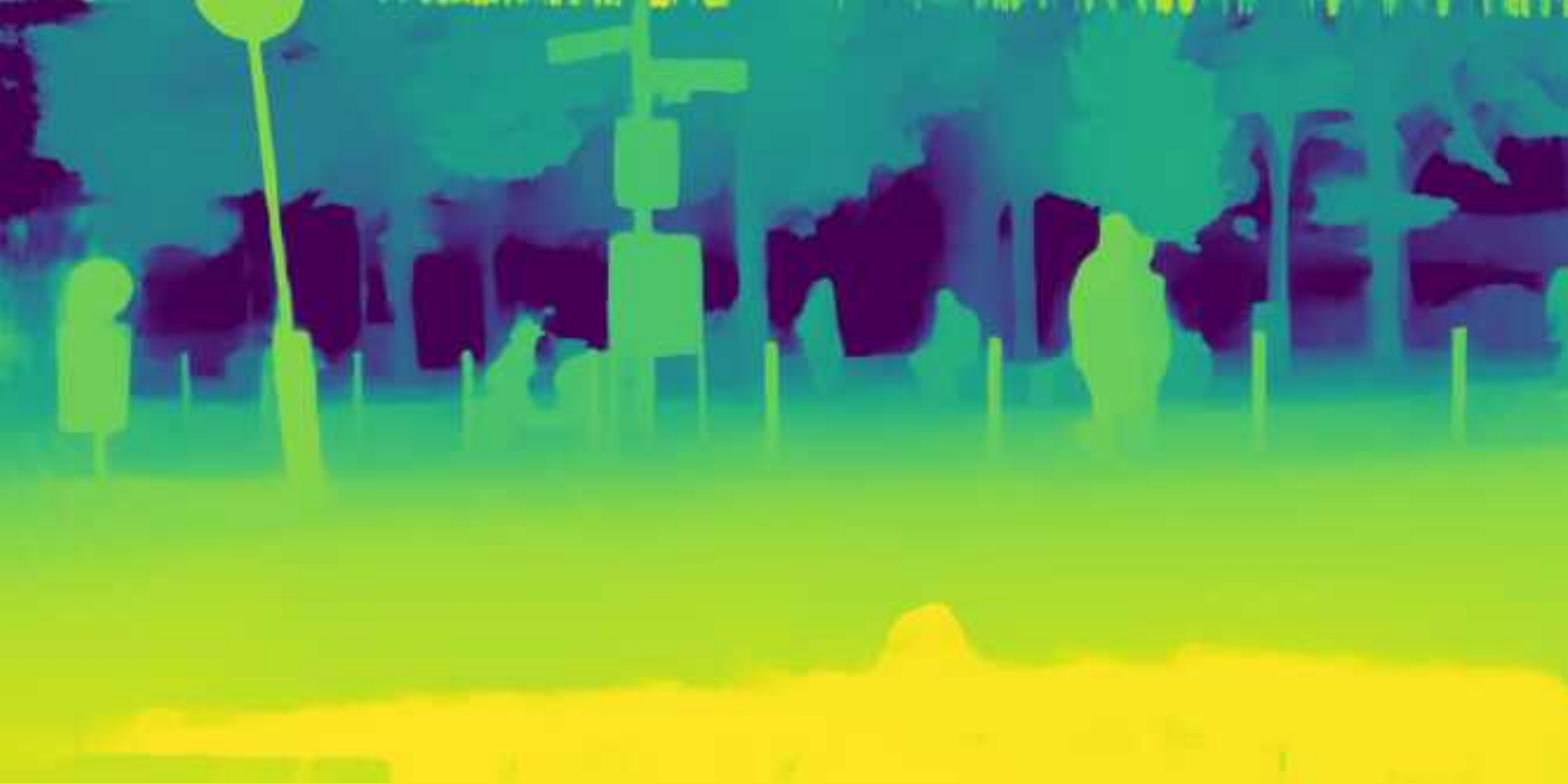}
        \caption*{Estimated Depth}
    \end{subfigure}    \begin{subfigure}{.2\textwidth}
        \centering
        \includegraphics[width=.98\linewidth]{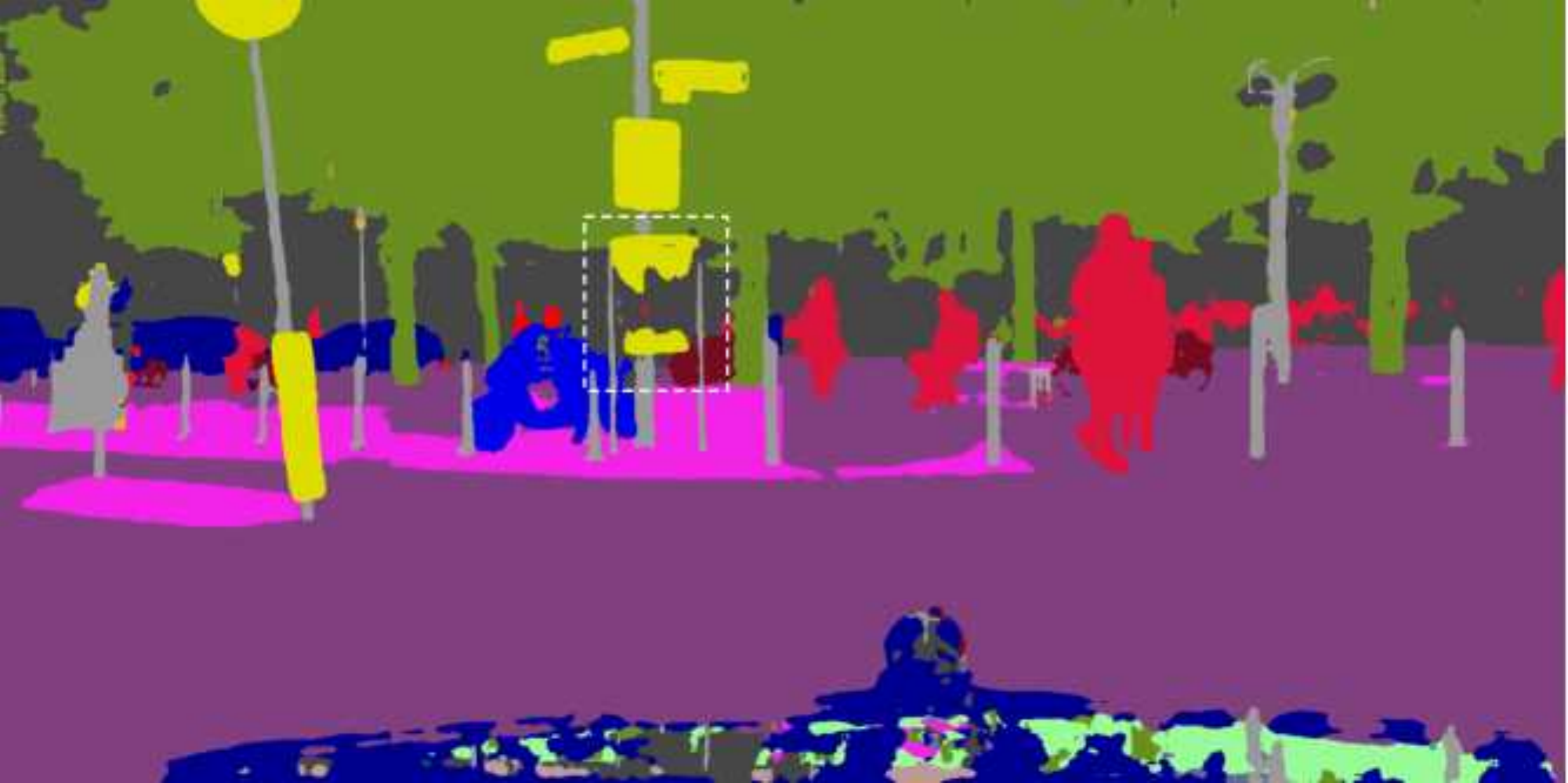}
        \caption*{HRDA~\cite{HRDA}}
    \end{subfigure}    \begin{subfigure}{.2\textwidth}
        \centering
        \includegraphics[width=.98\linewidth]{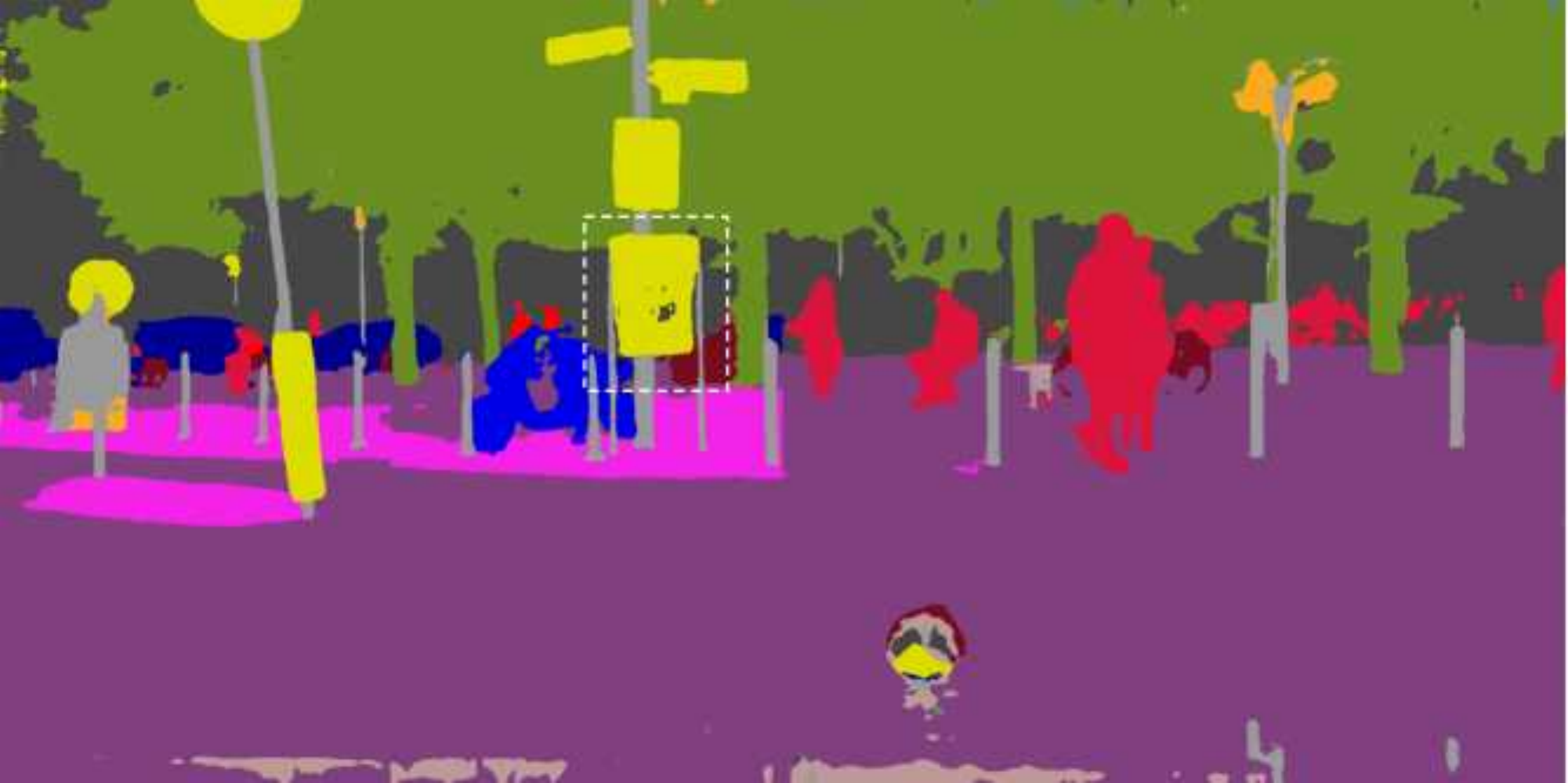}
        \caption*{\method\ (ours)}
    \end{subfigure}    \begin{subfigure}{.2\textwidth}
        \centering
        \includegraphics[width=.98\linewidth]{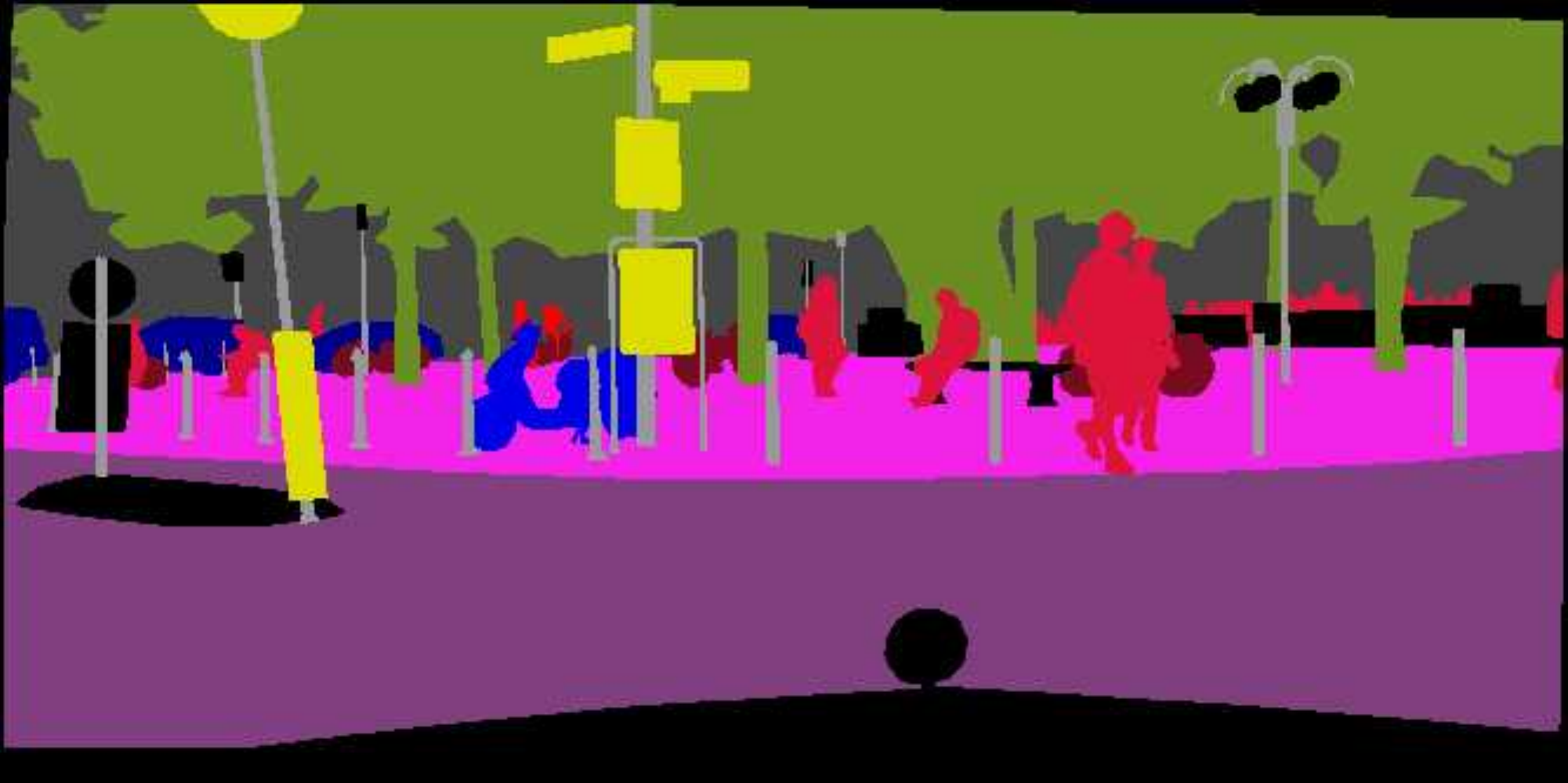}
        \caption*{Ground Truth}
    \end{subfigure}}

\centering

\caption{\textbf{Qualitative results using HRDA~\cite{HRDA}.}
In rows 4 and 5, depth cues correct mislabeled patches as part of larger structures.
Rows 1, 3, and 6 obtain more accurate \textit{traffic sign} segmentation with depth cues. 
Row 2 shows improvements to a single large \textit{pole}, while row 3 and 5 improve on smaller thin poles.
}
\label{fig:qualitative_supp_hrda}
\end{figure*}

In ~\cref{fig:qualitative_supp_daf,fig:qualitative_supp_mic_daf,fig:qualitative_supp_hrda}, we show further qualitative examples of our method using various UDA methods. We find that generally, improvements hold for fine structure \eg poles across different architectures. We also observe that our method can exploit depth information to rectify incoherent segmentations.

\clearpage

\clearpage

\clearpage
\end{document}